\documentclass[fleqn,twoside]{article}
\usepackage{espcrc1}

\newcommand{\AmS}{{\protect\the\textfont2
  A\kern-.1667em\lower.5ex\hbox{M}\kern-.125emS}}
\title{TOWARDS COMBINATORIAL CLUSTERING:
   Preliminary Research Survey
  }

\author{Mark Sh. Levin
\address{
 Inst. for Information Transmission Problems, Russian Academy of
 Sciences\\
 19 Bolshoj Karetny Lane, Moscow 127994, Russia\\
 E-mail: mslevin@acm.org
 }
 }

\begin{document}

\maketitle

\begin{abstract}

 The paper describes clustering problems
 from the combinatorial viewpoint.
 A brief systemic
  survey is presented
 including the following:
 (i) basic clustering problems
 (e.g., classification, clustering, sorting,
 clustering with an order over cluster),
 (ii) basic approaches to
 assessment of objects and
 object proximities
 (i.e., scales, comparison, aggregation issues),
 (iii) basic approaches to evaluation of
 local quality characteristics for clusters
  and
 total quality characteristics for clustering solutions,
 (iv)
 clustering as multicriteria optimization problem,
 (v) generalized modular clustering framework,
 (vi) basic clustering models/methods
 (e.g., hierarchical clustering, k-means clustering,
 minimum spanning tree based clustering,
 clustering as assignment,
 detection of clisue/quasi-clique based clustering,
 correlation clustering,
 network communities based clustering),
 Special attention is targeted to
 formulation of clustering as
 multicriteria optimization models.
%

%
%
 Combinatorial optimization models are used as auxiliary problems
 (e.g., assignment, partitioning, knapsack problem,
 multiple choice problem,
  morphological clique problem,
 searching for consensus/median for structures).

 Numerical examples illustrate
 problem formulations, solving methods, and applications.
%
 The material can be used as follows:
 (a) a research survey,
 (b) a fundamental for designing the structure/architecture
    of composite modular clustering software,
 (c) a bibliography reference collection,
  and
 (d) a tutorial.

~~~~~~~~~~~


\vspace{1pc}
\end{abstract}

\newcounter{cms}
\setlength{\unitlength}{1mm}

\tableofcontents

\newpage
\section{Introduction}

 Recently clustering/classificataion problems
 have been widely used
 in many
 domains
%
%
%
 (Table 1.1).


\begin{center}
 {\bf Table 1.1.} Main application  domains of clustering-like models/problems \\
\begin{tabular}{| c | l | l | l|}
\hline
 No.  & Some applied domains & Basic applied problem(s) & Source(s)  \\
\hline

 1. &Design/analysis of information  &Document clustering/classification,
 & \cite{bae99,berry99,fan09,man08}\\

 & systems, information retieval& design of hierarchy/ontology/menu, &
 \cite{mur95,schen04,stein00,van79}\\

   & & information retrieval &  \cite{voor86,willet88,zhong05}\\


 2. & Web systems, web services& Clustering of Web sites, design of
      & \cite{boley99,brin98,brod97,carp09,flor98}\\

   &  & hierarchical Web systems, search & \cite{man08,schen04,sinka02,zamir99}\\

 3. & Data mining\&knowledge discovery &
 Detection of information objects, &
  \cite{berk06,cook00,fish87,halg05,han01}\\

  &
  &associations, rules, structures&
 \cite{hand01,he07,mirkin05,mul09}\\


 4. & Medical/technical diagnostics, &
  Definition of solution classes, & \cite{fur13,fursok11,hir00,kor04} \\

   &system testing, maintenance of &  diagnostics (assignment/multiple
   & \cite{lei08,lev08b,mas99,pat89} \\
  & systems & assignment of patients/system    & \cite{szo82,wal06,yang00,zogg06} \\
  & & components into solution classes), &  \\

    & & clustering of symptoms/faults, etc.  &  \\

 5. & Graph partitioning (network
   &Grouping of graph vertices
   & \cite{alp95,aug04,dutt96,gar79,hag92} \\
   &  design, VLSI design, etc.) && \cite{karyp99,ker70,trif08,xux07}\\

 6. & Cell formation in industrial/
  &  Grouping of machines & \cite{chu89,gold13,selim98,srin94} \\
  & manufacturing  systems  &   &  \\

 7. & Anomaly detection (networks,  &
  Finding patterns in data that do
  &\cite{cha09,frilast07,pat07,zhong05a}\\
 & distributed systems, etc.)& not conform to expected behavior &\\


 8. & Computer vision  (images/scenes, &
    Clustering/segmentation of images,
    &  \cite{jain99,jolion91,shap01,sri05} \\

   & object trajectories)
   & shape analysis, detection of events & \cite{wu93}\\

 9. &
  Trajectory
  clustering/classification &
  Tracking; traces initialization/
  & \cite{berin06,chenw04,fu05,guha00a}\\

  &(e.g., air-traffic control), system  &maintenance; classification/
      & \cite{has14,last02,lee07,lev12clique}\\

 & testing/maintenance, monitoring
 & clustering/fusion of streams&  \cite{lev15,li02,li06}\\

 10. & Chemistry, biology, gene expression
 & Classification/clustering of &
   \cite{abu13,baldi02,gus97,chu12} \\

 &data clustering in DNA microarray
  &chemical objects (elements,&
     \cite{down02,jia04,mad04,monti03} \\
   &  technology   &  structures), detection of natural &
   \cite{saeed12,saeed12a,schuf07,wells12}\\

   & & structures, interesting patterns   &
  \cite{willett86,yeu01,xu01}\\

 11. & Communication/sensor systems/ & Management, clustering of nodes,
 & \cite{abbas07,an01,band03,band04} \\

 &networks, computer networks
  & clustering based routing,
   & \cite{baner01,bens11,cham09,chatt02} \\

 && detection of cluster heads,
 &\cite{chen02,for09,gerla95,ghia02}\\

 && design of hierarchical network & \cite{klein77,klein80,shama04,yu07}  \\

  12. &Management,  planning, marketing,
   & Hierarchical management, design
  & \cite{arabie94,berry96,cai00,chiu02,dewa00}\\

  & evaluation of economical objects
  & of management hierarchy, team
  & \cite{dick87,gross97,har85,harris91}\\

  & (e.g., financial instruments, tasks, & design, segmentation of market,
  & \cite{ket96,lev09bi,lev15,mola04}\\
   &firms, countries)& segmentation of customers
   &\cite{ngai09,ronen85,simcha85,wang09a}\\

 13. & Social sciences, political marketing,
 &Clustering of social/phychological
 & \cite{bata14,duan12,dore04,gran89} \\

  &  social network analysis, recognition
  & objects, analysis of networks/ & \cite{handc07,levfim09,mis07,ops09}\\

   & of communities, econometrics, etc.
   & hierarchies, evaluation, planning &  \cite{osuagwu08,rob76,vanr77,was94}\\

 14.& Education (evaluation, course  &
 Evaluation (students, courses),&  \cite{bid09,burke94,cheng05,comm88}\\
 &design, cluster grouping, planning)
 &  clustering of students, timetabling,  &  \cite{gent09,nem98,qu09,romero07}\\
 &  &  educational data mining  &  \cite{romero10,romero10a,wineb08}\\

\hline
\end{tabular}
\end{center}

%
 The significance of
 clustering/classification
  is essentially increased,
  for example, in
 the following contemporary fields:
 data analysis,
 management and decision making,
 communication systems, engineering,
 chemistry,
 biomedicine,
 information retrieval,
 system monitoring,
 social sciences,
 network modeling and analysis
 (e.g.,
 \cite{agrawal05,band04,chatt02,day86,ghosh11,jain99,mirkin05,monti03,roy96,rocha13a,stein00,zop02})
 (Fig. 1.1).
%

%
%
 In two recent decades, excellent well-known surveys and books on clustering
 problems and methods have been published (e.g.,
 \cite{jain99,mirkin05,xu05,xu09,zop02}).
%
%
 Many research publications
 including surveys and books are targeted to special clustering approaches,
 for example:
 clustering based on fuzzy data
 (e.g.,
   \cite{crespo05,dur06}),
%
 support vector clustering (e.g., \cite{ben01}),
  cross-entropy based clustering (e.g., \cite{jung07,kroe07}),
 online clustering
 (e.g., \cite{barb08,berin06,chan07,last02}),
 dynamic clustering (e.g., \cite{bens11,chenw04}),
 consensus clustering (e.g., \cite{chu12,gue11}),
 graph-based clustering
 (e.g., \cite{fel11,gra05,koch05}),
 clustering ensembles
 (e.g., \cite{vega11,ghosh11,he05}),
 clustering based on hesitant fuzzy information
 (e.g.,  \cite{nchen14,zhang15}),
 multicriteria clustering  (e.g., \cite{desmet09}),
 correlation clustering (e.g., \cite{dem06}).

\begin{center}
\begin{picture}(110,90)
\put(15,00){\makebox(0,0)[bl]{Fig. 1.1. Clustering
 as basic support problem}}

\put(24.5,45){\oval(49,28)}

\put(02,53){\makebox(0,0)[bl]{Clustering (``hard''
 problems,}}

\put(02,49){\makebox(0,0)[bl]{``soft'' problems):}}
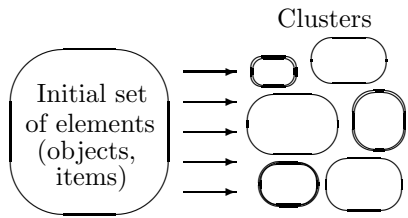
\put(02,45){\makebox(0,0)[bl]{clustering as partitioning,}}
\put(02,41){\makebox(0,0)[bl]{classification (e.g., diagnosis),}}
\put(02,37){\makebox(0,0)[bl]{sorting (stratification), }}
\put(02,33){\makebox(0,0)[bl]{multiple clustering}}


\put(49,55){\vector(1,2){16}}

\put(49,53){\vector(2,3){16}}

\put(49,51){\vector(1,1){16}}

\put(49,49){\vector(2,1){16}}

\put(49,45){\vector(1,0){16}}

\put(49,41){\vector(2,-1){16}}

\put(49,38){\vector(1,-1){16}}

\put(49,36){\vector(2,-3){16}}


\put(66,86){\makebox(0,0)[bl]{Social networks analysis}}

\put(65,84){\line(1,0){40}} \put(65,90){\line(1,0){40}}
\put(65,84){\line(0,1){6}} \put(105,84){\line(0,1){6}}


\put(66,76){\makebox(0,0)[bl]{Pattern recognition}}

\put(65,74){\line(1,0){40}} \put(65,80){\line(1,0){40}}
\put(65,74){\line(0,1){6}} \put(105,74){\line(0,1){6}}

\put(66,66){\makebox(0,0)[bl]{Decision making}}

\put(65,64){\line(1,0){40}} \put(65,70){\line(1,0){40}}
\put(65,64){\line(0,1){6}} \put(105,64){\line(0,1){6}}

\put(66,56){\makebox(0,0)[bl]{Information retrieval}}

\put(65,54){\line(1,0){40}} \put(65,60){\line(1,0){40}}
\put(65,54){\line(0,1){6}} \put(105,54){\line(0,1){6}}

\put(66,46){\makebox(0,0)[bl]{Knowledge discovery }}
\put(66,42){\makebox(0,0)[bl]{and data mining}}

\put(65,40){\line(1,0){40}} \put(65,50){\line(1,0){40}}
\put(65,40){\line(0,1){10}} \put(105,40){\line(0,1){10}}

\put(66,32){\makebox(0,0)[bl]{Large scale data analysis}}

\put(65,30){\line(1,0){40}} \put(65,36){\line(1,0){40}}
\put(65,30){\line(0,1){6}} \put(105,30){\line(0,1){6}}

\put(66,22){\makebox(0,0)[bl]{Machine learning}}

\put(65,20){\line(1,0){40}} \put(65,26){\line(1,0){40}}
\put(65,20){\line(0,1){6}} \put(105,20){\line(0,1){6}}

\put(66,12){\makebox(0,0)[bl]{Systems monitoring}}

\put(65,10){\line(1,0){40}} \put(65,16){\line(1,0){40}}
\put(65,10){\line(0,1){6}} \put(105,10){\line(0,1){6}}


\put(79,07){\makebox(0,0)[bl]{{\bf . ~. ~.}}}

\end{picture}
\end{center}
%



 In recent decade, the significance of combinatorial approaches to clustering
 have been increased.
 In Table 1.2,
  some
   research efforts in
  combinatorial approaches to clustering are pointed out.


 This material contains an author's ``architectural'' engineering glance to
 combinatorial clustering problems.
 A special attention is targeted to
 formulation of clustering problems as
 multicriteria optimization models
 which are based on usage
 of various quality parameters for clustering solutions
 (e.g., clusterings as partitions, hierarchical clusterings).
%
%
 Combinatorial optimization models are used as auxiliary problems
 (e.g.,  partitioning, assignment, knapsack problem,
 multiple choice problem, matching problem).

 In recent years,
 various fuzzy clustering methods
 have been widely studied and used
  (e.g.,
 \cite{bar99,chen13,nchen14,dur06,grave10,hop99,jain99,kim04,kris95,miy90,oli07,sato06,xu14a,zhang15})
 and these approaches are not considered in the material.
%
%
%

%

%
 The presented materials can be useful for educational courses
 and student projects
 in computer science, engineering, management,
 social sciences applications
 (e.g., \cite{lev09edu,levcsedu11,lev11ed}).

\newpage

\begin{center}
 {\bf Table 1.2.} Combinatorial approaches to clustering \\
\begin{tabular}{| c | l |l|}
\hline
 No.  & Approaches, algorithmic schemes & Source(s)  \\
\hline

 1.& Some surveys: &\\
 1.1. & General  &
  \cite{gar79,jain99,mirkin99} \\

 1.2.& Graph clustering& \cite{scha07}\\

 1.3.& Approximate graph partitioning &
  \cite{even99}\\

 1.4.& Cross-entropy method for clustering, partitioning &
  \cite{kroe07,rub02,tabor14}\\

 1.5.&
 Cell formation (in industrial engineering)
 & \cite{gold13,selim98,srin94} \\

 1.6.& Clustering ensemble algorithms &
  \cite{vega11}\\

 1.7.& Multicriteria classification and sorting methods &
 \cite{roy96,zop02}\\

\hline
 2. &
 Basic combinatorial optimization problems: & \\

 2.1.& Minimal spanning tree approach &
 \cite{gry06,mul12,paiv05,peter10,trif08,wang09,xu01,zhong10} \\

 2.2.& Partitioning based clustering &
 \cite{aug04,boley99,condon01,ding01,even99,spiel13} \\

 2.3.& Assignment/location based clustering &  \cite{goldb08} \\

 2.4.& Graph matching
  &
  \cite{schen04}\\

 2.5.& Dominant set based clustering  &
 \cite{chen02,han07,li06,pav07,youn06} \\

 2.6.& Covering based clustering &
 \cite{agrawal05,mul09,salz09} \\

 2.7.& Clique based clustering &
 \cite{agrawal05,berk06,but06,duan12,gra05,koch05,shamir02}  \\

  2.8.& Structural clustering (detection of communities) &
 \cite{agar08,new06,new04,port09,xux07}  \\

\hline
 3.& Correlation clustering  &
 \cite{achtert07,bansal04,dem06,kri09,swam04}  \\

\hline
 4.& Graph-based data clustering with overlaps&
 \cite{fel11}\\

\hline
 5. & Segmentation problems & \cite{klein98}\\

\hline
 6.& Cluster graph modification problems&
 \cite{shamir02,sham04}\\

\hline
 7.& Multi-criteria decision making in  clustering-sorting
  &\cite{fur13,rocha13a,rocha13,zop02}\\

\hline
 8.& Consensus clustering: &\\

 8.1.& Voting-based consensus of cluster ensembles&
 \cite{ayad10,saeed12a}\\

 8.2.& Consensus partitions&
 \cite{gue11}\\

\hline
 9. & Algorithmic schemes: &\\

 9.1.& Enumerative methods: & \\

 9.1.1. & Branch-and-bound methods &
 \cite{cheng95}\\

 9.1.2. & Dynamic programming & \cite{yeh86}\\


 9.2. & Local optimization heuristics: &\\

 9.2.1.& Simulated annealing algorithms & \cite{brown92,osman94,selim91}\\

 9.2.2.& Tabu search algorithms& \cite{osman94,sung00} \\

 9.2.3.&
 Ant colonies algorithms
 & \cite{jov15,yang06}\\

 9.2.4.& PSO methods & \cite{chenc04,till02,vand03} \\

 9.2.5.&Variable neighborhood search & \cite{hansen97a,hansen07,hansen07b,hansen09} \\

 9.3. &Genetic algorithms, evolutionary strategies
 & \cite{babu94,cow99,hru09,ozy09,tseng01}\\

 9.4. & Hyper-heuristic approach & \cite{cobos11,kumari13,tsai12}\\

\hline
\end{tabular}
\end{center}

\newpage

\section{General Preliminary Glance}

\subsection{Preliminaries}

%
 From the structured viewpoint,
 the following basic clustering problem formulations can be pointed out
 \cite{day86,doro71,ghosh11,gord99,har75,jain88,jain99,kau90,kot04,kot07,mirkin96,mirkin05,monti03,mull15,rocha13,roy96,zop02}:
 (i) set partitioning clustering (Fig. 2.1),
 (ii) classification
 (i.e., solution classes are predefined as in diagnostics)
  (Fig. 2.2),
  (iii) sorting (group ranking, stratification) problem
   (the obtained clusters are linear ordered) (Fig. 2.3),
  (iv) hierarchical clustering (Fig. 2.4),
 (v) multiple clustering (i.e., obtaining \(n\) different clustering solutions
  while taking into account \(n\) goals/models; alternative clustering)
  (Fig. 2.5), and
  (vi) consensus clustering (aggregation
  of clustering solutions;
  clustering ensembles) (Fig. 2.6).

%
 Table 2.1 contains initial data for an illustrative numerical example:
 (a list of item/students
 and their skill estimates upon parameters/criteria):
 (1) formal approaches, modeling (i.e., mathematics, physical modeling) \(C_{1}\),
 (2) applied computer science, computing (i.e., software development,
 software implementation, computing) \(C_{2}\),
 (3) engineering science domain (i.e., information transmission, radio channels,
 antenna  devices, sender/receiver devices, networking) \(C_{3}\),
 (4) measurement radio techniques \(C_{4}\),  and
 (5) preparation 
  of the technical documentations
  (reports, papers, presentations \(C_{5}\).
 Here ordinal scale \([3,4,5]\) is used:
 excellent (\(``5''\)), good (\(``4''\)), sufficient (\(``3''\)).

\begin{center}
\begin{picture}(67,35)

\put(00,00){\makebox(0,0)[bl]{Fig. 2.1. Clustering as
 partitioning}}

\put(10.5,16){\oval(21,22)}


\put(3.5,20){\makebox(0,0)[bl]{Initial set}}
\put(2,16){\makebox(0,0)[bl]{of elements}}
\put(03.5,11.5){\makebox(0,0)[bl]{(objects,}}
\put(06,08){\makebox(0,0)[bl]{items)}}

\put(23,24){\vector(1,0){07}} \put(23,20){\vector(1,0){07}}
\put(23,16){\vector(1,0){07}} \put(23,12){\vector(1,0){07}}
\put(23,08){\vector(1,0){07}}


\put(35.5,30){\makebox(0,0)[bl]{Clusters}}



\put(37,09){\oval(08,6)} \put(37,09){\oval(07.7,5.7)}
\put(37,09){\oval(07.4,5.4)}

\put(47,09){\oval(10,7)}


\put(37.5,17){\oval(12,8)}

\put(49,17.5){\oval(07,8)} \put(49,17.5){\oval(06.6,7.6)}

\put(45,25.5){\oval(10,6)}


\put(35,24){\oval(06,4)} \put(35,24){\oval(05.5,3.5)}

\end{picture}
%
\begin{picture}(62.5,34.5)

\put(04,00){\makebox(0,0)[bl]{Fig. 2.2. Classification
 problem}}

\put(10.5,16){\oval(21,22)}


\put(3.5,20){\makebox(0,0)[bl]{Initial set}}
\put(2,16){\makebox(0,0)[bl]{of elements}}
\put(03.5,11.5){\makebox(0,0)[bl]{(objects,}}
\put(06,08){\makebox(0,0)[bl]{items)}}

\put(21.5,23){\vector(1,2){02.5}} \put(22,19){\vector(1,1){04}}
\put(22,14){\vector(1,0){05}} \put(22,08){\vector(1,0){05}}


\put(42,25){\makebox(0,0)[bl]{Predefined}}
\put(46,22){\makebox(0,0)[bl]{solution}}
\put(48,19){\makebox(0,0)[bl]{classes}}


\put(24.5,30.5){\makebox(0,0)[bl]{Class \(1\)}}

\put(30,31.5){\oval(13,6)}


\put(28,23){\makebox(0,0)[bl]{Class \(2\)}}

\put(34,24){\oval(13,6)}


\put(31,19){\makebox(0,0)[bl]{{. . .}}}


\put(29,12.3){\makebox(0,0)[bl]{Class \((k-1)\)}}

\put(39,14){\oval(21,6)}


\put(29,06){\makebox(0,0)[bl]{Class \(k\)}}

\put(35,07){\oval(13,5.5)}

\end{picture}
\end{center}

\begin{center}
\begin{picture}(72,50)

\put(00,00){\makebox(0,0)[bl]{Fig. 2.3. Sorting (stratification)
 problem}}

\put(10.5,22){\oval(21,22)}

\put(3.5,26){\makebox(0,0)[bl]{Initial set}}
\put(2,22){\makebox(0,0)[bl]{of elements}}
\put(03.5,17.5){\makebox(0,0)[bl]{(objects,}}
\put(06,14){\makebox(0,0)[bl]{items)}}

\put(22,30){\vector(1,1){05}} \put(22,26){\vector(2,1){05}}
\put(22,22){\vector(1,0){05}} \put(22,18){\vector(2,-1){05}}
\put(22,14){\vector(1,-1){05}}


\put(28,45){\makebox(0,0)[bl]{Ranking (linear}}
\put(28,42){\makebox(0,0)[bl]{ordered clusters)}}


\put(42,37){\makebox(0,0)[bl]{Layer \(1\)}}

\put(35,38){\oval(08,4)} \put(35,38){\oval(7.5,3.5)}
\put(35,38){\oval(7,3)}


\put(35,36){\vector(0,-1){4}}

\put(42,29){\makebox(0,0)[bl]{Layer \(2\)}}

\put(35,30){\oval(11,4)}


\put(35,28){\vector(0,-1){4}}

\put(32,22){\makebox(0,0)[bl]{{. . .}}}


\put(35,21){\vector(0,-1){4}}

\put(42,14){\makebox(0,0)[bl]{Layer \((k-1)\)}}

\put(35,15){\oval(11,4)}


\put(35,13){\vector(0,-1){4}}

\put(42,06){\makebox(0,0)[bl]{Layer \(k\)}}

\put(35,07){\oval(06,4)} \put(35,7){\oval(05.5,3.5)}

\end{picture}
%
\begin{picture}(70,55)

\put(01,00){\makebox(0,0)[bl]{Fig. 2.4. Example of
 hierarchical clustering}}

\put(00,06){\makebox(0,0)[bl]{Step \(0\)}}
\put(17,07){\makebox(0,0)[bl]{\(1\)}}

\put(15,10){\line(1,0){5}} \put(15,06){\line(1,0){5}}
\put(15,06){\line(0,1){4}} \put(20,06){\line(0,1){4}}

\put(27,07){\makebox(0,0)[bl]{\(2\)}}

\put(25,10){\line(1,0){5}} \put(25,06){\line(1,0){5}}
\put(25,06){\line(0,1){4}} \put(30,06){\line(0,1){4}}
\put(37,07){\makebox(0,0)[bl]{\(3\)}}

\put(35,10){\line(1,0){5}} \put(35,6){\line(1,0){5}}
\put(35,06){\line(0,1){4}} \put(40,6){\line(0,1){4}}
\put(47,7){\makebox(0,0)[bl]{\(4\)}}

\put(45,10){\line(1,0){5}} \put(45,6){\line(1,0){5}}
\put(45,6){\line(0,1){4}} \put(50,6){\line(0,1){4}}

\put(57,7){\makebox(0,0)[bl]{\(5\)}}

\put(55,10){\line(1,0){5}} \put(55,6){\line(1,0){5}}
\put(55,6){\line(0,1){4}} \put(60,6){\line(0,1){4}}

\put(67,7){\makebox(0,0)[bl]{\(6\)}}

\put(65,10){\line(1,0){5}} \put(65,6){\line(1,0){5}}
\put(65,6){\line(0,1){4}} \put(70,6){\line(0,1){4}}

\put(00,16){\makebox(0,0)[bl]{Step \(1\)}}
\put(17,17){\makebox(0,0)[bl]{\(1\)}}

\put(15,20){\line(1,0){5}} \put(15,16){\line(1,0){5}}
\put(15,16){\line(0,1){4}} \put(20,16){\line(0,1){4}}
\put(17.5,10){\vector(0,1){6}}
\put(27,17){\makebox(0,0)[bl]{\(2\)}}

\put(25,20){\line(1,0){5}} \put(25,16){\line(1,0){5}}
\put(25,16){\line(0,1){4}} \put(30,16){\line(0,1){4}}
\put(27.5,10){\vector(0,1){6}}
\put(37,16.5){\makebox(0,0)[bl]{\(3\)}}

\put(35,20){\line(1,0){5}} \put(35,16){\line(1,0){5}}
\put(35,16){\line(0,1){4}} \put(40,16){\line(0,1){4}}
\put(37.5,10){\vector(0,1){6}}
\put(47,17){\makebox(0,0)[bl]{\(4\)}}

\put(45,20){\line(1,0){5}} \put(45,16){\line(1,0){5}}
\put(45,16){\line(0,1){4}} \put(50,16){\line(0,1){4}}
\put(47.5,10){\vector(0,1){6}}
\put(57,16.5){\makebox(0,0)[bl]{\(5,6\)}}

\put(55,20){\line(1,0){9}} \put(55,16){\line(1,0){9}}
\put(55,16){\line(0,1){4}} \put(64,16){\line(0,1){4}}

\put(57.5,10){\vector(0,1){6}} \put(67.5,10){\vector(-1,1){6}}

\put(00,26){\makebox(0,0)[bl]{Step \(2\)}}
\put(17,27){\makebox(0,0)[bl]{\(1\)}}

\put(15,30){\line(1,0){5}} \put(15,26){\line(1,0){5}}
\put(15,26){\line(0,1){4}} \put(20,26){\line(0,1){4}}
\put(17.5,20){\vector(0,1){6}}
\put(27,26.5){\makebox(0,0)[bl]{\(2,4\)}}

\put(25,30){\line(1,0){9}} \put(25,26){\line(1,0){9}}
\put(25,26){\line(0,1){4}} \put(34,26){\line(0,1){4}}
\put(27.5,20){\vector(0,1){6}}

\put(47.5,20){\vector(-3,1){18}}
\put(37,26.5){\makebox(0,0)[bl]{\(3\)}}

\put(35,30){\line(1,0){5}} \put(35,26){\line(1,0){5}}
\put(35,26){\line(0,1){4}} \put(40,26){\line(0,1){4}}
\put(37.5,20){\vector(0,1){6}}

\put(57,26.5){\makebox(0,0)[bl]{\(5,6\)}}

\put(55,30){\line(1,0){9}} \put(55,26){\line(1,0){9}}
\put(55,26){\line(0,1){4}} \put(64,26){\line(0,1){4}}
\put(57.5,20){\vector(0,1){6}}

\put(00,36){\makebox(0,0)[bl]{Step \(3\)}}
\put(14,36.5){\makebox(0,0)[bl]{\(1,5,6\)}}

\put(13,40){\line(1,0){11}} \put(13,36){\line(1,0){11}}
\put(13,36){\line(0,1){4}} \put(24,36){\line(0,1){4}}
\put(17.5,30){\vector(0,1){6}}

 \put(19.5,32){\vector(0,1){4}}

 \put(54.5,32){\line(-1,0){35}}

 \put(58.5,30){\line(-2,1){4}}

\put(27,36.5){\makebox(0,0)[bl]{\(2,4\)}}

\put(25,40){\line(1,0){9}} \put(25,36){\line(1,0){9}}
\put(25,36){\line(0,1){4}} \put(34,36){\line(0,1){4}}
\put(27.5,30){\vector(0,1){6}}
\put(37,36.5){\makebox(0,0)[bl]{\(3\)}}

\put(35,40){\line(1,0){5}} \put(35,36){\line(1,0){5}}
\put(35,36){\line(0,1){4}} \put(40,36){\line(0,1){4}}
\put(37.5,30){\vector(0,1){6}}

\put(26.6,50){\oval(26,5)}

\put(17.5,48.2){\makebox(0,0)[bl]{\(1,2,3,4,5,6\)}}

\put(21.5,45){\makebox(0,0)[bl]{{\bf . ~. ~.}}}

\put(19.5,40){\vector(0,1){4}} \put(29.5,40){\vector(0,1){4}}
\put(37.5,40){\vector(0,1){4}}

\end{picture}
\end{center}

\begin{center}
\begin{picture}(115,44)

\put(24,00){\makebox(0,0)[bl]{Fig. 2.5. Illustration for multiple
 clustering}}

\put(58,08){\oval(100,07)} \put(58,08){\oval(99,06)}

\put(29,06){\makebox(0,0)[bl]{Initial set of elements
 (objects, items)}}

\put(28,11.5){\vector(-1,1){08.5}}
\put(58,11.5){\vector(0,1){08.5}}
\put(88,11.5){\vector(1,1){08.5}}

\put(10.5,16){\makebox(0,0)[bl]{Model}}
\put(10.5,12){\makebox(0,0)[bl]{(goal) \(1\)}}

\put(45,16){\makebox(0,0)[bl]{Model}}
\put(45,12){\makebox(0,0)[bl]{(goal) \(k\)}}

\put(95.5,16){\makebox(0,0)[bl]{Model}}
\put(95.5,12){\makebox(0,0)[bl]{(goal) \(n\)}}


\put(02,39){\makebox(0,0)[bl]{Clustering solution \(1\)}}

\put(18,29){\oval(34,18)}

\put(11,33){\oval(08,6)} \put(11,33){\oval(07.7,5.7)}

\put(24,34){\oval(08,5)} \put(25,25){\oval(10,7)}
\put(10,25){\oval(10,8)}


\put(36.2,28.5){\makebox(0,0)[bl]{{\bf ...}}}


\put(42,39){\makebox(0,0)[bl]{Clustering solution \(k\)}}

\put(58,29){\oval(34,18)}

\put(58,33){\oval(20,5)} \put(66,25){\oval(10,7)}
\put(49,25){\oval(10,8)}


\put(76.2,28.5){\makebox(0,0)[bl]{{\bf ...}}}


\put(82,39){\makebox(0,0)[bl]{Clustering solution \(n\)}}

\put(98,29){\oval(34,18)}

\put(88,33){\oval(07,6)} \put(88,33){\oval(06.7,5.7)}

\put(98,34){\oval(08,5)}

\put(108,32.5){\oval(08,5)}

\put(107,25){\oval(09,7)}

\put(88,25){\oval(10,8)}

\put(97.5,25.5){\oval(04,8.5)}

\end{picture}
\end{center}

\begin{center}
\begin{picture}(115,85)

\put(13,00){\makebox(0,0)[bl]{Fig. 2.6. Illustration for consensus
 clustering, summarizing}}

\put(58,09){\oval(100,08)} \put(58,09){\oval(99,07)}

\put(30,07){\makebox(0,0)[bl]{Initial set of elements
 (objects, items)}}

\put(28,15){\vector(-1,1){09}} \put(58,15){\vector(0,1){09}}
\put(88,15){\vector(1,1){09}}

\put(08,18.5){\makebox(0,0)[bl]{Model/}}
\put(08,15.5){\makebox(0,0)[bl]{method \(1\)}}

\put(42.5,18.5){\makebox(0,0)[bl]{Model/}}
\put(42.5,15.5){\makebox(0,0)[bl]{method \(k\)}}

\put(96.2,18.5){\makebox(0,0)[bl]{Model/}}
\put(96.2,15.5){\makebox(0,0)[bl]{method \(n\)}}


\put(09,44){\makebox(0,0)[bl]{Clustering \(1\)}}

\put(18,34){\oval(34,18)}

\put(11,38){\oval(08,6)} \put(11,38){\oval(07.7,5.7)}

\put(24,39){\oval(08,5)}

\put(25,30){\oval(10,7)} \put(10,30){\oval(10,8)}


\put(36.2,33.5){\makebox(0,0)[bl]{{\bf ...}}}


\put(41,44){\makebox(0,0)[bl]{Clustering  ~\(k\)}}

\put(58,34){\oval(34,18)}

\put(58,38){\oval(20,5)} \put(66,30){\oval(10,7)}
\put(49,30){\oval(10,8)}


\put(76.2,33.5){\makebox(0,0)[bl]{{\bf ...}}}


\put(88.5,44){\makebox(0,0)[bl]{Clustering \(n\)}}

\put(98,34){\oval(34,18)}

\put(88,38){\oval(07,6)} \put(88,38){\oval(06.7,5.7)}

\put(98,39){\oval(08,5)}

\put(108,37.5){\oval(08,5)}

\put(107,30){\oval(09,7)}

\put(88,30){\oval(10,8)}

\put(97.5,30.5){\oval(04,8.5)}


\put(28,44){\vector(1,1){05}} \put(58,44){\vector(0,1){05}}
\put(88,44){\vector(-1,1){05}}

\put(20,49.5){\line(1,0){76}} \put(20,57){\line(1,0){76}}
\put(20,49.5){\line(0,1){07.5}} \put(96,49.5){\line(0,1){07.5}}
\put(21,49.5){\line(0,1){07.5}} \put(95,49.5){\line(0,1){07.5}}

\put(27.5,53.5){\makebox(0,0)[bl]{Aggregation of clusterings,
 summarizing}}

\put(41,50){\makebox(0,0)[bl]{(clustering ensembles)}}

\put(58,57.5){\vector(0,1){04}}


\put(28,81){\makebox(0,0)[bl]{Consensus (e.g., median)
 of clusterings}}

\put(58,71){\oval(34,18)}

\put(48,75){\oval(07,6)} \put(48,75){\oval(06.7,5.7)}

\put(68,75){\oval(07,5)}

\put(67,67){\oval(09,7)}

\put(48,67){\oval(08.5,8)}

\put(57.5,71){\oval(05,15)}

\end{picture}
\end{center}
%


 This numerical example illustrates
 the pointed out 5 types of clustering problems
 on the basis of the same initial data (Table 2.2,
 for 12 students):
 (i) clustering  as partitioning (Fig. 2.7a),
 (ii) classification problem (``soft'' problem formulation) (Fig. 2.7b),
 (iii) sorting problem (stratification, multicriteria ranking)
  (Fig. 2.7c),
 (iv) hierarchical clustering (Fig. 2.8), and
 (v) consensus clustering (Fig. 2.9).
 Note, consensus clustering is based on three initial clustering
 solutions:

 (a) five clusters (from partitioning problem, Fig. 2.7a):
 \( \{ 6,9,10 \}\), \( \{ 1,8,3 \}\),
 \( \{ 2,5,7,11 \}\), \( \{ 12,14 \}\), \( \{ 3,4 \}\);

 (b) four clusters (from sorting problem, Fig. 2.7c):
  \( \{ 6,9 \}\), \( \{ 1,3,10 \}\),
 \( \{ 4,8,13,14 \}\), \( \{ 2,5,7,11,12 \}\);

 (c) four clusters (from hierarchical clustering, Fig. 2.8):
  \( \{ 6,9,10 \}\), \( \{ 1,3 \}\),  \( \{ 4,8,13,14 \}\),
 \( \{ 2,5,7,11,12 \}\).

 In recent decade, the significance of
 the fifth type of clustering problem has been increased:
 network clustering (e.g., detection of community structures) (Fig. 2.10).
 Here initial information consists in
 description of network:
 i.e., set of vertices/nodes and
 set of edges.
 Weights of vertices and/or edges can be used as well.

\begin{center}
{\bf Table 2.2.} Items,  estimates upon criteria   \\
\begin{tabular}{| c | c | c | c | c | c |}
\hline
 Item (student)  & \(C_{1}\) & \(C_{2}\) & \(C_{3}\) & \(C_{4}\) & \(C_{5}\)   \\
\hline
 Student 1   & 4 & 4 & 5 & 5 & 4 \\
 Student 2   & 3 & 3 & 3 & 4 & 3 \\
 Student 3   & 4 & 4 & 4 & 5 & 4 \\
 Student 4   & 5 & 4 & 4 & 3 & 5 \\
 Student 5   & 3 & 3 & 3 & 4 & 3 \\
 Student 6   & 5 & 5 & 5 & 5 & 5 \\
 Student 7   & 3 & 3 & 3 & 4 & 3 \\
 Student 8   & 4 & 3 & 4 & 4 & 3 \\
 Student 9   & 5 & 5 & 5 & 5 & 5 \\
 Student 10  & 5 & 5 & 5 & 4 & 5 \\
 Student 11  & 3 & 3 & 3 & 5 & 3 \\
 Student 12  & 3 & 5 & 3 & 3 & 3 \\
 Student 13  & 5 & 3 & 4 & 3 & 3 \\
 Student 14  & 3 & 5 & 3 & 5 & 4 \\

\hline
\end{tabular}
\end{center}

\begin{center}
\begin{picture}(47,50)

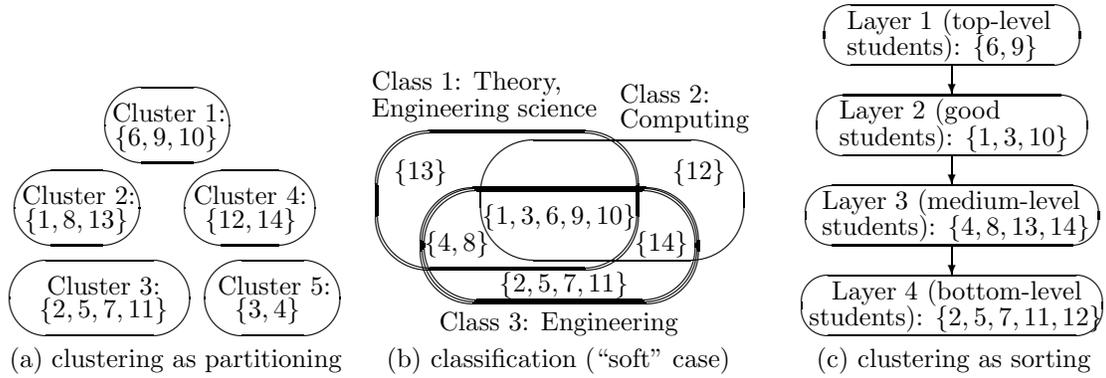
\put(019,00){\makebox(0,0)[bl]{Fig. 2.7. Numerical examples of
 clustering/classification problems}}

\put(00,05){\makebox(0,0)[bl]{(a) clustering as partitioning}}


\put(13.7,38.5){\makebox(0,0)[bl]{Cluster \(1\):}}
\put(14,34.5){\makebox(0,0)[bl]{\(\{ 6,9,10 \}\)}}

\put(21,38){\oval(16.6,10)}


\put(01.7,27.5){\makebox(0,0)[bl]{Cluster \(2\):}}
\put(02,23.5){\makebox(0,0)[bl]{\(\{ 1,8,13 \}\)}}

\put(09,27){\oval(16.6,10)}


\put(05,15.5){\makebox(0,0)[bl]{Cluster \(3\):}}
\put(04,11.5){\makebox(0,0)[bl]{\(\{ 2,5,7,11 \}\)}}

\put(12,15){\oval(24,10)}


\put(24.5,27.5){\makebox(0,0)[bl]{Cluster \(4\):}}
\put(26,23.5){\makebox(0,0)[bl]{\( \{ 12,14 \} \)}}

\put(32,27){\oval(17.4,10)}


\put(27.7,15.5){\makebox(0,0)[bl]{Cluster \(5\):}}
\put(30,11.5){\makebox(0,0)[bl]{\( \{ 3,4 \} \)}}

\put(35,15){\oval(18,10)}

\end{picture}
%
\begin{picture}(56,46)


\put(02,05){\makebox(0,0)[bl]{(b) classification
 (``soft'' case)}}


\put(09,10){\makebox(0,0)[bl]{Class \(3\): Engineering}}

\put(25,22){\oval(37,15.5)} \put(25,22){\oval(36.5,15)}
\put(25,22){\oval(36,14.5)}

\put(17,15){\makebox(0,0)[bl]{\(\{ 2,5,7,11 \}\)}}


\put(00,42){\makebox(0,0)[bl]{Class \(1\): Theory,}}
\put(00,38.5){\makebox(0,0)[bl]{Engineering science}}

\put(18,28){\oval(35,18.5)} \put(18,28){\oval(34.5,18)}

\put(03,30){\makebox(0,0)[bl]{\(\{13 \}\)}}

\put(14.7,24){\makebox(0,0)[bl]{\(\{ 1,3,6,9,10 \}\)}}

\put(07,20.4){\makebox(0,0)[bl]{\(\{ 4,8 \}\)}}


\put(33,41){\makebox(0,0)[bl]{Class \(2\):}}
\put(33,37){\makebox(0,0)[bl]{Computing}}

\put(32,28){\oval(35,16)}

\put(35,20.4){\makebox(0,0)[bl]{\(\{ 14 \}\)}}

\put(40,30){\makebox(0,0)[bl]{\(\{ 12 \}\)}}

\end{picture}
%
\begin{picture}(40,54)


\put(02,05){\makebox(0,0)[bl]{(c) clustering as sorting}}


\put(06,50){\makebox(0,0)[bl]{Layer \(1\) (top-level}}

\put(20,50){\oval(34,08)}

\put(06,46.5){\makebox(0,0)[bl]{students): \(\{6,9\}\)}}

\put(20,46){\vector(0,-1){4}}


\put(05,38){\makebox(0,0)[bl]{Layer \(2\) (good }}

\put(20,38){\oval(36,08)}

\put(05,34.5){\makebox(0,0)[bl]{students): \(\{1,3,10\}\)}}

\put(20,34){\vector(0,-1){4}}


\put(02.7,26){\makebox(0,0)[bl]{Layer \(3\) (medium-level}}

\put(20,26){\oval(39,08)}

\put(02.7,22.5){\makebox(0,0)[bl]{students): \(\{4,8,13,14\}\)}}

\put(20,22){\vector(0,-1){4}}


\put(04,14){\makebox(0,0)[bl]{Layer \(4\) (bottom-level }}

\put(20,14){\oval(40,08)}
\put(01,10.5){\makebox(0,0)[bl]{students): \(\{2,5,7,11,12\}\)}}

\end{picture}
\end{center}

\begin{center}
\begin{picture}(136,61.5)

\put(27,00){\makebox(0,0)[bl]{Fig. 2.8. Numerical example of
 hierarchical clustering}}


\put(03,08){\oval(06,06)} \put(02,7){\makebox(0,0)[bl]{\(1\)}}
\put(03,11){\vector(2,3){09.4}}

\put(13,08){\oval(06,06)} \put(12,7){\makebox(0,0)[bl]{\(2\)}}
\put(13,11){\line(1,1){7}} \put(20,18){\vector(1,0){17}}

\put(23,08){\oval(06,06)} \put(22,7){\makebox(0,0)[bl]{\(3\)}}
\put(23,11){\vector(-2,3){09.4}}

\put(33,08){\oval(06,06)} \put(32,7){\makebox(0,0)[bl]{\(4\)}}
\put(33,11){\line(0,1){11}} \put(33,22){\line(1,0){29}}
\put(62,22){\vector(1,1){4}}

\put(43,08){\oval(06,06)} \put(42,7){\makebox(0,0)[bl]{\(5\)}}
\put(43,11){\vector(0,1){4}}

\put(53,08){\oval(06,06)} \put(52,7){\makebox(0,0)[bl]{\(6\)}}
\put(53,11){\line(1,1){7}} \put(60,18){\vector(1,0){18}}

\put(63,08){\oval(06,06)} \put(62,7){\makebox(0,0)[bl]{\(7\)}}
\put(62,11){\vector(-2,1){13}}

\put(73,08){\oval(06,06)} \put(72,7){\makebox(0,0)[bl]{\(8\)}}
\put(73,11){\vector(0,1){14}}

\put(83,08){\oval(06,06)} \put(82,7){\makebox(0,0)[bl]{\(9\)}}
\put(83,11){\vector(0,1){4}}

\put(93,08){\oval(06,06)} \put(91,7){\makebox(0,0)[bl]{\(10\)}}
\put(94,11){\vector(-1,4){4}}

\put(103,08){\oval(06,06)} \put(101,7){\makebox(0,0)[bl]{\(11\)}}
\put(103,11){\line(-2,1){4}} \put(99,13){\line(-1,0){35.5}}
\put(63.5,13){\vector(-1,1){12.7}}

\put(113,08){\oval(06,06)} \put(111,7){\makebox(0,0)[bl]{\(12\)}}
\put(113,11){\line(-1,1){21}} \put(92,32){\line(-1,0){34}}
\put(58,32){\vector(-1,1){4}}

\put(123,08){\oval(06,06)} \put(121,7){\makebox(0,0)[bl]{\(13\)}}
\put(123,11){\vector(0,1){24}}

\put(133,08){\oval(06,06)} \put(131,7){\makebox(0,0)[bl]{\(14\)}}
\put(133,11){\vector(-1,4){6}}

\put(43,18){\oval(12,06)}

\put(37,16){\makebox(0,0)[bl]{\( \{2,5,7\}\)}}
\put(43,21){\vector(0,1){4}}

\put(83,18){\oval(10,06)}
\put(78.5,16){\makebox(0,0)[bl]{\(\{6,9\}\)}}
\put(83,21){\vector(0,1){4}}

\put(13,28){\oval(09.5,06)}
\put(08.5,26){\makebox(0,0)[bl]{\(\{1,3\}\)}}
\put(13,31){\line(1,1){11}} \put(24,42){\line(1,0){41}}
\put(65,42){\vector(1,1){4}}

\put(43,28){\oval(18,06)}
\put(34.5,26){\makebox(0,0)[bl]{\(\{2,5,7,11\}\)}}
\put(43,31){\vector(0,1){4}}

\put(70,28){\oval(09,06)}
\put(65.6,26){\makebox(0,0)[bl]{\(\{4,8\}\)}}
\put(72,31){\vector(3,2){6}}

\put(73,31){\line(2,1){4}} \put(77,33){\line(1,0){32}}
\put(109,33){\vector(1,1){4}}

\put(83,28){\oval(14,06)}
\put(76.3,26){\makebox(0,0)[bl]{\(\{6,9,10\}\)}}
\put(83,31){\vector(0,1){4}}

\put(43,38){\oval(23,06)}
\put(32,36){\makebox(0,0)[bl]{\(\{2,5,7,11,12\}\)}}
\put(43,41){\vector(1,1){14}}


\put(83,38){\oval(22,06)}
\put(73,36){\makebox(0,0)[bl]{\(\{4,6,8,9,10\}\)}}
\put(83,41){\vector(0,1){4}}

\put(123,38){\oval(20,06)}
\put(113.5,36){\makebox(0,0)[bl]{\(\{4,8,13,14\}\)}}

\put(123,41){\vector(-1,1){14}}


\put(83,48){\oval(29,06)}
\put(69.4,46){\makebox(0,0)[bl]{\(\{1,3,4,6,8,9,10\}\)}}
\put(83,51){\vector(0,1){4}}

\put(83,58){\oval(78,06)} \put(83,58){\oval(77,05)}
\put(83,58){\oval(75,04)}


\put(48.5,56){\makebox(0,0)[bl]{Root
 (\{ \(1,2,3,4,5,6,7,8,9,10,11,12,13,14\)\})}}

\end{picture}
\end{center}

\begin{center}
\begin{picture}(154,10)
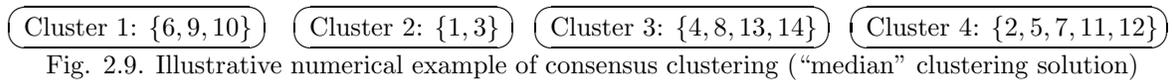
\put(05,00){\makebox(0,0)[bl]{Fig. 2.9. Illustrative numerical
 example of consensus clustering (``median'' clustering solution)}}


\put(02,05){\makebox(0,0)[bl]{Cluster \(1\): \(\{ 6,9,10 \}\)}}

\put(17,07){\oval(34,05.4)}


\put(40,05){\makebox(0,0)[bl]{Cluster \(2\): \( \{ 1,3 \} \)}}

\put(52.5,07){\oval(29,05.4)}


\put(72,05){\makebox(0,0)[bl]{Cluster \(3\): \(\{ 4,8,13,14 \}\)}}

\put(89.5,07){\oval(39,05.4)}


\put(114,05){\makebox(0,0)[bl]{Cluster \(4\): \( \{ 2,5,7,11,12 \}
\)}}

\put(133,07){\oval(42,05.4)}

\end{picture}
\end{center}

\begin{center}
\begin{picture}(73,28.5)
\put(03.5,00){\makebox(0,0)[bl]{Fig. 2.10. Illustration for
 network clustering}}



\put(07,12.5){\oval(14,15)}


\put(03,18){\circle*{1.2}} \put(03,18){\line(1,-1){4}}

\put(11,18){\circle*{1.2}} \put(11,18){\line(-1,-1){4}}

\put(03,18){\line(1,0){8}}

\put(07,14){\circle*{1.2}} \put(03,10){\circle*{1.2}}
\put(11,10){\circle*{1.2}} \put(07,06){\circle*{1.2}}
\put(07,06){\circle*{1.2}} \put(07,10){\circle*{1.2}}

\put(07,06){\line(-1,1){4}} \put(07,06){\line(1,1){4}}
\put(07,06){\line(0,1){8}} \put(03,10){\line(1,0){8}}
\put(07,14){\line(-1,-1){4}} \put(07,14){\line(1,-1){4}}

\put(03,10){\line(0,1){8}}


\put(11,18.3){\line(2,-1){12.4}}




\put(26.5,14.5){\oval(11,07)}

\put(24,12){\circle*{0.7}} \put(29,12){\circle*{0.7}}
\put(24,17){\circle*{0.7}} \put(29,17){\circle*{0.7}}

\put(24,12){\circle{1.2}} \put(29,12){\circle{1.2}}
\put(24,17){\circle{1.2}} \put(29,17){\circle{1.2}}

\put(24,12){\line(1,0){5}} \put(24,12){\line(0,1){5}}
\put(24,17){\line(1,0){5}} \put(29,12){\line(0,1){5}}
\put(24,12){\line(1,1){5}} \put(29,12){\line(-1,1){5}}


\put(29,17){\line(1,1){6}}

\put(24,12){\line(1,0){19}}






\put(36.5,27){\circle*{1.1}}

\put(36.5,27){\line(-1,-2){2}} \put(36.5,27){\line(1,-2){2}}

\put(36.5,23){\oval(6,9.5)}


\put(36.5,19){\circle*{1.1}}

\put(34.5,23){\circle*{1.1}} \put(38.5,23){\circle*{1.1}}

\put(34.5,23){\line(1,0){4}}

\put(36.5,19.5){\line(-1,2){2}} \put(36.5,19.5){\line(1,2){2}}

\put(38.5,23){\line(1,-1){6}}


\put(48.5,14.5){\oval(15,07)}


\put(44,12){\circle*{0.9}} \put(49,12){\circle*{0.9}}
\put(44,17){\circle*{0.9}} \put(49,17){\circle*{0.9}}
\put(46.5,14.5){\circle*{0.9}} \put(54,14.5){\circle*{0.9}}

\put(44,12){\circle{1.5}} \put(49,12){\circle{1.5}}
\put(44,17){\circle{1.5}} \put(49,17){\circle{1.5}}
\put(46.5,14.5){\circle{1.5}} \put(54,14.5){\circle{1.5}}

\put(54,14.5){\line(-2,1){4.3}} \put(54,14.5){\line(-2,-1){4.3}}

\put(44,12){\line(1,0){5}} \put(44,12){\line(0,1){5}}
\put(44,17){\line(1,0){5}} \put(49,12){\line(0,1){5}}
\put(44,12){\line(1,1){5}} \put(44,12){\line(1,1){5}}

\put(51,08){\line(-1,2){2}}

\put(53.5,08){\oval(9,05)}

\put(56,08){\circle*{1.1}}  \put(51,08){\circle*{1.1}}

\put(51,08){\line(1,0){5}}

\put(56,08){\line(1,0){8}} \put(56,08){\line(3,2){8}}
\put(56,08){\line(-1,3){2}}




\put(67.5,12.5){\oval(11,14)}

\put(64,08){\circle*{0.7}} \put(69,08){\circle*{0.7}}
\put(64,13){\circle*{0.7}} \put(69,13){\circle*{0.7}}

\put(64,08){\circle{1.2}} \put(69,08){\circle{1.2}}
\put(64,13){\circle{1.2}} \put(69,13){\circle{1.2}}

\put(64,08){\line(1,0){5}} \put(64,08){\line(0,1){5}}
\put(64,13){\line(1,0){5}} \put(69,08){\line(0,1){5}}

\put(64,08){\line(1,1){5}} \put(69,08){\line(-1,1){5}}

\put(64,13){\line(1,2){2}}

\put(66,17){\circle*{0.7}} \put(66,17){\circle{1.2}}
\put(66,17){\line(1,0){5}}

\put(69,13){\line(1,2){2}}

\put(71,17){\circle*{0.7}} \put(71,17){\circle{1.2}}


\end{picture}
\end{center}

  A generalized scheme of clustering processes consists of the
  following \cite{and73,dy04,har75,jain88,jain99,kumar14,mirkin05,rocha13,xu05,xu09}
 (Fig. 2.11):

~~

 {\it Phase 1.} Collection of initial data on the applied
 situation.

 {\it Phase 2.}  Analysis of applied situation(s)
  and formulation/structuring of clustering problem:

 (2.1.) generation/detection and description of
 initial objects/elements;

 (2.2.) generation/detection of element parameters
 (i.e., feature selection or extraction);

 (2.3.) selection/design of proximity measures
 (for elements, for clusters)
 and types for inter-cluster criterion and
 for intra-cluster criterion; and

 (2.4.) selection of basic clustering model(s)
 (i.e., hierarchical clustering, partitioning clustering).

 {\it Phase 3.} Selection/design of clustering solving scheme
 (solving method/procedure).

 {\it Phase 4.} Implementation of clustering procedure(s).

 {\it Phase 5.} Analysis of clustering solution(s) (i.e., clusters validation).


\begin{center}
\begin{picture}(124,55)

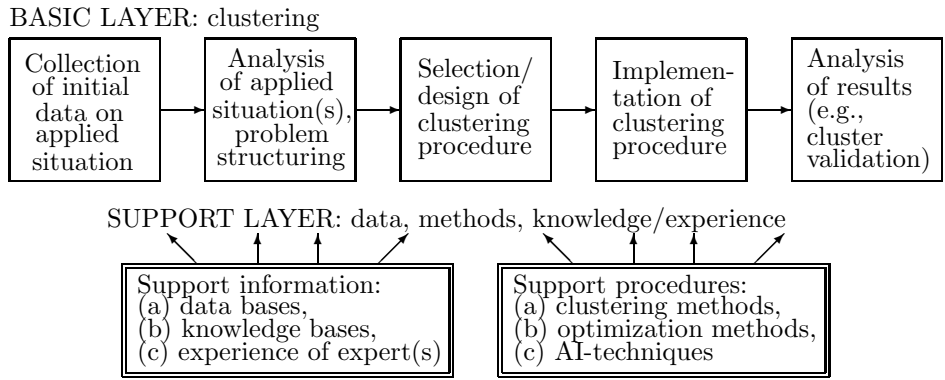
\put(23.5,00){\makebox(0,0)[bl] {Fig. 2.11. System
 two-layer scheme for clustering}}

\put(00,51){\makebox(0,8)[bl]{BASIC LAYER: clustering}}


\put(00,31){\line(1,0){20}} \put(00,50){\line(1,0){20}}
\put(00,31){\line(0,1){19}} \put(20,31){\line(0,1){19}}

\put(02,45.5){\makebox(0,8)[bl]{Collection}}
\put(03,42.5){\makebox(0,8)[bl]{of initial}}
\put(03.5,39){\makebox(0,8)[bl]{data on }}
\put(03.5,35.5){\makebox(0,8)[bl]{applied }}
\put(03,32.5){\makebox(0,8)[bl]{situation}}

\put(20,40.5){\vector(1,0){6}}


\put(26,31){\line(1,0){20}} \put(26,50){\line(1,0){20}}
\put(26,31){\line(0,1){19}} \put(46,31){\line(0,1){19}}

\put(29,45.5){\makebox(0,8)[bl]{Analysis }}
\put(28,42.5){\makebox(0,8)[bl]{of applied}}
\put(27,38.5){\makebox(0,8)[bl]{situation(s), }}
\put(30,35.5){\makebox(0,8)[bl]{problem }}
\put(27.6,32.5){\makebox(0,8)[bl]{structuring}}

\put(46,40.5){\vector(1,0){6}}


\put(52,31){\line(1,0){20}} \put(52,50){\line(1,0){20}}
\put(52,31){\line(0,1){19}} \put(72,31){\line(0,1){19}}

\put(54.3,44){\makebox(0,8)[bl]{Selection/ }}
\put(54.3,40.6){\makebox(0,8)[bl]{design of}}
\put(54.3,37){\makebox(0,8)[bl]{clustering }}
\put(54.3,34){\makebox(0,8)[bl]{procedure }}
\put(54,33){\makebox(0,8)[bl]{}}

\put(72,40.5){\vector(1,0){6}}


\put(78,31){\line(1,0){20}} \put(78,50){\line(1,0){20}}
\put(78,31){\line(0,1){19}} \put(98,31){\line(0,1){19}}

\put(80.2,44){\makebox(0,8)[bl]{Implemen-}}
\put(80.2,41){\makebox(0,8)[bl]{tation of }}
\put(80.2,37){\makebox(0,8)[bl]{clustering  }}
\put(80.2,34){\makebox(0,8)[bl]{procedure}}
\put(80.2,31){\makebox(0,8)[bl]{}}

\put(98,40.5){\vector(1,0){6}}


\put(104,31){\line(1,0){20}} \put(104,50){\line(1,0){20}}
\put(104,31){\line(0,1){19}} \put(124,31){\line(0,1){19}}

\put(106,45.5){\makebox(0,8)[bl]{Analysis }}
\put(106,42.5){\makebox(0,8)[bl]{of results}}
\put(106,38.5){\makebox(0,8)[bl]{(e.g.,  }}
\put(106,35.8){\makebox(0,8)[bl]{cluster }}
\put(106,32){\makebox(0,8)[bl]{validation)}}

\put(013,24){\makebox(0,8)[bl]{SUPPORT LAYER: data, methods,
 knowledge/experience}}

\put(015,05){\line(1,0){44}} \put(15,20){\line(1,0){44}}
\put(015,05){\line(0,1){15}} \put(59,05){\line(0,1){15}}

\put(15.5,05.5){\line(1,0){43}} \put(15.5,19.5){\line(1,0){43}}
\put(15.5,05.5){\line(0,1){14}} \put(58.5,05.5){\line(0,1){14}}

\put(17,15.4){\makebox(0,8)[bl]{Support information:}}
\put(17,12.4){\makebox(0,8)[bl]{(a) data bases,}}
\put(17,09.4){\makebox(0,8)[bl]{(b) knowledge bases,}}
\put(17,06.4){\makebox(0,8)[bl]{(c) experience of expert(s)}}


\put(25,20){\vector(-1,1){4}} \put(33,20){\vector(0,1){4}}
\put(41,20){\vector(0,1){4}} \put(49,20){\vector(1,1){4}}

\put(75,20){\vector(-1,1){4}} \put(83,20){\vector(0,1){4}}
\put(91,20){\vector(0,1){4}} \put(99,20){\vector(1,1){4}}


\put(65,05){\line(1,0){44}} \put(65,20){\line(1,0){44}}
\put(65,05){\line(0,1){15}} \put(109,05){\line(0,1){15}}

\put(65.5,05.5){\line(1,0){43}} \put(65.5,19.5){\line(1,0){43}}
\put(65.5,05.5){\line(0,1){14}} \put(108.5,05.5){\line(0,1){14}}

\put(67,15.4){\makebox(0,8)[bl]{Support procedures:}}
\put(67,12.4){\makebox(0,8)[bl]{(a) clustering methods,}}
\put(67,09.4){\makebox(0,8)[bl]{(b) optimization methods,}}
\put(67,06.4){\makebox(0,8)[bl]{(c) AI-techniques}}

\end{picture}
\end{center}

 Fig. 2.12 depicts a structural description
 for main machine learning paradigms
 (e.g., \cite{ghah04,kae96,kot07,mohri12,sut98,wang12,vap00}):
 (a) unsupervised learning,
 (b) reinforcement learning, and
 (c) supervised learning.
 Basic clustering problems/models corresponds to unsupervised learning.
 Evidently, in complex cases, it is possible to analyze clustering
 results to correct the clustering solving process
 (i.e., as in reinforcement learning or in supervised learning).

\begin{center}
\begin{picture}(52,30)
\put(42,00){\makebox(0,0)[bl] {Fig. 2.12. Three
 learning paradigms \cite{wang12}}}

\put(19,05){\makebox(0,0)[bl] {(a)}}

\put(00,23.5){\makebox(0,0)[bl] {Input}}
\put(00,20){\makebox(0,0)[bl] {(signal,}}
\put(00,16.5){\makebox(0,0)[bl] {data)}}

\put(00,15){\vector(1,0){09}}

\put(09,10){\line(1,0){24}} \put(09,20){\line(1,0){24}}
\put(09,10){\line(0,1){10}} \put(33,10){\line(0,1){10}}

\put(09.5,10.5){\line(1,0){23}} \put(09.5,19.5){\line(1,0){23}}
\put(09.5,10.5){\line(0,1){09}} \put(32.5,10.5){\line(0,1){09}}

\put(10.5,15.5){\makebox(0,0)[bl] {Unsupervised}}
\put(15,12){\makebox(0,0)[bl] {learning}}

\put(33,15){\vector(1,0){11}}

\put(33.5,22){\makebox(0,0)[bl] {Output}}
\put(33.5,19){\makebox(0,0)[bl] {(signal,}}
\put(33.5,16){\makebox(0,0)[bl] {data,}}
\put(33.5,11){\makebox(0,0)[bl] {structure,}}
\put(33.5,08){\makebox(0,0)[bl] {model)}}

\end{picture}
%
\begin{picture}(52,37)

\put(19,05){\makebox(0,0)[bl] {(b)}}

\put(00,23.5){\makebox(0,0)[bl] {Input}}
\put(00,20){\makebox(0,0)[bl] {(signal,}}
\put(00,16.5){\makebox(0,0)[bl] {data)}}


\put(00,15){\vector(1,0){09}}


\put(36.5,32){\oval(21,09)}

\put(27,32.7){\makebox(0,0)[bl] {Performance}}
\put(28.7,29){\makebox(0,0)[bl] {evaluation}}

\put(10,33){\makebox(0,0)[bl] {Reward}}

\put(21,32){\line(1,0){05}} \put(21,32){\vector(0,-1){12}}

\put(09,10){\line(1,0){24}} \put(09,20){\line(1,0){24}}
\put(09,10){\line(0,1){10}} \put(33,10){\line(0,1){10}}

\put(09.5,10.5){\line(1,0){23}} \put(09.5,19.5){\line(1,0){23}}
\put(09.5,10.5){\line(0,1){09}} \put(32.5,10.5){\line(0,1){09}}

\put(10,15.5){\makebox(0,0)[bl] {Reinforcement}}
\put(15,12){\makebox(0,0)[bl] {learning}}

\put(33,15){\vector(1,0){11}}

\put(33.5,22){\makebox(0,0)[bl] {Output}}
\put(33.5,19){\makebox(0,0)[bl] {(signal,}}
\put(33.5,16){\makebox(0,0)[bl] {data,}}
\put(33.5,11){\makebox(0,0)[bl] {structure,}}
\put(33.5,08){\makebox(0,0)[bl] {model)}}

\end{picture}
%
\begin{picture}(48,41.5)

\put(19,05){\makebox(0,0)[bl] {(c)}}

\put(00,23.5){\makebox(0,0)[bl] {Input}}
\put(00,20){\makebox(0,0)[bl] {(signal,}}
\put(00,16.5){\makebox(0,0)[bl] {data)}}


\put(00,15){\vector(1,0){09}}


\put(18,31){\makebox(0,0)[bl] {Error}}

\put(21,30){\line(1,0){11}} \put(21,30){\vector(0,-1){10}}

\put(34,38.5){\oval(25,05)}

\put(22.5,36.5){\makebox(0,0)[bl] {Teacher/target}}

\put(34,36){\vector(0,-1){04}}

\put(34,30){\oval(4,4)} \put(34,30){\oval(3,3)}
\put(34,30){\oval(2,2)}

\put(34,15){\vector(0,1){13}}


\put(09,10){\line(1,0){24}} \put(09,20){\line(1,0){24}}
\put(09,10){\line(0,1){10}} \put(33,10){\line(0,1){10}}

\put(09.5,10.5){\line(1,0){23}} \put(09.5,19.5){\line(1,0){23}}
\put(09.5,10.5){\line(0,1){09}} \put(32.5,10.5){\line(0,1){09}}

\put(12.6,15.5){\makebox(0,0)[bl] {Supervised}}
\put(15,12){\makebox(0,0)[bl] {learning}}

\put(33,15){\vector(1,0){11}}

\put(34.5,22){\makebox(0,0)[bl] {Output}}
\put(34.5,19){\makebox(0,0)[bl] {(signal,}}
\put(34.5,16){\makebox(0,0)[bl] {data,}}
\put(34.5,11){\makebox(0,0)[bl] {structure,}}
\put(34.5,08){\makebox(0,0)[bl] {model)}}
\end{picture}
\end{center}

\subsection{Objects, evaluation, problems}

 \subsubsection{Examined objects, parameters}

 It is reasonable to point out the basic examined objects:

 {\it 1.} item (i.e., element, point),

 {\it 2.} set of items (e.g., initial set of items),

 {\it 3.}  subset of items, cluster (e.g., a subset of initial items),

 {\it 4.} clustering solution as
 partitioning of the initial item set into set of clusters,
 and

 {\it 5.} clustering solution and an order over its clusters:

 {\it 5.1.} linear order over the clusters (sorting problem),

 {\it 5.2.} hierarchy over the clusters,

 {\it 5.3.} poset over the clusters.

 Significant auxiliary problems consist in measurement of
 proximity for pair of objects:
 {\it 1.} item and item,
 {\it 2.} item and cluster,
 {\it 3.} cluster and cluster,
 {\it 4.} clustering solution and clustering solution,
 {\it 5.} order over clusters and order over clusters,
 and
 {\it 6.}  clustering solution, order over its clusters
 and
  clustering solution, order over its clusters.

 In recent years, a special research attention is targeted to
 aggregation of clustering solutions and evaluation of
 the obtained aggregated clustering solution(s).


 Generally, basic problem types, parameters, and criteria
 are pointed out in Table 2.2
 (e.g.,
 \cite{agar08,acker08,arabie96,das05,dubes79,gonz85,jain88,jain99,klein02,lev07e,millig96,mirkin05,murtagh92,new04a,new06,new04,rand71,rask99,vanr77,wall68,xu05,xu09,xux07,zah07}).

\newpage

\begin{center}
{\bf Table 2.2.} Basic parameters in clustering/classification  problems   \\
\begin{tabular}{| c | l | l |}
\hline
 No.  & Parameters, requirements & Description(s) (e.g., type, evaluation scale) \\
\hline

 I. & Elements (items/objects):
     &\\
 1.1. & Type of element(s)
    & One-type/multi-type element(s), whole\\

  & &  element(s) or structured/composite element(s)\\

  1.2.&Description of element
    & Quantitative, nominal, ordinal estimate(s);\\

 && fuzzy, multiset estimates, vector estimates,\\

 && binary relation(s) over elements\\

 1.3. & Proximity for elements pair
    & Metric/proximity, ordinal estimate, fuzzy\\
 &&   estimate, multiset estimate,  vector estimate \\

  1.4. & Proximity between element and
    & Metric/proximity, ordinal estimate, fuzzy\\

  &element set (e.g., cluster)&estimate, multiset estimate, vector estimate \\

  1.5. & Proximity between element set and
    & Metric/proximity, ordinal estimate, fuzzy\\

  &element set (e.g., two clusters)&estimate, multiset estimate, vector estimate \\


 II.& Clusters/classes (clustering solution): & \\
 2.1.& Definition type for  clusters/classes
  & 1.Predefined clusters (classification problem) \\
 & &
 2.Clusters are defend under solving process\\
 && (clustering problem)\\

 2.2. & Constraints for clusters & Number of clusters, number of
 cluster elements \\

  2.3. & Order over clusters/classes
  & Independent clusters, linear order/chain, ranking \\

 &(binary relation(s))& (layered structure), hierarchy (e.g., tree), poset \\

  2.4. & Clustering solution validity: & Quality, correspondence to requirements, etc.\\

  2.4.1&Basic criteria (e.g., \cite{acker08,dubes79,rask99,zah07}): &   \\

  2.4.1.1&Compactness (minimization): & Uniqueness of objects in each cluster:  \\

   & (i) intra-cluster ``distance'', & e.g., closeness to cluster centroid in cluster,\\
   & (ii) object positioning. & maximum distance between objects in each\\
   &&cluster, closeness of object  to cluster centroid\\
  && ``good'' correspondence of object to cluster \\

 2.4.1.2&Isolation or separability (maximization): &
   Well-separated clusters:  e.g., maximum \\

  &  inter-cluster ``distance'' &
 distances between cluster centroids \\

  2.4.1.3&General (maximization):  & \\

   & (i) number of correctly positioned objects,& \\
   & (ii) number of ``good'' clusters (Fig. 2.13),&  \\
   & (iii) connectedness. & similar objects are neighboring\\

  2.4.2&Quality of  structure  over clusters&
   Similarity to predefined structure \\

 2.5.& Modularity (community structure based
   & Maximization  \\

 & network clustering, Fig. 2.14
  \cite{agar08,new06,port09}) &
  \\


 III.& Fuziness/softness of problem/model:& \\

 3.1.& Hard problem& Assignment of each item to the only one cluster  \\

 3.2.& Fuzzy/soft problem &
  Assignment of each item to many clusters\\


 IV.& Problem time mode:&  \\

 4.1.&Off-line (statical) mode & Collection of initial data and processing \\

 4.2.&On-line mode (dynamics) &On-line processing (e.g., stream data) \\

 V.& Complexity of clustering process:&  \\

 5.1.& Algorithmic complexity  &  Estimate \\

 5.2.& Volume of required data  &  Estimate of required data \\

 5.3.& Volume of required  expert's work  &  Estimate of  expert/expert knowledge  \\

\hline
\end{tabular}
\end{center}


\begin{center}
\begin{picture}(70,19)

\put(00,00){\makebox(0,0)[bl]{Fig. 2.13. Illustration for
 ``good'' clusters}}

\put(15,05){\makebox(0,0)[bl]{``Good'' clusters}}

\put(14.5,06){\vector(-1,0){4}}

\put(22,08){\vector(-3,2){4}}


\put(05,07){\oval(10,05.5)}

\put(01.5,07){\circle*{0.7}} \put(05,07){\circle*{0.7}}
\put(08.5,07){\circle*{0.7}}

\put(05,09){\circle*{0.7}}\put(05,05){\circle*{0.7}}

\put(03,08.5){\circle*{0.7}} \put(07,08.5){\circle*{0.7}}
\put(03,05.5){\circle*{0.7}} \put(07,05.5){\circle*{0.7}}


\put(18,14){\oval(12,05.5)}

\put(13.5,14){\circle*{0.7}} \put(17,14){\circle*{0.7}}
\put(19,14){\circle*{0.7}} \put(22.5,14){\circle*{0.7}}

\put(17,16){\circle*{0.7}}\put(17,12){\circle*{0.7}}
\put(19,16){\circle*{0.7}}\put(19,12){\circle*{0.7}}

\put(15,15.5){\circle*{0.7}} \put(21,15.5){\circle*{0.7}}
\put(15,12.5){\circle*{0.7}} \put(21,12.5){\circle*{0.7}}


\put(38,14){\oval(21,05.5)}

\put(29,14){\circle*{0.7}} \put(31,14){\circle*{0.7}}
\put(33,14){\circle*{0.7}} \put(35,14){\circle*{0.7}}
\put(37,14){\circle*{0.7}} \put(39,14){\circle*{0.7}}
\put(41,14){\circle*{0.7}} \put(43,14){\circle*{0.7}}
\put(45,14){\circle*{0.7}} \put(47,14){\circle*{0.7}}


\put(56,10){\oval(10,11)}

\put(55,06){\circle*{0.7}}

\put(58,14){\circle*{0.7}}

\put(52,09){\circle*{0.7}} \put(52,11){\circle*{0.7}}


\put(59,07){\circle*{0.7}} \put(60,09){\circle*{0.7}}

\end{picture}
%
\begin{picture}(73,27)
\put(0.6,00){\makebox(0,0)[bl]{Fig. 2.14. Illustration for
 community structures}}

 \put(00,23){\makebox(0,0)[bl]{Outlier}}

\put(03,18.7){\line(1,2){2}}


\put(07,12.5){\oval(14,15)}


\put(03,18){\circle*{1.2}} \put(03,18){\line(1,-1){4}}

\put(07,14){\circle*{1.2}} \put(03,10){\circle*{1.2}}
\put(11,10){\circle*{1.2}} \put(07,06){\circle*{1.2}}
\put(07,06){\circle*{1.2}} \put(07,10){\circle*{1.2}}

\put(07,06){\line(-1,1){4}} \put(07,06){\line(1,1){4}}
\put(07,06){\line(0,1){8}} \put(03,10){\line(1,0){8}}
\put(07,14){\line(-1,-1){4}} \put(07,14){\line(1,-1){4}}


\put(17.5,08){\circle*{0.6}} \put(17.5,08){\circle*{1.0}}
\put(17.5,08){\circle{2}}

\put(17.5,08){\line(-3,1){6}} \put(17.5,08){\line(3,2){6}}


\put(18.5,04.2){\makebox(0,0)[bl]{Bridge}}


\put(26.5,14.5){\oval(11,07)}

\put(24,12){\circle*{0.7}} \put(29,12){\circle*{0.7}}
\put(24,17){\circle*{0.7}} \put(29,17){\circle*{0.7}}

\put(24,12){\circle{1.2}} \put(29,12){\circle{1.2}}
\put(24,17){\circle{1.2}} \put(29,17){\circle{1.2}}

\put(24,12){\line(1,0){5}} \put(24,12){\line(0,1){5}}
\put(24,17){\line(1,0){5}} \put(29,12){\line(0,1){5}}
\put(24,12){\line(1,1){5}}


\put(36.5,14.5){\circle*{0.6}} \put(36.5,14.5){\circle*{1.0}}
\put(36.5,14.5){\circle{1.8}} \put(36.5,14.5){\circle{2.3}}

\put(36.5,14.5){\line(-3,1){7.5}} \put(36.5,14.5){\line(3,1){7.5}}
\put(36.5,14.5){\line(-3,-1){7.5}}
\put(36.5,14.5){\line(3,-1){7.5}}



\put(36.5,12.5){\line(0,-1){4}}

\put(33,6.5){\makebox(0,0)[bl]{Hub}}


\put(36.5,21.5){\oval(6,07)}

\put(36.5,14.5){\line(0,1){4.5}}

\put(36.5,19){\circle*{1.1}}

\put(34.5,23){\circle*{1.1}} \put(38.5,23){\circle*{1.1}}

\put(34.5,23){\line(1,0){4}}

\put(36.5,19.5){\line(-1,2){2}} \put(36.5,19.5){\line(1,2){2}}


\put(48.5,14.5){\oval(15,07)}

\put(44,12){\circle*{0.9}} \put(49,12){\circle*{0.9}}
\put(44,17){\circle*{0.9}} \put(49,17){\circle*{0.9}}
\put(46.5,14.5){\circle*{0.9}} \put(54,14.5){\circle*{0.9}}

\put(44,12){\circle{1.5}} \put(49,12){\circle{1.5}}
\put(44,17){\circle{1.5}} \put(49,17){\circle{1.5}}
\put(46.5,14.5){\circle{1.5}} \put(54,14.5){\circle{1.5}}

\put(54,14.5){\line(-2,1){4.3}} \put(54,14.5){\line(-2,-1){4.3}}

\put(44,12){\line(1,0){5}} \put(44,12){\line(0,1){5}}
\put(44,17){\line(1,0){5}} \put(49,12){\line(0,1){5}}
\put(44,12){\line(1,1){5}} \put(44,12){\line(1,1){5}}


\put(56,08){\circle*{0.6}} \put(56,08){\circle*{1.0}}
\put(56,08){\circle{2}}

\put(56,08){\line(1,0){8}} \put(56,08){\line(3,2){8}}
\put(56,08){\line(-1,3){2}}

\put(46,04.2){\makebox(0,0)[bl]{Bridge}}



\put(67.5,12.5){\oval(11,14)}

\put(64,08){\circle*{0.7}} \put(69,08){\circle*{0.7}}
\put(64,13){\circle*{0.7}} \put(69,13){\circle*{0.7}}

\put(64,08){\circle{1.2}} \put(69,08){\circle{1.2}}
\put(64,13){\circle{1.2}} \put(69,13){\circle{1.2}}

\put(64,08){\line(1,0){5}} \put(64,08){\line(0,1){5}}
\put(64,13){\line(1,0){5}} \put(69,08){\line(0,1){5}}

\put(64,08){\line(1,1){5}} \put(69,08){\line(-1,1){5}}


\put(69,13){\line(1,2){2}}

\put(71,17){\circle*{0.7}} \put(71,17){\circle{1.2}}

 \put(63,22){\makebox(0,0)[bl]{Outlier}}

\put(71,18){\line(-1,2){2}}

\end{picture}
\end{center}
%

\subsubsection{Basic scales}

 Usually, the following main approaches (i.e., scales, types of estimates)
 for assessment of vector component (i.e., \(x_{i}, i=\overline{1,m}\)) are used
 (e.g., \cite{har75,jain88,jain99,lev07e,lev15,mirkin05,xu05,xu09}):
 (i) quantitative estimate (Fig. 2.15a),
 (ii) ordinal estimate (Fig. 2.15b),
 (iii) nominal estimate,
 (iv) poset-like scale (Fig. 2.15c),
 (v) fuzzy estimate,
 (vi)  multiset based scales:
 (a) multiset estimate for evaluation of composite system
 (Fig. 2.16, \cite{lev98,lev06,lev12multiset,lev15}),
 (b) interval multiset estimate
 \cite{lev12multiset,lev15},
 and
 (vii) vector-like estimate (i.e., multicriteria description)
 (e.g., \cite{lev06,lev15}).
 Here, traditional fuzzy estimates and hesitant fuzzy estimates are not
 considered
 (e.g., \cite{nchen14,xu14a,zadeh65,zhang15}).

\begin{center}
\begin{picture}(55,63)

\put(00,00){\makebox(0,0)[bl]{Fig. 2.15. Illustration for scales}}

\put(08,05){\makebox(0,0)[bl]{(c) poset-like scale}}

\put(00,12){\makebox(0,0)[bl]{Worst}}
\put(00,09){\makebox(0,0)[bl]{points}}

\put(06,22){\circle*{1.0}}

\put(12,22){\vector(-1,0){5}} \put(12,22){\circle*{1.5}}


\put(06,16){\circle*{1.0}}

\put(12,16){\vector(-1,0){5}} \put(12,16){\vector(-1,1){5}}

\put(12,16){\circle*{1.5}} \put(18,16){\vector(-1,1){5}}


\put(18,16){\vector(-1,0){5}} \put(18,16){\circle*{1.5}}
\put(24,16){\vector(-1,0){5}} \put(30,16){\vector(-1,0){5}}

\put(24,16){\circle*{1.5}}

\put(30,16){\circle*{1.5}} \put(30,22){\circle*{1.5}}

\put(30,22){\vector(-1,-1){5}}


\put(36,22){\vector(-1,0){5}} \put(36,22){\circle*{1.5}}

\put(42,22){\vector(-1,0){5}}

\put(42,22){\circle*{1.5}} \put(42,22){\circle{2.4}}


\put(36,16){\vector(-1,0){5}} \put(36,16){\circle*{1.5}}

\put(42,16){\vector(-1,0){5}} \put(42,22){\vector(-1,-1){5}}

\put(42,16){\circle*{1.5}} \put(42,16){\circle{2.4}}

\put(38,12){\makebox(0,0)[bl]{Best}}
\put(36.5,09){\makebox(0,0)[bl]{points}}



\put(08,26){\makebox(0,0)[bl]{(b) ordinal scale}}

\put(00,39){\makebox(0,0)[bl]{Worst}}
\put(00,36){\makebox(0,0)[bl]{point}}

\put(06,35){\circle*{1.2}}

\put(14,35){\vector(-1,0){7}} \put(14,35){\circle*{1.5}}

\put(22,35){\vector(-1,0){7}}

\put(23,34.5){\makebox(0,0)[bl]{{\bf ...}}}

\put(35,35){\vector(-1,0){7}} \put(35,35){\circle*{1.5}}

\put(43,35){\vector(-1,0){7}}

\put(43,35){\circle*{1.5}} \put(43,35){\circle{2.4}}

\put(39,39){\makebox(0,0)[bl]{Best}}
\put(39,36){\makebox(0,0)[bl]{point}}

\put(05,31){\makebox(0,0)[bl]{\(\kappa\)}}
\put(09,30){\makebox(0,0)[bl]{\((\kappa -1)\)}}
\put(34,31){\makebox(0,0)[bl]{\(2\)}}
\put(42,30.8){\makebox(0,0)[bl]{\(1\)}}


\put(08,44){\makebox(0,0)[bl]{(a) quantitative  scale}}

\put(06,53){\vector(1,0){38}}

\put(06,51){\line(0,1){4}}

\put(00,58){\makebox(0,0)[bl]{Worst}}
\put(00,55){\makebox(0,0)[bl]{point}}

\put(10,53){\circle*{1.2}}
\put(09,48.5){\makebox(0,0)[bl]{\(\theta_{1}\)}}

\put(37,48.5){\makebox(0,0)[bl]{\(\theta_{2}\)}}
\put(38,53){\circle*{1.5}} \put(38,53){\circle{2.4}}

\put(34,58){\makebox(0,0)[bl]{Best}}
\put(34,55){\makebox(0,0)[bl]{point}}

\end{picture}
%
\begin{picture}(53,75.5)

\put(00,00){\makebox(0,0)[bl] {Fig. 2.16. Multiset based scale}}


\put(05,71){\makebox(0,0)[bl]{\(<3,0,0>\) }}

\put(21.6,71){\makebox(0,0)[bl]{Best (ideal)}}
\put(21.6,68){\makebox(0,0)[bl]{point}}

\put(12,67){\line(0,1){3}}
\put(05,62){\makebox(0,0)[bl]{\(<2,1,0>\)}}


\put(12,55){\line(0,1){6}}
\put(05,50){\makebox(0,0)[bl]{\(<2,0,1>\) }}

\put(12,43){\line(0,1){6}}
\put(05,38){\makebox(0,0)[bl]{\(<1,1,1>\) }}

\put(12,31){\line(0,1){6}}
\put(05,26){\makebox(0,0)[bl]{\(<1,0,2>\) }}


\put(12,19){\line(0,1){6}}
\put(05,14){\makebox(0,0)[bl]{\(<0,1,2>\) }}


\put(12,10){\line(0,1){3}}

\put(05,05){\makebox(0,0)[bl]{\(<0,0,3>\) }}

\put(21.6,08){\makebox(0,0)[bl]{Worst}}
\put(21.6,05){\makebox(0,0)[bl]{point}}

\put(14,58){\line(0,1){3}} \put(30,58){\line(-1,0){16}}
\put(30,55){\line(0,1){3}}

\put(23,50){\makebox(0,0)[bl]{\(<1,2,0>\) }}

\put(30,49){\line(0,-1){3}} \put(30,46){\line(-1,0){16}}
\put(14,46){\line(0,-1){3}}
\put(32,43){\line(0,1){6}}
\put(23,38){\makebox(0,0)[bl]{\(<0,3,0>\) }}

\put(14,34){\line(0,1){3}} \put(30,34){\line(-1,0){16}}
\put(30,31){\line(0,1){3}}

\put(32,31){\line(0,1){6}}
\put(23,26){\makebox(0,0)[bl]{\(<0,2,1>\) }}

\put(30,25){\line(0,-1){3}} \put(30,22){\line(-1,0){16}}
\put(14,22){\line(0,-1){3}}

\end{picture}
%
\begin{picture}(45,38)

\put(00,00){\makebox(0,0)[bl] {Fig. 2.17.
 3-component system}}

\put(4,05){\makebox(0,8)[bl]{\(X_{3}(1)\)}}
\put(4,09){\makebox(0,8)[bl]{\(X_{2}(1)\)}}
\put(4,13){\makebox(0,8)[bl]{\(X_{1}(2)\)}}

\put(19,09){\makebox(0,8)[bl]{\(Y_{2}(1)\)}}
\put(19,13){\makebox(0,8)[bl]{\(Y_{1}(3)\)}}

\put(34,05){\makebox(0,8)[bl]{\(Z_{3}(3)\)}}
\put(34,09){\makebox(0,8)[bl]{\(Z_{2}(1)\)}}
\put(34,13){\makebox(0,8)[bl]{\(Z_{1}(1)\)}}
\put(3,19){\circle{2}} \put(18,19){\circle{2}}
\put(33,19){\circle{2}}

\put(0,19){\line(1,0){02}} \put(15,19){\line(1,0){02}}
\put(30,19){\line(1,0){02}}

\put(0,19){\line(0,-1){13}} \put(15,19){\line(0,-1){09}}
\put(30,19){\line(0,-1){13}}

\put(30,14){\line(1,0){01}} \put(30,10){\line(1,0){01}}
\put(30,06){\line(1,0){01}}

\put(32,14){\circle{2}} \put(32,14){\circle*{1}}
\put(32,10){\circle{2}} \put(32,10){\circle*{1}}
\put(32,06){\circle{2}} \put(32,06){\circle*{1}}

\put(15,10){\line(1,0){01}} \put(15,14){\line(1,0){01}}
\put(17,10){\circle{2}} \put(17,10){\circle*{1}}
\put(17,14){\circle{2}} \put(17,14){\circle*{1}}

\put(0,06){\line(1,0){01}} \put(0,10){\line(1,0){01}}
\put(0,14){\line(1,0){01}}

\put(2,10){\circle{2}} \put(2,14){\circle{2}}
\put(2,10){\circle*{1}} \put(2,14){\circle*{1}}
\put(2,06){\circle{2}} \put(2,06){\circle*{1}}
\put(3,24){\line(0,-1){04}} \put(18,24){\line(0,-1){04}}
\put(33,24){\line(0,-1){04}}

\put(3,24){\line(1,0){30}} \put(06,24){\line(0,1){11}}

\put(06,34){\circle*{2.6}}

\put(4,20.5){\makebox(0,8)[bl]{X}}
\put(14,20.5){\makebox(0,8)[bl]{Y}}
\put(29,20.5){\makebox(0,8)[bl]{Z}}

\put(09,33){\makebox(0,8)[bl]{\(S = X\star Y\star Z\)}}

\put(08,29){\makebox(0,8)[bl] {\(S_{1}=X_{2}\star Y_{1}\star
Z_{2}\)}}

\put(08,25){\makebox(0,8)[bl] {\(S_{2}=X_{3}\star Y_{2}\star
Z_{3}\)}}

\end{picture}
\end{center}



 Some fundamentals of multisets and comparison of sets and multisets
  have been described
 in (\cite{hallez09,knuth98,yager86}).
 Our brief description of multiset estimates
 is the following
 (e.g., \cite{lev98,lev06,lev12multiset,lev15}).
 The approach consists in assignment of elements (\(1,2,3,...\))
 into an ordinal scale \([1,2,...,l]\).
%
%
 As a result, a multiset based estimate is obtained,
 where a basis set involves all levels of the ordinal scale:
 \(\Omega = \{ 1,2,...,l\}\) (the levels are linear ordered:
 \(1 \succ 2 \succ 3 \succ ...\)) and
 the assessment problem (for each alternative)
 consists in selection of a multiset over set \(\Omega\) while taking into
 account two conditions:


 {\it 1.} cardinality of the selected multiset equals a specified

 number of elements \( \eta = 1,2,3,...\)
 (i.e., multisets of cardinality \(\eta \) are considered);

 {\it 2.} ``configuration'' of the multiset is the following:
 the selected elements of \(\Omega\) cover an interval over scale \([1,l]\)
 (i.e., ``interval multiset estimate'').

 Thus, an estimate \(e\) for an alternative \(A\) is
 (scale \([1,l]\), position-based form or position form):
 \(e(A) = (\eta_{1},...,\eta_{\iota},...,\eta_{l})\),
 where \(\eta_{\iota}\) corresponds to the number of elements at the
 level \(\iota\) (\(\iota = \overline{1,l}\)), or
 \(e(A) = \{ \overbrace{1,...,1}^{\eta_{1}},\overbrace{2,...2}^{\eta_{2}},
 \overbrace{3,...,3}^{\eta_{3}},...,\overbrace{l,...,l}^{\eta_{l}}
 \}\).
 The number of multisets of cardinality \(\eta\),
 with elements taken from a finite set of cardinality \(l\),
 is called the
 ``multiset coefficient'' or ``multiset number''
  (\cite{knuth98,yager86}):
 ~~\( \mu^{l,\eta} =
   \frac{l(l+1)(l+2)... (l+\eta-1) } {\eta!}
   \).
%
%
%
%
 This number corresponds to possible estimates
 (without taking into account interval condition 2).
 The basic multiset estimate
 (i.e., without taking into account condition 2)
 can be used as integration of ordinal estimates
 to obtain a resultant estimate for a composite system
 (when system components are evaluated
 via ordinal scale  \([1,2,3]\))
 (e.g., \cite{lev98,lev06,lev15}) (Fig. 2.16, Fig. 2.17).

 In the case of condition 2 (i.e., interval mutliset estimate),
 the number of estimates is decreased.
 Generally, assessment problems based on interval multiset estimates
 can be denoted as follows: ~\(P^{l,\eta}\).
 A poset-like scale of interval multiset estimates for assessment problem \(P^{3,4}\)
 is presented in Fig. 2.18.
%
 Calculation
 of multiset  estimate is be based on
 transformation of vector ordinal estimate.
 An illustrative numerical example for obtaining
 multiset vector estimate is the following:

  \(\overline{x} = (0, 3, 1, 0, 2, 1) \) \(\ \Longrightarrow \)
 \(e(\overline{x}) = (e_{o},e_{1},e_{2},e_{3}) =   (2,2,1,1) \),

 where \(e_{k}\) equals the number of ordinal estimate \(k\)
 in ordinal vector estimate  \(\overline{x}\).


\begin{center}
\begin{picture}(70,140)

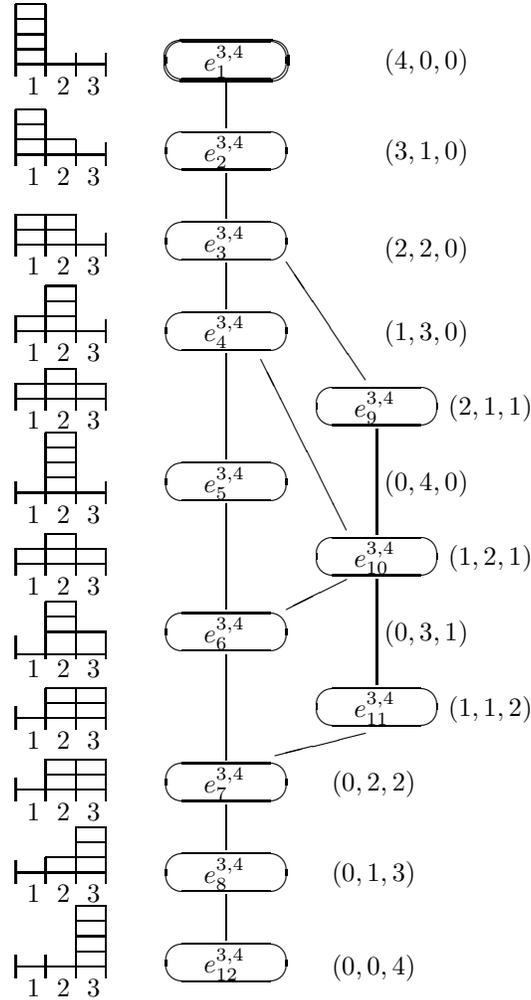
\put(00,00){\makebox(0,0)[bl] {Fig. 2.18. Scale, estimates
 (\(P^{3,4}\)) \cite{lev12multiset,lev15}}}




\put(25,126.7){\makebox(0,0)[bl]{\(e^{3,4}_{1}\) }}

\put(28,129){\oval(16,5)} \put(28,129){\oval(16.5,5.5)}


\put(49,127){\makebox(0,0)[bl]{\((4,0,0)\) }}

\put(00,128.5){\line(0,1){08}} \put(04,128.5){\line(0,1){08}}

\put(00,130.5){\line(1,0){4}} \put(00,132.5){\line(1,0){4}}
\put(00,134.5){\line(1,0){4}} \put(00,136.5){\line(1,0){4}}

\put(00,128.5){\line(1,0){12}}

\put(00,127){\line(0,1){3}} \put(04,127){\line(0,1){3}}
\put(08,127){\line(0,1){3}} \put(12,127){\line(0,1){3}}

\put(01.5,124.5){\makebox(0,0)[bl]{\(1\)}}
\put(05.5,124.5){\makebox(0,0)[bl]{\(2\)}}
\put(09.5,124.5){\makebox(0,0)[bl]{\(3\)}}


\put(28,120){\line(0,1){6}}


\put(25,114.7){\makebox(0,0)[bl]{\(e^{3,4}_{2}\) }}

\put(28,117){\oval(16,5)}


\put(49,115){\makebox(0,0)[bl]{\((3,1,0)\) }}

\put(00,116.5){\line(0,1){06}} \put(04,116.5){\line(0,1){06}}
\put(08,116.5){\line(0,1){02}}

\put(00,118.5){\line(1,0){8}} \put(00,120.5){\line(1,0){4}}
\put(00,122.5){\line(1,0){4}}

\put(00,116.5){\line(1,0){12}}

\put(00,115){\line(0,1){3}} \put(04,115){\line(0,1){3}}
\put(08,115){\line(0,1){3}} \put(12,115){\line(0,1){3}}

\put(01.5,112.5){\makebox(0,0)[bl]{\(1\)}}
\put(05.5,112.5){\makebox(0,0)[bl]{\(2\)}}
\put(09.5,112.5){\makebox(0,0)[bl]{\(3\)}}


\put(28,108){\line(0,1){6}}


\put(25,102.7){\makebox(0,0)[bl]{\(e^{3,4}_{3}\) }}

\put(28,105){\oval(16,5)}


\put(49,102){\makebox(0,0)[bl]{\((2,2,0)\) }}

\put(00,104.5){\line(0,1){04}}

\put(04,104.5){\line(0,1){04}} \put(08,104.5){\line(0,1){04}}
\put(00,106.5){\line(1,0){8}} \put(00,108.5){\line(1,0){8}}

\put(00,104.5){\line(1,0){12}}

\put(00,103){\line(0,1){3}} \put(04,103){\line(0,1){3}}
\put(08,103){\line(0,1){3}} \put(12,103){\line(0,1){3}}

\put(01.5,100.5){\makebox(0,0)[bl]{\(1\)}}
\put(05.5,100.5){\makebox(0,0)[bl]{\(2\)}}
\put(09.5,100.5){\makebox(0,0)[bl]{\(3\)}}


\put(28,96){\line(0,1){6}}


\put(25,90.7){\makebox(0,0)[bl]{\(e^{3,4}_{4}\) }}

\put(28,93){\oval(16,5)}


\put(49,91){\makebox(0,0)[bl]{\((1,3,0)\) }}

\put(00,93){\line(0,1){02}}

\put(04,93){\line(0,1){06}} \put(08,93){\line(0,1){06}}

\put(00,95){\line(1,0){8}} \put(04,97){\line(1,0){4}}
\put(04,99){\line(1,0){4}}

\put(00,93){\line(1,0){12}}

\put(00,91.5){\line(0,1){3}} \put(04,91.5){\line(0,1){3}}
\put(08,91.5){\line(0,1){3}} \put(12,91.5){\line(0,1){3}}

\put(01.5,89){\makebox(0,0)[bl]{\(1\)}}
\put(05.5,89){\makebox(0,0)[bl]{\(2\)}}
\put(09.5,89){\makebox(0,0)[bl]{\(3\)}}


\put(28,76){\line(0,1){14}}

\put(25,70.7){\makebox(0,0)[bl]{\(e^{3,4}_{5}\) }}

\put(28,73){\oval(16,5)}


\put(49,71){\makebox(0,0)[bl]{\((0,4,0)\) }}

\put(04,71.5){\line(0,1){08}} \put(08,71.5){\line(0,1){08}}

\put(04,73.5){\line(1,0){4}} \put(04,75.5){\line(1,0){4}}
\put(04,77.5){\line(1,0){4}} \put(04,79.5){\line(1,0){4}}

\put(00,71.5){\line(1,0){12}}

\put(00,70){\line(0,1){3}} \put(04,70){\line(0,1){3}}
\put(08,70){\line(0,1){3}} \put(12,70){\line(0,1){3}}

\put(01.5,67.5){\makebox(0,0)[bl]{\(1\)}}
\put(05.5,67.5){\makebox(0,0)[bl]{\(2\)}}
\put(09.5,67.5){\makebox(0,0)[bl]{\(3\)}}


\put(28,56){\line(0,1){14}}

\put(25,50.7){\makebox(0,0)[bl]{\(e^{3,4}_{6}\) }}

\put(28,53){\oval(16,5)}


\put(49,51){\makebox(0,0)[bl]{\((0,3,1)\) }}

\put(04,51){\line(0,1){06}} \put(08,51){\line(0,1){06}}
 \put(12,51){\line(0,1){02}}

\put(04,53){\line(1,0){8}} \put(04,55){\line(1,0){4}}
\put(04,57){\line(1,0){4}}

\put(00,50){\line(1,0){12}}

\put(00,49.5){\line(0,1){3}} \put(04,49.5){\line(0,1){3}}
\put(08,49.5){\line(0,1){3}} \put(12,49.5){\line(0,1){3}}

\put(01.5,47){\makebox(0,0)[bl]{\(1\)}}
\put(05.5,47){\makebox(0,0)[bl]{\(2\)}}
\put(09.5,47){\makebox(0,0)[bl]{\(3\)}}


\put(28,36){\line(0,1){14}}

\put(25,30.7){\makebox(0,0)[bl]{\(e^{3,4}_{7}\) }}

\put(28,33){\oval(16,5)}


\put(42,31){\makebox(0,0)[bl]{\((0,2,2)\) }}

\put(04,32){\line(0,1){04}}

\put(08,32){\line(0,1){04}} \put(12,32){\line(0,1){04}}
\put(04,34){\line(1,0){8}} \put(04,36){\line(1,0){8}}

\put(00,32){\line(1,0){12}}

\put(00,30.5){\line(0,1){3}} \put(04,30.5){\line(0,1){3}}
\put(08,30.5){\line(0,1){3}} \put(12,30.5){\line(0,1){3}}

\put(01.5,28){\makebox(0,0)[bl]{\(1\)}}
\put(05.5,28){\makebox(0,0)[bl]{\(2\)}}
\put(09.5,28){\makebox(0,0)[bl]{\(3\)}}


\put(28,24){\line(0,1){6}}

\put(25,18.7){\makebox(0,0)[bl]{\(e^{3,4}_{8}\) }}

\put(28,21){\oval(16,5)}


\put(42,19){\makebox(0,0)[bl]{\((0,1,3)\) }}

\put(04,21){\line(0,1){02}}

\put(08,21){\line(0,1){06}} \put(12,21){\line(0,1){06}}

\put(04,23){\line(1,0){8}} \put(08,25){\line(1,0){4}}
\put(08,27){\line(1,0){4}}

\put(00,21){\line(1,0){12}}

\put(00,19.5){\line(0,1){3}} \put(04,19.5){\line(0,1){3}}
\put(08,19.5){\line(0,1){3}} \put(12,19.5){\line(0,1){3}}

\put(01.5,17){\makebox(0,0)[bl]{\(1\)}}
\put(05.5,17){\makebox(0,0)[bl]{\(2\)}}
\put(09.5,17){\makebox(0,0)[bl]{\(3\)}}


\put(28,12){\line(0,1){6}}


\put(25,06.7){\makebox(0,0)[bl]{\(e^{3,4}_{12}\) }}

\put(28,09){\oval(16,5)}


\put(42,6.5){\makebox(0,0)[bl]{\((0,0,4)\) }}

\put(08,8.5){\line(0,1){08}} \put(12,8.5){\line(0,1){08}}
\put(08,10.5){\line(1,0){4}} \put(08,12.5){\line(1,0){4}}
\put(08,14.5){\line(1,0){4}} \put(08,16.5){\line(1,0){4}}

\put(00,8.5){\line(1,0){12}}

\put(00,07){\line(0,1){3}} \put(04,07){\line(0,1){3}}
\put(08,07){\line(0,1){3}} \put(12,07){\line(0,1){3}}

\put(01.5,4.5){\makebox(0,0)[bl]{\(1\)}}
\put(05.5,4.5){\makebox(0,0)[bl]{\(2\)}}
\put(09.5,4.5){\makebox(0,0)[bl]{\(3\)}}


\put(46.5,86.5){\line(-2,3){10.5}}




\put(45,80.7){\makebox(0,0)[bl]{\(e^{3,4}_{9}\) }}

\put(48,83){\oval(16,5)}


\put(57.5,81){\makebox(0,0)[bl]{\((2,1,1)\) }}

\put(00,84){\line(0,1){02}} \put(04,84){\line(0,1){04}}
\put(08,84){\line(0,1){04}} \put(12,84){\line(0,1){02}}

\put(00,86){\line(1,0){12}} \put(04,88){\line(1,0){4}}

\put(00,84){\line(1,0){12}}

\put(00,82.5){\line(0,1){3}} \put(04,82.5){\line(0,1){3}}
\put(08,82.5){\line(0,1){3}} \put(12,82.5){\line(0,1){3}}

\put(01.5,80){\makebox(0,0)[bl]{\(1\)}}
\put(05.5,80){\makebox(0,0)[bl]{\(2\)}}
\put(09.5,80){\makebox(0,0)[bl]{\(3\)}}


\put(44,66.5){\line(-1,2){11.4}}


\put(44,60){\line(-2,-1){8}}

\put(48,66){\line(0,1){14}}


\put(45,60.7){\makebox(0,0)[bl]{\(e^{3,4}_{10}\) }}

\put(48,63){\oval(16,5)}


\put(57.5,61){\makebox(0,0)[bl]{\((1,2,1)\) }}

\put(00,62){\line(0,1){02}} \put(04,62){\line(0,1){04}}
\put(08,62){\line(0,1){04}} \put(12,62){\line(0,1){02}}

\put(00,64){\line(1,0){12}} \put(04,66){\line(1,0){4}}

\put(00,62){\line(1,0){12}}

\put(00,60.5){\line(0,1){3}} \put(04,60.5){\line(0,1){3}}
\put(08,60.5){\line(0,1){3}} \put(12,60.5){\line(0,1){3}}

\put(01.5,58){\makebox(0,0)[bl]{\(1\)}}
\put(05.5,58){\makebox(0,0)[bl]{\(2\)}}
\put(09.5,58){\makebox(0,0)[bl]{\(3\)}}



\put(46.5,39.5){\line(-4,-1){12}}

\put(48,46){\line(0,1){14}}


\put(45,40.7){\makebox(0,0)[bl]{\(e^{3,4}_{11}\) }}

\put(48,43){\oval(16,5)}


\put(57.5,41){\makebox(0,0)[bl]{\((1,1,2)\) }}

\put(04,41.5){\line(0,1){04}}

\put(08,41.5){\line(0,1){04}} \put(12,41.5){\line(0,1){04}}
\put(04,43.5){\line(1,0){8}} \put(04,45.5){\line(1,0){8}}

\put(00,41.5){\line(1,0){12}}

\put(00,40){\line(0,1){3}} \put(04,40){\line(0,1){3}}
\put(08,40){\line(0,1){3}} \put(12,40){\line(0,1){3}}

\put(01.5,37.5){\makebox(0,0)[bl]{\(1\)}}
\put(05.5,37.5){\makebox(0,0)[bl]{\(2\)}}
\put(09.5,37.5){\makebox(0,0)[bl]{\(3\)}}


\end{picture}
\end{center}


%
 Brief descriptions of the scales above
 and their transformations (e.g., mapping, integration)
 are presented in \cite{lev12multiset,lev13eva,lev15}.
 The basic types of operations over estimates above are the following
 (Table 2.3) (e.g., \cite{lev13eva,lev15}):

 {\it 1.} transformation of an estimate
 (including transformation into an estimate of
 another type),


 {\it 2.}
 calculation
 the difference (distance/proximity)
  for two estimates,

 {\it 3.} integrations
 (e.g., summarization,
 average estimate or median-like estimate).

 A basic approach to integration of qualitative estimates
 (including vector qualitative estimates)
 consists in summarization or
 calculation of
 an average value.
 On the other hand, integration of of ordinal estimates,
 poset-like estimates, and multiset estimates
 (i.e., ``structural'' estimates)
 is usually considered as
 calculation of
  a median-like
  estimate (i.e., as agreement/consensus structure).
 This kind of problems is formulated as an optimization
 (or multicriteria optimization)
 (e.g., \cite{lev98,lev11agg,lev15}).
 Sometimes, the resultant integrated estimate
 can be considered as a ``fuzzy'' structure
 (e.g., interval-like integrated ranking for integration
 of rankings in
 \cite{lev98,lev15}).

 The above-mentioned operations
 (integration as summarization,
 calculation
 of proximity,
 calculation of
  a median estimate)
  for multiset estimates are presented in
   \cite{lev12multiset,lev15}.

\newpage
\begin{center}
{\bf Table 2.3.} Basic operations over estimate(s) \\
\begin{tabular}{| c | l | l | l | l| }
\hline
 No. & Type of initial&Transformation&Difference (proximity)&Integration\\
 &  estimate(s)& of estimate& of two estimates&of several estimates\\
  &    &(resultant estimate)&(resultant estimate)&(resultant estimate)\\
\hline
 1.&Quantitative&1.Qualitative&1.Quantitative&1.Quantitative  \\
   &            &             &              &  (average value)\\
   &            &2.Ordinal    & 2.Ordinal    &2.Vector estimate\\
 \hline
 2.&Ordinal     &1.Ordinal   &1.Ordinal      &1.Ordinal  \\
   &            &            &               &  (average/median)\\
   &            &            &               &2.Vector estimate\\
   &            &            &               &3.Multiset estimate\\
 \hline
 3.&Nominal     &1.Nominal   &1.Ordinal      &1.Vector estimate\\
 \hline
 4.&Vector      &1.Vector    &1.Qualitative&1.Vector  \\
   &            &2.Ordinal   &2.Ordinal    &  (average/median)\\
   &            &3.Multiset  &3.Vector     &  \\
 \hline
 4.&Poset-like  &1.Poset-like&1.Qualitative&1.Poset-like  \\
   &            &2.Vector    &2.Ordinal     &(average/median)\\
   &            &3.Multiset  &3.Vector      & 2.Multiset    \\
   &            &4.Ordinal   &              &     \\

 \hline

 5.&Multiset    &1.Multiset &1.Ordinal &1.Multiset  \\
   &            &2.Ordinal  &2.Multiset&(average/median)\\
   &            &           &3.Vector  &     \\
 \hline
\end{tabular}
\end{center}

\subsubsection{Assessment/evaluation problems}

%

 Generally,
  the following measurement approaches are under examination
  (for solutions as clustering, ranking, consensus clustering, and their components):
 {\it 1.} metrics/proximities:
 (i) for objects/clusters,
 (ii) for orders over clusters,
 (iii) for clustering solutions;
 {\it 2.} total measures for clustering solution(s)
 (total quality);
 {\it 3.} measure for an aggregation structure
 (i.e., median, consensus, agreement structure, covering
 structure).
%
%
 Table 2.4 involves a list of basic types of the assessment/evaluation
 problems.



\subsubsection{Calculation of
 objects/clusters proximities}

  The initial set of items (objects) under examination is
  \(A = \{a_{1},...,a_{j},...,a_{n}\}\).
 There are \(m\) parameters of \(x \in A\)
 and  vector estimate
 \( \overline{x}= (x_{1},..., x_{1},...,x_{m})\) (Fig. 2.19).
 Here,
 the following cases of metrics/proximities for objects/clusters
 are considered:

 {\it Case 1.} objects/points - object/point (Fig. 2.20, Fig. 2.21a).

 {\it Case 2.} object - cluster (subset) (Fig. 2.21b).

 {\it Case 3.} Intra-cluster proximity (distance):
  for all elements in cluster.

 {\it Case 4.} Inter-cluster proximity (distance):
  cluster - cluster (Fig. 2.21c).


  Further, the above-mentioned cases are examined.

~~

 {\bf Case 1.} Metrics/proximity for two objects (points) (Fig. 2.20, Fig. 2.21a).

   Two objects are examined while taking into account  \(m\) parameters
   (criteria):
%
%
 (i) the first object
   \(x \in A\): ~vector estimate
   ~\(\overline{x} = (x_{1},...,x_{i},...,x_{m}) \),
  (ii) the second object
  \(y \in A\): ~vector estimate
  ~\(\overline{y} = (y_{1},...,y_{i},...,y_{m}) \).
 Here, \(x_{i},y_{i}\) \((i=\overline{1,m})\)  are real numbers.
 The basic types of proximities/distances
 (as an integration of vector-like difference between vector
 element estimate upon  quantitative scales)
 between two objects  \(x\) and \(y\)
 (\(D( x,y )\))
 are the following
 (e.g.,
 \cite{gord99,har75,jain88,jain99,kau90,kot07,lev15,mirkin96,mirkin05,pedr05,xu05,xu09,zop02}):

 {\it 1.} Euclidean distance:~
 \(D(x,y)=[ ~ \sum^{m}_{i=1} ~  | x_{i} - y_{i} |^{2} ~]^{1/2}\).

 {\it 2.} Minkowski distance:~
 \(D^{mink}(x,y)=[ ~ \sum^{m}_{i=1} ~  | x_{i} - y_{i} |^{r} ~]^{1/r}\) ~(\(r > 0\)).

 {\it 3.} Manhattan distance:~
 \(D^{manh}(x,y)= ~ \sum^{m}_{i=1} ~ | x_{i} - y_{i} | \).

 {\it 4.} Tchebyschev distance:~
 \(D^{cheb}(x,y)= ~ \max_{i \in \{ 1,...,m\} } ~ | x_{i} - y_{i} |\).
%

 {\it 5.} Canberra distance:~
 \(D^{cam}(x,y)=~\sum^{m}_{i=1}  \frac{ |x_{i} - y_{i}|}{  |x_{i} + y_{i}|} \) ~~(\(x_{i} > 0\) and \(y_{i} > 0\)).

 {\it 6.} Vector proximity:~
 \(\overline{D( x,y)}= ~
 ( |x_{1} - y_{1}|,...,|x_{i} - y_{i}|,...,|x_{m} - y_{m}|)\);

 {\it 7.} Ordinal estimate (for example, scale \([0,1,2,3,4]\):~
  \(0\) corresponds to the same objects or equivalent ones,
 \(1\) corresponds to the ``very close'' objects,
 \(2\) corresponds to the ``close'' objects,
 \(3\) corresponds to the ``different'' objects,
 \(4\) corresponds to the ``very different'' objects).


\begin{center}
{\bf Table 2.4.} Assessment/evaluation problems
 \\
\begin{tabular}{| c | l | l | l | l| }
\hline
 No.  & Analyzed object(s)
  & Goal & Approach& Estimate type (scales)  \\
\hline

 1. & Element (item)& Description & Assessment
                   &  Quantitative, ordinal, \\
 & && (e.g., expert,  &  nominal, fuzzy, multiset, \\

 &&& statistics) & vector \\



 2.& Two elements/items
 & Proximity/distance &
 Calculation
 &Quantitative, ordinal,  \\

 & &&  &fuzzy, multiset, vector  \\


 3. & Element,
 & Proximity/distance &
 Calculation
 &Quantitative, ordinal,\\

 & cluster (subset)&& & fuzzy,  multiset, vector  \\



 4. & Two clusters
 & Proximity/distance &
  Calculation &Quantitative, ordinal, \\

 &  (subsets)&& & fuzzy, multiset, vector  \\




 5. & All elements of
 & Intra-cluster proximity&
  Calculation&Quantitative, ordinal, \\

 & cluster (subset)&(quality of cluster as
 & &fuzzy,  multiset, vector \\

 && element proximity)&& \\



 6.&All elements in all &Total intra-cluster&
   Calculation &Quantitative, ordinal, \\

 & clusters (clustering& proximity (generalized  &
  &fuzzy, multiset, vector  \\

 &solution) & element proximity&&  \\

 & &in all clusters)&& \\


 7. & All clusters of
 & Inter-cluster proximity&
  Calculation&Quantitative, ordinal, \\

 & clustering solution &
 (quality of solution as &&fuzzy,  multiset, vector \\

 && integrated proximity&&   \\

 && between clusters)&&  \\


 8.&Clustering solution&Criteria, requirements/  &Constraint&Quantitative, ordinal, \\

 &              & constraints        &satisfaction/
 & fuzzy, multiset, vector \\

 &&                                   &optimization   & \\



 9. & Two rankings (two&Proximity/distance&
  Calculation&Quantitative, ordinal, \\

 &  sorting solutions)&&&fuzzy, multiset, vector \\



 10.&Ranking (sorting)&Proximity to standard&Constraint&Quantitative, ordinal, \\

 & solution  & solution, correspon- &satisfaction/ & fuzzy, multiset, vector \\

 & & dence to requirements&optimization&   \\

  & & (quality of solution)&problem&   \\


 11.&Two hierarchies (e.g.,&Proximity/distance&
  Calculation&Quantitative, ordinal, \\

 & trees) over clusters && &fuzzy,  multiset, vector \\



 12.&Hierarchy over&Proximity to standard&Constraint&Quantitative, ordinal, \\

 & clusters (solution) & solution, correspon- &satisfaction/&fuzzy,  multiset, vector \\

 & & dence to requirements&optimization&  \\



 13.&Aggregated item(s):& Quality of aggregated &
  Calculation&Quantitative, ordinal,  \\
      &``median item/set'',      & item(s) (value,         &&fuzzy,  multiset, vector \\
      &``center'', covering&   vector)         &&  \\
    & object (e.g., ellipsoid)&            &&  \\


 14.&Consensus clustering& Quality of consensus&Constraint&Quantitative, ordinal, \\
      &(median/agreement)    & solution (value,   &satisfation/ &fuzzy,  multiset, vector \\
      &  &     vector)    &optimization &  \\


 15. & Consensus ranking/& Quality of consensus&Constraint&Quantitative, ordinal,\\
      & sorting solution   & solution (value,  &satisfaction/ &fuzzy, multiset, vector \\
      &                   &  vector)      &optimization& \\

 16. & Consensus hierarchi-&Quality of consensus&Constraint&Quantitative, ordinal, \\
      & cal clustering       &  solution &satisfaction/ &fuzzy, multiset, vector  \\
     &   solution            &                 & optimization &  \\
\hline
\end{tabular}
\end{center}

 {\it 8.} Fuzzy estimate of object proximity
 and/or vector fuzzy estimate of object proximity.
 Fuzzy estimates are widely used in
 clustering methods (e.g.,
 \cite{bar99,chen13,dur06,grave10,hop99,jain99,kim04,kris95,miy90,oli07,sato06})
 including
 clustering based on hesitant fuzzy estimates
  (e.g., \cite{nchen14,xu14a,zhang15})
  (here they are not considered).
%

 {\it 9.} Multiset
 estimate of object proximity
 \cite{lev12multiset,lev15}.

{\it 10.} Angular separation (e.g., \cite{pedr05}, Fig. 2.20):~
 \(D^{angular} (x,y) = \frac
 {\sum_{i=1}^{m} x_{i} y_{i}}
 { [ \sum_{i=1}^{m} x^{2}_{i} \sum_{i=1}^{m} y^{2}_{i} ]^{1/2}}
 \).

 The similarity measure corresponds to  the angle
 between the item vectors in
 directions of \(x\) and \(y\)
 \cite{pedr05}.

\begin{center}
\begin{picture}(76,27)

\put(01.5,00){\makebox(0,0)[bl]{Fig. 2.19. Illustration for
 object/item}}

\put(04,09){\vector(0,1){17.5}} \put(04,09){\vector(1,0){55}}

\put(04,09){\vector(-1,-1){4}}

\put(26,17.5){\oval(41,15)}

\put(25,23){\circle*{0.7}}  \put(16,23){\circle*{0.7}}
\put(41,22){\circle*{0.7}} \put(09,22){\circle*{0.7}}
\put(20,22){\circle*{0.7}}

\put(28,14){\circle*{0.7}} \put(19,13){\circle*{0.7}}
\put(37,15){\circle*{0.7}}


\put(07,16.4){\makebox(0,0)[bl]{\(a:
 \overline{x}= (x_{1},..., x_{1},...,x_{m})\)}}

\put(10,15.5){\circle*{1.0}} \put(10,15.5){\circle{1.8}}

\put(48,22){\makebox(0,0)[bl]{Set of}}
\put(48,18.5){\makebox(0,0)[bl]{objects/}}
\put(48,16){\makebox(0,0)[bl]{items}}

\put(51,12){\makebox(0,0)[bl]{\(A\)}}

\end{picture}
%
%
%
\begin{picture}(52,28)
\put(005,00){\makebox(0,0)[bl]{Fig. 2.20.
 Angle proximity
  }}

\put(04,09){\vector(0,1){17.5}} \put(04,09){\vector(1,0){47}}

\put(04,09){\vector(-1,-1){4}}


\put(22,19){\oval(34,18)}

\put(04,09){\circle*{1.3}}

 \put(02,10){\makebox(0,0)[bl]{\(\overline{0}\)}}

\put(04,09){\vector(1,3){5}}

 \put(04,09){\vector(4,1){27.5}}


\put(30,21){\circle*{0.7}} \put(22,16){\circle*{0.7}}
\put(32,23){\circle*{0.7}} \put(33,12){\circle*{0.7}}

\put(11,25){\makebox(0,0)[bl]{\(x\)}}

\put(09.5,25.5){\circle*{1.7}}



\put(34,15){\makebox(0,0)[bl]{\(y\)}}

\put(32,16){\circle*{0.7}} \put(32,16){\circle{1.5}}

\put(12.5,17){\vector(-1,1){4}} \put(12.5,17){\vector(1,-1){4}}

\put(17,21){\line(-1,-1){04}}
\put(13.5,21){\makebox(0,0)[bl]{\(D(x,y)\)}}

\put(40,20){\makebox(0,0)[bl]{Objects}}
\put(42,16){\makebox(0,0)[bl]{set \(A\)}}

\end{picture}
\end{center}

\begin{center}
\begin{picture}(40,36)
\put(29.5,00){\makebox(0,0)[bl]{Fig. 2.21.
 Object/cluster proximities}}


\put(00,05){\makebox(0,0)[bl]{(a) object \(x\), object \(y\)}}

\put(16.5,20){\oval(33,20)}

\put(13,27){\circle*{0.7}} \put(17.5,28){\circle*{0.7}}
\put(14.5,24){\circle*{0.7}} \put(4,26){\circle*{0.7}}
\put(27,15){\circle*{0.7}} \put(29,27){\circle*{0.7}}
\put(30,21){\circle*{0.7}}


\put(05,21){\circle*{0.7}} \put(05,15){\circle*{0.7}}
\put(08,23){\circle*{0.7}} \put(09,14){\circle*{0.7}}

\put(06.5,19){\makebox(0,0)[bl]{\(x\)}}
\put(07.5,17.5){\circle*{1.6}}


\put(24,27){\circle*{0.7}} \put(21.5,18){\circle*{0.7}}
\put(20,25){\circle*{0.7}} \put(25,20){\circle*{0.7}}

\put(21.5,24){\makebox(0,0)[bl]{\(y\)}}
\put(22.5,22.5){\circle*{0.7}} \put(22.5,22.5){\circle{1.7}}


\put(10.5,18.5){\vector(3,1){10.8}}
\put(19.5,21.5){\vector(-3,-1){10.8}}

\put(19,14.5){\line(-2,3){03.3}}
\put(14,11){\makebox(0,0)[bl]{\(D(x,y)\)}}

\put(06,31){\makebox(0,0)[bl]{Objects set \(A\)}}

\end{picture}
%
\begin{picture}(40,36)



\put(00,05){\makebox(0,0)[bl]{(b) object \(x\), cluster \(Y\)}}

\put(16.5,20){\oval(33,20)}

\put(13,27){\circle*{0.7}} \put(17.5,28){\circle*{0.7}}
\put(14.5,24){\circle*{0.7}} \put(4,26){\circle*{0.7}}
\put(27,15){\circle*{0.7}} \put(29,27){\circle*{0.7}}
\put(30,21){\circle*{0.7}}


\put(05,21){\circle*{0.7}} \put(05,15){\circle*{0.7}}
\put(08,23){\circle*{0.7}} \put(09,14){\circle*{0.7}}

\put(06.5,19){\makebox(0,0)[bl]{\(x\)}}
\put(07.5,17.5){\circle*{1.6}}

\put(22.5,22.5){\oval(08,12)}

\put(24,27){\circle*{0.7}} \put(21.5,18){\circle*{0.7}}
\put(20,25){\circle*{0.7}} \put(25,20){\circle*{0.7}}

\put(24,27){\circle{1.7}} \put(21.5,18){\circle{1.7}}
\put(20,25){\circle{1.7}} \put(25,20){\circle{1.7}}

\put(20.5,24){\makebox(0,0)[bl]{\(Y\)}}
\put(22.5,22.5){\circle*{0.7}} \put(22.5,22.5){\circle{1.7}}


\put(10.5,18.5){\vector(3,1){07.4}}
\put(16.5,20.5){\vector(-3,-1){07.8}}

\put(19,14.5){\line(-2,3){03.3}}
\put(14,11){\makebox(0,0)[bl]{\(D(x,Y)\)}}

\put(06,31){\makebox(0,0)[bl]{Objects set \(A\)}}

\end{picture}
%
\begin{picture}(36.5,36)


\put(00,05){\makebox(0,0)[bl]{(c) cluster \(X\), cluster \(Y\)}}

\put(16.5,20){\oval(33,20)}

\put(13,27){\circle*{0.7}} \put(17.5,28){\circle*{0.7}}
\put(14.5,24){\circle*{0.7}} \put(4,26){\circle*{0.7}}
\put(27,15){\circle*{0.7}} \put(29,27){\circle*{0.7}}
\put(30,21){\circle*{0.7}}

\put(07.5,18.5){\oval(08,12)}

\put(05,21){\circle*{1.7}} \put(05,15){\circle*{1.7}}
\put(08,23){\circle*{1.7}} \put(09,14){\circle*{1.7}}

\put(06.5,19){\makebox(0,0)[bl]{\(X\)}}
\put(07.5,17.5){\circle*{1.6}}

\put(22.5,22.5){\oval(08,12)}

\put(24,27){\circle*{0.7}} \put(21.5,18){\circle*{0.7}}
\put(20,25){\circle*{0.7}} \put(25,20){\circle*{0.7}}

\put(24,27){\circle{1.7}} \put(21.5,18){\circle{1.7}}
\put(20,25){\circle{1.7}} \put(25,20){\circle{1.7}}

\put(20.5,24){\makebox(0,0)[bl]{\(Y\)}}
\put(22.5,22.5){\circle*{0.7}} \put(22.5,22.5){\circle{1.7}}


\put(13.5,19.5){\vector(3,1){04.4}}
\put(16.5,20.5){\vector(-3,-1){04.8}}

\put(19,14.5){\line(-2,3){03.3}}
\put(14,11){\makebox(0,0)[bl]{\(D(X,Y)\)}}

\put(06,31){\makebox(0,0)[bl]{Objects set \(A\)}}

\end{picture}
\end{center}


 Illustrative numerical examples are the following.

~~

 {\bf Example 2.1.}
 An ordinal vector proximity can be used for
  calculation of
 the proximity of two ordinal vectors \(\overline{x}\) and
 \(\overline{y}\)
   (ordinal scale \([1,5]\) for vector estimates,
    ordinal scale \([0,4]\) for vector proximity):

 (a) ordinal vector estimates are:
   \(x: \overline{x} = (3,4,1,1,2,5) \),
   \(y: \overline{y} = (3,1,2,1,4,4) \);

 (b) ordinal proximities are:

 (i) \(\overline{D(x,y)} = (x_{1} -y_{1},x_{2}-y_{2},x_{3}-y_{3},x_{4}-y_{4},x_{5}-y_{5},x_{6}-y_{6} )
 =   (0,3,-1,0,-2,1) \),

 (ii) \(\overline{D'(x,y)} = (|x_{1}-y_{1}|,|x_{2}-y_{2}|,|x_{3}-y_{3}|,|x_{4}-y_{4}|,|x_{5}-y_{5}|,|x_{6}-y_{6}| )
 =   (0,3,1,0,2,1) \).

~~

 {\bf Example 2.2.}
 An ordinal vector proximity can be used for
  calculation of the proximity
  of two quantitative vectors \(\overline{x}\) and
 \(\overline{y}\)
   (quantitative scale \((0,5)\) for vector estimates,
    ordinal scale \([0,4]\) for vector proximity).
 Two initial quantitative vector estimates are:
  (a)
   \(\overline{x} = (0.3, 3.5, 1.4, 1.5, 2.3, 4.9) \),
  (b)
  \(\overline{y} = (0.3, 0.8, 2.2, 1.6, 3.9, 4.1) \).
 Examples for calculation of proximities are:

 (1) The quantitative proximity is:

 \(\overline{D(x,y)} = (|x_{1}-y_{1}|,|x_{2}-y_{2}|,|x_{3}-y_{3}|,|x_{4}-y_{4}|,|x_{5}-y_{5}|,|x_{6}-y_{6}| )
 =   (0.0, 2.7, 0.8, 0.1, 1.6, 0.8) \).

 (2)
  Calculation of the vector ordinal proximity
 \(\overline{D'(x,y)}=(d_{1},d_{2},d_{3},d_{4},d_{5},d_{6})\)
  can be based on
  rule:
%
    \[d_{i} = \left\{ \begin{array}{ll}
               0, & \mbox{if $  ~0.0 \leq |x_{i}-y_{i}| \leq 0.2, $}\\
               1, & \mbox{if $  ~0.2 < |x_{i}-y_{i}| \leq 0.5, $}\\
               2, & \mbox{if $ ~0.5 < |x_{i}-y_{i}| \leq 0.8, $}\\
               3, & \mbox{if $ ~0.8 < |x_{i}-y_{i}| \leq 3.5, $}\\
              4, & \mbox{if $ ~3.5 < |x_{i}-y_{i}| \leq 5.0. $}
               \end{array}
               \right. \]
 The resultant vector ordinal proximity is (based on \(\overline{D(x,y)}\) ):~~
  \(\overline{D'(x,y)} = (0, 3, 2, 0, 3, 2) \).

~~

  {\bf Example 2.3.} In the case when
  quantitative vector estimates are transformed
  into ordinal vector estimates, the example 1 is obtained.

~~

 {\bf Case 2.} Object-cluster (subset) (Fig. 2.21b).
 The following initial information is considered:

 (i) object \(x\in A\): \(\overline{x} = (x_{1},...,x_{i},...,x_{m})\)
  (vector estimate);

 (ii) elements of cluster
 \(Y = \{y^{1},...,y^{\zeta},...,y^{\eta}\} \subset A\),
  \( y^{\zeta}\):
  \(\overline{y^{\zeta}} = (y^{\zeta}_{1},...,y^{\zeta}_{i},...,y^{\zeta}_{m}) \)
 (vector estimate)
  (\(\forall y^{\zeta}\in Y\)).

 Proximity
 \(D(x,Y)\)
 between object \(x\) and cluster \(Y\) can be
 considered, for example, as the following:

 (1)~ \(D^{min}(x,Y) = \min_{\forall y^{\zeta} \in Y} ~D(x,y^{\zeta})\);

 (2)~ \(D^{max}(x,Y) = \max_{\forall y^{\zeta} \in Y} ~D(x,y^{\zeta})\);

 (3)~ \(D^{av}(x,Y) =1/\eta ~\sum_{\forall y^{\zeta}\in Y} ~D(x,y^{\zeta})\);

 (4)~ \(D^{cent}(x,Y) =  D(x, \widehat{y} )\),
 where \(\widehat{y}\) is centroid (or median point)  of cluster \(Y\).
 Clearly, various measurement approaches (i.e., metric/proximity)
 described  for case 1 can be used.

~~

 {\bf Example 2.4.} Let us consider item
 \(x\): \(\overline{x} = (0.3,3.5,1.4,1.5 ) \)
 and cluster \(Y\):
 \(\overline{y^{1}} = (1.1,4.0,3.2,4.3 ) \),
 \(\overline{y^{2}} = (2.0,5.1,2.5,5.2 ) \),
 \(\overline{y^{3}} = (1.3,4.7,4.2,1.6 ) \).

 For four case above,
 the following proximities are obtained
 (Euclidean distance of two items is used):

 (1) \(D^{min}(x,Y) = \min_{\forall y^{\zeta} \in Y} ~D(x,y^{\zeta})\) =
 \(\min \{ D(x,y^{1}), D(x,y^{2}),D(x,y^{3})) = \min \{ 3.4, 4.5, 3.3)\}
 = 3.3\);

 (2) \(D^{max}(x,Y) = \max_{\forall y^{\zeta} \in Y}
 ~D(x,y^{\zeta})\)=
 \(\max \{ D(x,y^{1}), D(x,y^{2}),D(x,y^{3})) = \max \{ 3.4, 4.5, 3.3)\}
 = 4.5\);

 (3) \(D^{av}(x,Y)= \frac{1}{3} ~\sum_{\forall y^{\zeta}\in Y} ~D(x,y^{\zeta})\)=
 \( \frac{1}{3}  ( D(x,y^{1}) + D(x,y^{2}) + D(x,y^{3}))\) =
  \( 1/3  (3.4 + 4.5 + 3.3 ) = 3.7\);

 (4) \(D^{cent}(x,Y) =  D(x, \widehat{y} )\) =
 \( D( (0.3 , 3.5 , 1.4 , 1.5 ),( 1.16, 1.1, 1.9, 2.2 ))  = 2.7\).

~~

 In the case of vector proximity
 \(\overline{D}(x,y^{\zeta})\)
%
%
 the following
  calculation schemes can be used:

 {\it Scheme 1.} Preliminary transformation of vector proximities
 \(D(x,y^{\zeta})\) (\(\forall y^{\zeta}\))
 into  ordinal vector proximity
 and usage of the
 measurement methods above
 (e.g., proximity to the closest point of cluster);

 {\it Scheme 2.} Integration of proximities \(D(x,y^{\zeta})\) (\(\forall y^{\zeta}\))
 to obtain a general vector proximity \(D(x,Y)\)
 and transformation the general vector proximity into
  a number (quantitative or ordinal).

~~

 {\bf Example 2.5.} Numerical examples for the above-mentioned two cases (i) and (ii)
 are the following:

 {\it Scheme 1}:
 \(\overline{D}(x,y^{\zeta}) =
 (|x_{1}-y^{\zeta}_{1}|,|x_{2}-y^{\zeta}_{2}|,|x_{3}-y^{\zeta}_{3}|,|x_{4}-y^{\zeta}_{4}| )\).

 Thus,
 \(\overline{D}(x,y^{1}) = (0.8,0.5,1.8,2.8) \),
 \(\overline{D}(x,y^{2}) = (1.7,1.6,1.1,3.7) \),
 \(\overline{D}(x,y^{3}) = (1.0,1.2,2.8,0.1) \).

  Transformed vector proximities (into ordinal vector estimates, rules from example 2) are:

  \(\overline{D'}(x,y^{1}) = (2,1,3,3) \),
  \(\overline{D'}(x,y^{2}) = (3,3,3,4) \),
  \(\overline{D'}(x,y^{3}) = (3,3,3,0) \).

 Evidently,
 point  \(y^{1}\) can be considered as the closest point
 for item \(x\)
 (version of proximity between point \(x\) and cluster \(Y\)
 as proximity of
 point \(x\) and the closest point of the cluster \(Y\)),
  i.e.,
 \( \overline{D'}(x,Y) =  \overline{D'}(x,y^{1}) = (2,1,3,3) \).

 {\it Scheme 2}:
 Let us integrated general vector proximity is:
 \(D(x,Y) = \overline{D'}(x,Y) = (2,1,3,3)\)
 (from scheme 1 above).
 A transformation process of the ordinal vector into an ordinal estimate
 can be based on the following simplified rule
 (for ordinal vector \((\beta_{1},...,\beta_{i},...,\beta_{m}  ) \)):
    \[\alpha' = \left\{ \begin{array}{ll}
               0, & \mbox{if $  ~0 \leq  ~\sum_{i=\overline{1,m}}  \beta_{i}~ \leq m, $}\\
               1, & \mbox{if $  ~m < ~\sum_{i=\overline{1,m}}  \beta_{i}~ \leq 2m, $}\\
               2, & \mbox{if $ ~2m < ~\sum_{i=\overline{1,m}} \beta_{i}~ \leq 3m, $}\\
               3, & \mbox{if $ ~3m < ~\sum_{i=\overline{1,m}} \beta_{i}~ \leq 4m, $}\\
              4, & \mbox{if $ ~4m < ~\sum_{i=\overline{1,m}} \beta_{i}~.  $}
               \end{array}
               \right. \]
 Here, \(m=4\) and the resultant ordinal estimate  for \(\overline{D'}(x,Y)\) is:
 \(2 \).
 In general, it may be reasonable to use a multiset estimate,
  for example,
 \(e(x,Y) = (0,1,1,2 )\)
 because in  \(\overline{D'}(x,Y) = (2,1,3,3)\):~
 \(0\) estimates at the level \(0\),
 \(1\) estimate at the level \(1\),
 \(1\) estimate at the level \(2\),
 \(2\) estimate at the level \(3\).

~~

 {\bf Case 3.} Intra-cluster proximity (distance)
 (for all element pair in cluster).
  The following initial information is considered:
%
  ~elements of cluster
 \(Y = \{y^{1},...,y^{\zeta},...,y^{\eta}\} \subset A\),
  \( y^{\zeta}\):
  \(\overline{y^{\zeta}} = (y^{\zeta}_{1},...,y^{\zeta}_{i},...,y^{\zeta}_{m}) \)
 (vector estimate)
  (\(\forall y^{\zeta}\in Y\)).

 Intra-cluster proximity for cluster \(Y\):~
 \(I(Y), ~Y\subseteq A \)
  can be considered, for example, as the following:

 (1) minimum distance
  ~\(I^{intra,min}(Y)=\min_{\zeta=\overline{1,\eta},\xi=\overline{1,\eta},\zeta \neq
 \xi} ~D(y^{\zeta},y^{\xi})\);

 (2) maximum distance
 ~\(I^{intra,max}(Y)=\max_{\zeta=\overline{1,\eta},\xi=\overline{1,\eta},\zeta \neq
 \xi} ~D(y^{\zeta},y^{\xi})\);

 (3) average distance
 \(I^{intra,av}(Y) =  \frac{1}{\eta} ~\sum_{\zeta=\overline{1,\eta},
 \xi=\overline{1,\eta}, \zeta \neq \xi } ~D(y^{\zeta},y^{\xi})\).

 Various measurement approaches (i.e., metric/proximity)
 from case 1 can be used as well.

~~

 {\bf Example 2.6.} Let us consider cluster
 \(Y = \{y^{1},y^{2},y^{3}\}\) (from example 2.4).

 Euclidean distances between cluster elements are:
  ~\(D(y^{1},y^{2}) = 1.8\),
  ~\(D(y^{1},y^{3}) = 3.0\),
  ~\(D(y^{2},y^{3}) = 4.0\).

 Thus, versions of intra-cluster proximities for cluster \(Y\) are:

   \(I^{intra,min}(Y) = 1.8\),
  ~\(I^{intra,max}(Y) = 4.0\),
    ~\(I^{intra,av}(Y) = 2.9\).

 On the other hand, element distances

 \(\overline{D}(y^{1},y^{2}) = (0.9, 1.1, 0.7, 0.9)\),
 ~\(\overline{D}(y^{1},y^{3}) = (0.2,0.7,1.0,2.7)\),
 ~\(\overline{D}(y^{2},y^{3}) = (0.7,0.4,1.7,3.6)\)

 can be transformed into
 ordinal vector proximities (by rule above)

 \(\overline{D'}(y^{1},y^{2}) = (3,3,2,3)\),
 ~\(\overline{D'}(y^{1},y^{3}) = (0,2,3,3)\),
 ~\(\overline{D'}(y^{2},y^{3}) = (2,1,3,4)\).

 After transformation of ordinal vector proximities into
 integrated ordinal estimates,
 the following estimates are obtained (by rule above):
 ~\(\overline{D'}(y^{1},y^{2}) \Longrightarrow 2\),
 ~\(\overline{D'}(y^{1},y^{3}) \Longrightarrow 1\),
 ~\(\overline{D'}(y^{2},y^{3}) \Longrightarrow 2\).
 Here, minimum ordinal intra-cluster proximity equals \(1\),
 maximum ordinal intra-cluster proximity equals \(2\).

 Transformation of
  ordinal vector proximities into multiset estimates leads to the
  following estimates:
 ~\(e(y^{1},y^{2}) = (0,1,1,3,0)\),
 ~\(e(y^{1},y^{3}) = (1,0,1,2,0)\),
 ~\(e(y^{2},y^{3}) = (0,1,1,1,1)\).
 Note, the average multiset estimate can be computer as
 a median estimate
  \cite{lev12multiset,lev15}.

 ~~

 {\bf Case 4.} Inter-cluster proximity (distance)
 (for two clusters, Fig. 2.21c).
 The following initial information is considered:
%
%
 ~(i) elements of cluster
 \(X = \{x^{1},...,x^{\xi},...,x^{\phi}\} \subset A\),
  \( x^{\xi}\):
  \(\overline{x^{\xi}} = (x^{\xi}_{1},...,x^{\xi}_{i},...,x^{\xi}_{m}) \)
 (vector estimate)
  (\(\forall x^{\xi}\in X\));
 ~(ii) elements of cluster
 \(Y = \{y^{1},...,y^{\zeta},...,y^{\eta}\} \subset A\),
  \( y^{\zeta}\):
  \(\overline{y^{\zeta}} = (y^{\zeta}_{1},...,y^{\zeta}_{i},...,y^{\zeta}_{m}) \)
 (vector estimate)
  (\(\forall y^{\zeta}\in Y\)).
 In the main, the following basic types of inter-cluster distances
 are considered:
 (a)  minimum element proximity (distance) (``single link''),
 (b) maximum element proximity (distance) (``complete link''),
 (c) average element proximity (distance),
 (d) median element proximity (distance),
 (e) proximity (distance) between cluster centroids.
 For example, inter-cluster proximity for two clusters \(X,Y\):
 \(I^{inter}(X,Y),~ X,Y\subseteq A \)
  can be considered as the following:

  (1) \(I^{inter,min}(X,Y) = \min_{\xi=\overline{1,\phi},\zeta=\overline{1,\eta}} ~D(x^{\xi},y^{\zeta})\);

  (2) \(I^{inter,max}(X,Y) = \max_{\xi=\overline{1,\phi},\zeta=\overline{1,\eta}} ~D(x^{\xi},y^{\zeta})\);

  (3) \(I^{inter,av}(X,Y) =  \frac{1}{\phi \times \eta}
  ~\sum_{\xi=\overline{1,\phi},\zeta=\overline{1,\eta}}
  ~D(x^{\xi},y^{\zeta})\).

 Clearly, various measurement approaches (i.e., metric/proximity)
 described  for case 1 can be used.
 In the case of vector proximity \(D(x^{\xi},y^{\zeta})\), the following
  calculation scheme can be used:

  (a) preliminary transformation of vector proximity
 \(D(x^{\xi},y^{\zeta})\) (\(\forall x^{\xi}, y^{\zeta}\))
 into a number (quantitative or ordinal)
 and usage of the above-mentioned measurement methods;

 (b) integration of proximities \(D(x^{\xi},y^{\zeta})\) (\(\forall x^{\xi}, y^{\zeta}\))
 to obtain a general vector proximity \(D(X,Y)\)
 and transformation the general vector proximity into
  a number (quantitative or ordinal).

~~

{\bf Example 2.7.} Let us consider cluster
 \(Y = \{y^{1},y^{2},y^{3}\}\) (from example 2.4)
 and cluster
  and cluster \(X\):
 \(\overline{x^{1}} = (0.1, 1.0, 0.2, 0.3 ) \),
 \(\overline{x^{2}} = (0.3, 0.9, 0.5, 0.6 ) \).
%
 Distances between elements of \(X\) and \(Y\)
 are presented
 in Table 2.5 (\(\overline{D}(x^{\xi},y^{\zeta})\)),
 Table 2.6 (Euclidean distance \(D(x^{\xi},y^{\zeta})\)),
 and Table 2.7 (ordinal distance  \(\overline{D'}(x^{\xi},y^{\zeta})\), by rule above).

\begin{center}
{\bf Table 2.5.} Vector distance \(\overline{D}(x^{\xi},y^{\zeta})\)   \\
\begin{tabular}{| l | c | c |c|}
\hline
   & \(y^{1}\) & \(y^{2}\)& \(y^{3}\) \\
\hline
 \(x^{1}\)&\((1.0,3.0,3.0,4.0)\)&\((1.9,4.1,2.3,4.9)\)&\((1.2,3.7,4.0,1.3)\)\\
 \(x^{2}\)&\((0.8,3.1,2.7,3.7)\)&\((1.7,4.2,2.0,4.6)\)&\((1.0,3.8,3.7,1.0)\)\\
\hline
\end{tabular}
\end{center}

\begin{center}
{\bf Table 2.6.} Euclidean distance \(D(x^{\xi},y^{\zeta})\)   \\
\begin{tabular}{| l | c | c |c|}
\hline
   & \(y^{1}\) & \(y^{2}\)& \(y^{3}\) \\
\hline
 \(x^{1}\) & \(6.0\)  &  \(7.0\)  &\(5.8 \) \\
 \(x^{2}\) & \(5.7\)  &  \(6.8\)  &\(5.4 \) \\
\hline
\end{tabular}
\end{center}

\begin{center}
{\bf Table 2.7.} Ordinal vector distance \(\overline{D'}(x^{\xi},y^{\zeta})\)   \\
\begin{tabular}{| l | c | c |c|}
\hline
   & \(y^{1}\) & \(y^{2}\)& \(y^{3}\) \\
\hline
 \(x^{1}\) & \((3,3,3,4)\)  &  \((3,4,3,4)\)  &\((3,4,4,3) \) \\
 \(x^{2}\) & \((2,3,3,4)\)  &  \((3,4,3,4)\)  &\((3,4,4,3) \) \\

\hline
\end{tabular}
\end{center}

 Thus, versions of inter-cluster proximities for clusters
 \(X\) and \(Y\) are:

   \(I^{inter,min}(X,Y) = 5.4\),
  ~\(I^{inter,max}(X,Y) = 7.0\),
    ~\(I^{inter,av}(X,Y) = 6.1\).

 Results of
 transformation of ordinal vector proximities into
 integrated ordinal estimates
 (by rule above)
 are presented in Table 2.8.
 Here, minimum ordinal inter-cluster proximity equals \(2\),
 maximum ordinal inter-cluster proximity equals \(3\).
 Resultant multiset estimates are presented in Table 2.9.


\begin{center}
{\bf Table 2.8.} Integrated ordinal estimates    \\
\begin{tabular}{| l | c | c |c|}
\hline
   & \(y^{1}\) & \(y^{2}\)& \(y^{3}\) \\
\hline
 \(x^{1}\) & \(3\)  &  \(3\)  &\(3 \) \\
 \(x^{2}\) & \(2\)  &  \(3\)  &\(3 \) \\

\hline
\end{tabular}
\end{center}

\begin{center}
{\bf Table 2.9.} Resultant multiset estimates  \(e(x^{\xi},y^{\zeta})\)   \\
\begin{tabular}{| l | c | c |c|}
\hline
   & \(y^{1}\) & \(y^{2}\)& \(y^{3}\) \\
\hline
 \(x^{1}\) & \((0,0,0,3,1)\)  &  \((0,0,0,2,2)\)  &\((0,0,0,2,2) \) \\
 \(x^{2}\) & \((0,0,1,2,1)\)  &  \((0,0,0,2,2)\)  &\((0,0,0,2,2) \) \\

\hline
\end{tabular}
\end{center}

\subsubsection{Quality  of clustering solution}

 Here ``hard'' clustering problem is examined.
 Consider initial  items/elements of
 element set \(A = \{a_{1},...,a_{j},...,a_{n}\}\)
  (Fig. 2.19).
 Two types of initial information for clustering can be examined:
 {\it 1.}
%
%
%
 there are \(m\) parameters/criteria and measurement of \(a\)
 is based on vector estimate
 \( \overline{x}= (x_{1},..., x_{1},...,x_{m})\)
 (Fig. 2.19);
 {\it 2.} binary relation(s) over element set \(A\)
 (including weighted binary relation(s);
 this is a structure over obtained clusters of a graph).
 Note,
 the first type of initial information can be transformed into
 the second type.
%
%
 A clustering solution consists of the following two parts:

 (1) Clusters
 \(\widehat{X} = \{X_{1},...,X_{\iota},...X_{\lambda} \}\),
 i.e. dividing set \(A\) into clusters:
 \( X_{\iota} \subseteq A ~~\forall \iota =\overline{1,\lambda}\);
 ~ \(\eta_{\iota} = | X_{\iota} |\) is the cluster size
 (i.e., cardinality for cluster
 \( X_{\iota}\), \(\iota =\overline{1,\lambda}\)).

 (2) Structure over clusters (if needed).
 Let \(\Gamma ( \widehat{X}) \) be a structure over the clusters
 of the clustering solution \( \widehat{X} \),
 i.e., there exists
  digraph
 \(G = \widehat{X}, \Gamma (\widehat{X})\).
 Let
 \(  \Gamma (X_{\iota}) \) be the structure
  over the  elements of cluster
  \(X_{\iota}\) ~(\(\forall X_{\iota} \in  \widehat{X}\) ).

 The list of basic quality characteristics is the following
 (Table 2.10):

\begin{center}
{\bf Table 2.10.} List of quality characteristics   \\
\begin{tabular}{| c | l |  c | l |}
\hline
 No. & Quality type & Notation & Description  \\
\hline

 I. &Cluster   &\(X_{\iota}\)  & \(1 \leq\iota \leq \lambda\) \\

 1.1.&Intra-cluster distance& \(I^{intra} (X_{\iota}) \)&Proximity between elements of cluster\\


 1.2.&Size of cluster &\(|X_{\iota}|\)& Number of elements in cluster \(X_{\iota}\) \\

 1.3.&Quality of cluster form&&Closeness to predefined form \\
  & &&  (e.g., ball, ellipsoid) (if needed)\\

 1.4.&Size of cluster region  & & Difference between ``max'' and ``min''
  \\
   & &&coordinates (by parameters) \\

 1.5.&Quality of cluster content/structure && Configuration of
 element types (if needed)\\

 1.6.&Quality of cluster structure && Proximity of structure over cluster\\
 &&& elements to predefined structure (if needed\\

 II. & Clustering solution &\(\widehat{X} \) &
 \(\widehat{X}=\{X_{1},..., X_{\iota},...,X_{\lambda}\}\)\\

 2.1.& Total intra-cluster quality &
 \(Q^{intra}(\widehat{X})\)&Integration of intra-cluster
 parameters\\

 &&& (\(I^{intra} (X_{\iota})\), by \(\iota = \overline{1,\lambda}\) ) \\

 2.2.& Total inter-cluster quality &
 \(Q^{inter}(\widehat{X})\)&Integration of inter-cluster
 parameters\\

 &&& (\(I^{inter} (X_{\iota_{1}}, X_{\iota_{1}})\), by \(\iota_{1},\iota_{2}\), \(\iota_{1} \neq \iota_{2}\)) \\

 2.3.& Number of clusters ~(\(\lambda\))&\(Q^{num}(\widehat{X})\)
 &Number of clusters in clustering solution\\

 2.4.&Closeness to cluster size & \(Q^{bal}(\widehat{X})\)& Balance by cluster
 size, \\
  &&& closeness to predefined balance vector \\

 2.5.& Quality by forms of clusters &\(Q^{form}(\widehat{X})\)&
 Integration of cluster form parameters\\

 2.6.& Parameter of cluster regions & \(Q^{reg}(\widehat{X})\)&
   Integration of cluster regions sizes\\
 &&& (by coordinates)\\

 2.7.& ``Correlation clustering functional''&
 \(Q^{corr}(\widehat{X})\)&
 Integration of maximum agreement\\
 &&& (in each cluster) and minimum\\
 &&& disagreements (between clusters)
  \cite{bansal02,bansal04}
 \\

 2.8.&Quality of modularity  &
   \(Q^{mod}(\widehat{X})\)&
    Parameter of network modularity
    \cite{girvan02,new06}
    \\

 III.& Quality of structure over clusters
  &\(Q^{struc}(\widehat{X})\)&
   Closeness to predefined structure\\

  IV.& Multicriteria quality &
    \(\overline{Q}(\widehat{X})\)&
   Integrated vector of quality (e.g., \(\overline{Q}(\widehat{X})=\)\\
  &&&\((Q^{intra}(\widehat{X}),Q^{inter}(\widehat{X}),Q^{bal}(\widehat{X}),Q^{reg}(\widehat{X}) ~)\)
   \\

\hline
\end{tabular}
\end{center}


 {\bf 1.} {\bf Quality of clusters}
 (i.e., local quality parameters in clustering solution):

~~

 {\bf 1.1.} Intra-cluster distance
 (i.e., general proximity between elements in each cluster):

 \(I^{intra} (X_{\iota})\) (\(\iota = \overline{1,\lambda}\)).

 {\it Version 1.} Quantitative parameter as integration of
 quantitative element proximities (distances) in the cluster
 (this is described in previous section, case 3).

 {\it Version 2.} Multiset parameter as integration of ordinal
 estimates of element proximities.
 The approach is illustrated by example.

 ~~

 {\bf Example 2.8.}
 Example for three clusters is depicted
 in Fig. 2.22:
 \(X_{1} = \{1,2,3,4\}\),
 \(X_{2} = \{5,6,7\}\),
 \(X_{3} = \{8,9,10,11,12\}\).
 Ordinal scale \([1,2,3]\) for estimates
 of element similarity is used:

 \(1\) corresponds to ``very similar'',

 \(2\) corresponds to ``medium level'',

 \(3\) corresponds to ``very different''
 (in this case the edge in Fig. 2.22 is absent).

 Ordinal proximities of edges are presented in  Table 2.11.
 The resultant multiset intra-cluster parameters
 for clusters are:
 ~\(I^{intra} (X_{1}) = (2,3,1) \),
 ~\(I^{intra} (X_{2}) = (1,1,1) \),
 ~\(I^{intra} (X_{3}) = (4,2,4) \).

\begin{center}
\begin{picture}(80,30)
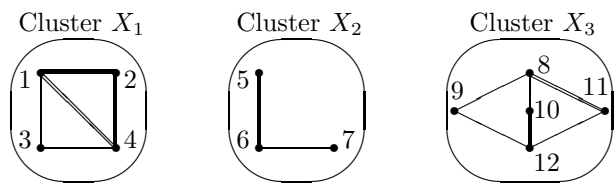
\put(02,00){\makebox(0,0)[bl]{Fig. 2.22. Local intra-cluster
 quality (for cluster)}}


\put(01,25){\makebox(0,0)[bl]{Cluster \(X_{1}\)}}

\put(09,15){\oval(18,19)}


\put(04,10){\circle*{1.0}} \put(14,10){\circle*{1.0}}
\put(04,20){\circle*{1.0}} \put(14,20){\circle*{1.0}}

\put(04,10){\line(0,1){10}} \put(14,10){\line(0,1){10}}
\put(04,10){\line(1,0){10}} \put(04,20){\line(1,0){10}}
\put(14,10.2){\line(-1,1){10}}

\put(04,20.3){\line(1,0){10}} \put(04,19.7){\line(1,-1){10}}
 \put(13.7,10){\line(0,1){10}}

\put(01,18){\makebox(0,0)[bl]{\(1\)}}
\put(01,10){\makebox(0,0)[bl]{\(3\)}}

\put(15,18){\makebox(0,0)[bl]{\(2\)}}
\put(15,10){\makebox(0,0)[bl]{\(4\)}}


\put(30,25){\makebox(0,0)[bl]{Cluster \(X_{2}\)}}

\put(38,15){\oval(18,19)}


\put(33,10){\circle*{1.0}} \put(43,10){\circle*{1.0}}
\put(33,20){\circle*{1.0}}

\put(32.8,10){\line(0,1){10}} \put(33,10){\line(1,0){10}}
\put(33.1,10){\line(0,1){10}}

\put(30,18){\makebox(0,0)[bl]{\(5\)}}
\put(30,10){\makebox(0,0)[bl]{\(6\)}}
\put(44,10){\makebox(0,0)[bl]{\(7\)}}


\put(61,25){\makebox(0,0)[bl]{Cluster \(X_{3}\)}}

\put(69,15){\oval(22,19)}

\put(69,20){\circle*{1.0}} \put(69,10){\circle*{1.0}}
 \put(59,15){\circle*{1.0}}
\put(69,15){\circle*{1.0}}
 \put(79,15){\circle*{1.0}}

\put(69,10){\line(-2,1){10}} \put(69,10){\line(2,1){10}}
\put(68.85,10){\line(0,1){05}} \put(69,15){\line(0,1){05}}

\put(69,20){\line(-2,-1){10}} \put(69,20){\line(2,-1){10}}
\put(69,19.6){\line(2,-1){10}} \put(69.15,10){\line(0,1){05}}

\put(70,20){\makebox(0,0)[bl]{\(8\)}}
\put(58.5,16.5){\makebox(0,0)[bl]{\(9\)}}
\put(69.5,14){\makebox(0,0)[bl]{\(10\)}}

\put(76,17){\makebox(0,0)[bl]{\(11\)}}
\put(69.5,07){\makebox(0,0)[bl]{\(12\)}}

\end{picture}
\end{center}

\begin{center}
{\bf Table 2.11.} Ordinal proximities (intra-cluster, edge \((i_{1},i_{2})\))   \\
\begin{tabular}{| c | c c c c  c  c   c  c c c|}
\hline
 \(i_{1}\) & \(i_{2}:\)& \(2\)&\(3\)&\(4\)&\(6\) &\(7\)&\(9\)&\(10\)&\(11\)&\(12\) \\
\hline
 1 &&\(2\)&\(1\)&\(2\)&&&&&&\\
 2 &&     &\(3\)&\(2\)&&&&&&\\
 3 &&     &     &\(1\)&&&&&&\\
 5 &&&&&\(2\)&\(3\)&&&&\\
 6 &&&&&     &\(1\)&&&&\\
 8 &&&&&&&\(1\)&\(1\) &\(2\)&\(3\) \\
 9 &&&&&&&     &\(3\)&\(3\)&\(1\)\\
 10 &&&&&&&    &     &\(3\)&\(2\)\\
 11 &&&&&&&    &     &     &\(1\)\\
\hline
\end{tabular}
\end{center}

 {\bf 1.2.} Number of elements in cluster
 (or in each cluster, i.e., cluster size)
 corresponds to constraints,
 for example:~
 \(\pi^{-} \leq \)
  \(\eta_{\iota}= |X_{\iota}| \)
 \(\leq \pi^{+}\)
 ~ (\(\pi^{-}, \pi^{+}\)
 are predefined limits of the cluster size)
 (\( \forall X_{\iota} \in \widehat{X}\)).

 The quality parameter corresponds to
 external requirement
 (from the viewpoint of applied problem(s), e.g., teams, communication systems).

~~

 {\bf 1.3.} Quality of cluster form (e.g., body, envelope, cover),
 for example:  sphere/ball, ellipsoid, globe
 (i.e., closeness to the required cluster form).

~~

 {\bf 1.4. } Quality as constraint for size of cluster region
  (limits for interval for coordinates of cluster elements).
 Let us consider cluster
 \(X = \{ x^{1},...,x^{\xi},...,x^{\phi}  \}\),
 parameter estimates
 of each cluster element
 \(x^{\xi}\)
  are
 (vector estimate,
 parameters \(i=\overline{1,m}\)):
  ~\(\overline{x^{\xi}} =
  (x_{1}^{\xi},...,x_{i}^{\xi},...,x_{m}^{\xi})\).
%
 Constraints are (by each parameter
 \(\forall i=\overline{1,\phi}\)) (Fig. 2.23):
  \[|\max_{\xi=\overline{1,\phi}} x_{i}^{\xi}-\min_{\xi=\overline{1,\phi}} x_{i}^{\xi} | \leq  d_{i}, ~~~ \forall i=\overline{1,m}.\]

  The quality parameter corresponds to
  external requirement (from the viewpoint of applied problem(s), e.g., communication systems).

~~

 {\bf 1.5.} Quality of the cluster contents/structure
 (if needed),
 for example (a composite ``team''):
 1 element of the 1st type,
 3 elements of the 2nd type,
 2 elements of the 3rd type,
 1 element of the 4th type.
 Here proximity of the obtained cluster content to the required
 content can be considered.

~~

 {\bf 1.6.} Quality of cluster structure (if needed)
 for cluster
 \(X_{\iota}\)
  (\(~\forall X_{\iota} \in \widehat{X}\) ),
   i.e., proximity
  ~\(\delta (\Gamma (X_{\iota}),\Gamma^{0} (X_{\iota})) \),
  where
  \(\Gamma^{0} (X_{\iota})\) is the predefined structure
  over the cluster elements.

~~

 {\bf 2.} {\bf Total quality for clustering solution}
 (i.e., for cluster set):

~~

  {\bf 2.1.} Total intra-cluster quality for clustering solution
  \(Q^{intra} ( \widehat{X} )\)
  is an integrated measure of intra-cluster parameters
 (\(I^{intra} (X_{\iota})\))
  of all clusters in clustering solution
  (i.e., \(\iota = \overline{1,\lambda}\)).

  {\it Version 1.} Total qualitative quality parameter for qualitative local
  estimates:
%
%
 \[Q^{intra}(\widehat{X})=\frac{1}{\lambda}~
 \sum_{\iota=\overline{1,\lambda}} I^{intra}(X_{\iota}).\]
 Note, integration process can be based on summarization and some
 other operations (maximization, minimization, etc.).

 {\it Version 2.} Total multiset quality parameter
  for multiset local  estimates.
 The approach is illustrated by example.

 ~~

 {\bf Example 2.9.}
 Example for three clusters is depicted
 in Fig. 2.24 (for simplification the cardinality of clusters is the same):
 \(X_{1} = \{1,2,3\}\),
 \(X_{2} = \{4,5,6\}\),
 \(X_{3} = \{7,8,9\}\);
  clustering solution is: \(\widehat{X} =\{X_{1},X_{2},X_{3}\} \).
 Ordinal scale \([1,2,3]\) for estimates
 of element similarity is used:
 \(1\) corresponds to ``very similar'',
 \(2\) corresponds to ``medium level'',
 \(3\) corresponds to ``very different''
 (in this case the corresponding edge is absent).
 Ordinal proximities of edges in clusters are presented in  Table
 2.12.

 The resultant multiset intra-cluster parameters
 for clusters are:

 \(I^{intra} (X_{1}) = (1,2,0) \),
 ~\(I^{intra} (X_{2}) = (2,1,0) \),
 ~\(I^{intra} (X_{3}) = (2,1,0) \).

 These multiset estimates correspond
 to poset (lattice) from Fig. 2.16.

 Integration of the above-mentioned intra-cluster multiset
 estimates can be based on two methods
 (e.g., \cite{lev12multiset,lev15}):

 (a) summarization (by the vector components):~
 \(Q^{intra}(\widehat{X}) = (5,4,0)\),
 the obtained integrate estimate corresponds to
 an extended lattice (not to lattice from Fig. 2.16);

 (b) searching for a median multiset estimate:~
 \(Q^{intra}(\widehat{X}) = (2,1,0)\)
 (lattice from Fig. 2.16).

\begin{center}
\begin{picture}(60,33)

\put(00,00){\makebox(0,0)[bl]{Fig. 2.23. Size of
  cluster region}}

\put(32,18){\makebox(0,0)[bl]{Cluster}}
\put(36,14){\makebox(0,0)[bl]{\(X\)}}

\put(23,16){\oval(16,18)}


\put(17,10){\circle*{1.0}} \put(18,20){\circle*{1.0}}
\put(23,15){\circle*{1.0}} \put(30,16){\circle*{1.0}}
\put(28,09){\circle*{1.0}} \put(28,23){\circle*{1.0}}

\put(17,05){\line(0,1){25}} \put(30,05){\line(0,1){25}}

\put(24,27){\vector(-1,0){7}} \put(24,27){\vector(1,0){6}}

\put(22,28){\makebox(0,0)[bl]{\(d^{i_{2}}\)}}


\put(07,23){\line(1,0){27}} \put(07,09){\line(1,0){27}}
\put(012,18){\vector(0,-1){9}} \put(012,18){\vector(0,1){5}}

\put(07,14){\makebox(0,0)[bl]{\(d^{i_{1}}\)}}

\end{picture}
%
\begin{picture}(80,40)
\put(05.6,00){\makebox(0,0)[bl]{Fig. 2.24. Intra- and
 inter-cluster qualities}}


\put(01,25){\makebox(0,0)[bl]{Cluster \(X_{1}\)}}

\put(09,15){\oval(18,19)}


\put(14,10){\circle*{1.0}} \put(04,20){\circle*{1.0}}
\put(14,20){\circle*{1.0}}

\put(14,10){\line(0,1){10}} \put(04,20){\line(1,0){10}}

\put(14,10.2){\line(-1,1){10}} \put(04,20.3){\line(1,0){10}}
\put(04,19.7){\line(1,-1){10}} \put(04,19.8){\line(1,0){10}}

\put(01,18){\makebox(0,0)[bl]{\(1\)}}

\put(11.3,16.5){\makebox(0,0)[bl]{\(2\)}}
\put(11,08){\makebox(0,0)[bl]{\(3\)}}

\put(14,19.8){\line(1,0){20}} \put(14,20.2){\line(1,0){20}}

\put(14,19.75){\line(2,1){20}} \put(14,20.25){\line(2,1){20}}


\put(31,35){\makebox(0,0)[bl]{Cluster \(X_{2}\)}}

\put(39,25){\oval(18,19)}


\put(34,20){\circle*{1.0}} \put(44,20){\circle*{1.0}}
\put(34,30){\circle*{1.0}}

\put(33.8,20){\line(0,1){10}} \put(34.2,20){\line(0,1){10}}

\put(34,30){\line(1,-1){10}} \put(34,20){\line(1,0){10}}

\put(35.5,29){\makebox(0,0)[bl]{\(4\)}}
\put(35,21){\makebox(0,0)[bl]{\(5\)}}
\put(44,21){\makebox(0,0)[bl]{\(6\)}}

\put(64,19.8){\line(-1,0){20}} \put(64,20.2){\line(-1,0){20}}

\put(64,9.75){\line(-2,1){20}} \put(64,10.25){\line(-2,1){20}}

\put(64,9.8){\line(-1,0){50}} \put(64,10.2){\line(-1,0){50}}

\put(61,25){\makebox(0,0)[bl]{Cluster \(X_{3}\)}}

\put(69,15){\oval(18,19)}


\put(64,10){\circle*{1.0}} \put(64,20){\circle*{1.0}}

\put(74,20){\circle*{1.0}}

\put(64,10){\line(0,1){10}} \put(64,20){\line(1,0){10}}

\put(64,20.3){\line(1,0){10}} \put(64,19.8){\line(1,0){10}}

\put(64,10){\line(1,1){10}}

\put(65,16.6){\makebox(0,0)[bl]{\(7\)}}

\put(75,18){\makebox(0,0)[bl]{\(8\)}}
\put(66,09){\makebox(0,0)[bl]{\(9\)}}

\end{picture}
\end{center}

\begin{center}
{\bf Table 2.12.} Ordinal proximities (intra-cluster, edge \((i_{1},i_{2})\))   \\
\begin{tabular}{| l | c c cc  c  c     c |}
\hline
 \(i_{1}\) & \(i_{2}:\)& \(2\)&\(3\)&\(5\) &\(6\)&\(8\)&\(9\) \\
\hline
 1 &&\(2\)&\(2\)&&&&\\
 2 &&     &\(1\)&&&&\\
 4 &&&&\(2\)&\(1\)&&\\
 5 &&&&     &\(1\)&&\\
 7 &&&&&&\(2\)&\(1\)\\
 8 &&&&&&     &\(1\)\\
\hline
\end{tabular}
\end{center}

  {\bf 2.2.} Total inter-cluster quality for clustering solution
  \(Q^{inter} ( \widehat{X} )\)
  is an integrated measure of inter-cluster parameters
 (\(I^{intra} (X_{\iota_{1}}, X_{\iota_{2}})\))
  of all cluster pairs in clustering solution
  (i.e., \(\iota_{1} = \overline{1,\lambda}\), \(\iota_{2} = \overline{1,\lambda}\),
 \(\iota_{1} \neq \iota_{1} \)).

  {\it Version 1.} Total qualitative quality parameter for qualitative local
  estimates as integration of
 all qualitative two-cluster inter-cluster proximities/distances
 (from previous section, case 4):
%
%
 \[Q^{inter} (\widehat{X})= \frac{1}{\lambda (\lambda-1)} ~
 \sum_{\iota_{1}=\overline{1,\lambda},\iota_{2}=\overline{1,\lambda},\iota_{1}\neq \iota_{2}  }
 I^{inter}(X_{\iota_{1}},X_{\iota_{2}}).\]
 Note, integration process can be based on summarization and some
 other operations (maximization, minimization, etc.).

 {\it Version 2.} Total multiset quality parameter
  for multiset local  estimates.
  The approach is illustrated by example.

 ~~

 {\bf Example 2.10.} The example is based on data from previous
 example 2.9 (i.e., Fig. 2.24).
  Table 2.13 contains inter-cluster ordinal proximities.
%

\begin{center}
{\bf Table 2.13.} Ordinal proximities (inter-cluster, edge \((i,j)\))   \\
\begin{tabular}{| l | c c   c  c   c  c   c |}
\hline
 \(i\) & \(j:\)& \(4\)&\(5\) &\(6\)&\(7\)&\(8\)&\(9\) \\
\hline
 1 &&\(3\)&\(3\)&\(3\)&\(3\)&\(3\)&\(3\)\\
 2 &&\(2\)&\(2\)&\(3\)&\(3\)&\(3\)&\(3\)\\
 3 &&\(3\)&\(3\)&\(3\)&\(3\)&\(3\)&\(2\)\\
 4 &&&&&\(3\)&\(3\)&\(3\)\\
 5 &&&&&\(3\)&\(3\)&\(3\)\\
 6 &&&&&\(2\)&\(3\)&\(2\)\\
\hline
\end{tabular}
\end{center}

 Inter-cluster multiset estimates are:

 \(I^{inter} (X_{1},X_{2})= (0,2,7)\),
 ~\(I^{inter} (X_{1},X_{3})= (0,1,8)\),
 ~\(I^{inter} (X_{2},X_{3})= (0,2,7)\).

 Integration of the above-mentioned intra-cluster multiset
 estimates can be based on two methods
 (e.g., \cite{lev12multiset,lev15}):

 (a) summarization (by the vector components):~
 \(Q^{inter}(\widehat{X}) = (0,5,22)\);

 (b) searching for a median multiset estimate:~
 \(Q^{inter}(\widehat{X}) = (0,2,7)\).

 ~~

 {\bf 2.3.} Total number of  clusters
 in clustering solution \(Q^{num}(\widehat{X})\),
 for example:~~
  \(\Upsilon^{-} \leq \)
  \(\lambda(\widehat{X} )  \)
   \(\leq \Upsilon^{+}\)
 ~~(\(\Upsilon^{-}, \Upsilon^{+}\) are predefined limits of the total cluster number).
 The quality parameter corresponds to
  external requirement (from the viewpoint of applied problem(s)).
 This is connected to {\it 1.2}

~~

  {\bf 2.4.} Closeness of element cluster sizes
 in clustering solution
 to the predefined cluster size constraints,
  i.e., balance (or imbalance) of cluster cardinalities
 \(Q^{bal} (\widehat{X})\),
 for example:~
 \(\pi^{-} \leq \)  \(   | X_{\iota}) |  \)  \(\leq \pi^{+}\)
 (\(\pi^{-}, \pi^{+}\)
 are general limits of each cluster size.
%
 Evidently, here the balance/imbalance
 (i.e., out-of-balance)
 estimate of a clustering solution
 can be consider as the number of clusters that
 corresponds (or does not correspond)
 to the constraints.
  The estimates can be examined as
 a vector-like estimate or a multiset estimate,
%
%
 for example:
 the number of ``good'' clusters
  (with ``good/right'' cluster size),
 the number of quasi-good clusters
 (with quasi-right cluster size), and
 the number of other clusters.
 This parameter corresponds to external requirement
 (from the viewpoint of applied problem(s)).
 Now let us describe the version of the vector-like estimate.
 The notations are as follows:
 ~(a) \(\pi^{0}(\widehat{X})\) is the  number of clusters in \(\widehat{X}\)
 in which the cluster size
 \( X_{\iota}\)  complies with the predefined limits,
 ~(b) \(\pi^{+l}(\widehat{X})\) is the  number of clusters in \( \widehat{X}\) where
 the cluster size
 \( X_{\iota}\)
 more then \(\widehat{\pi}^{+} \) (upper limit) by
 \(l\) elements,
  ~(c) \(\pi^{-l}(\widehat{X})\)
  be the  number of clusters in \(\widehat{X}\) where
 the cluster size
 \( X_{\iota}\)
 less then \(\widehat{\pi}^{-}\) (bottom limit) by
 \(l\) elements.
 As a result,
 the following vector estimate can be considered:~~

 \[Q^{bal} (\widehat{X}) =
 (\pi^{l^{-}}(\widehat{X}),...,\pi^{-1}(\widehat{X}),\pi^{0}(\widehat{X}),\pi^{1}(\widehat{X}),...,\pi^{l^{+}}(\widehat{X})).\]

 Note,
 a close type of the vector estimate (vector proximity) has been suggested for comparison of
 rankings in \cite{lev98}.
  The approach is illustrated by example.

~~

{\bf Example 2.11.}
 Initial set of objects is:
 \(A = \{1,2,3,4,5,6,7,8,9,10,11,12,13,14,15,16,17\}\),
 clustering solution is:~~
 \(\widehat{X}\):
 \(X_{1} = \{1,5,7\}\),
 \(X_{2} = \{2\}\),
 \(X_{3} = \{3,6,10,13,17\}\),
 \(X_{4} = \{11,12\}\),
 \(X_{5} = \{4,12,14,15\}\),
 \(X_{6} = \{8,16\}\).
 The following constrains for cluster size are considered:~
 \(\widehat{\pi}_{1} = 2\),  \(\widehat{\pi}_{2} = 3\).
 Vector estimate for balance of cluster cardinalities is:~~
 \(Q^{bal}(\widehat{X}) =\)
 \((\pi^{-1}(\widehat{X}),\pi^{0}(\widehat{X}),\pi^{1}(\widehat{X}),...,\pi^{2}(\widehat{X}))=\)
 \((  1,3,1,1 )\).

~~~

 The considered approach is close
 to \(\Upsilon\)-balanced partitioning
 (clustering solution \(\widehat{X}\))
  when size of each obtained cluster
 \(| X_{\iota}| \approx \frac{n}{\Upsilon (\widehat{X})}  \)
 (\(\forall  X_{\iota} \in \widehat{X} \))
 where \(\Upsilon( \widehat{X})\) (i.e., \(\lambda\))
 is the number of obtained clusters.


~~

 {\bf 2.5.} Total quality for balance (or imbalance) of cluster forms
 (i.e., cluster bodies/covers) in a clustering solution
  \(Q^{form} (\widehat{X})\),
 for example:
 majority of clusters of a clustering solution
 have the same (or about the same) bodies
 (e.g., spheres/balls, ellipsoids, globes).

 Evidently, it is possible to consider a measure of imbalance,
 analogically as in parameter {\it 2.3}.

~~

{\bf 2.6. } Total quality \(Q^{reg} (\widehat{X})\)
 as constraints for size of cluster regions
 (limits for interval of cluster element coordinates for each cluster).
 Let us consider cluster
 \(X_{\iota}=\{ x^{\iota,1},...,x^{\iota,\xi},...,x^{\iota,\phi_{\iota}}\}\).
 Parameter estimates
 of each cluster element
 \(x^{\iota,\xi}\)
  are
 (vector estimate,
 parameters \(i=\overline{1,m}\) and clusters
 \(\iota = \overline{1,\lambda}\)):
  ~\(\overline{x^{\iota,\xi}} =
  (x_{1}^{\iota,\xi},...,x_{i}^{\iota,\xi},...,x_{m}^{\iota,\xi})\).
%
 Constraints are (by each parameter
 \(\forall i=\overline{1,\phi_{\iota}}\)) (Fig. 2.23):
  \[|\max_{\xi=\overline{1,\phi_{\iota}}} x_{i}^{\xi}-\min_{\xi=\overline{1,\phi_{\iota}}}
   x_{i}^{\xi} | \leq  d_{i},
    ~~~ \forall i=\overline{1,m} ~~ (each~ coordinate/parameter),
    ~~~ \forall \iota=\overline{1,\lambda} ~~ (each~ cluster).\]
  The quality parameter corresponds to
  external requirement
  (from the viewpoint of applied problem(s), e.g., communication systems).

~~

 {\bf 2.7.}
 The ``correlation clustering functional''
 to maximize the intra-cluster
 agreement (attraction) and the inter-cluster
 disagreement (repulsion)
 has been proposed in
 \cite{bansal02,bansal04}~
  (\(Q^{corr} (\widehat{X})\)).
 Here,
 partitioning a fully connected labeled graph
 is examined
 (label ``+'' corresponds to edge between similar vertices,
 label ``-'' corresponds to edge between different vertices).
 The optimization functional  \(Q^{corr} (\widehat{X})\)
 is an integration (i.e., summarization) of two components:
  (i)
    the maximizing number of ``-'' edges between clusters
     (i.e., minimizing disagreements),
  (b) the number of ``+'' edges insides the clusters
  (i.e., maximizing agreements)
 (e.g.,
 \cite{achtert07,bagon11,bansal02,bansal04,dem06,kri09,swam04,zimek08}).
 Weighted versions of the
   ``correlation clustering functional''
 are considered as well
(e.g.,
 \cite{cgw03,cgw05,dem06}).

~~

 {\bf 2.8.} Modularity of clustering solution
 \(Q^{mod} (\widehat{X})\)
 is defined as follows
 (e.g., \cite{girvan02,new03,new06,new04})
 (Fig. 2.25).
%
%
 Let \(G = (A,E)\) be an initial graph, where
 \(A\) is the set of nodes, \(E\) is the  set of edges.
 Clustering solution for graph \(G\) is:
 \(\widehat{X} = \{X_{1},...,X_{\iota},...,X_{\lambda}\}\).
 Let \(A^{\iota}\) be the set of nodes in cluster \(X_{\iota}\)
 (\(\iota = \overline{1,\lambda}\) ).
 Let \(E^{\iota}\) be the set of internal edges in cluster
\(X_{\iota}\)
 (\(\iota = \overline{1,\lambda}\) ), i.e.,
 all corresponding nodes belong to \(A^{\iota}\).
 Let \(\widetilde{E}^{\iota}\)
 be the set of external edges for cluster \(X_{\iota}\)
 (\(\iota = \overline{1,\lambda}\) ),
 i.e., the only one
  corresponding node belong to \(A^{\iota}\).
 The definitions are illustrated in Fig. 2.25
 for a four cluster solution (cluster \(X_{3}\)).

 Thus, the following parameters for each cluster \(X_{\iota}\)
 can be used:

 (a)
 \(e_{\iota} = \frac{| E^{\iota} |}{|E| }\)
 ( \% edges in module \(\iota\)  ),

 (b)
 \(a_{\iota} = \frac{| \widetilde{E}^{\iota} | + | E^{\iota} |}
 {|E| }\)
( \% edges with at least one end in module \(\iota\)  ).

 Further,
 general modularity of clustering solution for graph \(G\) is:
 \[ Q^{mod}(\widehat{X}) = \sum_{\iota=1}^{\lambda} ~
 (e_{\iota}-(a_{\iota})^{2}).\]
 Clustering problem to maximize the modularity is NP-hard
 \cite{brand08}.
 The approach is illustrated by example.

~~

 {\bf Example 2.12.}
 Let us consider modularity parameters for clustering solution
 from Fig. 2.25.
 Here, \(|E| = 26\).
 Parameters for clusters are:

 (1) \(|E^{1}|=6\), \(|\widetilde{E}^{1}|=4\), \(e_{1}=0.23\), \(a_{1}=0.38\);
%
 (2) \(|E^{2}|=4\), \(|\widetilde{E}^{2}|=4\), \(e_{2}=0.15\), \(a_{2}=0.3\);

 (3) \(|E^{3}|=3\), \(|\widetilde{E}^{3}|=4\), \(e_{3}=0.115\), \(a_{3}=0.27\);
%
 (4) \(|E^{4}|=4\), \(|\widetilde{E}^{4}|=2\), \(e_{4}=0.15\), \(a_{4}=0.23\).

 The resultant modularity parameter for clustering solution is:~~
 \(Q^{mod}(\widehat{X}) =
 (0.23 - 0.14) + (0.15 - 0.09) + (0.115 - 0.073) + (0.15 - 0.053 )   =
 0.09 + 0.06 + 0.42 + 0.097 = 0.667 \).

\begin{center}
\begin{picture}(62,36)
\put(00,00){\makebox(0,0)[bl]{Fig. 2.25. Modularity in graph
 clustering}}



\put(01,16){\makebox(0,0)[bl]{\(X_{1}\)}}

\put(06,10){\oval(12,10)}


\put(06,14){\circle*{1.0}} \put(02,10){\circle*{1.0}}
\put(10,10){\circle*{1.0}} \put(06,06){\circle*{1.0}}
\put(06,06){\circle*{1.0}}


\put(06,06){\line(-1,1){4}} \put(06,06){\line(1,1){4}}
\put(06,06){\line(0,1){8}} \put(02,10){\line(1,0){8}}
\put(06,14){\line(-1,-1){4}} \put(06,14){\line(1,-1){4}}


\put(06,14){\line(1,1){06}} \put(15,20){\line(-1,-2){05}}



\put(12,26){\makebox(0,0)[bl]{\(X_{2}\)}}

\put(16,21.5){\oval(12,07)}

\put(12,20){\circle*{1.0}} \put(20,20){\circle*{1.0}}
\put(16,24){\circle*{1.0}} \put(15,20){\circle*{1.0}}

\put(12,20){\line(1,1){4}} \put(20,20){\line(-1,1){4}}
\put(12,20){\line(1,0){8}}

\put(42,21){\makebox(0,0)[bl]{\(X_{3}\)}}

\put(36,21.5){\oval(10,07)}

\put(27,32){\makebox(0,0)[bl]{Internal edges}}
\put(30,28.4){\makebox(0,0)[bl]{for \(X_{3}\):~\(E^{3}\) }}

\put(34,28){\vector(0,-1){5.6}} \put(37.5,28){\vector(0,-1){4.6}}
\put(38.5,28){\vector(0,-1){6}}


\put(33,13.5){\vector(-1,3){3.3}}

\put(32.5,13.5){\vector(-1,1){5.9}}

\put(32,13.5){\vector(-2,1){4}}

\put(40,13.5){\vector(1,1){4}}

\put(29,11){\makebox(0,0)[bl]{External}}
\put(22,05.6){\makebox(0,0)[bl]{edges for \(X_{3}\):
 \(\widetilde{E}^{3}\)}}

\put(32,20){\circle*{1.0}} \put(40,20){\circle*{1.0}}
\put(36,24){\circle*{1.0}}


\put(32,20){\line(1,1){4}} \put(40,20){\line(-1,1){4}}
\put(32,20){\line(1,0){8}}


\put(36,24){\line(-1,0){20}}

\put(32,20){\line(-1,0){12}}

\put(32,20){\line(-1,-1){6}}

\put(10,10){\line(1,1){4}}

\put(14,14){\line(1,0){12}}


\put(55,17){\makebox(0,0)[bl]{\(X_{4}\)}}

\put(55,12){\oval(12,08)}


\put(52,14){\circle*{1.0}} \put(58,14){\circle*{1.0}}
\put(52,10){\circle*{1.0}} \put(58,10){\circle*{1.0}}

\put(52,10){\line(1,0){6}} \put(52,14){\line(1,0){6}}
\put(52,10){\line(0,1){4}} \put(58,10){\line(0,1){4}}


\put(52,14){\line(-2,1){12}}


\put(52,10){\line(-1,0){42}}

\end{picture}
\end{center}


 {\bf 3.} {\bf Quality of structure over clusters} (e.g., tree, hierarchy)
 (if needed) ( \(Q^{struc}(\widehat{X}) \) ).
 Here a proximity of the obtained structure
 \( \Gamma ( \widehat{X} ) \)
 in clustering solution \(\widehat{X}\)
 and a predefined structure
 \(\Gamma^{0}\)
 is examined:~
 \(Q^{struc}(\widehat{X}) = \delta (\Gamma (\widehat{X}),\Gamma^{0})\).
 Clearly, various scales for assessment of the proximities can be
 used
 (e.g., qualitative, ordinal, vector-like,
 multiset)
(e.g., \cite{lev11agg,lev15}).

~~

 {\bf 4.} Generally, it is reasonable to consider
 multicriteria quality of clustering solutions that integrates
 the above-mentioned clustering characteristics, for example:~~
 \(Q(\widehat{X}) = (
  Q^{inter}(\widehat{X}),
  Q^{intra}(\widehat{X}),
  \pi(\widehat{X}) )
    \).

~~

 Evidently, general integrated vector-like estimate
 for assessment of total quality of clustering solution
  can be examined, for  example:~
 \( \overline{Q}(\widehat{X})= ( Q(\widehat{X}),\delta (\Gamma (\widehat{X}),\Gamma^{0})\).
 As a result,
  the clustering problem can be formulated
 as generalized multicriteria optimization problem
  (i.e., Pareto-efficient solutions have to be searched for),
  for example:
%
%
  \[
  \min ~ Q^{intra}(\widehat{X}), ~~~
  \max ~ Q^{inter}(\widehat{X}), ~~~~~~~
  s.t. ~~~ Q^{bal}(\widehat{X})  \preceq \pi^{0},  ~~~
  Q^{struc}=\delta (\Gamma (\widehat{X}),\Gamma^{0}) \leq \delta^{0}.\]
%
%

 In the case of multiset estimates,
 the multiple criteria optimization clustering problem can be considered
 on the basis of quality lattices (poset-like scales) as follows
 (i.e., Pareto-efficient solutions over posets have to be searched for):
  \[\min ~ Q^{intra}(\widehat{X}), ~~~~ (by~lattice, ~Fig.~2.26a)\]
  \[\max ~ Q^{inter}(\widehat{X}), ~~~~ (by~lattice, ~Fig.~2.26b) \]
  \[s.t. ~~~  I^{intra}(X_{\iota}) \succeq I^{0},~~ \forall \iota = \overline{1,\lambda},
  ~~~ I^{0}~is~reference~multiset~estimate~\forall X_{\iota}
  ~~ (by~lattice,~Fig.~2.26c),\]
  \[Q^{bal}(\widehat{X})  \preceq \pi^{0},
  ~~~  \pi^{0}~is~reference~multiset~estimate
  ~~ (balance~by~cluster~size,~by~lattice, ~Fig.~2.26d),\]
 \[Q^{struc}=\delta (\Gamma (\widehat{X}),\Gamma^{0}) \leq \delta^{0},
 ~~~~ (closeness~to~predefined~general~structure~ \Gamma^{0}).\]

 Thus, Fig. 2.26 illustrates the integrated
 ``discrete space'' (poset)
 for total multiset based vector quality of
 clustering solution \(\widehat{X} \).

\begin{center}
\begin{picture}(37,56)

\put(36,0){\makebox(0,0)[bl]{Fig. 2.26. Illustration for quality
 posets (lattices)}}

\put(00,09){\makebox(0,0)[bl]{(a) total intra-cluster}}
\put(05,05){\makebox(0,0)[bl]{quality
 \(Q^{intra}(\widehat{X})\)}}

\put(10,14){\line(0,1){40}} \put(10,14){\line(3,4){15}}
\put(10,54){\line(3,-4){15}}

\put(10,14){\circle*{1.0}} \put(10,14){\circle{1.9}}

\put(24.5,23.4){\makebox(0,0)[bl]{Ideal points}}

\put(28,23){\vector(-2,-1){016}} \put(38,27){\vector(1,3){08.5}}

\end{picture}
%
\begin{picture}(37,56)

\put(00,09){\makebox(0,0)[bl]{(b) total inter-cluster}}
\put(05,05){\makebox(0,0)[bl]{quality
 \(Q^{inter}(\widehat{X})\)}}

\put(10,14){\line(0,1){40}} \put(10,14){\line(3,4){15}}
\put(10,54){\line(3,-4){15}}

\put(10,54){\circle*{1.0}} \put(10,54){\circle{2.1}}
\put(10,54){\circle{3}}

\end{picture}
%
\begin{picture}(38,56)

\put(00,09){\makebox(0,0)[bl]{(c) local intra-cluster}}
\put(02,05){\makebox(0,0)[bl]{quality
 \(I^{intra}(X_{\iota}), \forall \iota\)}}

\put(10,14){\line(0,1){40}} \put(10,14){\line(3,4){15}}
\put(10,54){\line(3,-4){15}}

\put(13,32.6){\makebox(0,0)[bl]{\(I^{0}\)}}

\put(14,31){\circle*{1.8}}

\put(14,31){\line(-1,-1){10}} \put(14,31){\line(1,-1){07.5}}

\put(10,14){\circle*{1.0}} \put(10,14){\circle{1.9}}

\put(18,17){\makebox(0,0)[bl]{Ideal}}
\put(18,14){\makebox(0,0)[bl]{point}}

\put(17.5,16.2){\vector(-3,-1){05.7}}

\end{picture}
%
\begin{picture}(34,52)

\put(00,09){\makebox(0,0)[bl]{(d) total balance by}}
\put(02,05){\makebox(0,0)[bl]{cluster sizes
 \(Q^{bal}(\widehat{X})\)}}

\put(10,54){\circle*{1.1}} \put(10,54){\circle{2.2}}

\put(00,47){\makebox(0,0)[bl]{Ideal}}
\put(00,44){\makebox(0,0)[bl]{point}}

\put(04,50){\vector(3,2){04}}

\put(10,14){\line(0,1){40}} \put(10,14){\line(3,4){15}}
\put(10,54){\line(3,-4){15}}

\put(10,14){\circle*{1.0}} \put(10,14){\circle{1.6}}


\put(11.5,34){\makebox(0,0)[bl]{\(\pi^{0}\)}}

\put(12.5,38){\circle*{1.8}}

\put(12.5,38){\line(-3,1){12}} \put(12.5,38){\line(1,1){10}}

\end{picture}
\end{center}

 In the case of ``soft'' clustering problems,
 it is necessary to examine measures of solution ``softness''
 (e.g., total parameter for intersection of clusters).

%
%
%

\subsubsection{Comparison of two clustering solutions}

  Comparing methods for  clustering solutions (e.g., partitions,
  hierarchical clustering solutions)
 have been studied many years
 (e.g.,  \cite{bach06,ben06,fow83,hub85,huss14,meila12,mirkin96,rand71,zhou05}).
 The main methods are the following:
 (a) pair counting methods
 (i.e., how likely two solutions
 group an elements pair together, or, separate
 them in different clusters);
 (b) set matching,
 (c) variation of information.
 Here,
 the following kinds of clustering solution are considered:

 (1) basic clustering solution
 \(\widehat{X}\)
  as a set of clusters
  \(\widehat{X} = \{X_{1},..., X_{\iota},...,X_{\lambda}\}\)
  (i.e., partition of initial elements \(A = \{1,...,i,...,n\}\));

 (2) clustering and order over the cluster set \( \{X_{1},...,
 X_{\iota},...,X_{\lambda}\}\):

 (2.1) clustering and linear order over the cluster set, i.e, ranking \(R\),

 (2.2) hierarchy over the cluster set \(\widehat{H}= <H,\widehat{X}> \),
 where \(H\) is  hierarchy over clusters of basic clustering solution (partition of elements)
  \(\widehat{X}\).

 Evidently, general graph over the cluster set can be examined as
 well.

 Generally, two approaches can be examined for comparison of
 two clustering solutions:

 (a) difference by structures (``structural'' comparison),
 e.g., a cost of transformation:

 \(\widehat{X_{1}}  \Rightarrow  \widehat{X_{2}}\),
 \(R_{1} \Rightarrow R_{2}\),
 \(\widehat{H_{1}} \Rightarrow \widehat{H_{2}}\)).

 (b) difference by quality criterion (or criteria),
 for example:

 \(Q'(\widehat{X_{1}}) - Q'(\widehat{X_{2}})\),
 \(Q''(R_{1}) - Q''(R_{2})\),
 \(Q'''(\widehat{H_{1}}) - Q'''(\widehat{H_{2}})\).

 Here ``structural'' comparison is considered.
 The following comparison cases are examined:

 {\it Case 1.} Proximity of two clustering solutions
 \(\widehat{X}_{1}\) and \(\widehat{X}_{2}\) (Fig. 2.27):~
 \(D(\widehat{X}_{1},\widehat{X}_{2})\).

 {\it Case 2.} Proximity of two rankings
 \(R_{1}\) and \(R_{2}\) (Fig. 2.28):~ \(D(R_{1},R_{2})\).

 {\it Case 3.} Proximity (auxiliary ) of two hierarchies
 \(H_{1}\) and \(H_{2}\) (over clusters, Fig. 2.29):~
 \(D(H_{1},H_{2})\).


 {\it Case 4.} Proximity of two hierarchical clusterings
 \(\widehat{H}_{1}\) and \(\widehat{H}_{2}\)
 (e.g., vector; Fig. 2.30):~

 \(D(\widehat{H}_{1},\widehat{H}_{2}) =
 (D(H_{1},H_{2}), D(\widehat{X}_{1},\widehat{X}_{2})) \).

 Fig. 2.31 depicts an illustrative numerical example for 7 elements:
 \(A = \{1,2,3,4,5,6,7\}\)
 (``hard'' clustering problems).


\begin{center}
\begin{picture}(40,34)

\put(00,00){\makebox(0,0)[bl]{Fig. 2.27. Proximity for
  two clustering solutions}}

\put(16.5,18){\oval(33,20)}

\put(13,25){\circle*{0.7}} \put(17.5,26){\circle*{0.7}}
\put(14.5,22){\circle*{0.7}} \put(4,24){\circle*{0.7}}
\put(27,13){\circle*{0.7}} \put(29,25){\circle*{0.7}}
\put(30,19){\circle*{0.7}}

\put(06.5,18){\oval(08,15.5)}

\put(05,19){\circle*{0.7}} \put(05,13){\circle*{0.7}}
\put(08,21){\circle*{0.7}} \put(09,12){\circle*{0.7}}

\put(07.5,15.5){\circle*{0.7}}

\put(16,24){\oval(10,5)}

\put(26.4,25){\oval(07,3)}

\put(22.5,18.5){\oval(08,6)}

\put(30,18.5){\oval(03,5)} \put(27,13){\oval(04,3)}

\put(24,25){\circle*{0.7}} \put(21.5,16){\circle*{0.7}}
\put(20,23){\circle*{0.7}} \put(25,18){\circle*{0.7}}

\put(22.5,20.5){\circle*{0.7}}


\put(35,18){\vector(1,0){05}} \put(38,18){\vector(-1,0){4}}

\put(37,08.5){\line(0,1){09}}

\put(29,05){\makebox(0,0)[bl]{\(D(\widehat{X}_{1},\widehat{X}_{2})\)}}

\put(00,29){\makebox(0,0)[bl]{Clustering solution
 \(\widehat{X}_{1}\) }}

\end{picture}
%
\begin{picture}(42,34)

\put(16.5,18){\oval(33,20)}

\put(13,25){\circle*{0.7}} \put(17.5,26){\circle*{0.7}}
\put(14.5,22){\circle*{0.7}} \put(4,24){\circle*{0.7}}


\put(27,13){\circle*{0.7}}

\put(29,25){\circle*{0.7}} \put(30,19){\circle*{0.7}}

\put(07.5,16.5){\oval(08,12)}

\put(05,19){\circle*{0.7}} \put(05,13){\circle*{0.7}}
\put(08,21){\circle*{0.7}} \put(09,12){\circle*{0.7}}

\put(07.5,15.5){\circle*{0.7}}

\put(22.5,20.5){\oval(06.5,12)}

\put(29,22){\oval(04,8)}

\put(27,13){\oval(04,3)}

\put(04,24){\oval(04,3)}

\put(15.5,24){\oval(07,07)}

\put(24,25){\circle*{0.7}} \put(21.5,16){\circle*{0.7}}
\put(20,23){\circle*{0.7}} \put(25,18){\circle*{0.7}}


\put(22.5,20.5){\circle*{0.7}}


\put(00,29){\makebox(0,0)[bl]{Clustering solution
 \(\widehat{X}_{2}\) }}

\end{picture}
%
%
\begin{picture}(36,37)
\put(03,00){\makebox(0,0)[bl]{Fig. 2.28. Proximity
 of rankings}}

\put(01,33){\makebox(0,0)[bl]{Ranking
 \(R_{1}\) }}


\put(10,31){\oval(20,02)}
\put(08,31){\circle{1.3}} \put(12,31){\circle{1.3}}
\put(08,31){\circle{0.7}} \put(12,31){\circle{0.7}}
\put(08,31){\circle*{0.3}} \put(12,31){\circle*{0.3}}

\put(10,30){\vector(0,-1){4}}

\put(10,25){\oval(20,02)}

\put(05,25){\circle{1.0}} \put(10,25){\circle{1.0}}
\put(15,25){\circle{1.0}}

\put(05,25){\circle*{0.4}} \put(10,25){\circle*{0.4}}
\put(15,25){\circle*{0.4}}

\put(10,24){\vector(0,-1){4}}


\put(06.5,18){\makebox(0,0)[bl]{{\bf . . .}}}

\put(10,17){\vector(0,-1){4}}

\put(10,12){\oval(20,02)}

\put(04,12){\circle*{1.0}} \put(08,12){\circle*{1.0}}
\put(12,12){\circle*{1.0}} \put(17,12){\circle*{1.0}}

\put(10,11){\vector(0,-1){4}}

\put(10,06){\oval(20,02)}

\put(05,06){\circle*{0.7}} \put(10,06){\circle*{0.7}}
\put(15,06){\circle*{0.7}}


\put(27,15){\vector(1,0){07.5}} \put(28,15){\vector(-1,0){6}}

\put(28.5,15.5){\line(0,1){04}}

\put(20.6,20){\makebox(0,0)[bl]{\(D(R_{1},R_{2})\)}}

\end{picture}
%
\begin{picture}(20,34)

\put(01,33){\makebox(0,0)[bl]{Ranking
 \(R_{2}\) }}


\put(10,31){\oval(20,02)}
\put(08,31){\circle{1.3}} \put(08,31){\circle{0.7}}
\put(08,31){\circle*{0.3}}

\put(12,31){\circle{1.0}} \put(12,31){\circle*{0.4}}

\put(10,30){\vector(0,-1){4}}

\put(10,25){\oval(20,02)}

\put(05,25){\circle{1.0}} \put(05,25){\circle*{0.4}}

\put(10,25){\circle{1.0}} \put(10,25){\circle*{0.4}}

\put(15,25){\circle{1.3}} \put(15,25){\circle{0.7}}
 \put(15,25){\circle*{0.3}}

\put(10,24){\vector(0,-1){4}}


\put(06.5,18){\makebox(0,0)[bl]{{\bf . . .}}}

\put(10,17){\vector(0,-1){4}}

\put(10,12){\oval(20,02)}

\put(05,12){\circle*{1.0}} \put(10,12){\circle*{1.0}}
\put(15,12){\circle*{1.0}}


\put(10,11){\vector(0,-1){4}}

\put(10,06){\oval(20,02)}

\put(04,06){\circle*{0.7}} \put(08,06){\circle*{0.7}}
\put(12,06){\circle*{0.7}}

\put(17,06){\circle*{1.0}}

\end{picture}
\end{center}

\begin{center}
\begin{picture}(37,28)

\put(06,00){\makebox(0,0)[bl]{Fig. 2.29. Proximity
 of  hierarchies
  }}

\put(04.7,24){\makebox(0,0)[bl]{Hierarchy
 \(H_{1}\) }}


\put(14,22){\oval(04,02)}

\put(14,21){\vector(1,-1){6}} \put(14,21){\vector(-1,-1){6}}


\put(08,14){\oval(04,02)}

\put(08,13){\vector(0,-1){6}} \put(08,13){\vector(-1,-1){6}}

\put(20,14){\oval(04,02)}

\put(20,13){\vector(0,-1){6}} \put(20,13){\vector(-1,-1){6}}
\put(20,13){\vector(1,-1){6}}


\put(02,06){\oval(04,02)} \put(08,06){\oval(04,02)}
\put(14,06){\oval(04,02)} \put(20,06){\oval(04,02)}
\put(26,06){\oval(04,02)}


\put(31,13){\vector(1,0){10}} \put(34,13){\vector(-1,0){09}}

\put(33,13.5){\line(0,1){04}}

\put(25,18){\makebox(0,0)[bl]{\(D(H_{1},H_{2})\)}}

\end{picture}
%
\begin{picture}(36,28)

\put(04.7,24){\makebox(0,0)[bl]{Hierarchy
 \(H_{2}\) }}


\put(14,22){\oval(04,02)}

\put(14,21){\vector(1,-1){6}} \put(14,21){\vector(-1,-1){6}}


\put(08,14){\oval(04,02)}

\put(08,13){\vector(0,-1){6}} \put(08,13){\vector(-1,-1){6}}
\put(08,13){\vector(1,-1){6}}

\put(20,14){\oval(04,02)}

\put(20,13){\vector(0,-1){6}}

\put(20,13){\vector(-2,-1){11.5}}

\put(20,13){\vector(1,-1){6}}


\put(02,06){\oval(04,02)} \put(08,06){\oval(04,02)}
\put(14,06){\oval(04,02)} \put(20,06){\oval(04,02)}
\put(26,06){\oval(04,02)}

\end{picture}
\begin{picture}(37,37)

\put(00,00){\makebox(0,0)[bl]{Fig. 2.30. Proximity of
 hierarchical clusterings }}

\put(02.5,28){\makebox(0,0)[bl]{Hierarchical}}
\put(04,24){\makebox(0,0)[bl]{clustering
  \(\widehat{H}_{1}\) }}


\put(14,22){\circle{1.3}} \put(14,22){\circle{0.7}}
\put(14,22){\circle*{0.3}}

\put(14,22){\oval(04,02)}

\put(14,21){\vector(1,-1){6}} \put(14,21){\vector(-1,-1){6}}


\put(08,14){\oval(04,02)}

\put(07,14){\circle{1.0}} \put(07,14){\circle*{0.25}}
\put(09,14){\circle{1.0}} \put(09,14){\circle*{0.25}}

\put(08,13){\vector(0,-1){6}} \put(08,13){\vector(-1,-1){6}}

\put(20,14){\oval(04,02)}

\put(19,14){\circle*{1.3}} \put(21,14){\circle*{1.3}}

\put(20,13){\vector(0,-1){6}} \put(20,13){\vector(-1,-1){6}}
\put(20,13){\vector(1,-1){6}}


\put(02,06){\oval(04,02)}

\put(01,06){\circle*{0.7}} \put(03,06){\circle*{0.7}}


\put(08,06){\oval(04,02)}

\put(08,06){\circle*{1.1}}


\put(14,06){\oval(04,02)}

\put(13,06){\circle{1.3}} \put(13,06){\circle*{0.3}}

\put(15,06){\circle{1.3}} \put(15,06){\circle*{0.3}}


\put(20,06){\oval(04,02)}

\put(20,06){\circle*{0.5}}


\put(26,06){\oval(04,02)}

\put(25,06){\circle*{0.9}} \put(27,06){\circle*{0.9}}


\put(31,13){\vector(1,0){10}} \put(34,13){\vector(-1,0){09}}


\put(33,13.5){\line(0,1){18.5}}
\put(04.5,33){\makebox(0,0)[bl]{\(D(\widehat{H}_{1},\widehat{H}_{2})
  =
 (D(H_{1},H_{2}), D(\widehat{X}_{1},\widehat{X}_{2}))\)}}

\end{picture}
%
\begin{picture}(30,32)

\put(02.5,28){\makebox(0,0)[bl]{Hierarchical}}
\put(04,24){\makebox(0,0)[bl]{clustering
  \(\widehat{H}_{2}\) }}


\put(14,22){\oval(04,02)}

\put(14,21){\vector(1,-1){6}} \put(14,21){\vector(-1,-1){6}}

\put(13,22){\circle{1.3}} \put(13,22){\circle{0.7}}
\put(13,22){\circle*{0.3}}


\put(15,22){\circle*{1.3}}


\put(08,14){\oval(04,02)}

\put(07,14){\circle{1.0}} \put(07,14){\circle*{0.25}}
\put(09,14){\circle{1.0}} \put(09,14){\circle*{0.25}}

\put(08,13){\vector(0,-1){6}} \put(08,13){\vector(-1,-1){6}}
\put(08,13){\vector(1,-1){6}}

\put(20,14){\oval(04,02)}

\put(19,14){\circle*{1.3}}


\put(20,13){\vector(0,-1){6}}

\put(20,13){\vector(-2,-1){11.5}}

\put(20,13){\vector(1,-1){6}}


\put(02,06){\oval(04,02)}

\put(01,06){\circle*{0.7}} \put(03,06){\circle*{0.7}}


\put(08,06){\oval(04,02)}

\put(07,06){\circle*{1.1}}

\put(09,06){\circle{1.3}} \put(09,06){\circle*{0.3}}


\put(14,06){\oval(04,02)}

\put(13,06){\circle{1.3}} \put(13,06){\circle*{0.3}}



\put(20,06){\oval(04,02)}

\put(19,06){\circle*{0.5}}

\put(21,06){\circle*{0.9}}


\put(26,06){\oval(04,02)}

\put(25,06){\circle*{0.9}}


\end{picture}
\end{center}


 Further,
 some simplified versions of comparing methods
 for clusterings  solutions are
 described (for ``hard'' clustering).

 For case 1 (proximity for clustering solutions, i.e., partitions)
 the following main approaches are considered
 (e.g.,
 \cite{bar86,ben06,day86,hub85,huss14,meila12,mirkin79,mirkin96,mirkin70,rand71}):
  Misclassification Error distance,
 \(\chi^{2}\) distance,
 Hamming distance,
%
   Rand index,
  Mirkin metric,
%
 ordered sets,
 consensus.
%
 Generally, it is possible
 to use vector and multiset estimates as well.
 Let us consider a simplified edit proximity
 as number of steps for transformation (i.e., cost of
 transformation):

  \(D( \widehat{X}_{1} \Rightarrow \widehat{X}_{2})\):~~
 \(\widehat{X}_{1}=\{X_{1,1},...,X_{1,\iota_{1}},...,X_{1,\lambda_{1}}\}
  \Longrightarrow
 \widehat{X}_{2}=\{X_{2,1},...,X_{2,\iota_{2}},...,X_{2,\lambda_{2}}\}\).

 The value of proximity will be based on ordinal interval
 \([0,...,n]\)~
 (i.e., the number of relocated elements in \(\widehat{X}_{1}\)).
  The following simple algorithm (heuristic)
  can be used
  (i.e., \(\widehat{X}_{1} \Rightarrow  \widehat{X}'_{1} = \widehat{X}_{2} \)):

~~

 {\it Stage 1.} Definition:
 \(D( \widehat{X}_{1} \Rightarrow \widehat{X}_{2}) = 0. \)

 If \(\lambda_{1} < \lambda_{2}\) then extension of solution
 \(\overline{X}_{1}\) by
 \((\lambda_{2} - \lambda_{1})\) empty clusters.

 {\it Stage 2.} Calculation of  the number of common elements for clusters
 of the clustering solutions (the number corresponds to cluster
 proximity)
  \(X_{1,\iota_{1}}, ~\iota_{1} = \overline{1,\lambda_{1}}\)
 and
  \(X_{2,\iota_{2}}, ~\iota_{2} = \overline{1,\lambda_{2}}\)
  as intersection (as cardinality of the same elements set).
  As a result, the intersection matrix will be obtained:
  \(M(\widehat{X}_{1},\widehat{X}_{2}) = || \mu (X_{1,\iota_{1}},X_{2,\iota_{2}})||,
  ~\iota_{1}=\overline{1,\lambda_{1}},
  ~\iota_{2}=\overline{1,\lambda_{2}}\).

 {\it Stage 3.} Finding the maximum element in matrix \(M \):~
  \[\mu^{max}=  \mu ( X_{1,\iota'},X_{2,\iota''} ) =
  \max_{\iota_{1}=\overline{1,\lambda_{1}},\iota_{2}=\overline{1,\lambda_{2}}}~~
  \{ \mu (X_{1,\iota_{1}},X_{2,\iota_{2}}) \} .\]

 {\it Stage 4.} Selection the cluster \(X_{1,\iota'}\)
 in solution \(\widehat{X}_{1}\)
   as cluster of new (transformed from \(\widehat{X}_{1}\))
   solution  \(\widehat{X'}_{1}\)).
%

 {\it Stage 5.} Relocation for set
 \(X_{1,\iota'}\)
 the following elements:~
 \( \iota \in  X_{2,\iota''} | \iota \overline{\in} X_{2,\iota''} \)
 (with deletion of the elements from  other clusters of solution \(\widehat{X}_{1}\)).

 Increasing
 \(D( \widehat{X}_{1} \Rightarrow \widehat{X}_{2}) \)
 by the number of the relocated elements.


 {\it Stage 6.} Deletion of cluster  \(X_{1,\iota'}\)
 and
  \(X_{2,\iota''}\)
  from the examination and recalculation of matrix \(M\)
  (i.e., deletion of the corresponding line and column).

 {\it Stage 7.} If  matrix \(M\) is empty
 (i.e., resultant transformed cluster
 \(\widehat{X'}_{1}\)
   is built)
 then
 GO TO Stage 9.

 {\it Stage 8.} GO TO stage 3.

 {\it Stage 9.} Stop.

\begin{center}
\begin{picture}(22,44)

\put(33.5,00){\makebox(0,0)[bl]{Fig. 2.31. Illustrative
 numerical example}}

\put(00,05){\makebox(0,0)[bl]{(a) elements}}


\put(08.5,38.6){\makebox(0,0)[bl]{\(A\)}}

\put(10,23){\oval(08,28)}

\put(09,34){\makebox(0,0)[bl]{1}}
\put(09,30){\makebox(0,0)[bl]{2}}
\put(09,26){\makebox(0,0)[bl]{3}}
\put(09,22){\makebox(0,0)[bl]{4}}
\put(09,18){\makebox(0,0)[bl]{5}}
\put(09,14){\makebox(0,0)[bl]{6}}
\put(09,10){\makebox(0,0)[bl]{7}}

\end{picture}
%
\begin{picture}(27,42)

\put(00,05){\makebox(0,0)[bl]{(b) clusterings }}

\put(03.8,38){\makebox(0,0)[bl]{\(\widehat{X}_{1}\)}}

\put(05,23){\oval(10,28)}

\put(05,32){\oval(08,5)} \put(02.6,30){\makebox(0,0)[bl]{1,2}}

\put(05,26){\oval(08,5)} \put(02.6,24){\makebox(0,0)[bl]{3,4}}

\put(05,20){\oval(08,5)} \put(02.6,18){\makebox(0,0)[bl]{5,6}}

\put(05,14){\oval(08,5)}  \put(04,13){\makebox(0,0)[bl]{7}}


\put(15,38){\makebox(0,0)[bl]{\(\widehat{X}_{2}\)}}

\put(17,23){\oval(10,28)}

\put(17,32){\oval(08,5)} \put(16,31){\makebox(0,0)[bl]{1}}

\put(17,26){\oval(08,5)} \put(14.6,24){\makebox(0,0)[bl]{2,3}}

\put(17,20){\oval(08,5)} \put(16,19){\makebox(0,0)[bl]{4}}

\put(17,14){\oval(09,5)}  \put(13.6,12){\makebox(0,0)[bl]{5,6,7}}

\end{picture}
%
%
\begin{picture}(24,43)

\put(00,05){\makebox(0,0)[bl]{(c) rankings }}

\put(03,35){\makebox(0,0)[bl]{\(R_{1}\)}}


\put(05,32){\oval(08,4)} \put(02.6,30){\makebox(0,0)[bl]{1,6}}

\put(05,30){\vector(0,-1){4}}

\put(05,24){\oval(08,4)} \put(02.6,22){\makebox(0,0)[bl]{2,4}}

\put(05,22){\vector(0,-1){4}}

\put(05,16){\oval(09,4)} \put(01.6,14){\makebox(0,0)[bl]{3,5,7}}



\put(13,39){\makebox(0,0)[bl]{\(R_{2}\)}}


\put(15,36){\oval(08,4)} \put(12.6,34){\makebox(0,0)[bl]{1,2}}

\put(15,34){\vector(0,-1){4}}

\put(15,28){\oval(08,4)} \put(12.6,26){\makebox(0,0)[bl]{3,4}}

\put(15,26){\vector(0,-1){4}}

\put(15,20){\oval(08,4)} \put(12.6,18){\makebox(0,0)[bl]{5,6}}

\put(15,18){\vector(0,-1){4}}

\put(15,12){\oval(08,4)}

\put(14,11){\makebox(0,0)[bl]{7}}

\end{picture}
%
%
\begin{picture}(28,43)

\put(05,05){\makebox(0,0)[bl]{(d) hierarchies over clusters }}


\put(09,31.7){\makebox(0,0)[bl]{\(\widehat{H}_{1}\)}}



\put(11,28){\oval(06,4)} \put(10,27){\makebox(0,0)[bl]{1}}

\put(11,26){\vector(-1,-1){4}} \put(11,26){\vector(1,-1){4}}


\put(07,20){\oval(06,4)} \put(06,19){\makebox(0,0)[bl]{2}}

\put(07,18){\vector(-1,-1){4}} \put(07,18){\vector(1,-1){4}}

\put(15,20){\oval(06,4)} \put(14,19){\makebox(0,0)[bl]{3}}

\put(15,18){\vector(-1,-1){4}} \put(15,18){\vector(1,-1){4}}


\put(03,12){\oval(06,4)} \put(02,11){\makebox(0,0)[bl]{4}}

\put(11,12){\oval(06,4)} \put(10,11){\makebox(0,0)[bl]{5}}

\put(20,12){\oval(08,4)} \put(18,10){\makebox(0,0)[bl]{6,7}}

\end{picture}
%
\begin{picture}(24,43)


\put(09,31.7){\makebox(0,0)[bl]{\(\widehat{H}_{2}\)}}



\put(11,28){\oval(06,4)} \put(10,27){\makebox(0,0)[bl]{1}}

\put(11,26){\vector(-1,-1){4}} \put(11,26){\vector(1,-1){4}}


\put(07,20){\oval(06,4)} \put(06,19){\makebox(0,0)[bl]{2}}

\put(07,18){\vector(-1,-1){4}} \put(07,18){\vector(1,-1){4}}

\put(15,20){\oval(06,4)} \put(14,19){\makebox(0,0)[bl]{4}}

\put(15,18){\vector(-1,-1){4}} \put(15,18){\vector(1,-1){4}}

\put(15,18){\vector(-3,-1){11}}


\put(03,12){\oval(06,4)} \put(02,11){\makebox(0,0)[bl]{3}}

\put(11,12){\oval(06,4)} \put(10,11){\makebox(0,0)[bl]{5}}

\put(20,12){\oval(08,4)} \put(18,10){\makebox(0,0)[bl]{6,7}}

\end{picture}
\end{center}
%


 {\bf Example 2.13.} The usage of algorithm above for
 example from Fig. 2.31b is the following.

 Two  clustering solutions are under examination:

 (i) \(\widehat{X}_{1}=\{ X_{1,1},X_{1,2},X_{1,3},X_{1,4} \}\),
 \(X_{1,1}=\{1,2\}\), \(X_{1,2}=\{3,4\}\), \(X_{1,3}=\{5,6\}\), \(X_{1,4}=\{7\}\),

 (ii) \(\widehat{X}_{1}= \{X_{1,1},X_{1,2},X_{1,3},X_{1,4}\}\),
 \(X_{2,1}=\{1\}\), \(X_{2,2}=\{2,3\}\), \(X_{2,3}=\{4\}\),
 \(X_{2,4}=\{5,6,7\}\) ).

 Table 2.14 presents the number of common elements for cluster pairs.
 Fig. 2.32 depicts
 step-by-step building the clustering solution
 \(\widehat{X}'_{1} = \widehat{X}_{2} \).
 Thus,
 \(D ( \widehat{X}_{1} \Rightarrow \widehat{X}_{1} )=3\)
 (three elements have been re-located:
 \(7\), \(2\), \(4\)).

\begin{center}
{\bf Table 2.14.} Common elements for clusters  of solutions  \(\widehat{X}_{1}\), \(\widehat{X}_{1}\)  \\
\begin{tabular}{| l | c | c |c| c| }
\hline
   & \(X_{2,1}=\{1\}\) & \(X_{2,2}=\{2,3\}\)& \(X_{2,3}=\{4\}\)& \(X_{2,4}=\{5,6,7\}\) \\
\hline
 \(X_{1,1}=\{1,2\}\)&  1&1&0& 0   \\
 \(X_{1,2}=\{3,4\}\)&  0&1&1& 0   \\
 \(X_{1,3}=\{5,6\}\)&  0&0&0& 2   \\
 \(X_{1,4}=\{7\}\)  &  0&0&0& 1   \\

\hline
\end{tabular}
\end{center}

\begin{center}
\begin{picture}(75,40)
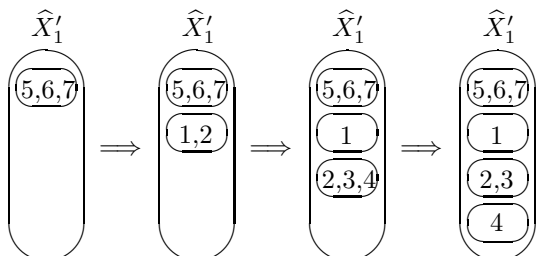
\put(001,00){\makebox(0,0)[bl]{Fig. 2.32. Steps for solution
 \(\widehat{X}_{1} \Rightarrow \widehat{X}'_{1}=\widehat{X}_{2}\)}}



\put(03,35){\makebox(0,0)[bl]{\(\widehat{X}'_{1}\)}}

\put(05,20){\oval(10,28)}

\put(05,29){\oval(08,5)} \put(01.6,27){\makebox(0,0)[bl]{5,6,7}}


\put(12,20){\makebox(0,0)[bl]{\(\Longrightarrow\)}}


\put(23,35){\makebox(0,0)[bl]{\(\widehat{X}'_{1}\)}}

\put(25,20){\oval(10,28)}

\put(25,29){\oval(08,5)} \put(21.6,27){\makebox(0,0)[bl]{5,6,7}}

\put(25,23){\oval(08,5)} \put(22.6,21){\makebox(0,0)[bl]{1,2}}

\put(32,20){\makebox(0,0)[bl]{\(\Longrightarrow\)}}


\put(43,35){\makebox(0,0)[bl]{\(\widehat{X}'_{1}\)}}

\put(45,20){\oval(10,28)}

\put(45,29){\oval(08,5)} \put(41.6,27){\makebox(0,0)[bl]{5,6,7}}

\put(45,23){\oval(08,5)} \put(44,21.5){\makebox(0,0)[bl]{1}}
\put(45,17){\oval(08,5)} \put(41.6,15){\makebox(0,0)[bl]{2,3,4}}

\put(52,20){\makebox(0,0)[bl]{\(\Longrightarrow\)}}


\put(63,35){\makebox(0,0)[bl]{\(\widehat{X}'_{1}\)}}

\put(65,20){\oval(10,28)}

\put(65,29){\oval(08,5)} \put(61.6,27){\makebox(0,0)[bl]{5,6,7}}

\put(65,23){\oval(08,5)} \put(64,21.5){\makebox(0,0)[bl]{1}}
\put(65,17){\oval(08,5)} \put(62.6,15){\makebox(0,0)[bl]{2,3}}
\put(65,11){\oval(08,5)}  \put(64,10){\makebox(0,0)[bl]{4}}

\end{picture}
\end{center}

 For case 2 (proximity for rankings),
 the following approaches are often considered
 (e.g., \cite{cook92,cook96,kem72,lev11agg,lev15,mirkin79}):
 {\it 1.} Kendall tau distance
 \cite{kendall62},
 {\it 2.} distances for partial rankings
 \cite{bansal09,fagin06}, and
 {\it 3.} vector-like proximity
 \cite{lev98}.

 Further, proximity measure as a transformation cost will be
 used:~
 \(D(R_{1},R_{2}) =  D (R_{1} \Rightarrow R_{2}) \)
 (i.e., the number of element re-allocations
 while taking into account linear ordering over clusters)
 (this is similar to Kendall tau distance).
 Let \(\alpha(i,R_{1})\) be the number of layer
 of element \(i \in A\) in ranking \(R_{1}\)
 and
 \(\alpha(i,R_{2})\) be the number of layer
 of element \(i \in A\) in ranking \(R_{2}\).
 Thus, re-location parameter
  for \(i \in A\) is:~
  \(\Delta (i, R_{1}\Rightarrow R_{2})=\alpha(i,R_{2})-\alpha(i,R_{1})\).
 Generally, the following proximity is obtained
 (an analogue of Kendall tau distance, ordinal scale \([0,(n-1)n]\)):
 \[D ( R_{1}\Rightarrow R_{2} ) = \sum_{i\in A}~ | \Delta (i, R_{1}\Rightarrow R_{2})| =
   \sum_{i\in A}~ |\alpha(i,R_{2})-\alpha(i,R_{1})|.\]

 For vector-like proximity
 (e.g., \cite{lev98,lev11agg,lev15}),
 the following parameter is considered:
 \(\beta^{\kappa}\)
 ( \(\kappa \in [-(n-1),(n-1)]  \))
  which  equals the number of elements with
 \(\Delta (i, R_{1}\Rightarrow R_{2})=\kappa\).
 Then, integrated vector-like proximity is:~~
 \( \overline{D} ( R_{1}\Rightarrow R_{2} ) =
  (\beta^{-(n-1)},...,\beta^{-1},\beta^{0},\beta^{1},...,\beta^{(n-1)}
  )\).

 ~~

 {\bf Example 2.14.} Example  of two rankings from Fig. 2.31
 is considered.
 Table 2.15  contains numbers of re-locations for each element
 of ranking \(R_{1}\) to obtain ranking \(R_{2}\).

\begin{center}
{\bf Table 2.15.} Re-location of elements for transformation   \(R_{1} \Rightarrow  R_{2}\)  \\
\begin{tabular}{| c | c | c| }
\hline
\(i \in A\) & Re-location &\(\Delta (i, R_{1}\Rightarrow R_{2})\)   \\
\hline
 \(1\)&  \(0\) & \(0\)   \\
 \(2\)&  \(1\) & \(1\)   \\
 \(3\)&  \(1\) & \(1\)   \\
 \(4\)&  \(0\) & \(0\)   \\
 \(5\)&  \(0\) & \(0\)   \\
 \(6\)&  \(1\) & \(-2\)  \\
 \(7\)&  \(1\) & \(-1\)  \\
\hline
\end{tabular}
\end{center}

 Thus, the resultant transformation cost
 (the number of re-location) is:~
 \(D(R_{1} \Rightarrow R_{2})=5\).
 For vector-like proximity
 \cite{lev98,lev11agg,lev15},
 the following result is obtained:~
 \(\overline{D}(R_{1} \Rightarrow R_{2})=(0,0,0,0,1,1,3,2,0,0,0,0,0 )\).

~~


 For case 3 (proximity for hierarchies, e.g., trees),
 the following main approaches are in use
 (e.g., \cite{boor73,gar79,lev11agg,lev15}):
 {\it 1.} metrics/distances
 (e.g.,
 alignment distance, top-down distance, bottom-up distance)
 (e.g., \cite{jiang95b,margush82,tanaka94,valiente02}),
 {\it 2.} tree edit distances (i.e., correction algorithms)
 (e.g., \cite{chen01,sasha89,tai79,tanaka88}),
 {\it 3.} largest common subtree, median tree or tree agreement/consensus
 (e.g., \cite{akatsu00,amir97,wang01}), and
 {\it 4.} vector proximity
  (e.g., \cite{lev11agg,lev15}).
 For the simplification,
 the following correction algorithm will be considered as
 the number of addition and deletion of edges/arcs
 to transform initial tree (hierarchy) \(H_{1}\)
  into the resultant hierarchy (tree) \(H_{2}\):~
 \(D(H_{1} \Rightarrow H_{2})\).
 For example in Fig. 2.31d,
 \(D(H_{1} \Rightarrow H_{2}) = 1\)
  (i.e., deletion of the only one arc).

 In case 4,
 it may be reasonable to use
 an integrated vector-like resultant
 including two main components:
  proximity for clustering solution and proximity for hierarchies.
 For example in Fig. 2.31d,
 the following initial information is examined:~
 \(\widehat{H}_{1}=<\widehat{X}_{1},H_{1}>\) and
 \(\widehat{H}_{2}=<\widehat{X}_{2},H_{2}>\).
 Thus,
 the following resultant two-component proximity  is obtained:~
 \(\overline{D}(\widehat{H}_{1}\Rightarrow \widehat{H}_{2}) =
  (D(\widehat{X}_{1}\Rightarrow \widehat{X}_{2}),D(H_{1}\Rightarrow H_{2})=
  (2,1)\).


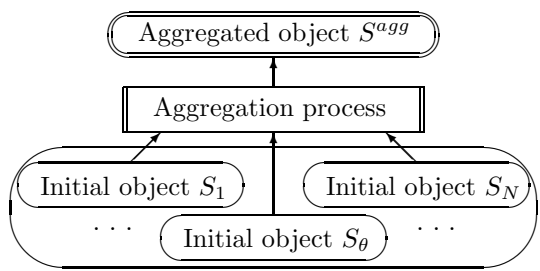
\subsubsection{Aggregation
 of objects/clustering solutions}

  The aggregation problem for  \(N\) initial objects
 is depicted in Fig. 2.33
 (e.g., \cite{lev11agg,lev15}):
 \[\{ S_{1},...,S_{\theta},...,S_{N} \} \Longrightarrow S^{agg}.\]
 The basic types of aggregation problems are listed in
 Table 2.16.
 Two notes can be pointed out:

 {\bf 1.} There are aggregation problems
 in which the resultant (agreement) structure
 has another type than the initial structures,
 for example:
 (a) trees are aggregated by graph,
 (b) trees are aggregated by forest,
 (c) rankings are aggregated by fuzzy ranking,
 (d) rankings are aggregated by poset
 (e.g., \cite{cha05,hallett07,lev15}).

 {\bf 2.} Separation of two-object aggregation problems
 is based some situations
 when these problems can have more simple level of
 algorithmic complexity than
 many object aggregation problems.

\begin{center}
\begin{picture}(69,40)
\put(08.8,00){\makebox(0,0)[bl]{Fig. 2.33. Aggregation
 of objects}}

\put(11,10){\makebox(0,0)[bl]{.~.~.}}
\put(54,10){\makebox(0,0)[bl]{.~.~.}}

\put(04,14){\makebox(0,0)[bl]{Initial object \(S_{1}\)}}
\put(16,16){\oval(30,6)} \put(16,19){\vector(1,1){04}}

\put(42,14){\makebox(0,0)[bl]{Initial object \(S_{N}\)}}
\put(54,16){\oval(30,6)} \put(54,19){\vector(-1,1){04}}

\put(23,07){\makebox(0,0)[bl]{Initial object \(S_{\theta}\)}}
\put(35,09){\oval(30,6)} \put(35,12){\vector(0,1){11}}

\put(35,13){\oval(70,16)}


\put(15,23){\line(1,0){40}} \put(15,29){\line(1,0){40}}
\put(15,23){\line(0,1){6}} \put(55,23){\line(0,1){6}}
\put(15.5,23){\line(0,1){6}} \put(54.5,23){\line(0,1){6}}

\put(19,24.5){\makebox(0,0)[bl]{Aggregation process}}

\put(35,29){\vector(0,1){4}}

\put(17,34.5){\makebox(0,0)[bl]{Aggregated object \(S^{agg}\)}}
\put(35,36){\oval(43,5)} \put(35,36){\oval(44,6)}

\end{picture}
\end{center}
%

\begin{center}
{\bf Table 2.16.} Types of aggregation problems \\
\begin{tabular}{|c|l|l|l|l| }
\hline
 No. & Type of objects& Result& Description &Some) \\

  & & & &source(s) \\
\hline
 1.& Two elements   (i.e., two&1.Aggregated element&1.Median
 &
  \\
  &parameters vectors)&2.Integration (cluster)&1.Integration&
  \\

 2.&\(N\) elements (i.e., \(N\) &1.Aggregated element&1.Centroid/median
  &\cite{jain88,jain99} \\
   &parameters vectors)&2.Integration (cluster) &2.Integration
   &
   \\

 3.&\(1\) element and cluster \(X\) &Extended cluster&Addition
 &
  \\

 4.&\(2\) clusters \(X_{1}\), \(X_{2}\),&1.Aggreement cluster&1.Median/agreement/
 &
 \\
  &&&consensus&\\

  &&2.Aggregated cluster&2.Integration&\\

 5.&\(N\) clusters \(\{X_{1},...,X_{N}\}\)&1.Agreement cluster&1.Median/agreement/
 &
   \\
    &&&consensus&\\

   &&2.Aggregated cluster&2.Integration&\\

 6.&\(2\) clustering solutions    &Aggregated clustering&Median/agreement/ &
  \\
   &\(\widehat{X}_{1}\), \(\widehat{X}_{2}\)&solution \(\widehat{X}^{agg}\)&consensus&\\

 7.&\(N\) clustering solutions&Aggregated clustering& Median/agreement/
 &\cite{ayad10,bar86,chen11,han13,hub85} \\
   &\(\{\widehat{X}_{1},...,\widehat{X}_{N}\}\)&solution \(\widehat{X}^{agg}\)
   &consensus& \cite{monti03,mull15,str02,topchy04} \\

 8.&\(2\) rankings  \(R_{1}\), \(R_{2}\)&Aggregated ranking &Median/agreement/
 &
 \\
   &&\(R^{agg}\)&consensus &\\

 9.&\(N\) rankings  \(\{R_{1},...,R_{N}\}\)&Aggregated ranking &Median, agreement/
 & \cite{beck83,cook92,cook96,tavana03} \\
   &&\(R^{agg}\) &consensus&\\

 10.&\(2\) hierarchies (trees)&Aggregated hierarchy&Median/agreement/
 &
   \\
   &\(H_{1}\), \(H_{2}\)  &(tree) \(H^{agg}\)&consensus&\\

 11.&\(N\) hierarchies (trees)&Aggregated hierarchy&Median/agreement/
 &\cite{akatsu00,amir97,farach95,wang01}\\
   &\(\{H_{1},...,H_{N}\}\)&(tree) \(H^{agg}\)&consensus&\\

 12.&\(2\) hierarchical clustering&Aggregated hierar- &Median/agreement/
   &
   \\
   & solutions \(\widehat{H}_{1}\), \(\widehat{H}_{2}\)
   & chical clustering &consensus (by cluste- &\\

 & &solution \(\widehat{H}^{agg}\) &ring,  by hierarchy)&\\

 13.&\(N\) hierarchical clustering &Aggregated hierar-
   &Median/agreement/ &
   \\
   &solutions \(\{\widehat{H}_{1},...,\widehat{H}_{N}\}\)
   &chical clustering  &consensus (by cluste-&\\
  & &solution \(\widehat{H}^{agg}\) &ring, by hierarchy)&\\

\hline
\end{tabular}
\end{center}

 The aggregation problems are formulated as a calculation  procedure (e.g., calculation of a median point)
 or as an optimization problem to find a set/structure which is
 median/agreement/consensus
 for the set of initial structures:~
%
 \( S^{agg} =   \arg \min_{\forall S}~  \sum_{\theta = \overline{1,N}} ~ D(S_{\theta} \Rightarrow S)
 \).
%
 On the other hand, the aggregation problem can be considered on the
  basis of optimization approach:

~~

 Find an aggregation object (set, structure) \(S^{agg}\)
 for the initial set of objects
 \( \{ S_{1},...,S_{\theta},...,S_{N} \}\)
 to obtain maximum/minimum for
 quality estimates of \(S^{agg}\)
 while taking into account requirements (as some constraints, e.g., transformation costs)
 for \(S^{agg}\).

~~

  Here, the problem is (\(D^{0}\) is a limit for transformation cost):
 \[  \max_{\forall S^{agg}}~ \overline{Q}(S^{agg}) ~~~~~~~
%
%
 s.t.~~~  D(S_{\theta} \Rightarrow S^{agg}) \leq D^{0}, ~~ \theta = \overline{1,N}.\]
 The basic cases for aggregation of \(N\) objects are the following:

~~

 {\it Case 1.} Aggregation of \(N\) objects (i.e.,
 points/clusters). Here, calculations of an average
 object/centroid or a covering object are usually used.

~~

 {\it Case 2.} Aggregation of \(N\) clustering solutions:~
 \(\{ \widehat{X}_{1},...,\widehat{X}_{\theta},...,\widehat{X}_{N} \} \Longrightarrow
 \widehat{X}^{agg}\).

~~

 {\it Case 3.} Aggregation of \(N\) rankings:~
 \(\{ R_{1},...,R_{\theta},...,R_{N} \} \Longrightarrow
 R^{agg}\).

 Here the following three methods can be used:
 (i) median consensus method based on assignment problem
 (e.g.,
 \cite{cook92,cook96}),
%
 (ii) heuristic approach
 (e.g., \cite{beck83,tavana03},
 and
 (iii) method based on multiple choice problem
 (e.g.,
 \cite{lev98}).

~~

 {\it Case 4.} Aggregation of \(N\) hierarchies (trees):~
 \(\{ H_{1},...,H_{\theta},...,H_{N} \} \Longrightarrow
 H^{agg}\).

 The following methods are often considered for trees:
 (1) maximum common subtree
 (e.g., \cite{akatsu00}),
%
 (2) median/agreement tree
 (e.g., \cite{amir97,farach95}),
%
 (3) compatible tree
 (e.g.,
   \cite{hamel96}), and
 (4) maximum agreement forest
 (e.g., \cite{cha05,hallett07}).
%
%

~~

 {\it Case 5.} Aggregation of \(N\) hierarchical clustering
 solutions:~
 \(\{ \widehat{H}_{1},...,\widehat{H}_{\theta},...,\widehat{H}_{N} \} \Longrightarrow
 \widehat{H}^{agg}\).

 Here, a composition of case 2 and case 4 can be considered.
 This aggregation problem is very prospective for future study.

~~

 Mainly
 (e.g., cases 2, 3, 4, 5),
 the aggregation problems above belong to class of NP-hard problems
 (e.g., \cite{cook92,hamel96}).
 A simplified illustrative numerical example
 for case 2 is as follows.

 ~~

 {\bf Example 2.15.}
 Three initial clustering solutions for set
 \(A = \{1,2,3,4,5,6,7\}\)
 are examined:

 (i) \(\widehat{X}_{1} = \{ X_{11}, X_{12},X_{13},X_{14}\}\),
 \(X_{11} = \{1,2\}\),
 \(X_{12} = \{3,4\}\),
 \(X_{13} = \{5,6\}\),
 \(X_{14} = \{7\}\) (Fig. 2.31b);

 (ii) \(\widehat{X}_{2} = \{ X_{21}, X_{22},X_{23},X_{24}\}\),
 \(X_{21} = \{1\}\),
 \(X_{22} = \{2,3\}\),
 \(X_{23} = \{4\}\),
 \(X_{24} = \{5,6,7\}\) (Fig. 2.31b);

 (iii) \(\widehat{X}_{3} = \{ X_{31}, X_{32},X_{33}\}\),
 \(X_{31} = \{1,2\}\),
 \(X_{32} = \{3,4\}\),
 \(X_{33} = \{5,6,7\}\).

 The following aggregation problem is under examination (with constraint for cluster size):
 \[\widehat{X}^{agg}= \arg \min_{2\leq |\widehat{X}|\leq3}~~ \sum_{\theta=\overline{1,N}} D(\widehat{X}_{\theta} \Rightarrow \widehat{X}).\]
%

 In our problem (\(N=3\)),
  the number of admissible partitions
 (i.e., clustering solutions)  equals 90
%
%
 (\(C^{2}_{6} \times C^{2}_{4}\)).
 For the simplified calculation,
 the following
 admissible aggregated clustering solution is
 considered:
 \(\widehat{X'}^{agg} = \{\widehat{X'}^{agg}_{1},
 \widehat{X'}^{agg}_{2},\widehat{X'}^{agg}_{3}\}\).
 Numbers of common elements for clusters of initial clustering solutions
 \(\widehat{X}_{1}\), \(\widehat{X}_{2}\), \(\widehat{X}_{3}\)
  and
 considered aggregated solution
 \(\widehat{X'}^{agg}\)
 are presented in Table 2.17, Table 2.18, Table 2.19.
 The transformation costs are
 (i.e., numbers of re-allocations):~

 \(D(\widehat{X}_{1} \Rightarrow \widehat{X'}^{agg}) = 1\),
 \(D(\widehat{X}_{2} \Rightarrow \widehat{X'}^{agg}) = 2\),
 \(D(\widehat{X}_{3} \Rightarrow \widehat{X'}^{agg}) = 0\).

\newpage
\begin{center}
{\bf Table 2.17.} Common elements for clusters  of solutions \(\widehat{X}_{1}\), \(\widehat{X'}^{agg}\)   \\
\begin{tabular}{| l | c | c |c| }
\hline
   & \(X'^{agg}_{1,1}=\{1,2\}\) & \(X'^{agg}_{3,4}=\{3,4\}\)& \(X'^{agg}_{1,3}=\{5,6,7\}\) \\
\hline
 \(X_{1,1}=\{1,2\}\)& 2&0&0 \\
 \(X_{1,2}=\{3,4\}\)& 0&2&0 \\
 \(X_{1,3}=\{5,6\}\)& 0&0&2 \\
 \(X_{1,4}=\{7\}\)  & 0&0&1 \\
\hline
\end{tabular}
\end{center}

\begin{center}
{\bf Table 2.18.} Common elements for clusters  of solutions  \(\widehat{X}_{2}\), \(\widehat{X'}^{agg}\)  \\
\begin{tabular}{| l | c |c| c| }
\hline
  &
   \(X'^{agg}_{1,1}=\{1,2\}\) & \(X'^{agg}_{3,4}=\{3,4\}\)& \(X'^{agg}_{1,3}=\{5,6,7\}\) \\
\hline
 \(X_{2,1}=\{1\}\)    &  1&0&0 \\
 \(X_{2,2}=\{2,3\}\)  &  1&1&0 \\
 \(X_{2,3}=\{4\}\)    &  0&1&0 \\
 \(X_{2,4}=\{5,6,7\}\)&  0&0&3 \\

\hline
\end{tabular}
\end{center}

\begin{center}
{\bf Table 2.19.} Common elements for clusters  of solutions  \(\widehat{X}_{3}\), \(\widehat{X'}^{agg}\)  \\
\begin{tabular}{| l | c|c| c| }
\hline
& \(X'^{agg}_{1,1}=\{1,2\}\) & \(X'^{agg}_{1,2}=\{3,4\}\)& \(X'^{agg}_{1,3}=\{5,6,7\}\) \\
\hline
 \(X_{3,1}=\{1,2\}\)  & 2&0&0 \\
 \(X_{3,2}=\{3,4\}\)  & 0&2&0 \\
 \(X_{3,3}=\{5,6,7\}\)& 0&0&3 \\
\hline
\end{tabular}
\end{center}


\subsection{Basic clustering models and general framework}

\subsubsection{Basic clustering problems/models}

 Table 2.20 contains the list of basic types of well-known clustering
 problems/models:
 (e.g.,
 \cite{arabie96,and73,cormen90,desmet09,duran74,gar79,gord99,har75,jain88,jain99,kau90,ker70,kot04,kri09,mirkin05,murtagh85,rocha13,scha07,vanr77,xu05,xu09}):

 (1) connectivity models (e.g., hierarchical clustering),

 (2) centroid models (e.g., k-means algorithms, i.e. {\it exclusive clustering}),

 (3) distribution models (based on statistical distribution),

 (4) subspace models (e.g., bi-clustering or two-mode clustering
 while taking into account elements and attributes),

 (5) graph-based models (e.g., detection of cliques or
 quasi-cliques/community structures, graph partitioning),
 etc.
%

  Note, clustering procedures based on combinatorial optimization
 problems and/or their composition are widely studied
 \cite{aug70,hansen97,jain88,jain99,mirkin99}:

 (i) spanning trees based clustering
 (e.g., \cite{gow69,gry06,mul12,paiv05,peter10,wang09,xu01,zhong10});

 (ii) assignment/location problems based clustering
  (e.g., \cite{goldb08,kearns98});

 (iii) set covering problem based clustering
 (e.g., \cite{agrawal05,mul09,salz09});

 (iv) partitioning problem based clustering
 (e.g., \cite{condon01,ding01,even99,ker70,spiel13,xux07})
 including
 correlation clustering
 (e.g., \cite{achtert07,dem06,kri09});

 (v) dominant sets/dominating sets  based clustering
 (e.g., \cite{chen02,han07,li06,pav07,youn06});
 and

 (vi) clique/community based clustering
 (e.g., \cite{agrawal05,berk06,dorn94,duan12,gra05,koch05,mehrot98,new04a,new06,new04,port09,shamir02}).

 Important contemporary clustering problems
  are targeted to clustering of complex
  (e.g., composite, modular, structured) objects,  for example:

%
 (a) words/chains/sequence clustering
 (e.g., in bioinformatics) \cite{enr00},

 (b) trajectory clustering \cite{fu05,lee07,li06},

 (c) data stream clustering \cite{has14,guha00a,lev12clique},

 (d) subspace clustering  \cite{agrawal05,kri09,mul09}, and

 (e) clustering of structured objects
 (e.g., trees, graph-based models)
 \cite{flores14,marx02,schen04,stepp86}.

 On the other hand, clustering is widely used in complex
 combinatorial optimization problems, for example:

  (1) clustering/partitioning of an initial problem for decreasing the problem dimension,

  (2) clustering as  local auxiliary problem(s)
  (e.g., \cite{kin07,kwa93,lev98,lev15}).

\newpage

\begin{center}
 {\bf Table 2.20.} Basic types of clustering  problems/models   \\
\begin{tabular}{| c | l | l |}
\hline
 No.  & Model type & Some sources \\
\hline

 I. & Basic problem formulations    &  \\

 1.1.& Connectivity models (hierarchical/agglomerative clustering
 & \cite{dahl00,har75,gowda77,kurita91,rocha13,olson95,yag00} \\


 1.2.& Centroid models (k-means algorithms, {\it exclusive clustering}) &

 \cite{dhil05,har79,jain88,jain99,jain10,kanu00,mirkin05} \\


 1.3.& Distribution models (based on statistical distribution) &
  \cite{jain88,jain00,jain99,mirkin05} \\

 1.4.& Subspace models (e.g., bi-clustering or two-mode clustering
 &  \cite{jain88,jain99,kri09,mad04} \\
 & while taking into account elements and attributes) & \\

 1.5. & Pattern-based clustering & \cite{alexe06,kri09,ozdal04,pei03,wang02}\\

 1.6. & Combinatorial optimization models in clustering: &
  \cite{cheng95,gar79,mirkin99}\\

  & (i)
  minimal spanning tree based clustering,
 &
 \cite{jain99,gow69,gry06,mul12,paiv05,peter10,wang09} \\

  && \cite{xu01,zhong10} \\

  & (ii) partitioning based clustering, & \cite{condon01,ding01,even99,ker70,mirkin99}\\

  & (iii) correlation clustering, & \cite{achtert07,bansal04,dem06,kri09,swam04}\\

 & (iv) detection of communities structures (clique, etc.),
   &
  \cite{gar79,dorn94,jain99,new06,new04,port09,tsuda06} \\


 & (v) assigment/location based clustering. & \cite{goldb08,hub87}\\

 1.7.& Overlapping clustering  & \cite{arabie81}\\

 1.8. & Modularity clustering &  \cite{agar08,brand08,new06,new04,xux07} \\

 1.9. &  Support vector clustering &\cite{ben01,chia03,li15} \\

 1.10.& Spectral clustering models &
  \cite{dhil05,hag92,kannan04,lux07} \\

 1.11.& Symbolic  approach in  clustering &
  \cite{billard07,bock00,diday88,gowda91,gowda92,lech06,sato11} \\

 1.12. & AI-based clustering (e.g., knowledge bases, heuristics,&
  \cite{al95,babu94,brown92,cow99,krish99,mur95,selim91} \\
  & evolutionary approaches)&  \cite{shekar87,sung00,tseng01}\\

 1.13.& Clustering based on Variable Neighborhood Search&
  \cite{hansen97a,hansen07,hansen09}\\

  1.14. & Neural networks based clustering &
  \cite{elt98,gab00,kam90} \\

  1.15. & Robust clustering &
  \cite{dave92,enr00,frig99,guha00,jolion91} \\

  1.16. & Clustering of structured objects &
   \cite{gus97,stepp86} \\


 II. & Fuzzy (soft) clustering problems/models & \\

 2.1.& Fuzzy clustering    &   \cite{bar99,grave10,hop99,jain99,kris95,miy90} \\

    &&   \cite{oli07,pedr05} \\

 2.2.&  Fuzzy k-means clustering  &  \cite{coppi02,dur06,hop99,ismail86}  \\

 2.3.&  Kernel-based fuzzy clustering   &  \cite{chen13,chia03,grave10,sato06}  \\

 2.4.& Fuzzy clustering for symbolic data/categorical data
 &\cite{el98,huang99,kim04} \\

 2.5.& Clustering based on hesitant fuzzy information & \cite{nchen13,nchen14,xu14a,zhang15}\\


 III.&Stochastic clustering & \\

 3.1.&Probabilistic clustering    &  \cite{bock96,bra91,jain99,kris95,lu07,taksar01}  \\

 3.2.&Probability-based graph partitioning, Markov random works&\cite{boley99,dutt96,lafon06,lov93,spiel04,szu01}\\

 3.3.&Cross-entropy method for clustering&\cite{debo05,kroe07,jung07,rub02,tabor14,tabor13}  \\


 IV. & Dynamic  clustering, online clustering, restructuring  & \\

 4.1. &Dynamic clustering & \cite{bens11,camp13,chenw04,diday73,kaneko94,yu07,zamir99} \\


 4.2. &Dynamic fuzzy clustering & \cite{crespo05,nguyen13,pal90,sato06a}\\

 4.3. & Online clustering & \cite{barb08,berin06,chan07,zhang04} \\

 4.4. & Restructuring in clustering (i.e., changing of clustering) & \cite{lev11restr,lev15}, this paper\\

 4.5. & Multistage clustering, cluster trajectories  & this paper\\


 V.& Very large clustering problems/models&\\

 5.1. &  Clustering of large data sets  &\cite{berk06,sheik00,zhang96,zup82}\\

 5.2. & K-means clustering for large data sets&\cite{huang98,huang05,jain10}\\


 VI.& Multiple clustering, framework-based clustering&\\

 6.1.& Multiple clustering, cluster ensembles, aggregation
 &\cite{abu13,ayad10,day86,ghosh11,gue11,he05,monti03}\\


  & clustering, consensus clustering
 &\cite{mull15,str02,vega11,yang06}   \\

 6.2.&Unified frameworks, parallel clustering, hybrid strategies&
   \cite{dahl00,ismail89,murtagh92,olson95,ozy09,sas14,zhong03} \\

\hline
\end{tabular}
\end{center}

\subsubsection{Systems problems}

  Generally, the following vital
   clustering system problems can be pointed out
 (e.g., \cite{ben99,fel11,jain88,jain99,mirkin05,sham04,shamir02,smith80,till02,vanr77,zait97})
 (Table 2.21).

\begin{center}
{\bf Table 2.21.}  Basic systems problems in clustering\\
\begin{tabular}{| c | l|l|}
\hline
  No. &Systems problem  &Some source(s)\\
\hline

 1.&Formulation/structuring of  clustering problem(s)
 &\cite{har75,jain88,jain99,mirkin79,mirkin96,mirkin05,vanr77}\\

 2.&Comparison of models/methods/techniques
 &\cite{meila01,rand71,stein00,zait97}\\

 3.&Selection/design of model/method/technique
 &\cite{cheungy05,fraley98,shamiro10}\\

 4.&Evaluation of clustering solution(s)
 &\cite{achtert07,bagon11,bansal02,bansal04,cgw03,cgw05,dem06}\\

   &&\cite{girvan02,kri09,new03,new06,swam04,zimek08}\\

 5.&Validation of clustering solution
 &\cite{bezdek98,dubes79,hal02,hal02a,lange04,mau02,murtagh92}\\

 6.&Stability of clustering solution
 &\cite{ben06,lange04,shamiro10,smith80}\\

 7.&Robustness of clustering solution
 &\cite{achtert07,chen13,dave92,enr00,frig99,guha00,jolion91}\\

 8.&Cluster editing, cluster graph modification,
 &\cite{boker09,boker11a,boker11,dama10,dehne06,guo09}\\

 &transformation of clustering solution
 &\cite{rah07,shamir02,sham04}\\

 9.&Identification/selection/assignment of cluster heads
 &\cite{butt09,chatt02,chenh06,chias04,hein02,jin05,till02,tripa12}\\
 &(e.g., sensor networks, mobile networks, target tracking)
 &\cite{tuba03,yang10} \\

 10.&Prospective clustering problems/approaches: &\\

 10.1.&Online clustering, clustering  data streams&  \cite{barb08,berin06,chan07,guha00a,last02,lev12clique}\\

 10.2.&Multiple clustering, consensus clustering
 &\cite{abu13,ayad10,day86,ghosh11,gue11,he05,monti03}\\

 10.3.&Hybrid (by methods, by data types) clustering methods&  \cite{cheu04,jans09,kamal81,kyper08,liu10,maji07,tilla05,tsai07}\\
 & composite, multistage/multilevel clustering methods&  \cite{surd05,wongm82,wong99,youn04}\\
 & (including adaptation modes)& \\

 10.4.&Expert knowledge based clustering/classification
 &\cite{clark94,clark98,fur13,fursok11,mur95,pedr05}\\

 & (including expert judgment based clustering)& \\

 10.5.&Multicriteria optimization clustering& \cite{desmet09,fer92,lef85,lev15,skou10,zop02} \\

 10.6.&Clustering with multi-type elements (each cluster is
 &this paper\\
 & composed by compatible elements of different types)& \\

 10.7.&Clustering with hesitant fuzzy sets data&\cite{xuz11,xu14a,nchen13} \\

 10.8.&Fast clustering methods
 &\cite{gar79,dorn94,jain99,new03,new06,new04}\\

\hline
\end{tabular}
\end{center}


 In addition,
 the following system stage can be  used:

~~~

  ``Modification of clustering process (if needed),
  for example, by the following ways:''~
%
%
 (a) modification of type(s) of element description(s),
 (b) modification of element parameters/features,
 (c) modification of element estimates types (or scales),
 (d) modification of criteria for inter-cluster proximity and intra-cluster proximity,
 (e) modification of criteria for quality of clustering solution(s),
 (f) modification of clustering method(s),
 and
 (g) searching for additional expert(s).


\subsubsection{General framework}

 Fig. 2.6 depicts an example of a clustering framework for multiple
 clustering.
 Another generalized clustering/classification framework
 (the viewpoint of a simplified information processing morphology )
 is presented in Fig. 2.34 (an extension of framework in Fig. 2.11):

~~

 {\it Stage 1.} Collection of initial data.

 {\it Stage 2.} Analysis of applied situation,
 problem structuring/formulation:
  (2.1) selection/generation of features/parameters/criteria,
  (2.2) definition and description (assessment)
  of the set of objects/items,
  (2.3) selection/design of basic clustering model(s).

 {\it Stage 3.} Preliminary data processing:
 (3.1)
  calculation of element proximities (distances),
  (3.2) definition of very close elements (i.e., definition of a small
  element proximity/distance),
  (3.3) definition of basic relation(s) over element set
  (a basic relation graph(s)),
  (3.4) revelation of basic preliminary groups of interconnected elements
  (i.e., some preliminary kernels of clusters).

 {\it Stage 4.} Basic clustering:
 (4.1) selection/definition of
 basic groups of interconnected elements
  (i.e., some candidates-clusters),
 (4.2) definition of the basic cluster set,
 (4.3.) extension of the basic cluster set
 (within framework of feedback).

 {\it Stage 5.} Classification (if needed):
 (5.1) assignment of elements into clusters,
 (5.2) multiple assignment of elements into clusters.

 {\it Stage 6.} Aggregation of cluster solutions
 (i.e., consensus clustering/clustering ensemble) (if needed).

 {\it Stage 7.} Analysis of clustering/classification results
 (clustering solution(s))
 (e.g., cluster validity).

 ~~

 Here,  a special support layer can include the following:
 {\it 1.} additional data,
 {\it 2.} expert(s) and expert(s) knowledge,
 {\it 3.} additional models (e.g.,
 vertex covering problem,
 assignment/matching problems,
 multiple criteria sorting/ranking problems,
 clique/quasi clique problem(s),
 multiple clique/quasi clique problem(s), median/consensus/agreement problems).

\begin{center}
\begin{picture}(153,92.5)

\put(035.5,00){\makebox(0,0)[bl] {Fig. 2.34. General framework
 for clustering/classification }}


\put(01,88){\makebox(0,0)[bl] {STAGE 1}}

\put(00,65){\line(1,0){17}} \put(00,86){\line(1,0){17}}
\put(00,65){\line(0,1){21}} \put(17,65){\line(0,1){21}}

 \put(17,74){\vector(1,0){04}}

\put(0.5,82){\makebox(0,0)[bl] {Collection}}
\put(0.5,79){\makebox(0,0)[bl] {of input}}
\put(0.5,76.5){\makebox(0,0)[bl] {data}}
\put(0.5,73){\makebox(0,0)[bl] {(elements,}}
\put(0.5,70){\makebox(0,0)[bl] {attributes,}}
\put(0.5,67){\makebox(0,0)[bl] {estimates)}}


\put(22,88){\makebox(0,0)[bl] {STAGE 2}}

\put(21,47){\line(1,0){17}} \put(21,86){\line(1,0){17}}
\put(21,47){\line(0,1){39}} \put(38,47){\line(0,1){39}}

 \put(38,74){\vector(1,0){04}}

\put(21.5,82){\makebox(0,0)[bl] {Problem}}
\put(21.5,79){\makebox(0,0)[bl] {analysis,}}
\put(21.5,76.5){\makebox(0,0)[bl] {statement}}
\put(21.5,73){\makebox(0,0)[bl] {(structu-}}
\put(21.5,70){\makebox(0,0)[bl] {ring):}}

\put(21.5,67){\makebox(0,0)[bl] {selection/}}
\put(21.5,64){\makebox(0,0)[bl] {design of}}
\put(21.5,61){\makebox(0,0)[bl] {criteria,}}
\put(21.5,58){\makebox(0,0)[bl] {evaluation,}}
\put(21.5,55){\makebox(0,0)[bl] {selection/}}
\put(21.5,52){\makebox(0,0)[bl] {design of}}
\put(21.5,49){\makebox(0,0)[bl] {model(s)}}

\put(45,88){\makebox(0,0)[bl] {STAGE 3}}

\put(42,34){\line(1,0){20}} \put(42,86){\line(1,0){20}}
\put(42,34){\line(0,1){52}} \put(62,34){\line(0,1){52}}

 \put(62,75){\vector(1,0){04}}

\put(43,82){\makebox(0,0)[bl] {Preliminary}}
\put(43,79){\makebox(0,0)[bl] {processing:}}
\put(43,76){\makebox(0,0)[bl] {proximity/}}
\put(43,73){\makebox(0,0)[bl] {distances,}}
\put(43,70){\makebox(0,0)[bl] {close }}
\put(43,67){\makebox(0,0)[bl] {element}}
\put(43,64){\makebox(0,0)[bl] {pairs,}}
\put(43,61){\makebox(0,0)[bl] {relations}}
\put(43,58){\makebox(0,0)[bl] {over}}
\put(43,55){\makebox(0,0)[bl] {elements,}}
\put(43,52){\makebox(0,0)[bl] {basic }}
\put(43,49){\makebox(0,0)[bl] {groups of}}
\put(43,46){\makebox(0,0)[bl] {close }}
\put(43,43){\makebox(0,0)[bl] {elements}}
\put(43,40){\makebox(0,0)[bl] {(as cluster}}
\put(43,37){\makebox(0,0)[bl] {kernels)}}


\put(68,88){\makebox(0,0)[bl] {STAGE 4}}

\put(66,64){\line(1,0){18}} \put(66,86){\line(1,0){18}}
\put(66,64){\line(0,1){22}} \put(84,64){\line(0,1){22}}

\put(84,75){\vector(1,0){04}}

\put(70,82){\makebox(0,0)[bl] {Basic}}
\put(67,78.5){\makebox(0,0)[bl] {clustering}}
\put(67,75.5){\makebox(0,0)[bl] {(including}}
\put(67.5,73){\makebox(0,0)[bl] {definition}}
\put(69,70){\makebox(0,0)[bl] {of basic }}
\put(70,67){\makebox(0,0)[bl] {cluster }}
\put(70.5,64){\makebox(0,0)[bl] {set(s)) }}

\put(74,60){\vector(0,1){04}}

\put(66,45){\line(1,0){18}} \put(66,60){\line(1,0){18}}
\put(66,45){\line(0,1){15}} \put(84,45){\line(0,1){15}}

\put(66.5,45.5){\line(1,0){17}} \put(66.5,59.5){\line(1,0){17}}
\put(66.5,45.5){\line(0,1){14}} \put(83.5,45.5){\line(0,1){14}}

\put(67,55.5){\makebox(0,0)[bl] {Extension/}}
\put(67,53){\makebox(0,0)[bl] {correction}}
\put(69,50){\makebox(0,0)[bl] {of basic }}
\put(67,47){\makebox(0,0)[bl] {cluster set}}

\put(92,88){\makebox(0,0)[bl] {STAGE 5}}

\put(88,64){\line(1,0){23}} \put(88,86){\line(1,0){23}}
\put(88,64){\line(0,1){22}} \put(111,64){\line(0,1){22}}
\put(88.5,64){\line(0,1){22}} \put(110.5,64){\line(0,1){22}}


\put(111,75){\vector(1,0){04}}

\put(89,81.5){\makebox(0,0)[bl] {Classification:}}
\put(91,78.5){\makebox(0,0)[bl] {assignment }}
\put(92,75){\makebox(0,0)[bl] {(multiple }}
\put(90.5,72){\makebox(0,0)[bl] {assignment) }}
\put(90.5,69.5){\makebox(0,0)[bl] {of elements}}
\put(90,66.6){\makebox(0,0)[bl] {into clusters}}

\put(116.5,88){\makebox(0,0)[bl] {STAGE 6}}

\put(132,75){\vector(1,0){04}}

\put(115,64){\line(1,0){17}} \put(115,86){\line(1,0){17}}
\put(115,64){\line(0,1){22}} \put(132,64){\line(0,1){22}}
\put(115.5,64){\line(0,1){22}} \put(131.5,64){\line(0,1){22}}


\put(116,82){\makebox(0,0)[bl] {}}

\put(116,79){\makebox(0,0)[bl] {Aggrega-}}
\put(116,76.5){\makebox(0,0)[bl] {tion of }}
\put(116,73){\makebox(0,0)[bl] {clustering }}
\put(116,70.5){\makebox(0,0)[bl] {solutions}}
\put(116,67){\makebox(0,0)[bl] {}}


\put(137,88){\makebox(0,0)[bl] {STAGE 7}}

\put(136,64){\line(1,0){17}} \put(136,86){\line(1,0){17}}
\put(136,64){\line(0,1){22}} \put(153,64){\line(0,1){22}}

\put(138,82){\makebox(0,0)[bl] {}}

\put(137.5,79){\makebox(0,0)[bl] {Analysis }}
\put(136.5,76.5){\makebox(0,0)[bl] {of results:}}
\put(136.5,73){\makebox(0,0)[bl] {clustering }}
\put(136.5,70){\makebox(0,0)[bl] {solution(s)}}
\put(136.5,67){\makebox(0,0)[bl] {}}


\put(141,64){\line(0,-1){12}} \put(141,52){\vector(-1,0){57}}

\put(143,64){\line(0,-1){24}} \put(143,40){\vector(-1,0){81}}

\put(146,64){\line(0,-1){31}} \put(146,33){\line(-1,0){116.5}}
\put(29.5,33){\vector(0,1){14}}

\put(148,64){\line(0,-1){32}} \put(148,32){\line(-1,0){139.5}}
\put(08.5,32){\vector(0,1){33}}

\put(18,26){\vector(0,1){4}} \put(33,26){\vector(0,1){4}}
\put(48,26){\vector(0,1){4}} \put(63,26){\vector(0,1){4}}
\put(78,26){\vector(0,1){4}} \put(93,26){\vector(0,1){4}}
\put(108,26){\vector(0,1){4}} \put(123,26){\vector(0,1){4}}
\put(138,26){\vector(0,1){4}}

\put(08,6){\line(1,0){140}} \put(08,26){\line(1,0){140}}
\put(08,6){\line(0,1){20}} \put(148,6){\line(0,1){20}}

\put(16,22){\makebox(0,0)[bl]{SUPPORT LAYER:}}
\put(16,19){\makebox(0,0)[bl]{1. Additional data}}
\put(16,15.5){\makebox(0,0)[bl]{2. Expert(s) and expert(s)
 knowledge}}

\put(16,12.5){\makebox(0,0)[bl]{3. Additional models:
 vertex covering problem, assignment/matching problems,}}

\put(16,09.5){\makebox(0,0)[bl]{multiple criteria sorting/ranking
 problems, clique/quasi clique problem(s)}}

\put(16,06.5){\makebox(0,0)[bl] {multiple clique/quasi clique
 problem(s), median/consensus/agreement problems}}

\end{picture}
\end{center}

 Note, the following can be used as alternative morphological components:

 (i) various problem analysis and formulation approaches
  (e.g., selection of well-known problem statement(s),
 design of a new problem formulation(s)),
 (ii) various element metrics/proximities,
 (iii) various cluster metrics/proximities,
 (iv) various item assessment techniques
 (e.g., usage of statistical data, usage of expert-based
 techniques),
 (v) various clustering methods,
 (v) various clustering solution aggregation methods.

\subsubsection{Example of solving morphological scheme}

 A simplified example of morphological scheme
 for clustering process presented in Fig. 2.34
 (an analogue of composite strategy for multicriteria
 ranking/sorting problem
 \cite{lev12b,lev15})
  is the following (Fig. 2.35):

~~

 {\bf 0.} Compressed solving framework \(S = H \star P \star M \star G \star Q \) :

  {\bf 1.} Analysis of situation,
 problem statement
  and structuring
 (i.e., parameters/criteria, scales, etc.),
  assessment~ \(H = X \star Y\) (stage 3):

  (1.1) problem formulation~  \(X\):
  classification (``hard'') \(X_{1}\),
  classification (``soft'') \(X_{2}\),
  clustering (i.e., partitioning) (``hard'') \(X_{3}\),
  clustering (i.e., partitioning) (``soft'') \(X_{4}\),
  sorting (``hard'') \(X_{5}\),
  sorting (``soft'') \(X_{6}\),
  a composite problem   \(X_{7}\).

 (1.2) assessment of objects/items~   \(Y\):
   usage of statistical data \(Y_{1}\),
   expert based procedures \(Y_{2}\),
   statistical data and expert based procedures \(Y_{3} = Y_{1}\&Y_{2}\).

 {\bf 2.} Criteria/proximities and preliminary processing~ \(P = U \star V\) (stage 3):

  (2.1) proximity/metric for element pair
  (i.e., similarity measure)~  \(U\):
   Euclidean distance (\(L_{2}\))   \(U_{1}\),
  ordinal estimate \(U_{2}\),
 multicriteria estimate \(U_{3}\),
 interval multiset estimate  \(U_{4}\),


 (2.2)
 intra-cluster quality
 (criterion  for intra-cluster ``distance'', to minimize)~
    \(V\):
 maximum of element pair proximity (single link) \(V_{1}\),
 maximum of all element pair proximities
 (all links or average link) \(V_{2}\);


 (2.3) criterion for inter-cluster ``distance'' (to maximize)~   \(W\):
   minumum ``distance'' between clusters  \(W_{1}\),
    average ``distance'' between clusters  \(W_{2}\).

  {\bf 3.} Clustering method/model~ \(M\) (stage 4):
   hierarchical clustering  \(M_{1}\),
   \(K\)-means clustering \(M_{2}\),
   spanning tree based clustering \(M_{3}\),
   graph method based on detection of cliques/quasi-cliques
   \(M_{4}\),
   correlation clustering
   \(M_{5}\),
   composite method (parallel processing)
    \(M_{6} = M_{2} \& M_{3} \& M_{4}\),
   composite method (parallel processing)
   \(M_{7} = M_{2} \& M_{4} \& M_{5}\).

 {\bf 4.} Aggregation of clustering solutions~ \(G\) (stage 6):
   none   \(G_{1}\),
   median-based solving process \(G_{2}\),
   extension of common clustering solution part (i.e., a solution kernel)  \(G_{3}\).

 {\bf 5.} Analysis of resultant clustering solution(s)
 (i.e., cluster validity)~ \(Q\) (stage 7):
   none   \(Q_{1}\),
   expert-based  process \(Q_{2}\),
   special
    calculation procedure(s)
   \(Q_{3}\).


\begin{center}
\begin{picture}(112.5,75)

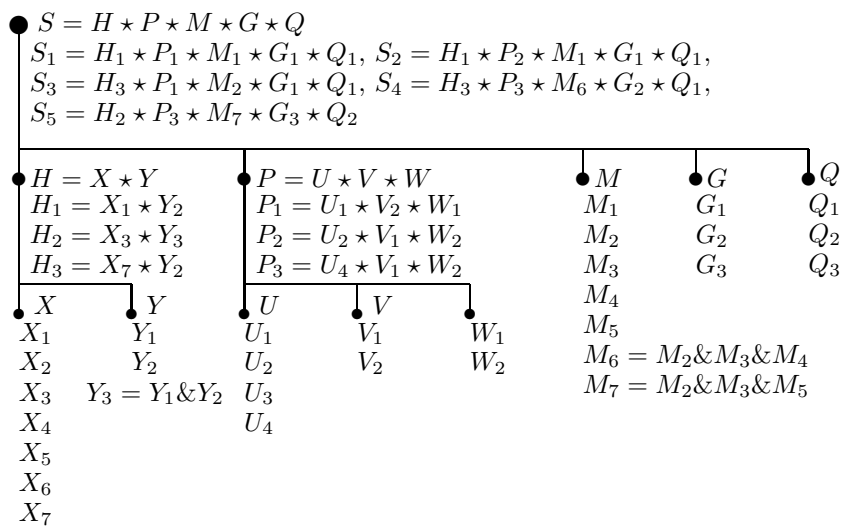
\put(08,00){\makebox(0,0)[bl] {Fig. 2.35. Illustration for
 morphological scheme of clustering }}

\put(00,55){\line(0,1){16.5}} \put(00,71.5){\circle*{2.5}}

\put(02.5,70){\makebox(0,0)[bl]{\(S =
 H \star P \star M \star G \star Q \)}}

\put(01.5,66){\makebox(0,0)[bl]{\(S_{1} =
  H_{1} \star P_{1}  \star M_{1}\star G_{1}\star Q_{1}\),

  \(S_{2} =
  H_{1} \star P_{2}  \star M_{1}\star G_{1}\star Q_{1}, \)}}

\put(01.5,62){\makebox(0,0)[bl]{\(S_{3} =
 H_{3} \star P_{1} \star M_{2}\star G_{1}\star Q_{1} \),
  \(S_{4} =
 H_{3} \star P_{3} \star  M_{6}
 \star G_{2}\star Q_{1} \), }}

\put(01.5,58){\makebox(0,0)[bl]{\(S_{5} =
 H_{2} \star P_{3} \star  M_{7}
 \star G_{3}\star Q_{2} \)}}

\put(00,55){\line(1,0){105}}

\put(00,51){\line(0,1){04}}

\put(00,37){\line(0,1){14}} \put(00,51){\circle*{1.9}}

\put(01.5,50){\makebox(0,0)[bl]{\(H = X \star Y\)}}
\put(01.5,46){\makebox(0,0)[bl]{\(H_{1} = X_{1} \star Y_{2}\)}}
\put(01.5,42){\makebox(0,0)[bl]{\(H_{2} = X_{3} \star Y_{3}\)}}
\put(01.5,38){\makebox(0,0)[bl]{\(H_{3} = X_{7} \star Y_{2}\)}}


\put(00,33){\line(0,1){04}} \put(00,33){\circle*{1.5}}
\put(15,33){\line(0,1){04}} \put(15,33){\circle*{1.5}}
\put(00,37){\line(1,0){15}}

\put(02,33){\makebox(0,0)[bl]{\(X\)}}
\put(17,33){\makebox(0,0)[bl]{\(Y\)}}

\put(00,29){\makebox(0,0)[bl]{\(X_{1}\)}}
\put(00,25){\makebox(0,0)[bl]{\(X_{2}\)}}
\put(00,21){\makebox(0,0)[bl]{\(X_{3}\)}}
\put(00,17){\makebox(0,0)[bl]{\(X_{4}\)}}
\put(00,13){\makebox(0,0)[bl]{\(X_{5}\)}}
\put(00,09){\makebox(0,0)[bl]{\(X_{6}\)}}
\put(00,05){\makebox(0,0)[bl]{\(X_{7}\)}}

\put(15,29){\makebox(0,0)[bl]{\(Y_{1}\)}}
\put(15,25){\makebox(0,0)[bl]{\(Y_{2}\)}}
\put(09,21){\makebox(0,0)[bl]{\(Y_{3}= Y_{1}\&Y_{2}\)}}


\put(30,51){\line(0,1){04}}

\put(30,37){\line(0,1){14}} \put(30,51){\circle*{1.9}}

\put(31.5,50){\makebox(0,0)[bl]{\(P = U \star V \star W\)}}
\put(31.5,46){\makebox(0,0)[bl]{\(P_{1} = U_{1} \star V_{2} \star
 W_{1} \)}}

\put(31.5,42){\makebox(0,0)[bl]{\(P_{2} = U_{2} \star V_{1} \star
 W_{2}\)}}

\put(31.5,38){\makebox(0,0)[bl]{\(P_{3} = U_{4} \star V_{1} \star
 W_{2}\)}}


\put(30,33){\line(0,1){04}} \put(30,33){\circle*{1.5}}
\put(45,33){\line(0,1){04}} \put(45,33){\circle*{1.5}}
\put(60,33){\line(0,1){04}} \put(60,33){\circle*{1.5}}
\put(30,37){\line(1,0){30}}

\put(32,33){\makebox(0,0)[bl]{\(U\)}}
\put(47,33){\makebox(0,0)[bl]{\(V\)}}

\put(30,29){\makebox(0,0)[bl]{\(U_{1}\)}}
\put(30,25){\makebox(0,0)[bl]{\(U_{2}\)}}
\put(30,21){\makebox(0,0)[bl]{\(U_{3}\)}}
\put(30,17){\makebox(0,0)[bl]{\(U_{4}\)}}

\put(45,29){\makebox(0,0)[bl]{\(V_{1}\)}}
\put(45,25){\makebox(0,0)[bl]{\(V_{2}\)}}

\put(60,29){\makebox(0,0)[bl]{\(W_{1}\)}}
\put(60,25){\makebox(0,0)[bl]{\(W_{2}\)}}

\put(75,51){\line(0,1){04}}

\put(75,51){\circle*{1.9}} \put(76.5,50){\makebox(0,0)[bl]{\(M\)}}

\put(75,46){\makebox(0,0)[bl]{\(M_{1}\)}}
\put(75,42){\makebox(0,0)[bl]{\(M_{2}\)}}
\put(75,38){\makebox(0,0)[bl]{\(M_{3}\)}}
\put(75,34){\makebox(0,0)[bl]{\(M_{4}\)}}
\put(75,30){\makebox(0,0)[bl]{\(M_{5}\)}}

\put(75,26){\makebox(0,0)[bl]{\(M_{6} =
  M_{2} \& M_{3} \& M_{4}\)}}

 \put(75,22){\makebox(0,0)[bl]{\(M_{7} =
  M_{2} \& M_{3} \& M_{5}\)}}


\put(90,51){\line(0,1){04}}

\put(90,51){\circle*{1.9}} \put(91.5,50){\makebox(0,0)[bl]{\(G\)}}

\put(90,46){\makebox(0,0)[bl]{\(G_{1}\)}}
\put(90,42){\makebox(0,0)[bl]{\(G_{2}\)}}
\put(90,38){\makebox(0,0)[bl]{\(G_{3}\)}}


\put(105,51){\line(0,1){04}}

\put(105,51){\circle*{1.9}}
\put(106.5,50){\makebox(0,0)[bl]{\(Q\)}}

\put(105,46){\makebox(0,0)[bl]{\(Q_{1}\)}}
\put(105,42){\makebox(0,0)[bl]{\(Q_{2}\)}}
\put(105,38){\makebox(0,0)[bl]{\(Q_{3}\)}}

\end{picture}
\end{center}

 Thus,  five illustrative alternative examples of the composite
 (series-parallel)
 solving strategies for clustering are the following (Fig. 2.35):

 \(S_{1} = H_{1} \star P_{1}  \star M_{1}\star G_{1}\star Q_{1}  = \)
 \( ( X_{1}\star Y_{2} ) \star ( U_{1} \star V_{2} \star W_{1} )
  \star M_{1}\star G_{1}\star Q_{1}\);

 \(S_{2} = H_{1} \star P_{2}  \star M_{1}\star G_{1}\star Q_{1} = \)
  \( ( X_{1}\star Y_{2} ) \star ( U_{2} \star V_{1} \star W_{2} )
  \star M_{1}\star G_{1}\star Q_{1}\);

 \(S_{3} = H_{3} \star P_{1} \star M_{2}\star G_{1}\star Q_{1} = \)
 \( ( X_{7}\star Y_{2} ) \star ( U_{1} \star V_{2} \star W_{1} )
  \star M_{2}\star G_{1}\star Q_{1}\);

  \(S_{4} = H_{3} \star P_{3} \star M_{6}\star G_{2}\star Q_{1} = \)
 \( ( X_{7}\star Y_{2} ) \star ( U_{4} \star V_{1} \star W_{2} )
  \star ( M_{2} \& M_{3} \& M_{4} ) \star G_{2}\star Q_{1}\);
  and

 \(S_{5} = H_{2} \star P_{3} \star M_{7}\star G_{3}\star Q_{2} = \)
  \( ( X_{3}\star ( Y_{1} \& Y_{2} ) ) \star ( U_{4} \star V_{1} \star W_{2} )
  \star ( M_{2} \& M_{3} \& M_{5} ) \star G_{3}\star Q_{1}\).

 In Fig. 2.36,
 a graphical illustration for three composite strategies
 above is depicted.

\begin{center}
\begin{picture}(106,42)

\put(13.5,00){\makebox(0,0)[bl] {Fig. 2.36. Examples of composite
  solving strategies}}


\put(00,36){\makebox(0,0)[bl]{\(S_{2}:\)}}

\put(10,36){\makebox(0,0)[bl]{\(X_{1}\)}}
\put(16,38){\vector(1,0){05}}

\put(23,36){\makebox(0,0)[bl]{\(Y_{2}\)}}
\put(29,38){\vector(1,0){05}}


\put(36,36){\makebox(0,0)[bl]{\(U_{1}\)}}
\put(42,38){\vector(1,0){05}}

\put(49,36){\makebox(0,0)[bl]{\(V_{2}\)}}
\put(55,38){\vector(1,0){05}}

\put(62,36){\makebox(0,0)[bl]{\(W_{2}\)}}
\put(68,38){\vector(1,0){05}}


\put(75,36){\makebox(0,0)[bl]{\(M_{1}\)}}
\put(81,38){\vector(1,0){05}}

\put(88,36){\makebox(0,0)[bl]{\(G_{1}\)}}
\put(94,38){\vector(1,0){05}}

\put(101,36){\makebox(0,0)[bl]{\( Q_{1}\)}}


\put(00,25){\makebox(0,0)[bl]{\(S_{4}:\)}}


\put(10,25){\makebox(0,0)[bl]{\(X_{7}\)}}
\put(16,27){\vector(1,0){05}}

\put(23,25){\makebox(0,0)[bl]{\(Y_{2}\)}}
\put(29,27){\vector(1,0){05}}



\put(36,25){\makebox(0,0)[bl]{\(U_{4}\)}}
\put(42,27){\vector(1,0){05}}

\put(49,25){\makebox(0,0)[bl]{\(V_{2}\)}}
\put(55,27){\vector(1,0){05}}

\put(62,25){\makebox(0,0)[bl]{\(W_{2}\)}}

\put(68,28){\vector(2,1){05}} \put(68,27){\vector(1,0){05}}
\put(68,26){\vector(2,-1){05}}


\put(75,29){\makebox(0,0)[bl]{\(M_{2}\)}}
\put(81,31){\vector(2,-1){05}}

\put(75,25){\makebox(0,0)[bl]{\(M_{3}\)}}
\put(81,27){\vector(1,0){05}}

\put(75,21){\makebox(0,0)[bl]{\(M_{4}\)}}
\put(81,23){\vector(2,1){05}}


\put(88,25){\makebox(0,0)[bl]{\(G_{2}\)}}
\put(94,27){\vector(1,0){05}}

\put(101,25){\makebox(0,0)[bl]{\( Q_{1}\)}}


\put(00,10){\makebox(0,0)[bl]{\(S_{5}:\)}}


\put(10,10){\makebox(0,0)[bl]{\(X_{3}\)}}



\put(16,13){\vector(2,1){05}} \put(16,11){\vector(2,-1){05}}


\put(23,14){\makebox(0,0)[bl]{\(Y_{1}\)}}
\put(29,16){\vector(2,-1){05}}

\put(23,06){\makebox(0,0)[bl]{\(Y_{2}\)}}
\put(29,08){\vector(2,1){05}}



\put(36,10){\makebox(0,0)[bl]{\(U_{4}\)}}
\put(42,12){\vector(1,0){05}}

\put(49,10){\makebox(0,0)[bl]{\(V_{2}\)}}
\put(55,12){\vector(1,0){05}}

\put(62,10){\makebox(0,0)[bl]{\(W_{2}\)}}

\put(68,13){\vector(2,1){05}} \put(68,12){\vector(1,0){05}}
\put(68,11){\vector(2,-1){05}}


\put(75,14){\makebox(0,0)[bl]{\(M_{2}\)}}
\put(81,16){\vector(2,-1){05}}

\put(75,10){\makebox(0,0)[bl]{\(M_{3}\)}}
\put(81,12){\vector(1,0){05}}

\put(75,06){\makebox(0,0)[bl]{\(M_{5}\)}}
\put(81,08){\vector(2,1){05}}


\put(88,10){\makebox(0,0)[bl]{\(G_{3}\)}}
\put(94,12){\vector(1,0){05}}

\put(101,10){\makebox(0,0)[bl]{\( Q_{1}\)}}

\end{picture}
\end{center}

\subsection{On clustering in large scale data sets/networks}

 In recent years,
 the significance of clustering in large-scale data bases and
 analysis and modeling in large networks has been increased,
 for example:

 (i)  clustering of large data sets
 (e.g., \cite{berk06,huang98,huang05,jain10,sheik00,zhang96,zup82});

 (ii) detection of communities in large networks
  (e.g.,
  \cite{clay04,gop13,hopcroft03,hopcroft04,les09,pons06,yangj15});

 (iii) detection of communities in mega-scale networks
  (e.g., \cite{blond08,wak07});

 (iv) tracking evolving communities in large networks
 (e.g., \cite{hopcroft04}).

 Table 2.22 illustrates some dimensional layers (classification)
 of data sets/networks.

\begin{center}
{\bf Table 2.22.} Dimensional layers of data sets/networks \\
\begin{tabular}{|c|l|c|l|l|}
\hline
 No.&Type of studied data&Number of&Examples of &Some \\
    &sets/networks       &objects/&applications&source(s)\\
    &                    &network nodes & &  \\

\hline

 1.&Simplified data sets/&\(\sim\)~10...60&(i) student group, &\cite{zach77}  \\
  &networks                             &&(ii) sport club network,&  \\
  &(e.g., small groups)                 &&(iii) laboratory group,  &  \\
  &                                     &&(iv) Web page structure,   &  \\
  &                                     &&(v) product assortment&  \\
  &                                     &&(product variety)&  \\

 2.&Simple data sets/ &\(\sim\)~100~&(i) university department, & \cite{new04} \\
    &networks                      &&(ii) animal network, &  \\
    &                              &&(iii) big firm department, &\\
    &                              &&(iv) department of&  \\
    &                              && government organization,&  \\
    &                              &&(v) network of  books/articles&\\
    &                              &&  (close by topic(s))       & \\
    &                              &&(vi) social network of& \\
    &                              &&bottlenose dolphins,  & \\
    &                              &&(vii) supply chain network,&  \\
    &                              &&(viii) network of software&\\
    &                              && system components,&\\
    &                              &&(ix) molecular structures, &\\
    &                              &&(x) manufacturing structures &\\

 3.&Traditional data&\(\sim\)~1~k&(i) citation networks,&\cite{girvan02}\\
   &sets/networks               &&(ii) university network,&  \\
   &                            &&(iii) collaboration network,&\\
   &                            &&(iv) urban systems,& \\
   &                            &&(v) consumers bases,    &\\
   &                            &&(vi) multiple server   &\\
   &                            &&  computer systems      &\\

 4.&Large data sets/&\(\sim\)~10~k&(i) research society network,&\cite{new04}\\
   &networks                     &&(ii) sensor networks,  &\\
   &                             &&(iii) manufacturing       &\\
   &                             && technology networks,&  \\
   &                             &&(iv) Microarrays     &\\

 5.&Very large data&\(\sim\)~100~k&(i) client bases,&\cite{clay04} \\
   &sets/networks                &&(ii) VLSI,  &  \\
   &                             &&(iii)  medical patients bases,    &\\

 6.&Mega-scale  data&\(\sim\)~ 1~M&(i) university library, &\cite{wak07} \\
   &sets/networks                &&(ii) bases of editorial houses,  &\\

 7.&Super-scale data&\(\sim\)~10~M&(i) library networks,&\cite{blond08}\\
   & sets/networks               &&(ii) Internet-based shops,&  \\
   &                             &&(iii) protein sequence databases &  \\

 8.&To-day's/prospective&\(\sim\)~100~M ...1~B&(i) World Wide Web,&\\
   &Web-based                         &&(ii) social networks&\\
   &data sets/networks                &&(e.g., Twitter, Facebook)&\\

\hline
\end{tabular}
\end{center}

\subsection{Note on multidimensional scaling}

 Many decades, multidimensional scaling approach is widely
 used in many domains
 (e.g., \cite{borg05,carrol80,cox00,david83,torg52,yuang13}).
 Here, an initial space of object parameters is transformed
 and simplified (by increasing its dimension, on the basis of optimization).
 As a result, obtained clusters are more ``good''.
 Table 2.23 contains some basic directions
 in multidimensional scaling researches.

\newpage

\begin{center}
 {\bf Table 2.23.}
 Directions in multidimensional scaling
 (methods, clustering, CS applications)\\
\begin{tabular}{| c | l |l|}
\hline
 No.  & Approaches, models & Source(s)  \\
\hline

 1.& Basic methods in multidimensional scaling:  &\\

 1.1.& Multidimensional scaling, general
 &  \cite{borg05,carrol80,cox00,david83,green89,krus64,torg52} \\

 1.2.& Nonmetric multidimensional scaling
 &  \cite{holman78,krus64a} \\

 1.3.& Least-squares multidimensional scaling
  &\cite{groen99}\\

 1.4.& Application of convex analysis to  multidimensional scaling
 & \cite{leew77} \\

 1.5.&Global optimization in multidimensional scaling
 &\cite{groen96,groen96}\\

 1.6.& Probabilistic multidimensional scaling & \cite{zin83} \\

 1.7.& Genetic algorithms, evolutionary methods  & \cite{tec12} \\
  &in multidimensional scaling &  \\

 1.8.& Multigrid multidimensional scaling  & \cite{bron06} \\

 1.9.& Configural synthesis in multidimensional scaling
 & \cite{green72}\\

 1.10.& Functional approach to data structure & \cite{cha88}\\
  & in multidimensional scaling& \\

 1.11.& Convergence of methods in multidimensional scaling
 & \cite{leew88}\\

 1.12.& Distributed multidimensional scaling
 &\cite{costa06}\\



\hline

 2. &Multidimensional clustering: &\\

 2.1.& Multidimensional clustering algorithms & \cite{murtagh85}\\

 2.2. &Multidimensional scaling and data clustering:&\\

 2.2.1.& Multidimensional scaling and data clustering&
  \cite{hof94}\\

 2.2.2.& Multidimensional scaling: tree-fitting, and clustering&
 \cite{shep80}\\

 2.3.& Multidimensional data clustering utilizing&
 \cite{ismail89}\\
  &hybrid search strategies&\\

\hline

  3.& Contemporary applications in CS: &\\

  3.1.& Multidimensional clustering in data mining &
 \cite{berk06}\\

 3.2.& Graph drawing by multidimensional scaling
 &\cite{klim13}\\

 3.3. & Visualization
  &\cite{wus05}\\

 3.4. &Multidimensional scaling in communication, sensor
 & \cite{cheung05,costa06,lat07}\\
   & networks (node localization, location, positioning, etc.)&\\


\hline
\end{tabular}
\end{center}

\newpage
\section{Basic Structured Clustering
 Combinatorial Schemes}


%

\subsection{Auxiliary problems}

 Table 3.1 contains a list of main auxiliary problems
 for combinatorial clustering methods/procedures.

\begin{center}
{\bf Table 3.1.}  Auxiliary problems \\
\begin{tabular}{|c|l|l|l|l|}
\hline
  No. &Problem &Clustering &Solving schemes &Some\\
  & &model(s)/stage(s) &  &source(s) \\
\hline

 1.&Transformation of &Data processing&1.Calculation& \\

 &scales&& 2.Expert judgment & \\


 2.&Calculation of &Main clustering& 1.Direct calculation &\\

 &proximity matrix&schemes& 2.Calculation with &\\

 &  &&scale transformation & \\

 3.&Multicriteria ranking/ &1.Data processing&1.Utility function &\cite{bou09,fis70,kee76,lev12b,lev15} \\
   &sorting&2.Clustering&2.Pareto approach& \cite{pareto71,roy96,zop02} \\
   &            &&3.Outranking technique& \\
   &     &       &4.Expert judgment, etc.& \\

 4.&Minimum spanning & Graph-based
 & 1.Kruskal's algorithms
 & \cite{aho74,cormen90,duhand99,gabow86,gar79} \\

 &tree&clustering& 2.Boruvka's algorithms
 & \cite{voss09,wang09,pet02,xu01,yao75} \\
 5.&Knapsack-like problems& k-means
 & 1.Dynamic programing& \cite{duhand99,gar79,ibarra75,keller04,mar90}\\

 &(basic problem, multi-&clustering,&2.Approximation
 & \cite{sahni75} \\

 &ple choice problem,&restructuring&(e.g., FPTAS)& \\

  &multicriteria problems)&&3.Heuristics& \\

 6.&Assignment problems&k-means
 & 1.Fast algorithms&
 \cite{alev09,burkard09,cela98,duhand99,gar79} \\

 &(basic problem,&clustering& 2.Heuristics
 & \cite{hub87,kearns98,kuh57,lev15,levpet10}\\

 &generalized problem,&&3.Enumerative methods
 & \cite{pardalos94,pardalos00,pit09,roy06,scar02}\\

  &multicriteria problem)&&& \\

 7.&Covering problems &Set covering
 &1.Enumerative methods
 & \cite{agrawal05,mul09,salz09}\\

 &&based clustering&2.Heuristics &\\

 8.&Dominating sets &Dominating set
 &1.Enumerative methods
 &\cite{chen02,han07,li06,pav07,youn06}\\

   &problem&based clustering&2.Heuristics &\\

9.&Partitioning problems:&&&\\

9.1.&Graph partitioning& Partitioning &1.Approximation
 &\cite{condon01,ding01,even99,ker70,spiel13}\\

 & &based clustering
 &2.Heuristics&
 \cite{xux07}\\

 9.2.&Graph &Correlation
 &1.Enumerative methods& \cite{achtert07,bagon11,bansal02,bansal04,nchen14}\\

 &partitioning &clustering&2.Heuristics& \cite{dem06,kri09,swam04,zimek08}\\
 10.&Communities detection:&&&\\

 10.1.&Detection of clique/&Clique based
 & 1.Enumerative methods& \cite{abe02,but06,dawande01,duan12,duhand99,gar79}\\

 &quasi-clique&clustering& 2.Heuristics& \cite{gra05,johnson96,koch05,mehrot98}\\
 10.2.&Detection of network&Communities
 & 1.Enumerative methods& \cite{agar08,girvan02,less01,new03,new04a}\\

 &communities&detection& 2.Heuristics& \cite{new06,new10,new04,paas14,port09}\\
 13.&Finding agreement/&Consensus &1.Approximation& \\
   &median/consensus for:        &clustering&2.Heuristics& \\

 13.1.&Partitions     & && \cite{ayad10,bar86,chen11,han13,hub85}\\
 &&&& \cite{monti03,mull15,str02,topchy04} \\

 13.2.&Rankings       & && \cite{beck83,cook92,cook96,lev15,tavana03}\\
 13.3.&Trees          & && \cite{akatsu00,amir97,farach95,wang01}\\

 14.&Morphological clique&Clustering of&1.Heuristic
    &\cite{dawande01,lev98,lev06,lev12morph,lev15}\\

  & problem &multi-type objects &2.Enumerative methods&\\


\hline
\end{tabular}
\end{center}


  Fig. 3.1 depicts main stages for processing of initial data:

 1. Collection of initial information,
  i.e., set of objects \(A = \{A_{1},...,A_{\iota},...,A_{n}\}\),
  set of parameters \(C = \{C_{1},...,C_{i},...,C_{m}\}\),
  estimates of objects upon parameters
  \(x_{\iota} = (x_{\iota,1},...,x_{\iota,i},...,x_{\iota,m} )\)
  (\(\iota = \overline{1,n}\)).

 2. Calculation of distance matrix
 \(Z = || z_{\iota_{1},\iota_{2}} ||\)
 (\(\iota_{1} = \overline{1,n}, \iota_{2} = \overline{1,n}\));
 (usually: complexity estimate equals \(O(m \times n^{2})\)).

 3. Transformation of distance matrix into a spanning graph
 (if needed):~

 \(Z = || z_{\iota_{1},\iota_{2}} ||  \Longrightarrow
 G(A,\Gamma)\);
%
%
 (usually: complexity estimate equals \(O(n^{2})\)).

\begin{center}
\begin{picture}(104,36.5)

\put(23,00){\makebox(0,0)[bl] {Fig. 3.1. Preliminary
 data processing}}

\put(01,31){\makebox(0,0)[bl]{Collection of }}
\put(01,28){\makebox(0,0)[bl]{information}}
\put(01,25){\makebox(0,0)[bl]{from }}
\put(01,22){\makebox(0,0)[bl]{databases,}}
\put(01,19){\makebox(0,0)[bl]{engineering/}}
\put(01,16){\makebox(0,0)[bl]{computing}}
\put(01,13){\makebox(0,0)[bl]{experiments,}}
\put(01,10){\makebox(0,0)[bl]{expert }}
\put(01,07){\makebox(0,0)[bl]{judgment}}

\put(00,05){\line(1,0){22}} \put(00,35){\line(1,0){22}}
\put(00,05){\line(0,1){30}} \put(22,05){\line(0,1){30}}

\put(22,20){\vector(1,0){4}}


\put(27,31){\makebox(0,0)[bl]{Basic data:}}
\put(27,27){\makebox(0,0)[bl]{1.Objects}}
\put(27,23){\makebox(0,0)[bl]{\(A = \{A_{1},...,A_{n} \}\)}}
\put(27,19){\makebox(0,0)[bl]{2.Parameters}}
\put(27,15){\makebox(0,0)[bl]{\(C = \{C_{1},...,C_{m}\}\)}}
\put(27,11){\makebox(0,0)[bl]{3.Estimates
 (\(\forall A_{\iota}\))}}
\put(27,07){\makebox(0,0)[bl]{\(x_{\iota} =
  (x_{\iota,1},...,x_{\iota,m})\)}}

\put(26,05){\line(1,0){30}} \put(26,35){\line(1,0){30}}
\put(26,05){\line(0,1){30}} \put(56,05){\line(0,1){30}}

\put(56,18.5){\vector(1,0){4}}


\put(61,27){\makebox(0,0)[bl]{Calculation}}
\put(61,23){\makebox(0,0)[bl]{of distance/}}
\put(61,19){\makebox(0,0)[bl]{proximity}}
\put(61,15){\makebox(0,0)[bl]{matrix}}
\put(61,10){\makebox(0,0)[bl]{\(Z =||z_{\iota_{1},\iota_{2}}||\)}}

\put(60,05){\line(1,0){20}} \put(60,35){\line(1,0){20}}
\put(60,05){\line(0,1){30}} \put(80,05){\line(0,1){30}}

\put(80,18.5){\vector(1,0){4}}


\put(85,27){\makebox(0,0)[bl]{Building}}
\put(85,23){\makebox(0,0)[bl]{of spanning}}
\put(85,19){\makebox(0,0)[bl]{graph}}
\put(85,15){\makebox(0,0)[bl]{\(G =(A,\Gamma)\)}}

\put(84,05){\line(1,0){20}} \put(84,35){\line(1,0){20}}
\put(84,05){\line(0,1){30}} \put(104,05){\line(0,1){30}}


\end{picture}
\end{center}

  Procedures for transformation of scales
 (e.g., transformation of ordinal scale into ordinal scale,
 transformation of vector-like scale into ordinals scale,
 transformation of vector scale into multiset  based scale)
  can be useful
 for processing of proximity matrix and
 for design of covering graph (to obtain a simple covering graph, e.g.,
  by the use of thresholds for egde/arcs estimates/weights).
 On the other hand, transformation of vector-like estimates into
 ordinal estimates can be based on multicriteria ranking
 (sorting problem).


 Minimum spanning tree problem
 is an important part of many effective solving schemes for many
 combinatorial optimization problems (e.g., \cite{cormen90,gar79}):
 (a) polynomial approximation of initial graph by tree,
 (b) effective solving a combinatorial problem over the obtained tree.
 Here, several well-known algorithms
 for design of minimum spanning tree
  can be used, for example:
 Borovka's algorithm
 Prim's algorithm,
 Kruskal's algorithm
 \cite{aho74,gabow86,gar79,cormen90,pet02,yao75}.
 Complexity estimate of the algorithms is:
 \(O(p \log n )\) (or less \cite{yao75})
 (\(p\) is the number of edges, \(n\) is the number of vertices).

 Other auxiliary combinatorial problems
 (Table 3.1)
  are more complicated
  (i.e., they belong to class of NP-hard problems).
 Only in some simple cases polynomial algorithms can be used:

  (i) polynomial algorithms for basic assignment problem
  (e.g., \cite{gar79,kuh57});

 (ii) polynomial approximate solving schemes
 for basic knapsack problem and multiple choice
 problem
 (e.g., \cite{gar79,ibarra75,keller04,mar90,sahni75});

 (iii) polynomial algorithms for some network partition
 problems (e.g., cores decomposition of networks
 \cite{bata03});

 (iv) polynomial approximate solving schemes
 for simple cases of partitioning problems,
 network community detection problems,
 covering problems.

 Thus, it is necessary
 to use polynomial heuristics or enumerative methods
 for the above-mentioned
 auxiliary combinatorial problems
 (i.e., generalized assignment problem,
 clique problems,
 morphological clique problem,
 dominating set problem,
 covering problems,
 graph partitioning problems,
 finding agreement/median/consensus problems,
 multicriteria combinatorial problems).
 In the case of fast clustering schemes,
 fast heuristics (e.g., some analogues of greedy algorithms)
 have to be used for auxiliary combinatorial problems.

 Some basic clustering models
 (e.g., hierarchical clustering, k-means clustering)
  are often
 used as auxiliary problems of multi-stage clustering
 schemes (e.g., for preliminary definition of
 a cluster set or cluster centroids).



 The significance of
 balanced clustering problems
 (by cluster size) has been increased in many domains
 (e.g., communication systems).
  As a result, balanced partition of tree problem
  can be useful components of contemporary clustering
 schemes.
 Generally, the problem of \(k\)-balanced partitioning a tree is
 NP-hard
 (\(k\) is the number of elements in each cluster of clustering solution)
 \cite{feld15}.
 Here, it may be reasonable to use
 \(k\)-balanced
 agglomerative algorithm over a tree
 (i.e., under restriction on cluster sizes in obtained solution).


\subsection{Hierarchical clustering}

 Hierarchical clustering is widely used in many domains
 (e.g.,
  \cite{dahl00,gowda77,har75,jain88,jain99,kurita91,lev07e,olson95,rocha13,yag00}).
%
%
%
 The approach consists in
  agglomerative (i.e., hierarchical, ``Bottom-Up'') scheme
 (e.g., \cite{har75,jain88,jain99,lev07e}):


 1. Calculate the proximity (distance) matrix between elements.

 2. Start with \(n\) clusters containing one element.

 3. Find the most similar pair of clusters from the proximity
 matrix and merge them into a single cluster

 4. Update the proximity matrix
 (reduce its order by one, by replacing the individual clusters
 with the merged clusters)

 5. Repeat steps  3,4 until a single cluster is obtained (i.e.,
 \(n-1\) times).

\subsubsection{Basic agglomerative algorithm}

 The basic agglomerative algorithm
 ({\it algorithm 1}) is the following
 (e.g., \cite{lev07e,lev15}):

~~

 {\it Stage 1.}
  Calculate the matrix of element pair
 \(\forall (A(i_{1}),A(i_{2}))\), ~\(A(i_{1})\in A\), ~\(A(i_{2})\in A \),
 ~\(i_{1} \neq i_{2}\)
 ``distances'' (a simple case, metric \(l_{2}\)):

 \[ z_{i_{1}i_{2}} = \sqrt{ \sum_{j=1}^{m} ( x_{i_{1},j} - x_{i_{2},j} )^{2} }. \]

  {\it Stage 2.} Searching for  the minimum element pair
  ``distance''~~
 \(z^{min}=\min_{i_{1},i{2} \in \{1,...,n\}} ~\{ z_{i_{1},i_{2}}\}\),
   integration of the corresponding two elements into a resultant
 ``integrated'' element,
  extension of the corresponding cluster.

 {\it Stage 3.}
 If all elements are processed than GO TO Stage 5.

 {\it Stage 4.}
 Recalculate
 the matrix of
 pair ``distances'' \(Z\)
 (initial element set is decreased by \(1\) element)
  and ~{\it Go To}~~  Stage 2.

 {\it Stage 5.} Stop.


~~

 As result,
 a tree-like
 structure for the element pair integration process (Bottom-Up) is
 obtained (one element pair integration at each integration step).
 A basic simplified procedure for aggregation of items
 (aggregation as average values) is as follows
 (\(J_{i_{1},i_{2}}= A_{i_{1}} \& A_{i_{2}}\)):
 ~\(\forall j=\overline{1,m} ~~ x_{J_{i_{1},i_{2}},j}=\frac{x_{i_{1},j} + x_{i_{2},j}} {2}
 \).

~~

  Complexity estimates for the above-mentioned version hierarchical clustering algorithm (by stages)
   is presented in Table 3.2.

\begin{center}
{\bf Table 3.2.}  Complexity estimates for stages of basic hierarchical clustering algorithm\\
\begin{tabular}{| l | l|c |}
\hline
  Stage &Description&Complexity estimate\\
   & &(running time)\\

\hline

 Stage 1 &
 Calculate
 the distance matrix \(Z\)&\(O (n^{2})\) \\
 Stage 2 & Searching for the minimum element pair ``distance'',&\(O (n^{2})\) \\
  & integration of the corresponding element pair,& \\
  & extension of the corresponding cluster& \\
 Stage 3.& Checking the condition for stopping &\(O(n)\)\\
        & (all elements are processed)&\\
 Stage 4 &
 Recalculate
  the ``distance'' matrix \(Z\)&\(O (n^{2})\) \\
 Stage 5.& Stopping        & \(O(1)\)\\

\hline
\end{tabular}
\end{center}

 Here there exists a computing cycle (stages 2,3,4)
 that can contain \((n-1)\) steps.
 Thus, the general complexity (running time)
 of this hierarchical clustering algorithm equals
 \(O (n^{3})\).
 Generally, hierarchical clustering methods have the following problems:
 (a) sensitivity to noise and outliers,
 (b) difficulty handling different sized clusters
 and convex shapes, and
 (c) breaking large clusters.

\subsubsection{Balancing by cluster size}

 Now let us consider a modified version of hierarchical
 clustering with a special requirement
 to cluster size
 (as balancing of cluster sizes)
 to obtain about the same (close) cluster sizes.
 Let \(B = \{B_{1},...,B_{\iota},...,B_{\kappa}\}\)
 be the obtained set of clusters.
 Let \(\alpha_{\iota} = | B_{\iota} |\)
 (\( \iota = \overline{1,\kappa}\))
 be the size (i.e., number of elements)
 for cluster \(B_{\iota}\).

 Thus for each cluster
 the following constraints are considered:
 \(\alpha' \leq  \alpha_{\iota}  \leq \alpha''\).
 For example:  \(\alpha' = 3\), \(\alpha'' = 4\).
 Evidently, one cluster of the obtained cluster set
 can contain less elements
 (i.e., \(1,2,..,(\alpha'-1)\)).
 Generally, the above-mentioned requirement leads
 to balanced clustering solution by cluster size.
 This is significant in many applications
 (e.g., local areas in communication networks,
 student teams in educational process).
 Our modified balanced by cluster size hierarchical clustering algorithm is:

~~

 {\it Stage 1.}
 Calculate
  the matrix of element pair
 \(\forall (A(i_{1}),A(i_{2}))\), ~\(A(i_{1})\in A\), ~\(A(i_{2})\in A \),
 ~\(i_{1} \neq i_{2}\)
 ``distances'' (a simple case, metric \(l_{2}\)):

 \[ z_{i_{1}i_{2}} = \sqrt{ \sum_{j=1}^{m} ( x_{i_{1},j} - x_{i_{2},j} )^{2} }. \]

 {\it Stage 2.} Searching for  the minimum element pair ``distance''~~
 \(z^{min}=\min_{i_{1},i{2} \in \{1,...,n\}} ~\{ z_{i_{1},i_{2}}\}\),
   integration of the corresponding two elements into a resultant
 ``integrated'' element,
  extension of the corresponding cluster.

 {\it Stage 3.} Analysis of the obtained extended cluster
 \(B_{\iota}\) by new size
 \(\alpha_{\iota}\).
 If  \(\alpha_{\iota} = \alpha''\) then
 deleting the cluster and its elements for the future processing
 (as a part of the resultant solution).

{\it Stage 4.} If  all elements are processed than GO TO
  Stage 6.

 {\it Stage 5.}
 Recalculate
 the matrix of  pair ``distances'' \(Z\)
 (initial element set is decreased by \(1\) element)
  and ~{\it Go To}~~
  Stage 2.

{\it Stage 6.}
 The other elements are organized as additional separated clusters
 (if needed).
 Stop.

~~

  Complexity estimates for the above-mentioned version hierarchical clustering algorithm (by stages)
   is presented in Table 3.3.

\begin{center}
{\bf Table 3.3.} Complexity estimates of stages for balanced
 (by cluster size) hierarchical
 algorithm\\
\begin{tabular}{| l | l|c |}
\hline
  Stage &Description &Complexity estimate   \\
  & &(running time)  \\
\hline
 Stage 1 &
 Calculate
  distance matrix \(Z\)&\(O (n^{2})\) \\
 Stage 2 & Searching for the minimum element pair ``distance'',
 &\(O (n^{2})\) \\
  & integration of the corresponding element pair,& \\
  & extension of the corresponding cluster& \\

 Stage 3.& Checking the condition for stopping &\(O(n)\)\\
 & (all elements are processed)&\\

 Stage 4.& Analysis of the obtained cluster by cluster size &\(O(1)\)\\
 Stage 5 &
 Recalculate
 the ``distance'' matrix \(Z\)&\(O (n^{2})\) \\
 Stage 6.& Stopping        & \(O(1)\)\\
\hline
\end{tabular}
\end{center}

 Here there exists a computing cycle (stages 2,3,4,5)
 that can contain \((n-1)\) steps.
 Thus, the general complexity
  (running time)
  of this hierarchical clustering algorithm
   equals
 \(O (n^{3})\).
 Note the average complexity estimate is less.

~~

 {\bf Example 3.1.} A numerical example of tree with weights of
 edges (i.e., proximity between element pairs)
 is presented in Fig. 3.2:
 tree \(T = (A,E)\), \(A=\{ 1,2,3,4,5,6,7,8,9,10,11,12,13,14\}\).
 The constraint for cluster size is: ~\( \leq 3\).
%
 Table 3.4. contains the corresponding proximity matrix, i.e., weights of edges
 (a very large proximity is denoted by symbol ``\(\star \)'',
 proximity is symmetric).

\begin{center}
\begin{picture}(60,44)
\put(02,00){\makebox(0,0)[bl]{Fig. 3.2. Balanced clustering of
tree}}


\put(35,40){\circle*{1}} \put(36,39){\makebox(0,8)[bl]{\(1\)}}

\put(10,30){\line(5,2){25}} \put(25,30){\line(1,1){10}}
\put(35,30){\line(0,1){10}} \put(40,30){\line(-1,2){5}}

\put(37.5,35){\oval(8,13.5)}

\put(42,39){\makebox(0,8)[bl]{Cluster \(X_{2}\)}}

\put(35,30){\circle*{1}} \put(36,29){\makebox(0,8)[bl]{\(4\)}}
\put(40,30){\circle*{1}} \put(41.8,29){\makebox(0,8)[bl]{\(5\)}}


\put(40,20){\line(0,1){10}} \put(50,20){\line(-1,1){10}}

\put(40,20){\circle*{1}} \put(41,19){\makebox(0,8)[bl]{\(11\)}}
\put(50,20){\circle*{1}} \put(51,19){\makebox(0,8)[bl]{\(12\)}}

\put(42,20){\oval(6,7)}

\put(35.5,13){\makebox(0,8)[bl]{Cluster}}
\put(39,09){\makebox(0,8)[bl]{\(X_{5}\)}}

\put(52,20){\oval(6,7)}

\put(49,13){\makebox(0,8)[bl]{Cluster}}
\put(52,09){\makebox(0,8)[bl]{\(X_{6}\)}}

\put(02,37){\makebox(0,8)[bl]{Cluster}}
\put(05,33){\makebox(0,8)[bl]{\(X_{3}\)}}

\put(7.5,25){\oval(8,13.5)}

\put(10,30){\circle*{1}} \put(07,29){\makebox(0,8)[bl]{\(2\)}}

\put(05,20){\line(1,2){05}} \put(10,20){\line(0,1){10}}

\put(05,20){\circle*{1}} \put(01.5,19){\makebox(0,8)[bl]{\(6\)}}
\put(10,20){\circle*{1}} \put(11.5,19){\makebox(0,8)[bl]{\(7\)}}

\put(11.8,28){\makebox(0,8)[bl]{Cluster}}
\put(17,24){\makebox(0,8)[bl]{\(X_{4}\)}}

\put(27.5,25){\oval(8,13.5)}

\put(25,30){\circle*{1}} \put(26.6,29){\makebox(0,8)[bl]{\(3\)}}

\put(20,20){\line(1,2){5}} \put(25,20){\line(0,1){10}}
\put(30,20){\line(-1,2){5}}

\put(25,20){\circle*{1}} \put(26,19){\makebox(0,8)[bl]{\(9\)}}
\put(30,20){\circle*{1}} \put(31.5,19){\makebox(0,8)[bl]{\(10\)}}

\put(10,05){\makebox(0,8)[bl]{Cluster \(X_{1}\)}}
\put(17.5,15){\oval(8,13.5)}

\put(20,20){\circle*{1}} \put(17,19){\makebox(0,8)[bl]{\(8\)}}

\put(15,10){\circle*{1}} \put(10,09){\makebox(0,8)[bl]{\(13\)}}
\put(20,10){\circle*{1}} \put(21.5,09){\makebox(0,8)[bl]{\(14\)}}

\put(15,10){\line(1,2){5}} \put(20,10){\line(0,1){10}}


\end{picture}
\end{center}

\begin{center}
{\bf Table 3.4.} Proximities for tree-like example  (edge \((i_{1},i_{2})\))   \\
\begin{tabular}{| c |cc c c c c c c c c c c c c|}
\hline
 \(i_{1}\) & \(i_{2}:\)& \(2\)&\(3\)&\(4\)&\(5\)&\(6\) &\(7\)&\(8\)&\(9\)&\(10\)&\(11\)&\(12\) &\(13\)&\(14\)\\
\hline
 1 &&\(1.5\)&\(1.7\)&\(0.5\)&\(0.2\) &\(\star\)&\(\star\) &\(\star\) & \(\star\)& \(\star\)& \(\star\)&\(\star\)& \(\star\) & \(\star\)\\
 2 &&&\(\star\)&\(\star\)&\(\star\)&\(0.1\)&\(0.6\)&\(\star\)&\(\star\)&\(\star\)&\(\star\)&\(\star\)&\(\star\)& \(\star\)\\
 3 &&&&\(\star\)&\(\star\)&\(\star\)&\(\star\)&\(3.1\)&\(1.0\)&\(0.9\)&\(\star\)&\(\star\)&\(\star\)& \(\star\)\\
 4 &&&&&\(\star\)&\(\star\)&\(\star\)&\(\star\)&\(\star\)&\(\star\)&\(\star\)&\(\star\)&\(\star\)& \(\star\)\\

 5 &&&&&&\(\star\)&\(\star\)&\(\star\)&\(\star\)&\(\star\)&\(2.5\)&\(2.1\)&\(\star\)& \(\star\)\\
 6 &&&&&&&\(\star\)&\(\star\)&\(\star\)&\(\star\)&\(\star\)&\(\star\)&\(\star\)& \(\star\)\\
 7 &&&&&&&&\(\star\)&\(\star\)&\(\star\)&\(\star\)&\(\star\)&\(\star\)& \(\star\)\\
 8 &&&&&&&&&\(\star\) &\(\star\)&\(\star\)&\(\star\)& \(0.3\)&\(0.4\)\\
 9 &&&&&&&&&&\(\star\)&\(\star\)&\(\star\)&\(\star\)&\(\star\)\\
 10 &&&&&&&&&&&\(\star\)&\(\star\)&\(\star\)&\(\star\)\\
 11 &&&&&&&&&&&&\(\star\)&\(\star\)&\(\star\)\\
 12 &&&&&&&&&&&&&\(\star\)&\(\star\)\\
 13 &&&&&&&&&&&&&&\(\star\)\\

\hline
\end{tabular}
\end{center}

 In the considered case (i.e., tree),
 proximity information can be presented as a list
 of \((n-1)\) components:
 number of vertex,  number of ``son''-vertex, weight of the
 corresponding edge (Table 3.5).

\begin{center}
{\bf Table 3.5.} Initial list if proximities \\
\begin{tabular}{| c |  c | c|}
\hline
  Vertex \(i_{1}\) &  Vertex \(i_{2}\)& Weight of \\
  & (``son'' of \(i_{1}\))  &edge  \((i_{1},i_{2})\)\\
\hline
 \(1\) & \(2\) & \(1.5\)\\
 \(1\) & \(3\) & \(1.7\)\\
 \(1\) & \(4\) & \(0.5\)\\
 \(1\) & \(5\) & \(0.2\)\\
 \(2\) & \(6\) & \(0.1\)\\
 \(2\) & \(7\) & \(0.6\)\\
 \(3\) & \(8\) & \(3.1\)\\
 \(3\) & \(9\) & \(1.0\)\\
 \(3\) & \(10\) & \(0.9\)\\
 \(5\) & \(11\) & \(2.5\)\\
 \(5\) & \(12\) & \(2.1\)\\
 \(8\) & \(13\) & \(0.3\)\\
 \(8\) & \(14\) & \(0.4\)\\
\hline
\end{tabular}
\end{center}

 In the example,
 proximity between element \(x\) and cluster  \(Y\)
 (or integrated element) \(D^{min}(x,Y)\) is used
 (case 2 from previous section 2).
 The steps of agglomerative algorithms to obtain the balanced clustering
 (cluster size is \( \leq 3\))
 are the following:

 {\it Step 1.} Integration of elements \(2\) and \(6\) into
 \(J(2,6)\). As a result, Table  3.6 is obtained.

\begin{center}
{\bf Table 3.6.} List of proximities after step 1\\
\begin{tabular}{| c |  c | c|}
\hline
  Vertex \(i_{1}\) &  Vertex \(i_{2}\)& Weight of \\
  & (``son'' of \(i_{1}\))  &edge  \((i_{1},i_{2})\)\\
\hline
 \(1\) & \(J(2,6)\) & \(1.5\)\\
 \(1\) & \(3\) & \(1.7\)\\
 \(1\) & \(4\) & \(0.5\)\\
 \(1\) & \(5\) & \(0.2\)\\
 \(J(2,6)\) & 7 & \(0.6\)\\
 \(3\) & \(8\) & \(3.1\)\\
 \(3\) & \(9\) & \(1.0\)\\
 \(3\) & \(10\) & \(0.9\)\\
 \(5\) & \(11\) & \(2.5\)\\
 \(5\) & \(12\) & \(2.1\)\\
 \(8\) & \(13\) & \(0.3\)\\
 \(8\) & \(14\) & \(0.4\)\\
\hline
\end{tabular}
\end{center}

  {\it Step 2.} Integration of elements \(1\) and \(5\) into
 \(J(1,5)\). As a result, Table  3.7 is obtained.

\begin{center}
{\bf Table 3.7.} List of proximities after step 2\\
\begin{tabular}{| c |  c | c|}
\hline
  Vertex \(i_{1}\) &  Vertex \(i_{2}\)& Weight of \\
  & (``son'' of \(i_{1}\))  &edge  \((i_{1},i_{2})\)\\
\hline
   \(J(1,5)\) & \(J(2,6)\) & \(1.5\)\\
 \(J(1,5)\) & \(3\) & \(1.7\)\\
 \(J(1,5)\) & \(4\) & \(0.5\)\\
 \(J(2,6)\) & 7 & \(0.6\)\\
 \(3\) & \(8\) & \(3.1\)\\
 \(3\) & \(9\) & \(1.0\)\\
 \(3\) & \(10\) & \(0.9\)\\
 \(J(1,5)\) & \(11\) & \(2.5\)\\
 \(J(1,5)\) & \(12\) & \(2.1\)\\
 \(8\) & \(13\) & \(0.3\)\\
 \(8\) & \(14\) & \(0.4\)\\
\hline
\end{tabular}
\end{center}

 {\it Step 3.} Integration of elements \(8\) and \(13\) into
 \(J(8,13)\). As a result, Table  3.8 is obtained.

\begin{center}
{\bf Table 3.8.} List of proximities after step 3\\
\begin{tabular}{| c |  c | c|}
\hline
  Vertex \(i_{1}\) &  Vertex \(i_{2}\)& Weight of \\
  & (``son'' of \(i_{1}\))  &edge  \((i_{1},i_{2})\)\\
\hline
   \(J(1,5)\) & \(J(2,6)\) & \(1.5\)\\
 \(J(1,5)\) & \(3\) & \(1.7\)\\
 \(J(1,5)\) & \(4\) & \(0.5\)\\
 \(J(2,6)\) & 7 & \(0.6\)\\
 \(3\) & \(J(8,13)\) & \(3.1\)\\
 \(3\) & \(9\) & \(1.0\)\\
 \(3\) & \(10\) & \(0.9\)\\
 \(J(1,5)\) & \(11\) & \(2.5\)\\
 \(J(1,5)\) & \(12\) & \(2.1\)\\
 \(J(8,13)\) & \(14\) & \(0.4\)\\
\hline
\end{tabular}
\end{center}

{\it Step 4.} Integration of elements \(J(8,13)\) and \(14\) into
 \(J(8,13,14)\).
 Thus, cluster 1 is designed \(X_{1} = \{8,13,14\}\).
 The corresponding elements
 (i.e., \(J(8,13),14\))
  can be deleted from the next analysis.
  As a result, Table  3.9 is obtained.

\begin{center}
{\bf Table 3.9.} List of proximities after step 4\\
\begin{tabular}{| c |  c | c|}
\hline
  Vertex \(i_{1}\) &  Vertex \(i_{2}\)& Weight of \\
  & (``son'' of \(i_{1}\))  &edge  \((i_{1},i_{2})\)\\
\hline
   \(J(1,5)\) & \(J(2,6)\) & \(1.5\)\\
 \(J(1,5)\) & \(3\) & \(1.7\)\\
 \(J(1,5)\) & \(4\) & \(0.5\)\\
 \(J(2,6)\) & 7 & \(0.6\)\\
 \(3\) & \(9\) & \(1.0\)\\
 \(3\) & \(10\) & \(0.9\)\\
 \(J(1,5)\) & \(11\) & \(2.5\)\\
 \(J(1,5)\) & \(12\) & \(2.1\)\\
\hline
\end{tabular}
\end{center}

{\it Step 5.} Integration of elements \(J(1,5)\) and \(4\) into
 \(J(1,4,5)\).
 Thus, cluster 2 is designed \(X_{2} = \{1,4,5\}\).
 The corresponding elements
 (i.e., \(J(1,5),4\))
  can be deleted from the next analysis.
  As a result, Table  3.10 is obtained.

\newpage
\begin{center}
{\bf Table 3.10.} List of proximities after step 5\\
\begin{tabular}{| c |  c | c|}
\hline
  Vertex \(i_{1}\) &  Vertex \(i_{2}\)& Weight of \\
  & (``son'' of \(i_{1}\))  &edge  \((i_{1},i_{2})\)\\
\hline
 \(J(2,6)\) & 7 & \(0.6\)\\
 \(3\) & \(9\) & \(1.0\)\\
 \(3\) & \(10\) & \(0.9\)\\
\hline
\end{tabular}
\end{center}

{\it Step 6.} Integration of elements \(J(2,6)\) and \(7\) into
 \(J(2,6,7)\).
 Thus, cluster 3 is designed \(X_{3} = \{2,6,7\}\).
 The corresponding elements
 (i.e., \(J(2,6),7\))
  can be deleted from the next analysis.
  As a result, Table  3.11 is obtained.

\begin{center}
{\bf Table 3.11.} List of proximities after step 6\\
\begin{tabular}{| c |  c | c|}
\hline
  Vertex \(i_{1}\) &  Vertex \(i_{2}\)& Weight of \\
  & (``son'' of \(i_{1}\))  &edge  \((i_{1},i_{2})\)\\
\hline
 \(3\) & \(9\) & \(1.0\)\\
 \(3\) & \(10\) & \(0.9\)\\
\hline
\end{tabular}
\end{center}

{\it Step 7.} Integration of elements \(3\) and \(10\) into
 \(J(3,10)\).
%
  As a result, Table  3.12 is obtained.

\begin{center}
{\bf Table 3.12.} List of proximities after step 7\\
\begin{tabular}{| c |  c | c|}
\hline
  Vertex \(i_{1}\) &  Vertex \(i_{2}\)& Weight of \\
  & (``son'' of \(i_{1}\))  &edge  \((i_{1},i_{2})\)\\
\hline
 \(J(3,10)\) & \(9\) & \(1.0\)\\
\hline
\end{tabular}
\end{center}

{\it Step 8.} Integration of elements \(J(3,10\) and \(9\) into
 \(J(3,9,10)\).
 Thus, cluster 4 is designed \(X_{4} = \{3,9,10\}\).
 The corresponding elements
 (i.e., \(J(3,10),9\))
  can be deleted from the next analysis.

 Two separated elements \(11\) and \(12\)
 can be organized as two clusters:
 \(X_{5} = \{11\}\) and
 \(X_{6} = \{12\}\).

\subsubsection{Improvements of hierarchical clustering scheme \cite{lev07e}}


 First, let us point out some properties
 of the considered clustering process as follows:

 {\it 1.} The matrix of element pair proximity
 can contain several ``minimal'' elements.
 Thus there are problems as follows:
  ~(i) selection of the best unit pair for integration;
  ~(ii) possible
  integration of several unit pair at each algorithm stage.

 {\it 2.} Computing the matrix of element pair
 distances often does not correspond to
 the problem context and it is reasonable to consider
 a ``softer'' approach for computing element pair proximity.

 {\it 3.} The obtained structure
 of the clustering process is a tree.
 Often the clustering problem is used to get
 a system structure that corresponds to the above-mentioned
 clustering process
 ({\it e.g.}, evolution trees, system architecture).
 Thus, it is often reasonable to organize the clustering process
 as a hierarchy, {\it e.g.}, for modular systems in which the same
 modules can be integrated into different system components/parts.

%
 Fig. 3.3 illustrates concurrent integration of element  pairs at the
 same step of the algorithm when some elements can be integrated into
 different system components parts, i.e., obtaining a hierarchical
 system structure
 (common modules/parts, {\it e.g.}, \(3\) and \(4\)).
 In this case, obtained clusters can have intersections
 (Fig. 3.4).
%
%

\begin{center}
\begin{picture}(105,49)

\put(20,00){\makebox(0,0)[bl]{Fig. 3.3. Illustration
 for hierarchy}}

\put(00,07){\makebox(0,0)[bl]{Step \(0\)}}
\put(17,08){\makebox(0,0)[bl]{\(1\)}}

\put(15,11){\line(1,0){5}} \put(15,07){\line(1,0){5}}
\put(15,07){\line(0,1){4}} \put(20,07){\line(0,1){4}}

\put(27,08){\makebox(0,0)[bl]{\(2\)}}

\put(25,11){\line(1,0){5}} \put(25,07){\line(1,0){5}}
\put(25,07){\line(0,1){4}} \put(30,07){\line(0,1){4}}
\put(37,08){\makebox(0,0)[bl]{\(3\)}}

\put(35,11){\line(1,0){5}} \put(35,07){\line(1,0){5}}
\put(35,07){\line(0,1){4}} \put(40,07){\line(0,1){4}}
\put(47,08){\makebox(0,0)[bl]{\(4\)}}

\put(45,11){\line(1,0){5}} \put(45,07){\line(1,0){5}}
\put(45,07){\line(0,1){4}} \put(50,07){\line(0,1){4}}
\put(57,08){\makebox(0,0)[bl]{\(5\)}}

\put(55,11){\line(1,0){5}} \put(55,07){\line(1,0){5}}
\put(55,07){\line(0,1){4}} \put(60,07){\line(0,1){4}}
\put(67,08){\makebox(0,0)[bl]{\(6\)}}

\put(65,11){\line(1,0){5}} \put(65,07){\line(1,0){5}}
\put(65,07){\line(0,1){4}} \put(70,07){\line(0,1){4}}
\put(77,08){\makebox(0,0)[bl]{\(7\)}}

\put(75,11){\line(1,0){5}} \put(75,07){\line(1,0){5}}
\put(75,07){\line(0,1){4}} \put(80,07){\line(0,1){4}}

\put(87,08){\makebox(0,0)[bl]{\(8\)}}

\put(85,11){\line(1,0){5}} \put(85,07){\line(1,0){5}}
\put(85,07){\line(0,1){4}} \put(90,07){\line(0,1){4}}


\put(00,17){\makebox(0,0)[bl]{Step \(1\)}}
\put(17,18){\makebox(0,0)[bl]{\(1\)}} \put(17.9,19){\oval(05,4)}


\put(17.5,11){\vector(0,1){6}}
\put(25.5,17.5){\makebox(0,0)[bl]{\(2,3\)}}
\put(28,19){\oval(08,4)}


\put(27.5,11){\vector(0,1){6}}
\put(37,17.5){\makebox(0,0)[bl]{\(3,4\)}}
\put(39.5,19){\oval(08,4)}


\put(37.5,11){\vector(0,1){6}} \put(37.5,11){\vector(-1,1){6}}


\put(47.5,11){\vector(-1,1){6}}
\put(56,17.5){\makebox(0,0)[bl]{\(4,5\)}}
\put(58.5,19){\oval(08,4)}


\put(57.5,11){\vector(0,1){6}} \put(49.5,11){\vector(1,1){6}}
\put(67,17.5){\makebox(0,0)[bl]{\(6,7,8\)}}
\put(71,19){\oval(12,4)}


\put(67.5,11){\vector(0,1){6}}

\put(77.5,11){\vector(-1,1){6}} \put(87.5,11){\vector(-2,1){12}}

\put(67.5,21){\vector(-4,1){24}}


\put(77.5,11){\vector(-1,1){6}}

\put(00,27){\makebox(0,0)[bl]{Step \(2\)}}

\put(17.5,21){\vector(3,1){18}}
\put(25,27.5){\makebox(0,0)[bl]{\(2,3\)}} \put(28,29){\oval(08,4)}


\put(27.5,21){\vector(0,1){6}} \put(27.5,31){\vector(1,2){2}}
\put(35.5,27.5){\makebox(0,0)[bl]{\(1,3,4,6,7,8\)}}
\put(45,29){\oval(20,4)}


\put(37.5,21){\vector(0,1){6}} \put(45,31){\vector(0,1){4}}

\put(58,27.5){\makebox(0,0)[bl]{\(4,5,6,7,8\)}}
\put(65.5,29){\oval(18,4)}


\put(57.5,21){\vector(1,2){3}} \put(67.5,21){\vector(0,1){6}}
\put(65,31){\vector(-1,2){2}}



\put(34.5,43){\makebox(0,0)[bl]{\(1,2,3,4,5,6,7,8\)}}
\put(46.5,45){\oval(36,6)} \put(46.5,45){\oval(35,5)}


\put(40,38){\makebox(0,0)[bl]{{\bf . ~. ~.}}}

\end{picture}
%
\begin{picture}(30,40)
\put(02,04){\makebox(0,0)[bl]{Fig. 3.4. Clusters}}
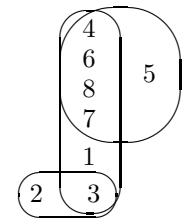
\put(02,00){\makebox(0,0)[bl]{for  Step 2 (Fig. 3.2)}}

\put(5,14){\makebox(0,0)[bl]{\(2 ~~~~~3\)}}
\put(10,15){\oval(13,6)}

\put(12,19){\makebox(0,0)[bl]{\(1\)}}

\put(12,24){\makebox(0,0)[bl]{\(7\)}}

\put(12,28){\makebox(0,0)[bl]{\(8\)}}
\put(12,32){\makebox(0,0)[bl]{\(6\)}}

\put(12,36){\makebox(0,0)[bl]{\(4\)}}

\put(13,26){\oval(8,27)}


\put(20,30){\makebox(0,0)[bl]{\(5\)}}

\put(17,31){\oval(16,18)}

\end{picture}
\end{center}

 Now, let us describe  some possible algorithm improvements.

~~

 {\bf Improvement 1} (algorithm 2):

~~

 {\it Stage 1.} Computing an ordinal distance/proximity
 (0 corresponds to equal or the more similar elements).
 Here it is possible to compute the pair distance/proximity
 via the previous approach and mapping
 the pair distance to the ordinal scale.

 {\it Stage 2.} Revelation of
 the smallest pair distance and integration of the corresponding
 elements.

 {\it Note 1.} It is possible to reveal several close element
 pairs and execution several pair integration.

 {\it Note 2.} It is possible to include
 the same element into different integrated pairs.

 {\it Stage 3.} The stage corresponds to stage 3 in {\it algorithm 1}.

~~

 Here, a hierarchical structure for the element pair integration
 (Bottom-Up) is obtained
 (several element pair integration at each integration step).
 The complexity of the problem may consist in
 revelation of many subcliques
 (in graph over elements and their proximity).
 In the process of computing the ordinal proximity it
 is reasonable to use a limited number of element pairs for each
 level of the proximity ordinal scale.
 As a result, the limited
 number of integrated element pairs
 (or complete subgraphs or cliques)
 will be revealed at each integration stage.
 This provides polynomial complexity of the algorithm
 (number of operations, volume of required memory)
 \(O(m~n^{2})\).

~~

 {\bf Improvement 2} (algorithm 3):
 This algorithm is close to {\it algorithm 2},
 but the computing process for the ordinal element pairs
 proximity is based on multicriteria analysis, {\it e.g.},
 Pareto-approach or outranking technique
 ({\it i.e.}, Electre-like methods).
 Complexity of the algorithm is \(O(m~n^{4})\).

 The algorithms 2 and 3 implement the following trend:

~~

 {\it from tree-like structure (of clustering process) to
 hierarchy.}

~~

 An analysis of obtained clique(s) can be included into the
 algorithms as well.


\subsection{K-means clustering}

 K-means clustering approach is widely used
%
 \cite{dhil05,har79,jain88,jain99,jain10,kanu00,mirkin05}.
  The basic simplified version of the algorithm is:

~~

 {\it Stage 1.} Select \(K\) points as initial centroids
 (e.g., mean points)
 (e.g., selection is based on random process).

 {\it Stage 2.} Cycle by all \(n\) points:

 (2.1) Form \(K\) clusters by assigning all points to the closest
 centroid.

 (2.2)
  Recalculate the centroid of each cluster.

 (2.3) If all points are assigned GO TO stage 3.

 (2.4) GO TO (2.1).

 {\it Stage 3.} Stop.

~~

  Complexity estimates for \(K\)-mean clustering algorithm (by stages)
   is presented in Table 3.13.


\begin{center}
{\bf Table 3.13.} Complexity estimates of stages
 of \(K\)-means clustering algorithm\\
\begin{tabular}{| l | l|c |}
\hline
  Stage &Description &Complexity estimate   \\
  & &(running time)  \\
\hline
 Stage 1 &
 Selection of \(K\) centroids
 &\(O (K)\) \\

 Stage 2.2 & Assignment of all \(n\) points to \(K\) centroids
  & \( O(n \times K \times m)\) \\

  & (by \(m\) attributes)& \\

 Stage 2.2
  &Recalculation of the centroids
  (for \(K\) clusters)
   &\(O (K \times n \times m)\) \\

 Stage 2.3.& Checking the condition for stopping &\(O(1)\)\\
 & (all elements are processed)&\\

 Stage 3.& Stopping        & \(O(1)\)\\
\hline
\end{tabular}
\end{center}

  Thus, the general complexity of this algorithm  equals
 \(O (K \times n \times m)\).

 This approach has some problems
 when initial object set contains ``outlier''-like point(s).
 (Fig. 2.14).

 A general framework of k-means clustering
 is shown in Fig. 3.5.

\begin{center}
\begin{picture}(146,50)
\put(27,0){\makebox(0,0)[bl]{Fig. 3.5.
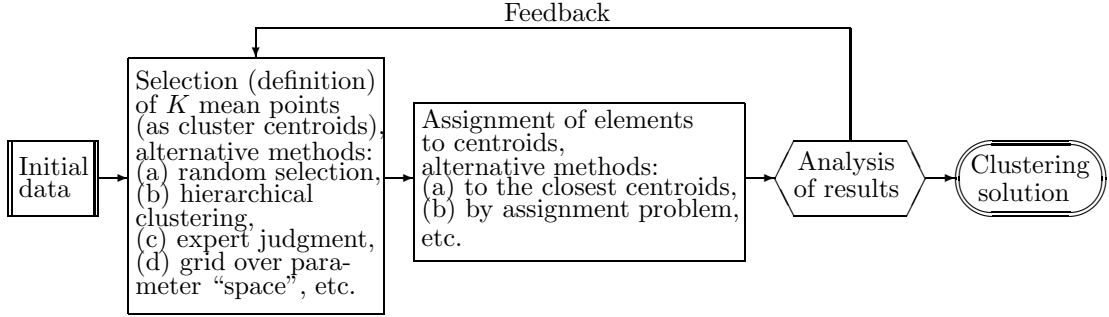
 General framework of k-means clustering process}}

\put(1.5,24.5){\makebox(0,0)[bl]{Initial}}
\put(1.5,21.5){\makebox(0,0)[bl]{data}}

\put(00,19){\line(1,0){12}} \put(00,29){\line(1,0){12}}
\put(00,19){\line(0,1){10}} \put(12,19){\line(0,1){10}}
\put(0.5,19){\line(0,1){10}} \put(11.5,19){\line(0,1){10}}
\put(12,24){\vector(1,0){4}}

\put(17,35){\makebox(0,0)[bl]{Selection (definition)}}
\put(17,32){\makebox(0,0)[bl]{of \(K\) mean points}}
\put(17,29){\makebox(0,0)[bl]{(as cluster centroids),}}
\put(17,26.5){\makebox(0,0)[bl]{alternative methods:}}
\put(17,23){\makebox(0,0)[bl]{(a) random selection,}}
\put(17,20){\makebox(0,0)[bl]{(b) hierarchical}}
\put(17,17){\makebox(0,0)[bl]{clustering,}}
\put(17,14){\makebox(0,0)[bl]{(c) expert judgment,}}
\put(17,11){\makebox(0,0)[bl]{(d) grid over para-}}
\put(17,08){\makebox(0,0)[bl]{meter ``space'', etc.}}

\put(16,06){\line(1,0){34}} \put(16,40){\line(1,0){34}}
\put(16,06){\line(0,1){34}} \put(50,06){\line(0,1){34}}

\put(50,24){\vector(1,0){4}}


\put(55,30){\makebox(0,0)[bl]{Assignment of elements}}
\put(55,27){\makebox(0,0)[bl]{to centroids, }}
\put(55,24.5){\makebox(0,0)[bl]{alternative methods: }}
\put(55,21){\makebox(0,0)[bl]{(a) to the closest centroids,}}
\put(55,18){\makebox(0,0)[bl]{(b) by assignment problem,}}
\put(55,15){\makebox(0,0)[bl]{etc.}}

\put(54,13){\line(1,0){44}} \put(54,34){\line(1,0){44}}
\put(54,13){\line(0,1){21}} \put(98,13){\line(0,1){21}}

\put(98,24){\vector(1,0){4}}


\put(105,24.5){\makebox(0,0)[bl]{Analysis }}
\put(104.5,21.5){\makebox(0,0)[bl]{of results}}

\put(104.5,19){\line(1,0){15}} \put(104.5,29){\line(1,0){15}}
\put(102,24){\line(1,2){2.5}} \put(102,24){\line(1,-2){2.5}}
\put(122,24){\line(-1,2){2.5}} \put(122,24){\line(-1,-2){2.5}}

\put(122,24){\vector(1,0){4}}


\put(66,45){\makebox(0,0)[bl]{Feedback}}

\put(33,44){\vector(0,-1){4}} \put(33,44){\line(1,0){79}}
\put(112,29){\line(0,1){15}}

\put(136,24){\oval(20,10)} \put(136,24){\oval(19,09)}

\put(128,24){\makebox(0,0)[bl]{Clustering}}
\put(129,21){\makebox(0,0)[bl]{solution}}

\end{picture}
\end{center}

\subsection{Clustering as assignment}

 Generally, the k-means clustering approaches involve
 a stage
 to assignment the items to preliminary defined clusters.
 Thus assignment problems can be used at this stage
 and, as a result,
 special assignment based clustering methods are obtained
 (e.g., \cite{goldb08,hub87}).

 In basic assignment problem
 (bipartite matching problem)
 there are the following:
 items/elements \(\{1,...,i,...,n\}\),
  agents \(\{1,...,j,...,\mu \}\),
 positive \(c_{ij}\) profit for assignment of item \(i\) to agent \(j\) ,
 binary variable \(x_{ij}\) equals \(1\) if item \(i\) is assigned to agent \(j\)
 and \(0\) otherwise.
 The basic assignment problem is (e.g., \cite{gar79}):
 \[ \max ~ \sum_{j=1}^{\mu} \sum_{i=1}^{n} ~ c_{ij} x_{ij}
%
 ~~ s.t.  ~ \sum_{i=1}^{n} a_{ij} x_{ij} \leq 1, ~ j =
\overline{1,\mu},
%
 ~ \sum_{j=1}^{\mu}  x_{ij} \leq 1, ~i=\overline{1,n}, ~ x_{ij} \in \{0,1\},
 ~i=\overline{1,n}, ~j=\overline{1,\mu}. \]
 Here each item has to be assigned to the only one cluster
 (agent).
 There exist several well-known polynomial
 algorithms for the problem above
 (e.g., \cite{edmo72,cormen90,dinic70,gar79,kuh57}).

 The generalized assignment problem GAP
 can be described as a multiple knapsack
 (or multiple agents)
  problem.
 Analogically,
 given \(n\) items/elements (\(i=\overline{1,n}\))
 and \(\mu\) knapsacks (agents)
 (\(j=\overline{1,\mu}\)).
 The following notations are used:
 \(c_{ij}\) is a profit of item \(i\) if it is assigned
 to knapsack (agent),
 \(a_{ij}\) is a weight (e.g., required resource) of item
 \(i\) if it assigned to knapsack (agent) \(j\),
 \(b_{j}\) is a capacity (volume of resource) of knapsack (agent)
 \(j\),
 binary variable \(x_{ij}\) equals \(1\) if item \(i\) is assigned to agent \(j\)
 and \(0\) otherwise.
  Clearly, the knapsack (agent) capacity can be considered as multiple
 recourse
 (i.e, a vector-like parameter)
 as well
 (e.g., \cite{gavish91}).
 The problem is \cite{gar79,mar90}:

~~

 {\it Assign each item to exactly one  knapsack so as to maximize the
 total profit assigned,
 without assigning to any knapsack a total weight greater than
 its capacity.}

~~

 The basic problem statement is:
 \[ \max  \sum_{j=1}^{\mu} \sum_{i=1}^{n} ~c_{ij} x_{ij}
%
 ~~~ s.t.  ~ \sum_{i=1}^{n} a_{ij} x_{ij} \leq b_{j}, ~j =
\overline{1,\mu},
%
 ~ \sum_{j=1}^{\mu}  x_{ij} \leq 1, ~ i=\overline{1,n},
 ~ x_{ij} \in \{0,1\}, ~i=\overline{1,n}, ~j=\overline{1,\mu}. \]
 The problem is known to be NP-hard
 (e.g., \cite{sahni76}).
 Evidently, the objective function can be minimized as well
 (e.g.,
 minumum cost assignment of a set of items/objects
 to a set of agents).
 In the case of \(a_{ij} =1 ~ \forall i, \forall j  \),
 each agent has an integer restristion of assigned elements as
 restriction for cluster size).
 In multiple assignment problem
 constraint
 \(\sum_{j=1}^{\mu}  x_{ij} \leq 1 ~ (i=\overline{1,n})\)
 is replaced by
 \(\sum_{j=1}^{\mu}  x_{ij} \leq \lambda_{i} ~ (i=\overline{1,n})\)
 where \(\lambda_{i}\) is restriction for the number of
 admissible assignment to different agents for
  elements \(i\).
 In applications,
 knapsack/agents can be considered as service centers
 (e.g., access points in communication networks)
 which have limited service resource(s)
 (e.g., \cite{levpet10,levpet10a}).

 Generally,
 the following kinds of algorithms have been used for
 basic generalized assignment problems
 (e.g., \cite{catt92,keller04,mar90}):
 (i) exact algorithms as enumerative methods (e.g..
 branch-and-bound algorithms)
 (e.g., \cite{raj13,ross75,savel97}),
 (ii) relaxation methods, reduction algorithms
 (e.g., relaxations to linear programming models,
 relaxation heuristics)
 (e.g., \cite{gavish91,lorena96,trick92}),
 (iii) approximation schemes
 (e.g., \cite{cohen06,dawande00,fle06,nutov06,shmo93}),
 (iv) various heuristics (e.g., \cite{had04,nauss03})
 including greedy algorithms (e.g., \cite{nauss03,romei00}),
 set partitioning heuristic (e.g., \cite{catt94})
 genetic algorithms (e.g., \cite{chu97}),
 tabu search algorithms (e.g., \cite{diaz01,higg01}).
 Fig. 3.6 illustrates the generalized assignment problem.

\begin{center}
\begin{picture}(75,62)

\put(01.5,00){\makebox(0,0)[bl] {Fig. 3.6.
 Illustration of generalized assignment}}


\put(42,58){\makebox(0,8)[bl]{Restriction}}
\put(42,54){\makebox(0,8)[bl]{for agent \(1\)}}
\put(51,54){\line(0,-1){4}}

\put(43,44.5){\line(1,0){15}} \put(43,49.5){\line(1,0){15}}
\put(43,44.5){\line(0,1){5}} \put(58,44.5){\line(0,1){5}}

\put(55,47){\circle*{1}} \put(55,47){\circle{2}}
\put(60,48){\makebox(0,8)[bl]{Agent \(1\)}}
\put(60,45){\makebox(0,8)[bl]{(knapsack,}}
\put(60,42){\makebox(0,8)[bl]{cluster)}}

\put(40,47){\vector(1,0){12}} \put(40,43){\vector(4,1){12}}


\put(53.5,38){\makebox(0,8)[bl]{{\bf ...}}}


\put(48,26){\line(1,0){10}} \put(48,30){\line(1,0){10}}
\put(48,26){\line(0,1){4}} \put(58,26){\line(0,1){4}}

\put(55,28){\circle*{1}} \put(55,28){\circle{2}}
\put(60,29){\makebox(0,8)[bl]{Agent \(j\)}}
\put(60,26){\makebox(0,8)[bl]{(knapsack,}}
\put(60,23){\makebox(0,8)[bl]{cluster)}}

\put(40,32){\vector(4,-1){12}} \put(40,28){\vector(1,0){12}}
\put(40,24){\vector(4,1){12}}


\put(53.5,17){\makebox(0,8)[bl]{{\bf ...}}}


\put(46,07){\line(1,0){12}} \put(46,13){\line(1,0){12}}
\put(46,07){\line(0,1){6}} \put(58,07){\line(0,1){6}}

\put(55,10){\circle*{1}} \put(55,10){\circle{2}}
\put(60,11){\makebox(0,8)[bl]{Agent \(\mu\)}}
\put(60,08){\makebox(0,8)[bl]{(knapsack,}}
\put(60,05){\makebox(0,8)[bl]{cluster)}}

\put(40,14){\vector(4,-1){12}} \put(40,10){\vector(1,0){12}}


\put(08,53){\makebox(0,8)[bl]{Items}}

\put(10,28.5){\oval(20,45)}

\put(06,46){\makebox(0,8)[bl]{\(1\)}}  \put(12,47){\circle*{1}}
\put(13,47){\vector(1,0){12}}

\put(06,37){\makebox(0,8)[bl]{\(2\)}} \put(12,38){\circle*{1}}
\put(13,38){\vector(4,1){12}}

\put(22.5,36){\makebox(0,8)[bl]{Local}}
\put(22.5,32.5){\makebox(0,8)[bl]{assignment}}
\put(22.5,30){\makebox(0,8)[bl]{solutions}}

\put(10,33){\makebox(0,8)[bl]{{\bf ...}}}

\put(06,27){\makebox(0,8)[bl]{\(i\)}}
\put(13,28){\vector(3,-1){12}} \put(12,28){\circle*{1}}

\put(10,14){\makebox(0,8)[bl]{{\bf ...}}}

\put(0.5,16){\makebox(0,8)[bl]{\((n-1)\)}}
\put(13,18){\vector(1,0){12}} \put(12,18){\circle*{1}}

\put(06,09){\makebox(0,8)[bl]{\(n\)}}
\put(13,10){\vector(4,1){12}} \put(12,10){\circle*{1}}

\end{picture}
\end{center}

 In multiple criteria generalized assignment problem,
 vector-like profit
  is considered for each item \(i\)
 (e.g., \cite{levpet10,lev15,scar02,zhang07}):~
 \(c_{ij}=(c^{1}_{ij},...,c^{l}_{ij},...,c^{k}_{ij})\)
  (criteria: \(\{C_{1},...,C_{l},...,C_{k}\}\)).

 A simplified multicriteria problem statement can be
 examined as follows:
 \[\max ~ \sum_{j=1}^{\mu} \sum_{i=1}^{n} ~ c^{1}_{ij} x_{ij},~...~,
 \max ~ \sum_{j=1}^{\mu} \sum_{i=1}^{n} ~ c^{l}_{ij} x_{ij},~...~,
  \max ~ \sum_{j=1}^{\mu} \sum_{i=1}^{n} ~ c^{k}_{ij} x_{ij}, ~~ l = \overline{1,k}\]
 \[s.t.  ~~ \sum_{i=1}^{n} a_{ij} x_{ij} \leq b_{i}, ~ j = \overline{1,\mu},
%
 ~~~ \sum_{j=1}^{\mu}  x_{ij} \leq 1, ~ i =\overline{1,n},
 ~~~ x_{ij} \in \{0,1\}, ~i=\overline{1,n}, ~j=\overline{1,\mu}.\]
 Here it is reasonable to search for
 Pareto-efficient solutions.

%

~~

 {\bf Example 3.2.} A modified numerical example
 for connection of end-users and access points
 in a wireless telecommunication network is based on the example from
 \cite{levpet10,levpet10a,lev15}.
 Four access points are considered as cluster centroids
 and \(14\) end-users are examined as initial objects
 (example from \cite{levpet10} is compressed, Fig. 3.7).

\begin{center}
\begin{picture}(68,50)
\put(00,00){\makebox(0,0)[bl]{Fig. 3.7.
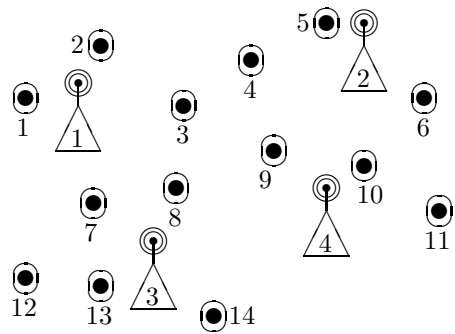
 Example: \(4\)  access points, \(14\) users}}

\put(15,12){\oval(3.2,4)} \put(15,12){\circle*{2}}
\put(13,07){\makebox(0,8)[bl]{\(13\)}}

\put(19,09){\line(1,0){6}} \put(19,09){\line(1,2){3}}
\put(25,09){\line(-1,2){3}} \put(22,15){\line(0,1){3}}
\put(22,18){\circle*{1}} \put(21,09.5){\makebox(0,8)[bl]{\(3\)}}

\put(22,18){\circle{2.5}} \put(22,18){\circle{3.5}}

\put(05,13){\oval(3.2,4)} \put(05,13){\circle*{2}}
\put(03,08){\makebox(0,8)[bl]{\(12\)}}


\put(05,37){\oval(3.2,4)} \put(05,37){\circle*{2}}
\put(03.8,32){\makebox(0,8)[bl]{\(1\)}}

\put(15,44){\oval(3.2,4)} \put(15,44){\circle*{2}}
\put(11,43){\makebox(0,8)[bl]{\(2\)}}

\put(09,30){\line(1,0){6}} \put(09,30){\line(1,2){3}}
\put(15,30){\line(-1,2){3}} \put(12,36){\line(0,1){3}}
\put(12,39){\circle*{1}} \put(11,30.5){\makebox(0,8)[bl]{\(1\)}}

\put(12,39){\circle{2.5}} \put(12,39){\circle{3.5}}

\put(14,23){\oval(3.2,4)} \put(14,23){\circle*{2}}
\put(12.8,18){\makebox(0,8)[bl]{\(7\)}}

\put(30,8){\oval(3.2,4)} \put(30,8){\circle*{2}}
\put(32,7){\makebox(0,8)[bl]{\(14\)}}

\put(38,30){\oval(3.2,4)} \put(38,30){\circle*{2}}
\put(36,25){\makebox(0,8)[bl]{\(9\)}}

\put(50,28){\oval(3.2,4)} \put(50,28){\circle*{2}}
\put(49,23){\makebox(0,8)[bl]{\(10\)}}

\put(42,16){\line(1,0){6}} \put(42,16){\line(1,2){3}}
\put(48,16){\line(-1,2){3}} \put(45,22){\line(0,1){3}}
\put(45,25){\circle*{1}} \put(44,16.5){\makebox(0,8)[bl]{\(4\)}}

\put(45,25){\circle{2.5}} \put(45,25){\circle{3.5}}

\put(60,22){\oval(3.2,4)} \put(60,22){\circle*{2}}
\put(58,17){\makebox(0,8)[bl]{\(11\)}}

\put(25,25){\oval(3.2,4)} \put(25,25){\circle*{2}}
\put(24,20){\makebox(0,8)[bl]{\(8\)}}

\put(26,36){\oval(3.2,4)} \put(26,36){\circle*{2}}
\put(25,31){\makebox(0,8)[bl]{\(3\)}}

\put(35,42){\oval(3.2,4)} \put(35,42){\circle*{2}}
\put(34,37){\makebox(0,8)[bl]{\(4\)}}

\put(45,47){\oval(3.2,4)} \put(45,47){\circle*{2}}
\put(41,46){\makebox(0,8)[bl]{\(5\)}}

\put(58,37){\oval(3.2,4)} \put(58,37){\circle*{2}}
\put(57,32){\makebox(0,8)[bl]{\(6\)}}

\put(47,38){\line(1,0){6}} \put(47,38){\line(1,2){3}}
\put(53,38){\line(-1,2){3}} \put(50,44){\line(0,1){3}}
\put(50,47){\circle*{1}} \put(49,38.5){\makebox(0,8)[bl]{\(2\)}}

\put(50,47){\circle{2.5}} \put(50,47){\circle{3.5}}

\end{picture}
\end{center}

 Let
 ~\( \{1,...,i,...,n \}\) be a set of users (here: \(14\))
 and
 ~\( \{1,...,j,...,\mu \}\) be a set of access points (here: \(4\)).

 Each user  \(i\) is described by  parameters
 (a compressed set of parameters):
 (i) coordinates \(( x_{i},y_{i},x_{i} )\);
 (ii) parameter corresponding to required frequency bandwidth
 (e.g., 1 Mbit/s ... 10 Mbit/s)
  \(f_{i}\);
 (iii) maximal possible number of access points for connection \( \kappa_{i} \leq \mu \)
 (here \(\kappa_{i} = 1 ~ \forall i\),
 i.e., each user is assigned to the only one access point/cluster);
%
%
 (iv) required reliability of information transmission
 \(r_{i}\).
 Note, in multi-assignment problem ~ \(\kappa_{i}  \geq 1\).
 Each access point is described by parameters
  (a compressed set of parameters):
 (a) coordinates of access point \(( x_{j},y_{j},x_{j} )\),
 (b) parameter corresponding to maximal possible traffic
 (i.e., maximum of possible bandwidth) \( f_{j} \),
 (c) maximal possible number of users under service
 \( k_{j} \),
 (d) reliability of channel for data transmission \( r_{j} \),
 (e) admissible proximity (distance) \( d_{j} \).
 Table 3.14 and Table 3.15 contains parameters estimates of the access points
 and users.

\begin{center}
{\bf Table 3.14.} Parameter estimates of access points  \\
\begin{tabular}{| c |c c c c| c| c| c|c|}
\hline

 Access & Coordinates: &\(x_{j}\)& \(y_{j}\)&\(z_{j}\)
 &Bandwidth &Number &Reliability&Admissible \\

 point &&&&&\(f_{j}\)&of users &\(r_{j}\)& distance \\

 \(j\) &&&&&         &\(n_{j}\)&         & \(d_{j}\)  \\
\hline
 1 &&\(50\) &\(157\)&\(10\)&\(30\) &\(4\) &\(10\) &\(10\) \\
 2 &&\(150\)&\(165\)&\(10\)&\(30\) &\(5\) &\(15\) &\(10\) \\
 3 &&\(72\) &\(102\)&\(10\)&\(42\) &\(6\) &\(10\) &\(6\) \\
 4 &&\(140\)&\(112\)&\(10\)&\(32\) &\(5\) &\(8\)  &\(9\) \\
\hline
\end{tabular}
\end{center}

\begin{center}
{\bf Table 3.15.} Parameter estimates of end users  \\
\begin{tabular}{| c |c c c c| c| c| }
\hline
 User  & Coordinates: &\(x_{j}\)& \(y_{j}\)&\(z_{j}\)
 &Bandwidth &Reliability \\

 \(i\) &&&&
 &\(f_{i}\)&\(r_{i}\)  \\
\hline
 1 &&\(30\) &\(165\)&\(5\) &\(10\) &\(5\) \\
 2 &&\(58\) &\(174\)&\(5\) &\(5\)  &\(9\)  \\
 3 &&\(88\) &\(156\)&\(0\) &\(6\)  &\(6\)  \\
 4 &&\(110\)&\(169\)&\(5\) &\(7\)  &\(5\) \\
 5 &&\(145\)&\(181\)&\(3\) &\(5\)  &\(4\) \\
 6 &&\(170\)&\(161\)&\(5\) &\(7\)  &\(4\) \\
 7 &&\(52\) &\(134\)&\(5\) &\(6\)  &\(8\)  \\
 8 &&\(86\) &\(134\)&\(3\) &\(6\)  &\(7\)  \\
 9 &&\(120\)&\(140\)&\(6\) &\(4\)  &\(6\)  \\
 10&&\(150\)&\(136\)&\(3\) &\(6\)  &\(7\) \\
 11&&\(175\)&\(125\)&\(1\) &\(8\)  &\(5\) \\
 12&&\(27\) &\(109\)&\(7\) &\(8\)  &\(5\)  \\
 13&&\(55\) &\(105\)&\(2\) &\(7\)  &\(10\) \\
 14&&\(98\) &\(89\) &\(3\) &\(10\) &\(10\) \\
\hline
\end{tabular}
\end{center}

 As a result, each pair ``user-access point''
%
%
  can be described by the following parameters
  (a compressed set of parameters):
 ~(1) proximity (e.g., Euclidean distance)
    ~\(d_{ij}\),
 ~(2) level of reliability
    ~\(r_{ij}\),
 ~(3) parameter of using  bandwidth
%
  ~\(f_{ij}\).

 Clearly,
 Euclidean distances
 between users and access points
 \(\{d_{ij}\}\)
 can be calculated on the basis coordinates from Table 3.14 and Table 3.15.
%
 Thus, the following parameter vector is obtained
%
%
 \( \widehat{c_{ij}} = (d_{ij},r_{ij},f_{ij} ) \)~
 (\(i=\overline{1,n}, j=\overline{1,\mu}\)).
 Further,
 the parameter vector can be transformed into a profit
  \(c_{ij}\)
%
 (i.e., mapping of vector estimate into ordinal scale \([1,2,3]\),
 \(3\) corresponds to the best level,
 multicriteria ranking based on outranking ELECTRE technique can be used).

 The assignment of user \( i\) to access point  \( j\)
 is defined by Boolean variable
 \(x_{ij}\) (\(x_{ij} =1\) in the case of assignment
  \( i\) to  \( j\)  and   \(x_{ij} =0\)  otherwise).
 The assignment solution
 is defined by Boolean matrix
 ~\(X = || x_{ij} ||, ~ i=\overline{1,n}, ~j=\overline{1,\mu}\).
 Finally, the problem  is:
 \[\max ~\sum_{j=1}^{\mu} ~\sum_{i=1}^{n} ~ c_{ij} ~x_{ij} \]
%
%
%
%
%
%
 \[ s.t.   ~~~~
 \sum_{i=1}^{n}  f_{ij} x_{ij}  \leq  f_{j}
 ~ \forall j =\overline{1,\mu};
  ~~~~ \sum_{i=1}^{n}  x_{ij}  \leq  k_{j}
 ~ \forall j=\overline{1,\mu};
 ~~~~ \sum_{j=1}^{\mu} x_{ij}\leq 1 ~\forall i=\overline{1,n};\]
%
%
  \[x_{ij} = 0
  ~~if~~ d_{ij} > d_{j} ~~\forall ~i=\overline{1,n}, ~~\forall
  ~j=\overline{1,\mu};
%
 ~~~~  x_{ij} = 0
  ~~if~~ r_{ij} > r_{j} ~~\forall ~i=\overline{1,n}, ~~\forall ~j=\overline{1,\mu};\]
%
%
  \[ x_{ij} \in \{0,1\} ~~\forall ~i=\overline{1,n},
  ~\forall ~j=\overline{1,\mu}.\]

 A numerical example of the assignment solution (i.e., clustering solution)
 is depicted in Fig. 3.8.
 In \cite{levpet10},
 the described problem is examined as
 generalized multiple assignment problem
 (extended version).

\begin{center}
\begin{picture}(68,50)
\put(00,00){\makebox(0,0)[bl]{Fig. 3.8.
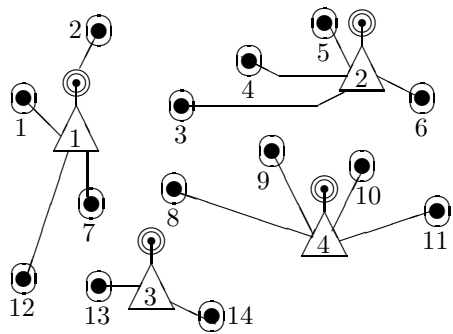
 Assignment of users to access points}}

\put(15,12){\oval(3.2,4)} \put(15,12){\circle*{2}}
\put(13,07){\makebox(0,8)[bl]{\(13\)}}
\put(15,12){\line(1,0){5.5}}

\put(19,09){\line(1,0){6}} \put(19,09){\line(1,2){3}}
\put(25,09){\line(-1,2){3}} \put(22,15){\line(0,1){3}}
\put(22,18){\circle*{1}} \put(21,09.5){\makebox(0,8)[bl]{\(3\)}}

\put(22,18){\circle{2.5}} \put(22,18){\circle{3.5}}
\put(24.5,10){\line(2,-1){6}}

\put(05,13){\oval(3.2,4)} \put(05,13){\circle*{2}}
\put(03,08){\makebox(0,8)[bl]{\(12\)}}

\put(11,30){\line(-1,-3){6}}

\put(05,37){\oval(3.2,4)} \put(05,37){\circle*{2}}
\put(03.8,32){\makebox(0,8)[bl]{\(1\)}}

\put(15,46){\oval(3.2,4)} \put(15,46){\circle*{2}}
\put(11,45){\makebox(0,8)[bl]{\(2\)}}

\put(15,46){\line(-1,-2){2.5}}


\put(09,30){\line(1,0){6}} \put(09,30){\line(1,2){3}}
\put(15,30){\line(-1,2){3}} \put(12,36){\line(0,1){3}}
\put(12,39){\circle*{1}} \put(11,30.5){\makebox(0,8)[bl]{\(1\)}}

\put(12,39){\circle{2.5}} \put(12,39){\circle{3.5}}
\put(10,32){\line(-1,1){5}}
\put(13.5,30){\line(0,-1){6}}

\put(14,23){\oval(3.2,4)} \put(14,23){\circle*{2}}
\put(12.8,18){\makebox(0,8)[bl]{\(7\)}}

\put(30,8){\oval(3.2,4)} \put(30,8){\circle*{2}}
\put(32,7){\makebox(0,8)[bl]{\(14\)}}

\put(38,30){\oval(3.2,4)} \put(38,30){\circle*{2}}
\put(36,25){\makebox(0,8)[bl]{\(9\)}}
\put(38,30){\line(1,-2){05.5}}

\put(50,28){\oval(3.2,4)} \put(50,28){\circle*{2}}
\put(49,23){\makebox(0,8)[bl]{\(10\)}}

\put(42,16){\line(1,0){6}} \put(42,16){\line(1,2){3}}
\put(48,16){\line(-1,2){3}} \put(45,22){\line(0,1){3}}
\put(45,25){\circle*{1}} \put(44,16.5){\makebox(0,8)[bl]{\(4\)}}

\put(45,25){\circle{2.5}} \put(45,25){\circle{3.5}}
\put(43.5,18.5){\line(-3,1){19}}
\put(46,19.4){\line(1,2){4.5}}
\put(47,18){\line(3,1){13}}

\put(60,22){\oval(3.2,4)} \put(60,22){\circle*{2}}
\put(58,17){\makebox(0,8)[bl]{\(11\)}}

\put(25,25){\oval(3.2,4)} \put(25,25){\circle*{2}}
\put(24,20){\makebox(0,8)[bl]{\(8\)}}

\put(26,36){\oval(3.2,4)} \put(26,36){\circle*{2}}
\put(25,31){\makebox(0,8)[bl]{\(3\)}}

\put(35,42){\oval(3.2,4)} \put(35,42){\circle*{2}}
\put(34,37){\makebox(0,8)[bl]{\(4\)}}

\put(45,47){\oval(3.2,4)} \put(45,47){\circle*{2}}
\put(44,42){\makebox(0,8)[bl]{\(5\)}}

\put(58,37){\oval(3.2,4)} \put(58,37){\circle*{2}}
\put(57,32){\makebox(0,8)[bl]{\(6\)}}

\put(47,38){\line(1,0){6}} \put(47,38){\line(1,2){3}}
\put(53,38){\line(-1,2){3}} \put(50,44){\line(0,1){3}}
\put(50,47){\circle*{1}} \put(49,38.5){\makebox(0,8)[bl]{\(2\)}}

\put(50,47){\circle{2.5}} \put(50,47){\circle{3.5}}
\put(48,38){\line(-2,-1){04}} \put(44,36){\line(-1,0){17}}
\put(48,40){\line(-1,0){09}} \put(39,40){\line(-2,1){5}}
\put(48.4,41.2){\line(-1,2){3}}
\put(52,40){\line(2,-1){5}}

\end{picture}
\end{center}

\subsection{Graph based clustering}

\subsubsection{Minimum spanning tree based clustering}

 The preliminary building of minimum trees
 is widely used in many combinatorial problems
 (e.g., \cite{gar79}).
 The algorithmic complexity estimate for
 this spanning problem over graph equals
 \(O (n\log n)\) (\(n\) is the number of graph vertices).
 Minimum spanning tree based clustering algorithms
 have been studied and applied by many researchers
 (e.g.,
 \cite{gow69,gry06,kumar05,mul12,paiv05,peter10,srin94,wang09,xu01,zhong10}).
 The basic stages of the algorithms are as follows:

~~

 {\it Stage 1.} Calculation of  distance/proximity matrix \(Z\).

 {\it Stage 2.} Design of the corresponding graph \(G\).

 {\it Stage 3.}  Building of the minimum spanning tree \(T\) for
 graph \(G\).

 {\it Stage 4.} Clustering of the vertices of tree \(T\)
 (e.g., by algorithm of deletion of branches, by algorithm of hierarchical clustering).

 {\it Stage 5.} Stopping.

~~

 Further, the usage of hierarchical clustering at stage 4 is
 considered.
%
%
%
%
  Complexity estimates for minimum spanning tree clustering algorithm  (by stages)
   are presented in Table 3.16.


\begin{center}
 {\bf Table 3.16.}  Complexity estimates of stages for
 minimum spanning tree based clustering
 \\
\begin{tabular}{| l | l|c |}
\hline
  Stage &Description &Complexity estimate   \\
  & &(running time)  \\
\hline
 Stage 1.& Calculate distance matrix \(Z\)&\(O (n^{2})\) \\
 Stage 2.& Design the corresponding graph \(G\)&\(O (n^{2})\) \\
 Stage 3.& Building the minimum spanning tree &\(O (n \log n)\) \\
 Stage 4.& Clustering of the tree vertices &\(O(n \log n)\)\\
 Stage 5.& Stopping        & \(O(1)\)\\
\hline
\end{tabular}
\end{center}

  Stages 3, 4, 5 correspond to the situation when
  a graph is examined as initial data.
  In this case, complexity of the algorithm equals \(O(n \log n)\).

~~

 {\bf Example 3.3.}
 A numerical example for elements
 \(A = \{1,2,3,4,5,6,7,8,9,10,11,12\}\)
  illustrates
 building of graph \(G=(A,E)\) (corresponding proximity matrix \(Z\)),
 minimum spanning tree \(T=(A,E')\),
 and clustering solution \(\widehat{X}= \{X_{1},X_{2},X_{3}\}\)
 (cluster size \( \leq 4\)).
 The following is implemented:
%

~~

 Proximity matrix \(Z\) \(\Rightarrow\) Graph \(G=(A,E)\)
 \(\Rightarrow\)
 Tree \(T=(A,E')\)
 \(\Rightarrow\) Clustering solution \(\widehat{X}=
 \{X_{1},X_{2},X_{3}\}\).

~~

 The method for building the spanning tree
 is used as in example 3.1.
 Table 3.17 contains proximity matrix
 (symbol ``\(\star\)'' corresponds to a very big value).

\begin{center}
{\bf Table 3.17.} Proximities for example (edge \((i_{1},i_{2})\)) \\
\begin{tabular}{| c |c c c c c c c c c c c c |}
\hline
 \(i_{1}\) & \(i_{2}:\)& \(2\)&\(3\)&\(4\)&\(5\)&\(6\) &\(7\)&\(8\)&\(9\)&\(10\)&\(11\)&\(12\) \\
\hline
 1 &&\(0.3\)&\(1.4\)&\(1.45\)&\(\star\) &\(\star\)&\(\star\) &\(\star\) & \(\star\)& \(\star\)& \(\star\)&\(\star\)\\
 2 &&&\(0.3\)&\(\star\)&\(\star\)&\(2.6\)&\(0.2\)&\(1.8\)&\(\star\)&\(\star\)&\(\star\)&\(\star\)\\
 3 &&&&\(0.4\)&\(\star\)&\(\star\)&\(1.65\)&\(0.25\)&\(\star\)&\(\star\)&\(\star\)&\(\star\)\\
 4 &&&&&\(0.4\)&\(\star\)&\(\star\)&\(0.45\)&\(1.9\)&\(\star\)&\(\star\)&\(\star\)\\

 5 &&&&&&\(\star\)&\(\star\)&\(\star\)&\(0.35\)&\(1.5\)&\(\star\)&\(\star\)\\
 6 &&&&&&&\(0.1\)&\(\star\)&\(\star\)&\(\star\)&\(1.4\)&\(\star\)\\
 7 &&&&&&&&\(0.41\)&\(\star\)&\(\star\)&\(0.4\)&\(\star\)\\
 8 &&&&&&&&&\(0.9\) &\(\star\)& \(2.1\)&\(\star\)\\
 9 &&&&&&&&&&\(0.15\)&\(\star\)&\(0.5\)\\
 10 &&&&&&&&&&&\(\star\)&\(2.0\)\\
 11 &&&&&&&&&&&&\(2.5\)\\

\hline
\end{tabular}
\end{center}

 Fig. 3.9 depicts corresponding graph,
  \(G=(A,E)\),
 Fig. 3.10 depicts spanning tree
  \(T=(A,E')\),
 and clustering solution.
 Generally, it is reasonable
 to point out threshold based modification
 of graph \(G = (A,E)\)
 over object set \(A\):
 deletion of edges by condition:
 the weight ``\(>\)'' the threshold.
 Decreasing the threshold leads to decreasing the
 cardinality of \(E\).
 This process
  can be very useful for analysis and processing of
 initial data in clustering.
 Let us consider an illustration of the above-mentioned process
 on the basis graph from example 3.3
 (basic proximity matrix from Table 3.17):
 (i) threshold equals \(2.6\):~ graph  \(G=(A,E)\) in Fig. 3.9;
 (ii) threshold equals \(1.4\):~ graph \(G^{1}=(A,E^{1})\)  in Fig. 3.11;
 (iii) threshold equals \(0.5\):~ graph \( G^{2}=(A,E^{2}) = T=(A,E')\) in Fig. 3.10 (here:
  spanning tree);
 (iv) threshold equals \(0.3\):~ graph \(G^{3}=(A,E^{3})\) in Fig. 3.12.

\begin{center}
\begin{picture}(65,44)

\put(04,00){\makebox(0,0)[bl]{Fig. 3.9. Graph
 \(G=(A,E)\)}}


\put(05,20){\line(1,1){10}}

\put(05,20){\line(1,-1){10}}

\put(15,30){\line(1,0){10}}

\put(25,40){\line(1,-1){10}}

\put(35,30){\line(1,-2){05}}

\put(25,30){\line(1,0){10}}

\put(35,30){\line(1,0){10}}

\put(15,30){\line(3,-2){15}}

\put(30,20){\line(3,-2){15}}

\put(15,30){\line(1,1){10}}

\put(15,10){\line(3,2){15}}

\put(15,10){\line(1,0){30}}

\put(15,20){\line(1,1){10}}

\put(50,20){\line(-1,2){05}}

\put(50,20){\line(-1,-2){05}}


\put(05,20){\circle*{1}} \put(02,19){\makebox(0,8)[bl]{\(6\)}}
\put(05,20){\line(1,0){10}}

\put(15,20){\circle*{1}} \put(12,21){\makebox(0,8)[bl]{\(7\)}}
\put(15,30){\circle*{1}} \put(11,29){\makebox(0,8)[bl]{\(2\)}}

\put(15,20){\line(0,1){10}} \put(15,20){\line(0,-1){10}}

\put(15,10){\circle*{1}} \put(11,08){\makebox(0,8)[bl]{\(11\)}}



\put(30,20){\circle*{1}} \put(31.5,20.6){\makebox(0,8)[bl]{\(8\)}}
\put(15,20){\line(1,0){15}}

\put(25,30){\circle*{1}} \put(26,30.5){\makebox(0,8)[bl]{\(3\)}}
\put(35,30){\circle*{1}} \put(35,31){\makebox(0,8)[bl]{\(4\)}}

\put(30,20){\line(-1,2){05}} \put(30,20){\line(1,2){05}}

\put(25,40){\circle*{1}} \put(26.5,39){\makebox(0,8)[bl]{\(1\)}}
\put(25,40){\line(0,-1){10}}



\put(30,20){\line(1,0){10}}

\put(40,20){\circle*{1}} \put(39,22){\makebox(0,8)[bl]{\(9\)}}

\put(50,20){\circle*{1}} \put(45,21){\makebox(0,8)[bl]{\(10\)}}

\put(40,20){\line(1,0){10}} \put(40,20){\line(1,2){05}}
\put(40,20){\line(1,-2){05}}

\put(45,30){\circle*{1}} \put(44,25){\makebox(0,8)[bl]{\(5\)}}
\put(45,10){\circle*{1}} \put(46,08){\makebox(0,8)[bl]{\(12\)}}

\end{picture}
%
\begin{picture}(57,44)
\put(00,00){\makebox(0,0)[bl]{Fig. 3.10. Spanning tree
 \(T=(A,E')\)
  }}

\put(03,33){\makebox(0,8)[bl]{Cluster \(X_{1}\)}}
\put(11.5,20){\oval(15,23)}


\put(05,20){\circle*{1}} \put(05,21){\makebox(0,8)[bl]{\(6\)}}
\put(05,20){\line(1,0){10}}

\put(15,20){\circle*{1}} \put(12,21){\makebox(0,8)[bl]{\(7\)}}
\put(15,30){\circle*{1}} \put(12,28.5){\makebox(0,8)[bl]{\(2\)}}

\put(15,20){\line(0,1){10}} \put(15,20){\line(0,-1){10}}

\put(15,10){\circle*{1}} \put(11,11){\makebox(0,8)[bl]{\(11\)}}

\put(36,40){\makebox(0,8)[bl]{Cluster \(X_{2}\)}}
\put(30,30.5){\oval(15,24)}


\put(30,20){\circle*{1}} \put(31.5,20.6){\makebox(0,8)[bl]{\(8\)}}
\put(15,20){\line(1,0){15}}

\put(25,30){\circle*{1}} \put(26,30.5){\makebox(0,8)[bl]{\(3\)}}
\put(35,30){\circle*{1}} \put(35,31){\makebox(0,8)[bl]{\(4\)}}

\put(30,20){\line(-1,2){05}} \put(30,20){\line(1,2){05}}

\put(25,40){\circle*{1}} \put(26.5,39){\makebox(0,8)[bl]{\(1\)}}
\put(25,40){\line(0,-1){10}}

\put(38.5,32.5){\makebox(0,8)[bl]{Cluster \(X_{3}\)}}
\put(45,20){\oval(13,23)}


\put(30,20){\line(1,0){10}}

\put(40,20){\circle*{1}} \put(39,22){\makebox(0,8)[bl]{\(9\)}}

\put(50,20){\circle*{1}} \put(47,21){\makebox(0,8)[bl]{\(10\)}}

\put(40,20){\line(1,0){10}} \put(40,20){\line(1,2){05}}
\put(40,20){\line(1,-2){05}}

\put(45,30){\circle*{1}} \put(44,25){\makebox(0,8)[bl]{\(5\)}}
\put(45,10){\circle*{1}} \put(44,13){\makebox(0,8)[bl]{\(12\)}}

\end{picture}
\end{center}

\begin{center}
\begin{picture}(65,42)

\put(02,00){\makebox(0,0)[bl]{Fig. 3.11. Graph
 \(G^{1}=(A,E^{1})\)}}



\put(05,20){\line(1,-1){10}}

\put(15,30){\line(1,0){10}}



\put(25,30){\line(1,0){10}}

\put(35,30){\line(1,0){10}}

\put(15,30){\line(3,-2){15}}


\put(15,30){\line(1,1){10}}







\put(05,20){\circle*{1}} \put(02,19){\makebox(0,8)[bl]{\(6\)}}
\put(05,20){\line(1,0){10}}

\put(15,20){\circle*{1}} \put(12,21){\makebox(0,8)[bl]{\(7\)}}
\put(15,30){\circle*{1}} \put(11,29){\makebox(0,8)[bl]{\(2\)}}

\put(15,20){\line(0,1){10}} \put(15,20){\line(0,-1){10}}

\put(15,10){\circle*{1}} \put(11,08){\makebox(0,8)[bl]{\(11\)}}



\put(30,20){\circle*{1}} \put(31.5,20.6){\makebox(0,8)[bl]{\(8\)}}
\put(15,20){\line(1,0){15}}

\put(25,30){\circle*{1}} \put(26,30.5){\makebox(0,8)[bl]{\(3\)}}
\put(35,30){\circle*{1}} \put(35,31){\makebox(0,8)[bl]{\(4\)}}

\put(30,20){\line(-1,2){05}} \put(30,20){\line(1,2){05}}

\put(25,40){\circle*{1}} \put(26.5,38){\makebox(0,8)[bl]{\(1\)}}
\put(25,40){\line(0,-1){10}}



\put(30,20){\line(1,0){10}}

\put(40,20){\circle*{1}} \put(39,22){\makebox(0,8)[bl]{\(9\)}}

\put(50,20){\circle*{1}} \put(47,21){\makebox(0,8)[bl]{\(10\)}}

\put(40,20){\line(1,0){10}} \put(40,20){\line(1,2){05}}
\put(40,20){\line(1,-2){05}}

\put(45,30){\circle*{1}} \put(44,31){\makebox(0,8)[bl]{\(5\)}}
\put(45,10){\circle*{1}} \put(46,08){\makebox(0,8)[bl]{\(12\)}}

\end{picture}
%
\begin{picture}(57,42)

\put(02,00){\makebox(0,0)[bl]{Fig. 3.12. Graph
 \(G^{3}=(A,E^{3})\)}}




\put(15,30){\line(1,0){10}}














\put(05,20){\circle*{1}} \put(02,19){\makebox(0,8)[bl]{\(6\)}}
\put(05,20){\line(1,0){10}}

\put(15,20){\circle*{1}} \put(12,21){\makebox(0,8)[bl]{\(7\)}}
\put(15,30){\circle*{1}} \put(11,29){\makebox(0,8)[bl]{\(2\)}}

\put(15,20){\line(0,1){10}}


\put(15,10){\circle*{1}} \put(11,08){\makebox(0,8)[bl]{\(11\)}}



\put(30,20){\circle*{1}} \put(31.5,20.6){\makebox(0,8)[bl]{\(8\)}}

\put(25,30){\circle*{1}} \put(26,30.5){\makebox(0,8)[bl]{\(3\)}}
\put(35,30){\circle*{1}} \put(35,31){\makebox(0,8)[bl]{\(4\)}}

\put(30,20){\line(-1,2){05}}


\put(25,40){\circle*{1}} \put(26.5,38){\makebox(0,8)[bl]{\(1\)}}
\put(25,40){\line(0,-1){10}}




\put(40,20){\circle*{1}} \put(39,21){\makebox(0,8)[bl]{\(9\)}}

\put(50,20){\circle*{1}} \put(47,21){\makebox(0,8)[bl]{\(10\)}}

\put(40,20){\line(1,0){10}}


\put(45,30){\circle*{1}} \put(44,31){\makebox(0,8)[bl]{\(5\)}}
\put(45,10){\circle*{1}} \put(46,08){\makebox(0,8)[bl]{\(12\)}}

\end{picture}
\end{center}
%


 As a result, a useful  structure can be found.
 The described by example procedure is an important auxiliary
 problem.
 An additional significant problem
 consists in analysis of the obtained graph, for example:
 (a) connectivity,
 (b) similarity to tree (or hierarchy, clique).
 Further,
 a modified version of adaptive minimum spanning tree clustering
 algorithm is examined as  follows.
 Initial data are:
 (a) set of objects/alternatives
 \(A = \{A_{1},...,A_{i},...,A_{n}\}\),
 (b) set of parameters/criteria
  \(\overline{C} = \{C_{1},...,C_{j},...,C_{m}\}\),
 (c) estimate matrix
  \(X = \{x_{ij} \}\), \(i=\overline{1,n}, j=\overline{1,m}\),
 where \(x_{ij}\) is estimate of \(A_{i}\) upon criterion
 \(C_{j}\) (a qualitative scale is considered).
 The algorithm consists of the following stages:

~~

 {\it Stage 1.}
 Calculation of  proximity matrix
 ~\(Z = \{z_{ik} \}\), \(i=\overline{1,n}, ~k=\overline{1,n}\),
 where \(z_{ik}\) is estimate of
 proximity (distance) between
 \(A_{i}\) and  \(A_{k}\)
 (e.g., Euclidean metric is used).
 Evidently,
 \(z_{ii} = 0,  \forall i = \overline{1,n} \).

 {\it Stage 2.} Transformation of matrix \(Z\) into ordinal matrix \(Y = \{y_{ik} \}\).
  Let us consider the maximum and minimum values of elements of
  matrix \(Z\):
 \(z^{min}=\min_{\forall i=\overline{1,n},i=\overline{1,k}} ~\{z_{ik}\}\),
 ~\(z^{max}=\max_{\forall i=\overline{1,n},i=\overline{1,k}} ~\{z_{ik}\}\).
 Thus an interval is obtained  \([z^{min},z^{max}]\) and
 \(d=z^{max}- z^{min}\).
 Now  an additional integer parameter \(\delta\)
 (e.g., \(3,4,5,6\)) is used.
 Let \(\delta = 5\).
 Then elements of new matrix \(Y\)
 (i.e, adjacency matrix)
 are based on the following calculation:
%
    \[y_{ik} = \left\{ \begin{array}{ll}
               0, & \mbox{if $  ~0.0 \leq z_{ik} \leq d /\delta , $}\\
               1, & \mbox{if $  ~d/\delta < z_{ik} \leq 2d/\delta, $}\\
               2, & \mbox{if $ ~2d/\delta < z_{ik} \leq 3d/\delta, $}\\
               3, & \mbox{if $ ~3d/\delta < z_{ik} \leq 4/\delta, $}\\
              4, & \mbox{if $ ~4d/\delta < z_{ik} \leq d. $}
               \end{array}
               \right. \]

 {\it Stage 3.} Obtaining an interconnected graph over elements \(A\)
 (iterative approach):

  Let \(\Delta = 1,2,...\) be an integer algorithmic parameter
  (for the algorithm cycle).

 {\it Step 3.1.} Initial value  \(\Delta = 1 \).

 {\it Step 3.2}
  Transformation of ordinal matrix \(Y\) into Boolean matrix
  \(B = \{b_{ik}\}\):

  \[b_{ik} = \left\{ \begin{array}{ll}
               1, & \mbox{if $  ~y_{ik} < \Delta , $}\\
               0, & \mbox{if $  ~y_{ik} \geq \Delta. $}
               \end{array}
               \right. \]

 {\it Step 3.3.} Building a graph over elements \(A\):
 \(G^{\Delta} =  (A,\Gamma^{\Delta} )\),
 where \(\Gamma^{\Delta}\) is the set of edges,
 edge \(( A_{i},A_{k}  )\) exists if
 \(b_{ik} = 1\).

 {\it Step 3.4.} Analysis of connectivity for graph
 \(G^{\Delta} =  (A,\Gamma^{\Delta} )\).
 If the graph is connected,
  then GOTO  {\it Step 3.6}.

 {\it Step 3.5.}  \(\Delta =\Delta + 1 \)
 and GOTO {\it Step 3.2}

 {\it Step 3.6.} Building of minimum spanning tree for graph
  \(G^{\Delta} =  (A,\Gamma^{\Delta} )\): ~
 \(T^{\Delta} =  (A,\widehat{E}^{\Delta} )\).

 Here, several well-known algorithms can be used, for example:
 Borovka's algorithm
 Prim's algorithm,
 Kruskal's algorithm
 \cite{aho74,gabow86,gar79,cormen90,pet02,yao75}.
 Complexity estimate of the algorithms is:
 \(O(p \log n )\) (or less \cite{yao75})
 (\(p\) is the number of edges, \(n\) is the number of vertices).
 It is necessary to take into account
 for each edge \( \gamma \in \Gamma\)
 its weight as follows:
 proximity value \(z_{ik}\) for corresponding element of \(Z\).

 {\it Step 3.7.}
 Clustering set \(A\) on the basis of spanning tree
 \(T^{\Delta} =  (A,\widehat{E}^{\Delta} )\)
 while taking into account
 an algorithmic parameter:
 a number of  elements \(\alpha \) in each obtained cluster
 \( \alpha' \leq  \alpha \leq \alpha'' \),
 for example
  \( \alpha' = 4\), \( \alpha'' = 6\).
  The constrains above have to be based
 on the engineering analysis of the applied problem.

 {\it Stage 4.} Stop.

~~

 Complexity estimates for the described adaptive algorithm (by stages) is presented
 in Table 3.18.
 Thus, the general complexity estimate
 (running time)
  of the described adaptive algorithm equals
 \(O (n^{2})\).

 Generally, the problem of \(k\)-balanced partitioning a tree is NP-hard
 (\(k\) is the number of elements in each cluster of clustering solution)
 \cite{feld15}.

 Note, the obtained clustering solution has a property:
 ``modularity''.  This can be very important for many applied problems
 (e.g., close cardinalities of clusters/groups:
 local region elements in communication network,
  student teams).


\newpage
\begin{center}
 {\bf Table 3.18.}  Complexity estimates for adaptive minimum spanning tree based algorithm\\
\begin{tabular}{| l |l| c |}
\hline
  Stage/step & Description&Complexity estimate   \\
  & &(running time)  \\
\hline

 Stage 1 &
 Calculate
 the distance matrix \(Z\)&\(O (n^{2})\) \\

 Stage 2 &Transformation of matrix \(Z\) into ordinal matrix \(Y\)  &\(O (n^{2})\) \\

 Stage 3 &Design of interconnected graph over elements \(A\) &\(O (n^{2})\) \\
 Step 3.1&Specifying the start of the cycle & \(O (1)\) \\
 Step 3.2&Transformation of matrix \(Y\) into Boolean matrix \(B\) & \(O (n^{2})\) \\
 Step 3.3 &Building the graph \(G\) that corresponds to matrix \(B\)& \(O (n^{2})\) \\
 Step 3.4 &Analysis of connectivity of graph \(G\)& \(O (n)\) \\
 Step 3.5 &Correction of cycle parameter& \(O (1)\) \\
 Step 3.6 &Building the minimum spanning tree \(T\) for graph \(G\)& \(O (p \log n)\) \\
 Step 3.7 &Clustering the vertices (elements \(A\)) of the spanning tree \(T\) & \(O (n)\) \\
  & while taking into account the constraints for cluster size &  \\

 Stage 4.&Stopping& \(O(1)\)\\

\hline
\end{tabular}
\end{center}

\subsubsection{Clique based clustering}

 Here, an initial graph  \(G = (A,E)\) is examined as initial data.
 In a clique (complete graph/subgraph),
 each vertex is connected to all other the vertices
 (Fig. 3.13).
 A quasi-clique can be examined, for example,
 as a clique without one-two edges.
 The cliques (or quasi-cliques)
 form a very strong clusters
 (from the viewpoint of interconnection).
 The problem of finding a maximal clique in a graph
 is a well-known NP-hard problem
 (e.g., \cite{gar79,karp72}).
 Thus, heuristics or enumerative methods have been used for the
 problem.

\begin{center}
\begin{picture}(130,24)

\put(37.5,00){\makebox(0,0)[bl] {Fig. 3.13.
 Illustration of cliques}}

\put(01,09){\makebox(0,8)[bl]{One-vertex}}
\put(05,05.5){\makebox(0,8)[bl]{clique}}

\put(10,17){\oval(17,07)} \put(10,17){\circle*{1}}

\put(26,09){\makebox(0,8)[bl]{Two-vertex}}
\put(30,05.5){\makebox(0,8)[bl]{clique}}

\put(35,17.5){\oval(17,10)}

\put(35,15){\circle*{1}} \put(35,20){\circle*{1}}

\put(35,15){\line(0,1){5}}

\put(51,09){\makebox(0,8)[bl]{Three-vertex}}
\put(56,05.5){\makebox(0,8)[bl]{clique}}

\put(60,17.5){\oval(17,10)}

\put(55,15){\circle*{1}} \put(60,20){\circle*{1}}
\put(65,15){\circle*{1}}

\put(55,15){\line(1,0){10}} \put(55,15){\line(1,1){5}}
\put(65,15){\line(-1,1){5}}

\put(76,09){\makebox(0,8)[bl]{Four-vertex}}
\put(81,05.5){\makebox(0,8)[bl]{clique}}

\put(85,17.5){\oval(17,10)}

\put(80,15){\circle*{1}} \put(80,20){\circle*{1}}
\put(90,15){\circle*{1}} \put(90,20){\circle*{1}}

\put(80,15){\line(1,0){10}} \put(80,15){\line(0,1){5}}
\put(90,15){\line(0,1){5}} \put(80,20){\line(1,0){10}}

\put(80,15){\line(2,1){10}} \put(90,15){\line(-2,1){10}}


\put(101,09){\makebox(0,8)[bl]{Five-vertex}}
\put(106,05.5){\makebox(0,8)[bl]{clique}}

\put(110,18){\oval(17,12)}

\put(110,23){\circle*{1}}

\put(110,23){\line(-1,-1){5}} \put(110,23){\line(1,-1){5}}
\put(110,23){\line(-1,-2){5}} \put(110,23){\line(1,-2){5}}

\put(105,13){\circle*{1}} \put(105,18){\circle*{1}}
\put(115,13){\circle*{1}} \put(115,18){\circle*{1}}

\put(105,13){\line(1,0){10}} \put(105,13){\line(0,1){5}}
\put(115,13){\line(0,1){5}} \put(105,18){\line(1,0){10}}

\put(105,13){\line(2,1){10}} \put(115,13){\line(-2,1){10}}


\put(125,17){\makebox(0,8)[bl]{{\bf ...}}}


\end{picture}
\end{center}

 Clique-based clustering process can be organized as
 a series of clique problems \cite{gar79}:

~~

 {\it Stage 1.} Finding the ``maximal clique''
 (or maximal ``quasi-clique'')
 in graph \(G = (A,E)\):~ subgraph \(H = (B,V)\)
 (\(H \subseteq A\), \(V \subseteq E\)).

 {\it Stage 2.} Forming a cluster from subgraph \(H\)
 and compression of initial graph \(G\):~
 \(G' = (A',E')\),
 (\(A' = A \backslash H\), \(E' = E \backslash \{ V \bigcup W \}\),
 where \(W\) is a set of external edges of clique,
 i.e., the only one vertex belongs to set \(H\))
 (Fig. 3.14).

 {\it Stage 3.} If \(G'\) is empty GO TO Stage 4
  otherwise GO TO Stage 1.

 {\it Stage 4.} Stop.


\begin{center}
\begin{picture}(61,35)

\put(00,00){\makebox(0,0)[bl] {Fig. 3.14.
 Illustration of clique in graph}}

\put(44,12){\makebox(0,8)[bl]{Graph }}
\put(41,08){\makebox(0,8)[bl]{\(G = (A,E)\)}}

\put(30.5,20){\oval(61,28)}


\put(09,21){\line(-1,0){04}} \put(09,21){\line(0,1){04}}
\put(09,21){\line(0,-1){04}} \put(09,21){\line(1,0){04}}

\put(09,11){\line(-1,0){04}} \put(09,11){\line(0,1){04}}
\put(09,11){\line(0,-1){04}} \put(09,11){\line(1,0){04}}

\put(29,21){\line(-1,0){04}} \put(29,21){\line(0,1){04}}
\put(29,21){\line(0,-1){04}} \put(29,21){\line(1,0){04}}

\put(29,11){\line(-1,0){04}} \put(29,11){\line(0,1){04}}
\put(29,11){\line(0,-1){04}} \put(29,11){\line(1,0){04}}

\put(09,21){\circle*{1}} \put(29,21){\circle*{1}}
\put(09,11){\circle*{1}} \put(29,11){\circle*{1}}

\put(06,29){\makebox(0,8)[bl]{External edges \(W\)}}
\put(14,29){\vector(-1,-1){04}} \put(24,29){\vector(1,-1){04}}

\put(19,16){\oval(24,12)}

\put(14,16){\makebox(0,8)[bl]{Clique}}
\put(10,12){\makebox(0,8)[bl]{\(H = (B,V)\)}}


\end{picture}
\end{center}

 The above-mentioned solving scheme is based on
 series of NP-hard problems.
 Evidently,
 it is possible to find several ``maximal cliques'' concurrently.
 Some sources on researches on clique finding and clique based clustering
 are presented in Table 3.19.

 Clique partitioning problem for a given graph \(G=(A,E)\) with
 edge weights consists in partitioning the graph into cliques such
 that the sum of the edge weights over all cliques formed is as
 large as possible
 (e.g., \cite{koch05,oost01}).

 There are some close problems over graphs/digraphs, for example,
 independent set problems and
 dominating set problems which are used in clustering as well
 (e.g., \cite{cheny05,cok06,cok07,corn84,han07,indu11,pav07}).
 Recently, the significance of dynamic problems over data streams
 has been increased
 including clique/quasi-clique finding in graph streams
 (e.g. \cite{aggar07,coble06,guha00a,lev12clique}).

 On the other hand,
 clique-based approaches can be considered as
 density-based and grid-based clustering methods.
 In some recent works,
 subgraph as clique/qiasi-clique
 is considered as one of  network community structures
 (network community based clustering
  \cite{girvan02,new06,new10,port09}).

\begin{center}
 {\bf Table 3.19.} Detection of cliques/quasi-cliques and clustering  \\
\begin{tabular}{| c |l | l| }
\hline
 No.& Research  & Source(s)  \\
\hline

 1.  &Detection and analysis of cliques in graphs:&\\
 1.1 & Cliques in graphs& \cite{johnson96,moon65} \\
 1.2.& Finding of clique/quasi-clique in graphs
 &\cite{abe02,alon98,but06,feig00,gar79,ost73,pardalos13}\\

 1.3.& Maximum-weight clique problem&\cite{babel94,balas87,bomze99,oster99,pard94}\\

 1.4.& Finding all cliques of undirected graph& \cite{bron73}\\
 1.5.& Enumeration maximal cliques of large graph & \cite{akko73}\\

 2.& Clustering based on cliques:&\\
 2.1. & Clique based clustering&
   \cite{agrawal05,berk06,duan12,gra05,jain99,mehrot98,mirkin99,shamir02,tsuda06}  \\

 2.2. & Clique partitioning problem &
   \cite{deamor92,koch05,oost01}  \\
   & (clique partition of maximum weight)& \\

 3. & Clique-based multiple clustering:&\\
 3.1. & Ensemble clustering with voting active clusters& \cite{tumer08}\\
 3.2. & Cliques for combining multiple clusterings& \cite{mim12} \\

 4.& Clique based methods over data streams:&\\
 4.1.& K-clique clustering in dynamic networks&\cite{duan12}\\
 4.2.& Clique-based fusion of graph streams&\cite{lev12clique}\\

\hline
\end{tabular}
\end{center}


 In recent decades,
 several new combinatorial problems as
 clique clustering in multipartite graphs
 have been suggested
 (e.g., \cite{char08,dawande01,har04,lev98,lev06,lev15,vas07}).
 Fig. 3.15. illustrates this kind of problems.
 Table 3.20 contains a list of the research
  directions in the above-mentioned field.

\begin{center}
\begin{picture}(64,54)
\put(01,00){\makebox(0,0)[bl]{Fig. 3.15. Cliques in four-partite
 graph}}

\put(07.5,44){\makebox(0,0)[bl]{Clique}}
\put(14,44){\vector(1,-1){11}}

\put(07.5,14){\makebox(0,0)[bl]{Clique}}
\put(12.7,17){\vector(1,1){04}}


\put(12,30){\oval(24,08)}

\put(04,30){\circle*{1.2}} \put(04,30){\circle{2.0}}

\put(08,29.5){\makebox(0,0)[bl]{{\bf ...}}}

\put(15,30){\circle*{1.2}} \put(15,30){\circle{2.0}}
\put(20,30){\circle*{1.2}} \put(20,30){\circle{2.0}}


\put(52,30){\oval(24,11)}

\put(44,30){\circle*{1.5}} \put(44,30){\circle{2.5}}
\put(49,30){\circle*{1.5}} \put(49,30){\circle{2.5}}

\put(53,29.5){\makebox(0,0)[bl]{{\bf ...}}}

\put(60,30){\circle*{1.5}} \put(60,30){\circle{2.5}}


\put(32,36){\line(0,-1){12}} \put(20,30){\line(1,0){24}}

\put(32,36){\line(-2,-1){12}} \put(32,36){\line(2,-1){12}}
\put(32,24){\line(-2,1){12}} \put(32,24){\line(2,1){12}}


\put(32,50){\line(-1,-1){12}} \put(20,38){\line(-2,-3){05}}

\put(32,50){\line(3,-2){28}}

\put(15,30){\line(1,-4){6}} \put(21,06){\line(1,0){33}}
\put(60,30){\line(-1,-4){6}}


\put(32,50){\line(1,-1){6}} \put(38,21){\line(0,1){23}}
\put(32,15){\line(1,1){6}}


\put(32,15){\line(-3,2){15}} \put(17,25){\line(-1,2){02.5}}
\put(32,15){\line(4,1){24}} \put(56,21){\line(1,2){04}}


\put(32,43){\oval(07,18)}

\put(32,36){\circle*{0.8}} \put(32,36){\circle{1.4}}
\put(32,41){\circle*{0.8}} \put(32,41){\circle{1.4}}

\put(30.5,45){\makebox(0,0)[bl]{{\bf ...}}}

\put(32,50){\circle*{0.8}} \put(32,50){\circle{1.4}}


\put(32,17){\oval(10,18)}

\put(32,10){\circle*{1.4}} \put(32,10){\circle{1.4}}
\put(32,15){\circle*{1.4}} \put(32,15){\circle{1.4}}

\put(30.5,19){\makebox(0,0)[bl]{{\bf ...}}}

\put(32,24){\circle*{1.4}} \put(32,24){\circle{1.4}}


\end{picture}
\end{center}

\begin{center}
 {\bf Table 3.20.} Research directions in multi-partite graphs \\
\begin{tabular}{| c |l | l| }
\hline
 No.& Research  & Source(s)  \\
\hline

 1.&Problem of compatible representatives&\cite{knu92} \\
 2.&Morphological clique (ordinal estimates)&\cite{lev98,lev06,lev15} \\
 3.&Morphological clique (multiset estimates)&\cite{lev12multiset,lev15} \\
 4.&Clustering in multipartite graph&\cite{char08,vas07} \\
 5.&Bipartite and multipartite clique problems &\cite{dawande01}\\

 6.&Morphological clique over graph streams &\cite{lev12clique} \\
 7.&Coreset problems &\cite{feld11,har04}\\
 8.&Coresets  in dynamic data streams&\cite{fra05}\\
\hline
\end{tabular}
\end{center}


\subsubsection{Correlation clustering}

  Correlation clustering
  provides a method for
   partitioning a fully connected labeled graph
  (label ``+'' corresponds to edge for similar vertices,
   label ``-'' corresponds to edge for different vertices)
  while taking into account  two objectives for the obtained clusters:

  (i) minimizing disagreements
  (i.e., minimizing the number of ``-'' edges within
    the clusters
 (\( Q^{disagr} (\widehat{X}) \rightarrow \min\))
    or  maximizing the number of ``-'' between clusters),

  (ii) maximizing agreements
  (i.e., the number of ``+'' edges insides the clusters)
 (\( Q^{agr} (\widehat{X}) \rightarrow \max \))
%
 (e.g.,
 \cite{achtert07,bagon11,bansal04,ben99,dem06,kri09,swam04,zimek08}).

 In the basic above-mentioned problem formulation,
 the objective functions are summarized.
 In other words,
  binary scale \([-1,+1]\) is used for each edge
 as a weight (zero value is not used).
 Here it is not necessary to specify the preliminary number of
 clusters (e.g., as in \(k\)-means clustering).
  The correlation clustering problem formulation is motivated from
   documents/web pages clustering.
  This combinatorial model belongs to NP-complete class
   (e.g., \cite{ailon08,bansal02,bansal04}).

 Various versions of correlation problem formulations are
 examined:

 (a) weighted versions of the
   ``correlation clustering functional''
 are considered as well
(e.g.,
 \cite{cgw03,cgw05,dem06}),

 (b) correlation clustering with partial information
 (e.g., \cite{demaine03}),

 (c) correlation clustering with noisy input
 (e.g., \cite{mat10}), etc.

 Let us consider the weighted version of the problem.
 Let \(A = \{A_{1},...,A_{j},...,A_{n}\}\)
 be the initial set of elements.
 As a result, \((n-1)^{2}\) elements pairs can be considered:
 \(G = \{g_{1},...,g_{(n-1)^{2}}\}\).
 Each element of \(G\) corresponds to
 element pair \((A_{j_{1}}, A_{j_{2}})\)
 and an element of
 proximity matrix \(Z =|| z_{j_{1},j_{2}}||\).
 Further, it is possible to replace  scale  \([-1,+1]\)
 for each edge (i.e., for each element from \(G\) or element of proximity matrix \(Z\))
  by two quantitative scales for weights:
 negative quantitative (or ordinal) scale
  \([-w{-},...,0)\) instead of ``\(-1\)''
 and
 positive quantitative scale
 \((0,...,w^{+}]\) instead of ``\(+1\)''.
 Evidently,
 element pair set
 is divided into two separated subsets
 \(G = G^{-} \bigcup G^{+}\)
 (without intersection, i.e.,
 \(|G^{-} \bigcap G^{+}| = 0\))
 where \(\forall g^{-} \in G^{-}\)
 weight estimate corresponds to negative quantitative scale above,
 where \(\forall g^{+} \in G^{+}\)
 weight estimate corresponds to negative quantitative scale above.
%
 The clustering solution is:~
 \(\widehat{X} = \{X_{1},...,X_{\iota},...,X_{\lambda}\}\).
 For this solution,
 two total quality parameters above are examined:

 (i) total agreements  quality as
 (summarization by all intra-cluster pairs with positive edge weight)
  \(Q^{agr} (\widehat{X})\)
  (maximization);

 (ii) total disagreements quality
 (summarization by all intra-cluster pairs with negative edge weight)
  \(Q^{disagr} (\widehat{X})\)
  (for minimization, by module).

 As a result, the weighted version of  correlation clustering problem is
 (Fig. 3.16):

~~

 Find clustering solution \(\widehat{X}\)
 such that:~
 (i) \( Q^{agr} (\widehat{X}) \rightarrow \max \)
 and
 (ii) \( |Q^{disagr} (\widehat{X})| \rightarrow \min\).



\begin{center}
\begin{picture}(80,45.5)
\put(12,00){\makebox(0,0)[bl]{Fig. 3.16. ``Space''
 of solution quality}}

\put(01,41){\makebox(0,0)[bl]{\((0,0)\)}}
\put(05,40){\circle*{1.3}}

\put(05,40){\vector(0,-1){30}} \put(05,40){\vector(1,0){60}}

\put(66,39){\makebox(0,0)[bl]{\(Q^{agr}(\widehat{X})\)}}
\put(00,05){\makebox(0,0)[bl]{\(Q^{disagr}(\widehat{X})\)}}

\put(60,40){\circle*{1.2}} \put(60,40){\circle{2.0}}
\put(68,35){\makebox(0,0)[bl]{Ideal}}
\put(68,32){\makebox(0,0)[bl]{point}}

\put(65,35){\line(-1,1){04}}

\put(05,12){\line(1,0){04}} \put(10,12){\line(1,0){04}}
\put(15,12){\line(1,0){04}} \put(20,12){\line(1,0){04}}
\put(25,12){\line(1,0){04}} \put(30,12){\line(1,0){04}}
\put(35,12){\line(1,0){04}} \put(40,12){\line(1,0){04}}
\put(45,12){\line(1,0){04}} \put(50,12){\line(1,0){04}}
\put(55,12){\line(1,0){04}}

\put(60,40){\line(0,-1){04}} \put(60,35){\line(0,-1){04}}
\put(60,30){\line(0,-1){05}} \put(60,24){\line(0,-1){05}}
\put(60,18){\line(0,-1){05}}


\put(15,37){\vector(2,-1){04}}

\put(20,34.7){\makebox(0,0)[bl]{{\bf ...}}}

\put(25,35){\vector(1,0){4}} \put(30,35){\circle*{1}}
\put(30,35){\vector(2,-1){4}} \put(35,32.5){\circle*{1}}
\put(35,32.5){\vector(1,0){4}} \put(40,32.5){\circle*{1}}
\put(41,32){\vector(2,-1){6}}

\put(48,29){\makebox(0,0)[bl]{{\bf ...}}}


\put(04,27){\line(1,0){05}} \put(11,27){\line(1,0){04}}
\put(17,27){\line(1,0){04}} \put(23,27){\line(1,0){04}}
\put(29,27){\line(1,0){04}} \put(35,27){\line(1,0){04}}
\put(41,27){\line(1,0){04}} \put(47,27){\line(1,0){04}}
\put(53,27){\line(1,0){04}} \put(59,27){\line(1,0){04}}

\put(62,24){\makebox(0,0)[bl]{Constraint}}
\put(62,19.5){\makebox(0,0)[bl]
 {\(q \geq | Q^{disagr}(\widehat{X})| \)}}

\put(28,24){\line(0,1){10}}

\put(09,21){\makebox(0,0)[bl]{Trajectory of vector objective}}
\put(06.8,18){\makebox(0,0)[bl]{function
 (by algorithm iterations)}}
\put(13,14){\makebox(0,0)[bl]{\((Q^{disagr}(\widehat{X}),Q^{agr}(\widehat{X}))\)}}

\end{picture}
\end{center}

%
  Heuristics and approximation algorithms (e.g., PTAS)
  have been proposed for the problem versions
  (e.g., \cite{bagon11,bansal02,bansal04,zimek08}).
 Clearly, agglomerative (hierarchical) clustering scheme
 (i.e., selection of an element pair from set \(B\) for next joining
 for improvement of a current clustering solution)
 can be used here as a simple greedy algorithm
 (Bottom-Up process of selection of element pair
  with the best improvement of objective vector function
  and corresponding joining the elements),
  for example (Fig. 3.16):

~~

 {\it Stage 1.}
  Calculation of the matrix of element pair
 \(\forall (A(j_{1}),A(j_{2}))\), ~\(A(j_{1})\in A\), ~\(A(j_{2})\in A \),
 ~\(j_{1} \neq j_{2}\)
 proximities (``distances'').

  {\it Stage 2.} Transformation of element pair proximities
 into positive
 (for similar elements)
 or negative
 (for dissimilar elements)
 weights (e.g., mapping).

 {\it Stage 3.} Specifying the initial clustering solution
 \(\widehat{X}^{0}\) as composition of initial elements,
 vector objective function
 \(\overline{f}^{0} = (Q^{disagr}(\widehat{X}^{0}),Q^{agr}(\widehat{X}^{0})) = (0,0)\)
 (initial value, initial index \(\gamma =0\)).

  {\it Stage 4.} Searching for the element pair
 with the best improvement of
 vector objective function \(\overline{f}\)
 (i.e., searching for Pareto-efficient point(s)).
 Integration  of the corresponding both elements into a cluster
 or inclusion of the corresponding element into
 the cluster with the second element
 (i.e., new clustering solution)
  \(\widehat{X}^{q}\)
  (\(q\) is parameter of algorithm iteration).
 Recalculation of the current value of vector objective function:
 \(\overline{f}^{\gamma} = (Q^{disagr}(\widehat{X}^{\gamma}),Q^{agr}(\widehat{X}^{\gamma})) \).

 {\it Stage 5.}
 If all elements are processed
 then GO TO Stage 7.

 {\it Stage 6.} Increasing index \(\gamma=\gamma+1\),
 while constraint  ~
  \(| Q^{disagr}(\widehat{X})|  \leq q\) ~
  is satisfied
  {\it Go To}   Stage 4, else GO TO Stage 7.

 {\it Stage 7.} Stop.


~~

  Complexity estimates of greedy heuristic above
  for two-objectives correlation clustering   (by stages)
   are presented in Table 3.21.


\begin{center}
{\bf Table 3.21.} Complexity estimates of stages for
 greedy agglomerative heuristic
 \\
\begin{tabular}{| l | l|c |}
\hline
  Stage &Description &Complexity estimate   \\
  & &(running time)  \\
\hline
 Stage 1.& Calculation of distance matrix \(Z\)&\(O (n^{2})\) \\

 Stage 2.& Calculation of positive/negative weights &\(O (n^{2})\) \\

 Stage 3.& Specifying the initial solution &\(O (1)\) \\

 Stage 4.& Searching for the best element pair &\(O (n^{2})\) \\

  &(by Pareto-efficient improvement of objective function)& \\


 Stage 5.& Analysis of algorithm end,
   recalculation of objective function&\(O(n)\)\\


 Stage 6.& Transition of computing process & \(O(1)\) \\

 Stage 7.& Stopping        & \(O(1)\)\\
\hline
\end{tabular}
\end{center}

~~

 {\bf Example 3.4.} The examined element set involves
 \(11\) elements: \(A = \{ A_{1},...,A_{j},...,A_{11}\}\).
 The weights of all edges are presented in Table 3.22.
 Here, two quantitative scales are used:
 \([-6.5,0)\) and \((0,3.5]\).

\begin{center}
{\bf Table 3.22.} Weights of edges  \(\{(A_{j_{1}},A_{j_{2}})\}\) \\
\begin{tabular}{| c |c c c c c c c c c c c  |}
\hline
 \(j_{1}\) & \(j_{2}:\)& \(2\)&\(3\)&\(4\)&\(5\)&\(6\) &\(7\)&\(8\)&\(9\)&\(10\)&\(11\) \\
\hline
 1 &&\(-3.0\)&\(1.4\)&\(-0.6\)&\(-5.1\)&\(-5.5\)&\(-3.5\)&\(-1.2\)&\(-3.4\)&\(-4.5\)&\(-6.5\)\\
 2 &&&\(-0.8\)&\(-1.3\)&\(-3.3\)&\(-1.4\)&\(3.1\)&\(-2.9\)&\(-3.6\)&\(-5.1\)&\(-4.9\)\\
 3 &&&&\(-1.1\)&\(-2.0\)&\(-2.7\)&\(-2.1\)&\(3.2\)&\(-3.0\)&\(-4.5\)&\(-4.1\)\\
 4 &&&&&\(-0.5\)&\(-3.7\)&\(-2.5\)&\(2.6\)&\(-0.9\)&\(-1.3\)&\(-2.2\)\\

 5 &&&&&&\(-6.5\)&\(-5.6\)&\(-1.1\)&\(2.8\)&\(-0.5\)&\(-6.1\)\\
 6 &&&&&&&\(3.5\)&\(0.4\)&\(-1.8\)&\(-3.2\)&\(-0.3\)\\
 7 &&&&&&&&\(0.5\)&\(-0.3\)&\(-0.8\)&\(2.9\)\\
 8 &&&&&&&&&\(1.0\) &\(-0.8\)& \(-2.8\)\\
 9 &&&&&&&&&&\(3.0\)&\(-5.5\)\\
 10 &&&&&&&&&&&\(-6.0\)\\

\hline
\end{tabular}
\end{center}

 Initial information is the following
 (iteration index equals \(\gamma=0\)):
 (a)
 \(\widehat{X}^{0} = \{X^{0}_{1},...,X^{0}_{\iota},...,X^{0}_{11}\}\) where
  \(X^{0}_{1} = \{A_{1}\}\),
 \(X^{0}_{2} = \{A_{2}\}\),
 \(X^{0}_{3} = \{A_{3}\}\),
 \(X^{0}_{4} = \{A_{4}\}\),
 \(X^{0}_{5} = \{A_{5}\}\),
 \(X^{0}_{6} = \{A_{6}\}\),
 \(X^{0}_{7} = \{A_{7}\}\),
 \(X^{0}_{8} = \{A_{8}\}\),
 \(X^{0}_{9} = \{A_{9}\}\),
 \(X^{0}_{10} = \{A_{10}\}\),
 \(X^{0}_{11} = \{A_{11}\}\);
 (b) \(\overline{f}^{0} (\widehat{X}^{0}) = (0,0)\);
 (c) improvement operations
 (i.e., inclusion of element (cluster) \(A_{j}\) into cluster \(X_{\iota}\))
 as
 \(O_{j,\iota} (A_{j} \rightarrow X_{\iota})\)
 (\(j = \overline{1,10}\), \(\iota=\overline{j,11} \))
 and corresponding improvements
 (by positive component or by negative component)
  of objective function \(\overline{f}\)
 as
  \(\Delta \overline{f} (O_{j,\iota})\)
  are presented in Table 3.23
  (component \(0\) of the vector is not pointed out).

\newpage
\begin{center}

{\bf Table 3.23.} Improvements of objective function \(\Delta
\overline{f} (O_{j,\iota}  )\)
 (iteration index \(\gamma = 0\))
  \\
\begin{tabular}{| c |c c c c c c c c c c c  |}
\hline
 \(A_{j}\)&\(X_{\iota}:\)&\(X_{2}\)&\(X_{3}\)&\(X_{4}\)&\(X_{5}\)&\(X_{6}\)&\(X_{7}\)&\(X_{8}\)&\(X_{9}\)&\(X_{10}\)&\(X_{11}\) \\
\hline
 \(A_{1}\) &&\(-3.0\)&\(1.4\)&\(-0.6\)&\(-5.1\)&\(-5.5\)&\(-3.5\)&\(-1.2\)&\(-3.4\)&\(-4.5\)&\(-6.5\)\\
 \(A_{2}\) &&&\(-0.8\)&\(-1.3\)&\(-3.3\)&\(-1.4\)&\(3.1\)&\(-2.9\)&\(-3.6\)&\(-5.1\)&\(-4.9\)\\
 \(A_{3}\) &&&&\(-1.1\)&\(-2.0\)&\(-2.7\)&\(-2.1\)&\(3.2\)&\(-3.0\)&\(-4.5\)&\(-4.1\)\\
 \(A_{4}\) &&&&&\(-0.5\)&\(-3.7\)&\(-2.5\)&\(2.6\)&\(-0.9\)&\(-1.3\)&\(-2.2\)\\

 \(A_{5}\) &&&&&&\(-6.5\)&\(-5.6\)&\(-1.1\)&\(2.8\)&\(-0.5\)&\(-6.1\)\\
 \(A_{6}\) &&&&&&&\(3.5\)&\(0.4\)&\(-1.8\)&\(-3.2\)&\(-0.3\)\\
 \(A_{7}\) &&&&&&&&\(0.5\)&\(-0.3\)&\(-0.8\)&\(2.9\)\\
 \(A_{8}\) &&&&&&&&&\(1.0\) &\(-0.8\)& \(-2.8\)\\
 \(A_{9}\) &&&&&&&&&&\(3.0\)&\(-5.5\)\\
 \(A_{10}\) &&&&&&&&&&&\(-6.0\)\\

\hline
\end{tabular}
\end{center}

 {\bf Iteration 1}. Selection of the best (Pareto-efficient) improvement operation
  \(O_{6,7}\)
  with the best  improvement
 \(\Delta \overline{f} (O_{j,\iota}  ) =(0,3.5)\).
 As a result,
 the following information is used for the next algorithm step:
(a)
 \(\widehat{X}^{1} = \{
  X^{1}_{1},X^{1}_{2},X^{1}_{3},X^{1}_{4}, X^{1}_{5},X^{1}_{7},X^{1}_{8},X^{1}_{9},
  X^{1}_{10},X^{1}_{11}\}\) where
  \(X^{1}_{1} = \{A_{1}\}\),
 \(X^{1}_{2} = \{A_{2}\}\),
 \(X^{1}_{3} = \{A_{3}\}\),
 \(X^{1}_{4} = \{A_{4}\}\),
 \(X^{1}_{5} = \{A_{5}\}\),
 \(X^{1}_{7} = \{A_{6},A_{7}\}\),
 \(X^{1}_{8} = \{A_{8}\}\),
 \(X^{1}_{9} = \{A_{9}\}\),
 \(X^{1}_{10} = \{A_{10}\}\),
 \(X^{1}_{11} = \{A_{11}\}\);
 (b) \(\overline{f}^{1} (\widehat{X}^{1}) = (0,3.5)\);
 (c) improvement operations
 (i.e., inclusion of element/cluster \(X_{\iota_{1}}\) into cluster \(X_{\iota_{2}}\))
 as
 \(O_{\iota_{1},\iota_{2}} (X_{\iota_{1}} \rightarrow X_{\iota_{2}})\)
%
%
 and corresponding improvements
 (by positive component or by negative component)
  of objective function \(\overline{f}\)
 as
  \(\Delta \overline{f} (O_{\iota_{1},\iota_{2}})\)
  are presented in Table 3.24
 (component \(0\) of the vector  is not pointed out).

\begin{center}
{\bf Table 3.24.} Improvements of objective function
 \(\Delta \overline{f} (O_{j,\iota}  )\)
 (iteration index \(\gamma= 1\))
  \\
\begin{tabular}{| l |c c c c c c c c c c |}
\hline
 \(X_{\iota_{1}}\)&\(X_{\iota_{2}}:\)&\(X^{1}_{2}\)&\(X^{1}_{3}\)&\(X^{1}_{4}\)&\(X^{1}_{5}\)&\(X^{1}_{7}\)
 &\(X^{1}_{8}\)&\(X^{1}_{9}\)&\(X^{1}_{10}\)&\(X^{1}_{11}\) \\
\hline
 \(X^{1}_{1}\) &&\(-3.0\)&\(1.4\)&\(-0.6\)&\(-5.1\)&\(-9.0\)&\(-1.2\)&\(-3.4\)&\(-4.5\)&\(-6.5\)\\
 \(X^{1}_{2}\) &&&\(-0.8\)&\(-1.3\)&\(-3.3\)&\((-1.4,3.1)\)&\(-2.9\)&\(-3.6\)&\(-5.1\)&\(-4.9\)\\
 \(X^{1}_{3}\) &&&&\(-1.1\)&\(-2.0\)&\(-4.8\)&\(3.2\)&\(-3.0\)&\(-4.5\)&\(-4.1\)\\
 \(X^{1}_{4}\) &&&&&\(-0.5\)&\(-6.2\)&\(2.6\)&\(-0.9\)&\(-1.3\)&\(-2.2\)\\

 \(X^{1}_{5}\) &&&&&&\(-12.2\)&\(-1.1\)&\(2.8\)&\(-0.5\)&\(-6.1\)\\
 \(X^{1}_{7} =\{A_{6},A_{7}\}\) &&&&&&&\(0.9\)&\(-2.1\)&\(-4.0\)&\((-0.3,2.9)\)\\
 \(X^{1}_{8}\) &&&&&&&&\(1.0\) &\(-0.8\)& \(-2.8\)\\
 \(X^{1}_{9}\) &&&&&&&&&\(3.0\)&\(-5.5\)\\
 \(X^{1}_{10}\) &&&&&&&&&&\(-6.0\)\\

\hline
\end{tabular}
\end{center}

 {\bf Iteration 2}. Selection of the best
 (Pareto-efficient)
 improvement operation
  \(O_{3,8}\)
  with the best  improvement
 \(\Delta \overline{f} (O_{j,\iota}  ) =(0,3.2)\).
 As a result,
 the following information is used for the next algorithm step:
(a)
 \(\widehat{X}^{2} = \{
  X^{2}_{1},X^{2}_{2},X^{2}_{4}, X^{2}_{5},X^{2}_{7},X^{2}_{8},X^{2}_{9},
  X^{2}_{10},X^{2}_{11}\}\) where
  \(X^{2}_{1} = \{A_{1}\}\),
 \(X^{2}_{2} = \{A_{2}\}\),
 \(X^{2}_{4} = \{A_{4}\}\),
 \(X^{2}_{5} = \{A_{5}\}\),
 \(X^{2}_{7} = \{A_{6},A_{7}\}\),
 \(X^{2}_{8} = \{A_{3},A_{8}\}\),
 \(X^{2}_{9} = \{A_{9}\}\),
 \(X^{2}_{10} = \{A_{10}\}\),
 \(X^{2}_{11} = \{A_{11}\}\);
 (b) \(\overline{f}^{2} (\widehat{X}^{2}) = (0,9.9)\);
 (c) improvement operations
 (i.e., inclusion of element/cluster \(X_{\iota_{1}}\) into cluster \(X_{\iota_{2}}\))
 as
 \(O_{\iota_{1},\iota_{2}} (X_{\iota_{1}} \rightarrow X_{\iota_{2}})\)
%
%
 and corresponding improvements
 (by positive component or by negative component)
  of objective function \(\overline{f}\)
 as
  \(\Delta \overline{f} (O_{\iota_{1},\iota_{2}})\)
  are presented in Table 3.25
 (component \(0\) of the vector  is not pointed out).

\begin{center}
{\bf Table 3.25.} Improvements of objective function
 \(\Delta \overline{f} (O_{j,\iota}  )\)
 (iteration index \(\gamma = 2\))
  \\
\begin{tabular}{| l |c c c c c c c c c   |}
\hline
 \(X_{\iota_{1}}\)&\(X_{\iota_{2}}:\)&\(X^{2}_{2}\)&\(X^{2}_{4}\)&\(X^{2}_{5}\)
 &\(X^{2}_{7}\)&\(X^{2}_{8}\)&\(X^{2}_{9}\)&\(X^{2}_{10}\)&\(X^{2}_{11}\) \\
\hline
 \(X^{2}_{1}\) &&\(-3.0\)&\(-0.6\)&\(-5.1\)&\(-9.0\)&\((-1.2,1.4)\)&\(-3.4\)&\(-4.5\)&\(-6.5\)\\
 \(X^{2}_{2}\) &&&\(-1.3\)&\(-3.3\)&\((-1.4,3.1) \)&\(-3.7\)&\(-3.6\)&\(-5.1\)&\(-4.9\)\\
 \(X^{2}_{4}\) &&&&\(-0.5\)&\(-6.2\)&\((-1.1,2.6)\)&\(-0.9\)&\(-1.3\)&\(-2.2\)\\

 \(X^{2}_{5}\) &&&&&\(-12.2\)&\(-3.1\)&\(2.8\)&\(-0.5\)&\(-6.1\)\\
 \(X^{2}_{7}= \{A_{6},A_{7}\}\) &&&&&&\((-4.8,0.9)\)&\(-2.1\)&\(-4.0\)&\((-0.3,2.9)\)\\
 \(X^{2}_{8}= \{A_{3},A_{8}\}\) &&&&&&&\(1.0\) &\(-0.8\)& \(-2.8\)\\
 \(X^{2}_{9}\) &&&&&&&&\(3.0\)&\(-5.5\)\\
 \(X^{2}_{10}\) &&&&&&&&&\(-6.0\)\\

\hline
\end{tabular}
\end{center}

 {\bf Iteration 3}. Selection of the best
 (Pareto-efficient)
 improvement operations:
  \(O_{2,7}\) and \(O_{9,10}\).
  The corresponding Pareto-efficient improvements are:
 \(\Delta \overline{f} (O_{2,7}  ) =(-1.4,3.1)\) and
 \(\Delta \overline{f} (O_{9,10}  ) =(0,3.0)\).
 Operation \(O_{9,10}\) is selected.
 As a result,
 the following information is used for the next algorithm step:
(a)
 \(\widehat{X}^{3} = \{
  X^{3}_{1},X^{3}_{2},X^{3}_{4}, X^{3}_{5},X^{3}_{7},X^{3}_{8},
  X^{3}_{10},X^{3}_{11}\}\) where
  \(X^{3}_{1} = \{A_{1}\}\),
 \(X^{3}_{2} = \{A_{2}\}\),
 \(X^{3}_{4} = \{A_{4}\}\),
 \(X^{3}_{5} = \{A_{5}\}\),
 \(X^{3}_{7} = \{A_{6},A_{7}\}\),
 \(X^{3}_{8} = \{A_{3},A_{8}\}\),
 \(X^{3}_{10} = \{A_{9},A_{10}\}\),
 \(X^{3}_{11} = \{A_{11}\}\);
 (b) \(\overline{f}^{3} (\widehat{X}^{3}) = (0,12.9)\);
 (c) improvement operations
 (i.e., inclusion of element/cluster \(X_{\iota_{1}}\) into cluster \(X_{\iota_{2}}\))
 as
 \(O_{\iota_{1},\iota_{2}} (X_{\iota_{1}} \rightarrow X_{\iota_{2}})\)
%
%
 and corresponding improvements
 (by positive component or by negative component)
  of objective function \(\overline{f}\)
 as
  \(\Delta \overline{f} (O_{\iota_{1},\iota_{2}})\)
  are presented in Table 3.26
 (component \(0\) of the vector  is not pointed out).

\begin{center}
{\bf Table 3.26.} Improvements of objective function
 \(\Delta \overline{f} (O_{j,\iota}  )\)
 (iteration index \(\gamma= 3\))
  \\
\begin{tabular}{| l |c c c c c c c c  |}
\hline
 \(X_{\iota_{1}}\)&\(X_{\iota_{2}}:\)&\(X^{3}_{2}\)&\(X^{3}_{4}\)&\(X^{3}_{5}\)&\(X^{3}_{7}\)&\(X^{3}_{8}\)&\(X^{3}_{10}\)&\(X^{3}_{11}\) \\
\hline
 \(X^{3}_{1}\) &&\(-3.0\)&\(-0.6\)&\(-5.1\)&\(-9.0\)&\((-1.2,1.4)\)&\(-7.9\)&\(-6.5\)\\
 \(X^{3}_{2}\) &&&\(-1.3\)&\(-3.3\)&\((-1.4,3.1)\)&\(-3.7\)&\(-8.7\)&\(-4.9\)\\
 \(X^{3}_{4}\) &&&&\(-0.5\)&\(-6.2\)&\((-1.1,2.6)\)&\(-2.2\)&\(-2.2\)\\

 \(X^{3}_{5}\) &&&&&\(-12.2\)&\(-3.1\)&\((-0.5,2.8)\)&\(-6.1\)\\
 \(X^{3}_{7} = \{A_{6},A_{7}\}\) &&&&&&\((-4.8,0.9)\)&\(-6.1\)&\((-0.3,2.9)\)\\
 \(X^{3}_{8}= \{A_{3},A_{8}\}\) &&&&&&&\((-8.3,1.0)\)& \(-2.8\)\\
 \(X^{3}_{10}= \{A_{9},A_{10}\}\) &&&&&&&&\(-8.8\)\\

\hline
\end{tabular}
\end{center}

 {\bf Iteration 4}. Selection of the best
 (Pareto-efficient)
 improvement operations:
  \(O_{2,7}\) and \(O_{7,11}\).
  The corresponding Pareto-efficient improvements are:
 \(\Delta \overline{f} (O_{2,7}  ) =(-1.4,3.1)\) and
 \(\Delta \overline{f} (O_{7,11}  ) =(-0.3,2.9)\).
 Operation \(O_{7,11}\) is selected.
 As a result,
 the following information is used for the next algorithm step:
(a)
 \(\widehat{X}^{4} = \{
  X^{4}_{1},X^{4}_{2},X^{4}_{4},X^{4}_{5},X^{4}_{8},
  X^{4}_{10},X^{4}_{11}\}\) where
 \(X^{4}_{1} = \{A_{1}\}\),
 \(X^{4}_{2} = \{A_{2}\}\),
 \(X^{4}_{4} = \{A_{4}\}\),
 \(X^{4}_{5} = \{A_{5}\}\),
 \(X^{4}_{8} = \{A_{3},A_{8}\}\),
 \(X^{4}_{10} = \{A_{9},A_{10}\}\),
 \(X^{4}_{11} = \{A_{6},A_{7},A_{11}\}\);
 (b) \(\overline{f}^{4} (\widehat{X}^{4}) = (-0.3,15.8)\);
 (c) improvement operations
 (i.e., inclusion of element/cluster \(X_{\iota_{1}}\) into cluster \(X_{\iota_{2}}\))
 as
 \(O_{\iota_{1},\iota_{2}} (X_{\iota_{1}} \rightarrow X_{\iota_{2}})\)
%
%
 and corresponding improvements
 (by positive component or by negative component)
  of objective function \(\overline{f}\)
 as
  \(\Delta \overline{f} (O_{\iota_{1},\iota_{2}})\)
  are presented in Table 3.27
 (component \(0\) of the vector  is not pointed out).

\begin{center}
 {\bf Table 3.27.} Improvements of objective function
 \(\Delta \overline{f} (O_{j,\iota}  )\)
 (iteration index \(\gamma = 4\))
  \\
\begin{tabular}{| l |c c c c c c c  |}
\hline
 \(X_{\iota_{1}}\)&\(X_{\iota_{2}}:\)&\(X^{4}_{2}\)&\(X^{4}_{4}\)&\(X^{4}_{5}\)&\(X^{4}_{8}\)&\(X^{4}_{10}\)
 &\(X^{4}_{11}= \{A_{6},A_{7},A_{11}\}\) \\
\hline
 \(X^{4}_{1}\) &&\(-3.0\)&\(-0.6\)&\(-5.1\)&\((-1.2,1.4)\)&\(-7.9\)&\(-15.5\)\\
 \(X^{4}_{2}\) &&&\(-1.3\)&\(-3.3\)&\(-3.7\)&\(-8.7\)&\((-6.3,3.1)\)\\
 \(X^{4}_{4}\) &&&&\(-0.5\)&\((-1.1,2.6)\)&\(-2.2\)&\(-8.4\)\\

 \(X^{4}_{5}\) &&&&&\(-3.1\)&\((-0.5,2.8)\)&\(-18.2\)\\
 \(X^{4}_{8} = \{A_{3},A_{8}\}\) &&&&&&\((-8.3,1.0)\)& \((-14.8,0.9)\)\\
 \(X^{4}_{10}= \{A_{9},A_{10}\}\) &&&&&&&\(-17.6\)\\

\hline
\end{tabular}
\end{center}

 {\bf Iteration 5}. Selection of the best
 (Pareto-efficient)
 improvement operations:
  \(O_{2,11}\) and \(O_{5,10}\).
  The corresponding Pareto-efficient improvements are:
 \(\Delta \overline{f} (O_{2,11}  ) =(-6.3,3.1)\) and
 \(\Delta \overline{f} (O_{5,10}  ) =(-0.5,2.8)\).
 Operation \(O_{5,10}\) is selected.
 As a result,
 the following information is used for the next algorithm step:
(a)
 \(\widehat{X}^{5} =
 \{X^{5}_{1},X^{5}_{2},X^{5}_{4},X^{5}_{8}, X^{5}_{10},X^{5}_{11}\}\)
   where
 \(X^{5}_{1} = \{A_{1}\}\),
 \(X^{5}_{2} = \{A_{2}\}\),
 \(X^{5}_{4} = \{A_{4}\}\),
 \(X^{5}_{8} = \{A_{3},A_{8}\}\),
 \(X^{5}_{10} = \{A_{5},A_{9},A_{10}\}\),
 \(X^{5}_{11} = \{A_{6},A_{7},A_{11}\}\);
 (b) \(\overline{f}^{5} (\widehat{X}^{5}) = (-0.8,18.6)\);
 (c) improvement operations
 (i.e., inclusion of element/cluster \(X_{\iota_{1}}\) into cluster \(X_{\iota_{2}}\))
 as
 \(O_{\iota_{1},\iota_{2}} (X_{\iota_{1}} \rightarrow X_{\iota_{2}})\)
%
%
 and corresponding improvements
 (by positive component or by negative component)
  of objective function \(\overline{f}\)
 as
  \(\Delta \overline{f} (O_{\iota_{1},\iota_{2}})\)
  are presented in Table 3.28
 (component \(0\) of the vector  is not pointed out).

\begin{center}
 {\bf Table 3.28.} Improvements of objective function
 \(\Delta \overline{f} (O_{j,\iota}  )\)
 (iteration index \(\gamma = 5\))
  \\
\begin{tabular}{| l |c c c c c c   |}
\hline
 \(X_{\iota_{1}}\)&\(X_{\iota_{2}}:\)&\(X^{5}_{2}\)&\(X^{5}_{4}\)&\(X^{5}_{8}\)&\(X^{5}_{10}\)
 &\(X^{5}_{11}= \{A_{6},A_{7},A_{11}\}\) \\
\hline
 \(X^{5}_{1}\) &&\(-3.0\)&\(-0.6\)&\((-1.2,1.4)\)&\(-13.0\)&\(-15.5\)\\
 \(X^{5}_{2}\) &&&\(-1.3\)&\(-3.7\)&\(-12.0\)&\((-6.3,3.1)\)\\
 \(X^{5}_{4}\) &&&&\((-1.1,2.6)\)&\(-2.7\)&\(-8.4\)\\

 \(X^{5}_{8}= \{A_{3},A_{8}\}\) &&&&&\((-11.4,1.0)\)& \((-14.8,0.9)\)\\
 \(X^{5}_{10}= \{A_{5},A_{9},A_{10}\}\) &&&&&&\(-35.8\)\\

\hline
\end{tabular}
\end{center}


 {\bf Iteration 6}. Selection of the best
 (Pareto-efficient)
 improvement operations:
  \(O_{2,11}\) and \(O_{4,8}\).
  The corresponding Pareto-efficient improvements are:
 \(\Delta \overline{f} (O_{2,11}  ) =(-6.3,3.1)\) and
 \(\Delta \overline{f} (O_{4,8}  ) =(-1.1,2.6)\).
 Operation \(O_{4,8}\) is selected.
 As a result,
 the following information is used for the next algorithm step:
(a)
 \(\widehat{X}^{6} =
 \{X^{6}_{1},X^{6}_{2},X^{6}_{8}, X^{6}_{10},X^{6}_{11}\}\)
   where
 \(X^{6}_{1} = \{A_{1}\}\),
 \(X^{6}_{2} = \{A_{2}\}\),
 \(X^{6}_{8} = \{A_{3},A_{4},A_{8}\}\),
 \(X^{6}_{10} = \{A_{5},A_{9},A_{10}\}\),
 \(X^{6}_{11} = \{A_{6},A_{7},A_{11}\}\);
 (b) \(\overline{f}^{6} (\widehat{X}^{6}) = (-1.9,21.2)\);
 (c) improvement operations
 (i.e., inclusion of element/cluster \(X_{\iota_{1}}\) into cluster \(X_{\iota_{2}}\))
 as
 \(O_{\iota_{1},\iota_{2}} (X_{\iota_{1}} \rightarrow X_{\iota_{2}})\)
%
%
 and corresponding improvements
 (by positive component or by negative component)
  of objective function \(\overline{f}\)
 as
  \(\Delta \overline{f} (O_{\iota_{1},\iota_{2}})\)
  are presented in Table 3.29
 (component \(0\) of the vector  is not pointed out).

\begin{center}
 {\bf Table 3.29.} Improvements of objective function
 \(\Delta \overline{f} (O_{j,\iota}  )\)
 (iteration index \(\gamma = 6\))
  \\
\begin{tabular}{| l |c c c c c |}
\hline
 \(X_{\iota_{1}}\)&\(X_{\iota_{2}}:\)&\(X^{6}_{2}\)&\(X^{6}_{8}\)&\(X^{6}_{10}\)
 &\(X^{6}_{11}= \{A_{6},A_{7},A_{11}\}\) \\
\hline
 \(X^{6}_{1}\) &&\(-3.0\)&\((-1.8,1.4)\)&\(-13.0\)&\(-15.5\)\\
 \(X^{6}_{2}\) &&&\(-5.7\)&\(-12.0\)&\((-6.3,3.1)\)\\
 \(X^{6}_{8}= \{A_{3},A_{4},A_{8}\}\) &&&&\((-14.1,1.0)\)& \((-20.1,0.9)\)\\
 \(X^{6}_{10}= \{A_{5},A_{9},A_{10}\}\) &&&&&\(-35.8\)\\

\hline
\end{tabular}
\end{center}

 {\bf Iteration 7}. Selection of the best
 (Pareto-efficient)
 improvement operations:
  \(O_{2,11}\) and \(O_{1,8}\).
  The corresponding Pareto-efficient improvements are:
 \(\Delta \overline{f} (O_{2,11}  ) =(-6.3,3.1)\) and
 \(\Delta \overline{f} (O_{1,8}  ) =(-1.8,1.4)\).
 Operation \(O_{1,8}\) is selected.
 As a result,
 the following information is used for the next algorithm step:
(a)
 \(\widehat{X}^{7} =
 \{X^{7}_{2},X^{7}_{8}, X^{7}_{10},X^{7}_{11}\}\)
   where
%
 \(X^{7}_{2} = \{A_{2}\}\),
 \(X^{7}_{8} = \{A_{1},A_{3},A_{4},A_{8}\}\),
 \(X^{7}_{10} = \{A_{5},A_{9},A_{10}\}\),
 \(X^{7}_{11} = \{A_{6},A_{7},A_{11}\}\);
 (b) \(\overline{f}^{7} (\widehat{X}^{7}) = (-3.7,22.6)\);
 (c) improvement operations
 (i.e., inclusion of element/cluster \(X_{\iota_{1}}\) into cluster \(X_{\iota_{2}}\))
 as
 \(O_{\iota_{1},\iota_{2}} (X_{\iota_{1}} \rightarrow X_{\iota_{2}})\)
%
%
 and corresponding improvements
 (by positive component or by negative component)
  of objective function \(\overline{f}\)
 as
  \(\Delta \overline{f} (O_{\iota_{1},\iota_{2}})\)
  are presented in Table 3.30
 (component \(0\) of the vector  is not pointed out).

\begin{center}
 {\bf Table 3.30.} Improvements of objective function
 \(\Delta \overline{f} (O_{j,\iota}  )\)
 (iteration index \(\gamma = 7\))
  \\
\begin{tabular}{| l |c c c c |}
\hline
 \(X_{\iota_{1}}\)&\(X_{\iota_{2}}:\)&\(X^{7}_{8}\)&\(X^{7}_{10}\)
 &\(X^{7}_{11}= \{A_{6},A_{7},A_{11}\}\) \\
\hline
 \(X^{7}_{2}\) &&\(-8.0\)&\(-12.0\)&\((-6.3,3.1)\)\\
 \(X^{7}_{8}= \{A_{1},A_{3},A_{4},A_{8}\}\) &&&\((-27.1,1.0)\)& \((-35.6,0.9)\)\\
 \(X^{7}_{10}= \{A_{5},A_{9},A_{10}\}\) &&&&\(-35.8\)\\

\hline
\end{tabular}
\end{center}

 {\bf Iteration 8}. Selection of the best
 (Pareto-efficient)
 improvement operation
  \(O_{2,11}\).
  The corresponding Pareto-efficient improvement is:
 \(\Delta \overline{f} (O_{2,11}  ) =(-6.3,3.1)\).
 As a result,
 the following information is used for the next algorithm step:
(a)
 \(\widehat{X}^{8} =
 \{X^{8}_{8}, X^{8}_{10},X^{8}_{11}\}\)
   where
 \(X^{8}_{8} = \{A_{1},A_{3},A_{4},A_{8}\}\),
 \(X^{8}_{10} = \{A_{5},A_{9},A_{10}\}\),
 \(X^{8}_{11} = \{A_{2},A_{6},A_{7},A_{11}\}\);
 (b) \(\overline{f}^{8} (\widehat{X}^{8}) = (-10.0,25.7)\);
 (c) improvement operations
 (i.e., inclusion of element/cluster \(X_{\iota_{1}}\) into cluster \(X_{\iota_{2}}\))
 as
 \(O_{\iota_{1},\iota_{2}} (X_{\iota_{1}} \rightarrow X_{\iota_{2}})\)
%
%
 and corresponding improvements
 (by positive component or by negative component)
  of objective function \(\overline{f}\)
 as
  \(\Delta \overline{f} (O_{\iota_{1},\iota_{2}})\)
  are presented in Table 3.31
 (component \(0\) of the vector  is not pointed out).

\begin{center}
{\bf Table 3.31.} Improvements of objective function
 \(\Delta \overline{f} (O_{j,\iota}  )\)
 (iteration index \(\gamma= 8\))
  \\
\begin{tabular}{| l |c c c c |}
\hline
 \(X_{\iota_{1}}\)&\(X_{\iota_{2}}:\)&\(X^{8}_{8}\)&\(X^{8}_{10}\)
 &\(X^{8}_{11}= \{A_{6},A_{7},A_{11}\}\) \\
\hline
 \(X^{8}_{8}= \{A_{1},A_{3},A_{4},A_{8}\}\) &&&\((-27.1,1.0)\)& \((-43.8,0.9)\)\\
 \(X^{8}_{10}= \{A_{5},A_{9},A_{10}\}\) &&&&\(-47.8\)\\

\hline
\end{tabular}
\end{center}

 Finally, it is reasonable to consider the result of iteration 8
 as the clustering solution:

  \(\widehat{X} = \widehat{X}^{8} =\{ X^{8}_{8}, X^{8}_{10}, X^{8}_{11} \}\).

~~

 Table 3.32 contains a list of main research directions in correlation
 clustering.


%
%
 On the other hand,
 it is possible to use a multiset based problem formulation.
%
%
 It is possible to replace  scale  \([-1,+1]\) (or two quantitative scales above)
 for each edge (i.e., for each element from \(G\) or element of proximity matrix \(Z\))
  by two ordinal scales:
 negative ordinal scale \([-k^{-},...,-1]\) instead of ``\(-1\)''
 and
 positive ordinal scale
 \([+1,...,k^{+}]\) instead of ``\(+1\)''.
 Note, calculation of edge weights upon
 the above-mentioned scales is sufficiently easy
 (e.g., mapping of the quantitative estimate into the ordinal scale).
%
%
 For the clustering solution
 \(\widehat{X} = \{X_{1},...,X_{\lambda}\}\)
  two total quality parameters can be calculated as follows:
 (i) total agreements  quality as multiset estimate
 (summarization by the component for all intra-cluster pairs with positive edge weight)
  \(Q^{agr} (\widehat{X})\)
  (maximization);
 (ii) total disagreements quality as multiset estimate
 (summarization by the component for all intra-cluster pairs
 with negative edge weight)
  \(Q^{disagr} (\widehat{X})\)
  (for minimization).
 As a result, the multiset based correlation clustering problem is
 (Fig. 3.17):

~~

 Find clustering solution \(\widehat{X}\)
 such that
 ~\( Q^{agr} (\widehat{X}) \rightarrow \max \)~
 and
 ~\( |Q^{disagr} (\widehat{X})| \rightarrow \min\).

\begin{center}
 {\bf Table 3.32.} Correlation clustering  \\
\begin{tabular}{| c |l | l| }
\hline
 No.& Research direction & Source(s)  \\
\hline

 1.& Basic problem formulations and complexity  &
   \cite{ailon08,bagon11,bansal02,bansal04,ben99,kri09,zimek08} \\

 2.& Surveys & \cite{ailon08,bansal04,kri09,zimek08} \\

 3.& Comparing methods for correlation clustering
 &\cite{els09}\\

 4.& Approximation algorithms (including PTAS)  & \cite{bagon11,bansal02,bansal04,giotis06,zimek08} \\

 5.& Weighted versions of correlation clustering problems
 &\cite{cgw03,cgw05,dem06}\\

 6. &Correlation clustering with fixed number of clusters
 &\cite{giotis06}\\

 7.& Maximizing agreements via semidefinite programming
  &\cite{swam04} \\

 8.& Minimizing disagreements on arbitrary weighted graphs
  &\cite{ema03} \\

 9.& Global correlation clustering
 &\cite{achtert08}\\

 10. &Correlation clustering with partial information
 &\cite{demaine03}\\

 11. &Correlation clustering with noisy input
 &\cite{mat10}\\

 12.& Error bounds for correlation clustering
 &\cite{joa05}\\

 13.& Robust correlation clustering
 &\cite{achtert07,krie08}\\


 14.& Correlation clustering in image segmentation
 &\cite{kim11}\\

\hline
\end{tabular}
\end{center}

\begin{center}
\begin{picture}(37,48)

\put(00,00){\makebox(0,0)[bl]{Fig. 3.17. Multiset quality
 posets (lattices)}}

\put(00,09){\makebox(0,0)[bl]{(a) disagreements}}
\put(03,05){\makebox(0,0)[bl]{quality
 \(|Q^{disagr}(\widehat{X})|\)}}

\put(10,14){\line(0,1){30}} \put(10,14){\line(1,1){15}}
\put(10,44){\line(1,-1){15}}

\put(10,14){\circle*{1.0}} \put(10,14){\circle{1.9}}

\put(24.5,23.4){\makebox(0,0)[bl]{Ideal points}}

\put(28,23){\vector(-2,-1){016}} \put(35.5,27){\vector(2,3){10.5}}

\end{picture}
%
\begin{picture}(32,46)

\put(00,09){\makebox(0,0)[bl]{(b) agreements}}
\put(03,05){\makebox(0,0)[bl]{quality
 \(Q^{agr}(\widehat{X})\)}}

\put(10,14){\line(0,1){30}} \put(10,14){\line(1,1){15}}
\put(10,44){\line(1,-1){15}}

\put(10,44){\circle*{1.0}} \put(10,44){\circle{2.1}}
\put(10,44){\circle{3}}

\end{picture}
\end{center}

\subsubsection{Network communities based clustering}

 In recent decades,
 ``network communities based clustering''
 as a new research direction has been organized
 (e.g., \cite{fortun10,girvan02,hopcroft03,les09,new03,new04a,new06,new10,new04,port09}).
%
%
 The largest connected components
  are examined as ``network communities'',
 for example:
 cliques, quasi-cliques,
 cliques/quasi-cliques with leaves,
 chains of cliques/quasi-cliques,
 integrated groups of clisues/quasi-cliques
 (Fig. 3.18).

\begin{center}
\begin{picture}(122,54)

\put(25.5,00){\makebox(0,0)[bl]{Fig. 3.18. Illustration
 for network communities}}

\put(00,48){\makebox(0,0)[bl]{Community 1}}
\put(04,45){\makebox(0,0)[bl]{(clique)}}

\put(10,38){\oval(15,13)}

\put(05,38){\circle*{0.7}} \put(15,38){\circle*{0.7}}
\put(10,33){\circle*{0.7}} \put(10,43){\circle*{0.7}}

\put(05,38){\circle{1.3}} \put(15,38){\circle{1.3}}
\put(10,33){\circle{1.3}} \put(10,43){\circle{1.3}}

\put(05,38){\line(1,0){10}} \put(10,33){\line(0,1){10}}
\put(05,38){\line(1,1){05}} \put(05,38){\line(1,-1){05}}
\put(15,38){\line(-1,1){05}} \put(15,38){\line(-1,-1){05}}


\put(10,33){\line(0,-1){15}}

\put(00,08){\makebox(0,0)[bl]{Community 2}}
\put(00,05){\makebox(0,0)[bl]{(clique with leave)}}

\put(10,16.5){\oval(15,10)}

\put(05,19){\circle*{1}} \put(10,19){\circle*{1}}
\put(05,14){\circle*{1}} \put(10,14){\circle*{1}}
\put(15,14){\circle*{1}}

\put(05,14){\line(1,0){10}} \put(05,19){\line(1,0){05}}

\put(05,14){\line(0,1){05}} \put(10,14){\line(0,1){05}}
\put(05,14){\line(1,1){05}} \put(10,14){\line(-1,1){05}}


\put(15,14){\line(3,1){12}} \put(27,18){\line(1,0){23}}
\put(35,33){\line(-4,1){20}}

\put(29,44){\makebox(0,0)[bl]{Community 3}}
\put(29,41){\makebox(0,0)[bl]{(quasi-clique)}}
\put(39.5,35.5){\oval(17,10)}

\put(34,38){\circle*{1.6}} \put(45,38){\circle*{1.6}}
\put(34,33){\circle*{1.6}} \put(45,33){\circle*{1.6}}

\put(34,33){\line(1,0){11}} \put(34,38){\line(1,0){11}}
\put(34,33){\line(0,1){05}} \put(45,33){\line(0,1){05}}
\put(34.3,33){\line(2,1){10}} 

\put(62,46.5){\makebox(0,0)[bl]{Community 4}}
\put(60,43.5){\makebox(0,0)[bl]{(chain of clique,}}
\put(62.5,40.5){\makebox(0,0)[bl]{quasi-clique)}}

\put(72.5,35.5){\oval(20,09)}

\put(65,38){\circle*{1.1}} \put(70,38){\circle*{1.1}}
\put(65,33){\circle*{1.1}} \put(70,33){\circle*{1.1}}

\put(65,33){\line(1,0){05}} \put(65,38){\line(1,0){05}}
\put(65,33){\line(0,1){05}} \put(70,33){\line(0,1){05}}
\put(70,33){\line(-1,1){05}} 

\put(75,38){\circle*{1.1}} \put(80,38){\circle*{1.1}}
\put(75,33){\circle*{1.1}} \put(80,33){\circle*{1.1}}

\put(75,33){\line(1,0){05}} \put(75,38){\line(1,0){05}}
\put(75,33){\line(0,1){05}} \put(80,33){\line(0,1){05}}
\put(80,33){\line(-1,1){05}}  \put(75,33){\line(1,1){05}}

\put(70,38){\line(1,0){05}}  \put(70,38){\line(1,-1){05}}
\put(65,38){\line(-1,0){20}}  \put(65,33){\line(-1,-1){10}}
\put(44,33){\line(-3,-2){29}} \put(55,23){\line(-1,1){10}}

\put(46.5,08){\makebox(0,0)[bl]{Community 5}}
\put(39.5,05){\makebox(0,0)[bl]{(quasi-clique with leave)}}

\put(57,18){\oval(19,13)}

\put(50,18){\circle*{0.9}} \put(60,18){\circle*{0.9}}
\put(55,13){\circle*{0.9}} \put(55,23){\circle*{0.9}}
\put(65,18){\circle*{0.9}}

\put(50,18){\circle{1.8}} \put(60,18){\circle{1.8}}
\put(55,13){\circle{1.8}} \put(55,23){\circle{1.8}}
\put(65,18){\circle{1.8}}

\put(50,18){\line(1,0){10}}

\put(60,18){\line(1,0){05}}

 \put(50,18){\line(1,-1){05}}
\put(60,18){\line(-1,1){05}} \put(60,18){\line(-1,-1){05}}

\put(86,08){\makebox(0,0)[bl]{Community 6 (group}}
\put(84,05){\makebox(0,0)[bl]{of cliques/quasi-cliques)}}

\put(101.5,23){\oval(26,23)}

\put(90,18){\circle*{1.4}} \put(100,18){\circle*{1.4}}
\put(95,13){\circle*{1.4}} \put(95,23){\circle*{1.4}}

\put(95,13){\line(0,1){10}}

\put(100,33){\circle*{1.4}} \put(105,33){\circle*{1.4}}
\put(100,28){\circle*{1.4}} \put(105,28){\circle*{1.4}}
\put(100,28){\line(0,1){05}}  \put(105,28){\line(0,1){05}}
\put(100,28){\line(1,0){05}} \put(100,33){\line(1,0){05}}

\put(100,28){\line(1,1){05}}

\put(100,28){\line(-1,-1){05}} \put(105,28){\line(1,-1){05}}

\put(100,33){\line(-1,0){20}}
\put(90,18){\line(-1,0){25}}

\put(105,18){\circle*{1.4}}\put(110,23){\circle*{1.4}}
\put(110,13){\circle*{1.4}}

\put(105,18){\line(1,1){05}}  \put(105,18){\line(1,-1){05}}
\put(110,13){\line(0,1){10}}


\put(90,18){\line(1,0){10}}

\put(100,18){\line(1,0){05}}

 \put(90,18){\line(1,-1){05}}  \put(90,18){\line(1,1){05}}
\put(100,18){\line(-1,1){05}} \put(100,18){\line(-1,-1){05}}

\end{picture}
\end{center}

 The network example in Fig. 3.18 does not contains overlaps
 (i.e., without intersection of community structures).
 Fig. 3.19 illustrates the overlaps.

\begin{center}
\begin{picture}(74,46)
\put(00,00){\makebox(0,0)[bl]{Fig. 3.19. Illustration
 for overlaps in structures}}


\put(15,32.5){\oval(25,18)}

\put(05,35){\circle*{0.7}} \put(15,35){\circle*{0.7}}
\put(10,30){\circle*{0.7}} \put(10,40){\circle*{0.7}}

\put(05,35){\circle{1.3}} \put(15,35){\circle{1.3}}
\put(10,30){\circle{1.3}} \put(10,40){\circle{1.3}}

\put(25,35){\circle*{0.7}} \put(25,35){\circle{1.3}}
\put(20,30){\circle*{0.7}} \put(20,30){\circle{1.3}}

\put(25,35){\line(-3,1){15}}

\put(15,35){\line(1,0){10}} \put(10,30){\line(1,0){10}}
\put(20,30){\line(1,1){05}} \put(20,30){\line(-1,1){05}}

\put(05,35){\line(1,0){10}} \put(10,30){\line(0,1){10}}
\put(05,35){\line(1,1){05}} \put(05,35){\line(1,-1){05}}
\put(15,35){\line(-1,1){05}} \put(15,35){\line(-1,-1){05}}


\put(10,30){\line(0,-1){5}} \put(20,30){\line(0,-1){5}}
\put(20,30){\line(-1,-1){5}} \put(10,30){\line(2,-1){10}}



\put(10,25){\circle*{1.7}} \put(15,25){\circle*{1.7}}
\put(20,25){\circle*{1.7}}

\put(10,25){\line(1,0){05}}


\put(10,25){\line(0,-1){10}} \put(15,25){\line(0,-1){10}}
\put(20,25){\line(0,-1){10}}

\put(10,25){\line(-1,-2){05}} \put(10,25){\line(1,-1){05}}

\put(15,25){\line(1,-1){05}}

\put(20,25){\line(-1,-1){05}}

\put(00,06){\makebox(0,0)[bl]{(a) Sparse overlaps}}

\put(15,19){\oval(25,15)}

\put(05,20){\circle*{1}} \put(10,20){\circle*{1}}
\put(05,15){\circle*{1}} \put(10,15){\circle*{1}}
\put(15,15){\circle*{1}} \put(20,15){\circle*{1}}
\put(15,20){\circle*{1}} \put(20,20){\circle*{1}}

\put(25,17.5){\circle*{1}}

\put(20,15){\line(2,1){05}} \put(20,20){\line(2,-1){05}}

\put(15,15){\line(0,1){05}} \put(20,15){\line(0,1){05}}

\put(15,15){\line(1,0){05}} \put(10,20){\line(1,0){10}}

\put(10,15){\line(1,1){05}} \put(15,15){\line(-1,1){05}}
\put(15,15){\line(1,1){05}} \put(20,15){\line(-1,1){05}}

\put(05,15){\line(1,0){10}} \put(05,20){\line(1,0){05}}

\put(05,15){\line(0,1){05}} \put(10,15){\line(0,1){05}}
\put(05,15){\line(1,1){05}} \put(10,15){\line(-1,1){05}}



\put(57.5,36){\oval(32,19)}

\put(45,43){\circle*{1.1}} \put(50,43){\circle*{1.1}}
\put(45,38){\circle*{1.1}} \put(50,38){\circle*{1.1}}

\put(45,38){\line(1,0){05}} \put(45,43){\line(1,0){05}}
\put(45,38){\line(0,1){05}} \put(50,38){\line(0,1){05}}
\put(50,38){\line(-1,1){05}} 

\put(55,43){\circle*{1.1}} \put(60,43){\circle*{1.1}}
\put(55,38){\circle*{1.1}} \put(60,38){\circle*{1.1}}

\put(55,38){\line(1,0){05}} \put(55,43){\line(1,0){05}}
\put(55,38){\line(0,1){05}} \put(60,38){\line(0,1){05}}
\put(60,38){\line(-1,1){05}}  \put(55,38){\line(1,1){05}}


\put(60,38){\line(1,0){05}} \put(60,43){\line(1,0){05}}
\put(60,38){\line(0,1){05}} \put(65,38){\line(0,1){05}}
\put(65,38){\line(-1,1){05}}  \put(60,38){\line(1,1){05}}


\put(65,43){\circle*{1.1}} \put(70,43){\circle*{1.1}}
\put(65,38){\circle*{1.1}} \put(70,38){\circle*{1.1}}

\put(65,38){\line(1,0){05}} \put(65,43){\line(1,0){05}}
\put(65,38){\line(0,1){05}} \put(70,38){\line(0,1){05}}
\put(70,38){\line(-1,1){05}}  \put(65,38){\line(1,1){05}}

\put(50,43){\line(1,0){05}}  \put(50,43){\line(1,-1){05}}



\put(55,30.5){\circle*{1.7}} \put(65,30.5){\circle*{1.7}}
\put(60,28){\circle*{1.7}} \put(60,33){\circle*{1.7}}

\put(60,28){\line(0,1){05}} \put(55,30.5){\line(1,0){10}}
\put(60,28){\line(-2,1){05}} \put(60,28){\line(2,1){05}}
\put(60,33){\line(-2,-1){05}} \put(60,33){\line(2,-1){05}}


\put(60,28){\line(0,-1){05}} \put(60,33){\line(0,1){05}}

\put(60,33){\line(-1,1){05}} \put(60,33){\line(1,1){05}}

\put(55,30.5){\line(0,1){07.5}} \put(65,30.5){\line(0,1){07.5}}

\put(55,30.5){\line(0,-1){12.5}} \put(65,30.5){\line(0,-1){12.5}}

\put(43,06){\makebox(0,0)[bl]{(b) Dense overlaps}}

\put(57.5,23){\oval(30,23)}

\put(55,18){\circle*{0.9}} \put(65,18){\circle*{0.9}}
\put(60,13){\circle*{0.9}} \put(60,23){\circle*{0.9}}
\put(70,18){\circle*{0.9}}

\put(55,18){\circle{1.8}} \put(65,18){\circle{1.8}}
\put(60,13){\circle{1.8}} \put(60,23){\circle{1.8}}
\put(70,18){\circle{1.8}}

\put(45,18){\circle*{0.9}} \put(45,18){\circle{1.8}}

\put(55,18){\line(1,1){05}}

\put(60,13){\line(0,1){10}}

\put(45,18){\line(1,0){10}}

\put(45,18){\line(3,-1){15}} \put(45,18){\line(3,1){15}}

\put(55,18){\line(1,0){10}}

\put(65,18){\line(1,0){05}}

 \put(55,18){\line(1,-1){05}}
\put(65,18){\line(-1,1){05}} \put(65,18){\line(-1,-1){05}}

\end{picture}
\end{center}

 The detection of
 ``network communities structures''
 corresponds to complex combinatorial
 optimization models
 (e.g., linear/nonlinear integer programming,
 mixed integer programming).
 The models belong to NP-hard problems
 (e.g., \cite{brand08,clay04,fortun10,new10}).
%
%
 Table 3.33 contains a list of basic research directions in
 community network based clustering.

 A list of basic algorithmic approaches  for finding communities
 involves the following (e.g., \cite{fortun10,new06,new10,new04}):
%
%
 (i) graph partitioning (e.g., minimum-cut method),
 (ii) hierarchical clustering
 (greedy agglomerative algorithms),
  (iii) Girvan-Newman algorithm (edge betweeness),
%
 (iv) modularity maximization approaches,
%
 (v) spectral clustering methods,
 (vi) methods based on statistical inference,
  (vii) clique based methods.

 Modularity of a graph can be defined as a normalized tradeoff between edges
 covered by clusters and squared cluster degree sums
 \cite{brand08,new04}.
 The problem
 is formulated as combinatorial optimization model.
 For the modularity maximization,
 several main algorithms are pointed out \cite{brand08}:
 (a) greedy agglomeration \cite{clay04,new03},
 (b) spectral division  \cite{new06,white05},
%
 (c) simulated annealing \cite{gui04,reich06},
%
 (d) extremal optimization \cite{duch05}.
%
%
 An example of modularity algorithm as greedy agglomerative heuristic
 is the following
  \cite{new03}:

 ~~

  {\it Stage 1.} Trivial clustering:
 each node corresponds to its own cluster.

  {\it Stage 2.} Cycle by cluster pairs:

 {\it Stage 2.1.} Calculation of possible increase of modularity
 for merging each cluster pairs.

  {\it Stage 2.2.} Merging  the two clusters
  with maximum possible increase.

  {\it Stage 2.3.} If increasing of modularity
  by merges of cluster pair
  is impossible then  GO TO Stage 3.

  {\it Stage 2.4.} Go To Stage 2.2.

  {\it Stage 3.} Stop.

~~

 In this
  algorithm,
 algorithmic complexity estimate equals
 \(O((p+n) n)\) or \(O(n^{2})\)
 \cite{new03}.

The general scheme of  Girvan-Newman (GN)
 algorithm based on edge betweenness
 is
  \cite{girvan02}:

~~

 {\it Step 1.} Calculation of the betweenness score for each the
 edges.

 {\it Step 2.} Deletion of the edges with the highest score.

 {\it Step 3.} Performance analysis for the network's components.

 {\it Step 4.} If all edges are deleted
 and the system breaks up into \(N\) non-connected nodes
 Go TO Step 5.
 Otherwise GO TO Step 1.

 {\it Step 5.} Stop.

~~


 Algorithmic complexity estimate of the algorithm
 equals \(O(p^{2} n)\)  (\(p\) is the number of edges)
 \cite{girvan02}).


%

\subsection{Towards fast clustering}


 Many
 applications based on
 very large data sets/networks require
  fast clustering
   approaches
 (e.g., \cite{clay04,new03,shio13,tsai10,wak07,zimek08}).
 In Table 3.34,  basic ideas for fast clustering
 schemes are pointed out.
%
 Generally, many  fast clustering
 schemes consist of two basic levels (global level and local level):
 (a) partition of the initial problems into local
 problems (i.e., decreased dimension, limited type of
 objects/elements) (global level),
 (b) clustering of local clustering problems (local level),
 (c) composition/integration of local clustering
 solutions into a resultant  global clustering
 solution (global level).

 In Table 3.35,
 a list of  basic fast local clustering algorithms
 (i.e., fast sub-algorithms) is presented.

\newpage
\begin{center}
 {\bf Table 3.33.} Community network based clustering  \\
\begin{tabular}{| c |l | l| }
\hline
 No.& Research direction & Source(s)  \\
\hline

 1.& Basic issues:&\\
 1.1.& Basic problem formulations
 &\cite{brand08,fortun10,girvan02,les09,new04,new06,new10,port09,wang13,yangj14a,yangj15} \\

 1.2.& Basic surveys & \cite{brand08,boc06,fortun10,les09,mitchell06,new04,new06,new10,port09} \\

 1.3.& Problems complexity  & \cite{brand08,clay04,fortun10,new10} \\

 1.4.& Overlapping (fuzzy) community structures
 &\cite{gop13,wang13,xie13,yangj13,yangj14,yangj14a}\\

  1.5.& Analysis/evaluation of community structures  & \cite{les09,new04,wang13,yangj15} \\

 2.& Main algorithms/solving schemes: &\\

 2.1.& Algorithm based on edge betweeness
 & \cite{girvan02}\\
 & (divisive algorithm)&\\

 2.2.& Modularity algorithm as greedy agglomerative
 & \cite{new03,ovel12}\\
 &  heuristic& \\

 2.3.& ``Karate Club'' algorithm &\cite{new04}\\

 2.4.& Kernighan-Lin method and variants &\cite{ker70}\\

 2.5.& Overlapping communities (clique percolation,
 &\cite{gop13,xie13,yangj13,yangj14,yangj14a}\\

 & local expansion, dynamic algorithms, etc.)&\\

 2.6. &Spectral clustering algorithms,
  modifications
 &\cite{yangj15}\\

 2.7.& Genetic algorithms  &\cite{liu07}  \\

 2.8.& Agent-based algorithms &\cite{gunes06}  \\



 3.& Modularity clustering (maximum modularity):& \\

 3.1.& Surveys& \cite{brand08,new06,ziv05,yangj15}\\

 3.2.& Tripartite modularity (three vertex types)&
 \cite{mur10a,mur10b} \\

 3.3.&Modularity in \(k\)-partite networks &
 \cite{liu11} \\

 3.4.& Greedy agglomeration algorithm & \cite{clay04,new03}\\

 3.5.& Spectral division algorithm  &\cite{new06,white05}\\

 3.6.& Simulated annealing algorithms & \cite{gui04,reich06}\\

 3.7.& Detecting communities by merging cliques
 &\cite{yan09}\\

 3.8.& Extremal optimization scheme& \cite{agar08,duch05}\\
 & (mathematical programming)&\\

 3.9.& Global optimization approach &\cite{med05}\\

 3.10.& Memetic algorithm &\cite{naeni15}\\

 3.11.& Random works algorithms &\cite{pons06}\\

 3.12.& Multi-level algorithms &\cite{dji06,noack08,rotta08}\\

 4.& Large networks:&\\

 4.1.& Communities in large networks& \cite{blond08,clay04,gop13,hopcroft03,hopcroft04,les09,pons06,yangj15}\\

 4.2.& Communities in mega-scale networks & \cite{wak07}\\

 4.3.& Communities in super-scale networks & \cite{blond08}\\

 4.4.& Tracking evolving communities in large networks
 &\cite{hopcroft04}\\

 5.& Applications: &\\
 5.1.&World Wide Web     & \cite{dji06,les09,mur10a} \\
 5.2.&Journal/article networks, citation networks, etc.
 & \cite{chenp10,fortun10,ros07} \\
 5.3.& Social networks (friendship,
 collaboration, etc.)
 &\cite{fortun10,girvan02,gui03,new03,new10,new04,wak07,yangj15} \\
 5.4.& Biological networks &\cite{fortun10,girvan02,new04} \\
 5.5.& Purchasing network &\cite{clay04}\\
 5.6.&
 CAD applications&\cite{new04} \\
 5.7.& Antenna-To-Antenna network
  &\cite{blond12,liu13}\\

 &(mobile phone network) &\\

\hline
\end{tabular}
\end{center}

\newpage
\begin{center}
 {\bf Table 3.34.} Main approaches to fast clustering \\
\begin{tabular}{| c |l | l| l|}
\hline
 No.& Approach &Solving schemes &Source(s)  \\
\hline

 1.& Aggregation of object/network &Hierarchical clustering (Bottom-Up,&\cite{jain88,jain99,shio13} \\
   & nodes                 & step-by-step node aggregation) & \\


 2.& Division of objects/network&Top-Down scheme &\\
  &nodes (partition/decomposition):&&\\

 2.1.& Pruning of objects/network & 1.Selection of basic edge betweenness &\cite{girvan02,shio13} \\
   & nodes  (Fig. 3.20)     &in  graph and decoupling  & \\
   &                                &(Top-Down  scheme) & \\
                          && 2.Clustering in each graph part&\\
                        &&(if needed)&\\

 2.2. & Multi-level schemes (partition,&1.Partition of object set/network&  \cite{tsai07,tsai10,zimek08}\\
   &clustering, integration of solutions): &2.Clustering of local regions&this paper\\
    &&3.Composition of local solutions&\\

 2.2.1.&``Basic'' objects (special ``key''&1.Detection of ``basic'' objects/nodes  & this paper\\
     &  objects/nodes) based clustering&(e.g., by filtering)&\\
    &  (Fig. 3.21)   & 2.Clustering of ``basic'' objects/nodes,  &\\
    &             & 3.Joining other elements/nodes &\\
    &                  & to obtained clusters&\\


 2.2.2.& Grid-based clustering &Dividing the space into cells&\cite{kri09,lin08,zimek08} \\
 2.2.3.& Grid-based clustering in data streams&Online clustering of data streams & \cite{luy05,park04}\\

 2.2.4.&Grid-based clustering (composition):&1.Grid over object ``space''/network
     &  this paper\\

 &multiple division of objects
  &2.Analysis of grid regions &\\

 & ``space''/network into cells/regions&3.Selection of ``non-empty'' regions
      &\\

 &(e.g., axis-parallel subspaces),&4.Clustering in ``dense'' regions&\\
 & region-based clustering, &5.Clustering in ``sparse'' regions&\\
 & composition of local solutions& (while taking into account &\\
 & (Fig. 3.22)                 & solutions in ``dense'' regions)&\\
 &                        &6.Composition of regions solutions&\\

 2.2.5.&Grid-based clustering (extension):
                          &1.Grid over object
                          ``space''/network   &this paper \\

 &multiple division of objects &2.Analysis of grid regions&\\

 &``space''/network and ``extension''&3.Selection of ``non-empty'' regions&\\

 &of clustering solutions &4.Clustering in ``dense'' regions&\\
 &(with condensing of clusters, &5.Extension of ``dense'' regions&\\
 &as in dynamic programming)& by neighbor region(s) and extension &\\
 &(Fig. 3.22)       & of clustering solution(s)&\\

 2.2.6.&Division of object ``space''/
               &1.Detection of objects by types&\cite{liu11,mur10a,mur10b}\\
   &network by types (\(k\)-partite &2.Clustering for each part &this paper\\
   &network) (close to 2.2.1)&3.Composition of clustering solutions&\\


 3.&Composite (multistage, concurrent,&Composition/combination  & \cite{kyper08}\\
  &multi-techniques) approaches &of various approaches&this paper\\

\hline
\end{tabular}
\end{center}


\begin{center}
\begin{picture}(80,42)

\put(06.5,00){\makebox(0,0)[bl]{Fig. 3.20.
 Edge betweenness for decoupling}}

\put(27,38){\makebox(0,0)[bl]{Initial set of
 objects/network}}

\put(40,22){\oval(80,30)}


\put(10,10){\makebox(0,0)[bl]{Part 1}}

\put(19,15){\oval(36,14)}

\put(33,19){\circle*{1.7}} \put(33,19){\line(3,2){12}}
\put(33,19){\line(-1,0){5}} \put(33,19){\line(-3,-1){5}}
\put(33,19){\line(-1,-1){4}} \put(33,19){\line(0,-1){5}}
\put(33,19){\line(-1,-3){1.8}}

\put(33,19){\line(3,2){12}}

\put(50,18){\makebox(0,0)[bl]{Edge of cut}}
\put(50,20){\vector(-2,1){8.17}}


\put(05,30){\makebox(0,0)[bl]{Decoupling}}
\put(11,27){\makebox(0,0)[bl]{line}}

\put(39,23){\line(-3,2){24}} \put(39,23){\line(3,-2){26}}
\put(39,22.7){\line(-3,2){24}} \put(39,22.7){\line(3,-2){26}}

\put(45,27){\circle*{1.1}} \put(45,27){\circle{2.1}}
\put(45,27){\line(0,1){5}}

\put(45,27){\line(3,2){5}} \put(45,27){\line(3,-1){5}}

\put(60,30){\oval(38,12)}

\put(60,30){\makebox(0,0)[bl]{Part 2}}

\end{picture}
\end{center}

\begin{center}
\begin{picture}(147,41)

\put(30,00){\makebox(0,0)[bl]{Fig. 3.21. ``Basic'' objects
 based  clustering framework}}

\put(09,20){\oval(18,22)}

\put(1.5,21){\makebox(0,0)[bl]{Initial set}}
\put(1.5,17){\makebox(0,0)[bl]{of objects}}

\put(18,20){\vector(1,0){04}}


\put(22,10){\line(1,0){13}} \put(22,30){\line(1,0){13}}
\put(22,10){\line(0,1){20}} \put(35,10){\line(0,1){20}}

\put(23,23){\makebox(0,0)[bl]{Detec-}}
\put(23,20){\makebox(0,0)[bl]{tion of}}
\put(23,17){\makebox(0,0)[bl]{``basic''}}
\put(23,14){\makebox(0,0)[bl]{objects}}

\put(35,20){\vector(1,0){04}}

\put(49,20){\oval(20,22)}

\put(49,20){\oval(18,16)} \put(49,20){\oval(17.5,15.5)}

\put(42.5,21){\makebox(0,0)[bl]{``Basic'' }}
\put(42.5,17){\makebox(0,0)[bl]{objects}}


\put(59,20){\vector(1,0){04}}


\put(63,10){\line(1,0){12}} \put(63,30){\line(1,0){12}}
\put(63,10){\line(0,1){20}} \put(75,10){\line(0,1){20}}

\put(65,24.5){\makebox(0,0)[bl]{Clus-}}
\put(64.5,21.5){\makebox(0,0)[bl]{tering}}
\put(67,18.5){\makebox(0,0)[bl]{of}}
\put(63.5,15.5){\makebox(0,0)[bl]{``basic'' }}
\put(63.5,12.5){\makebox(0,0)[bl]{objects}}

\put(75,20){\vector(1,0){04}}


\put(89,20){\oval(20,22)}

\put(84,26){\oval(6,6)} \put(84,26){\oval(6.5,6.5)}

\put(93,25){\oval(6,9)} \put(93,25){\oval(6.5,9.5)}

\put(84,19){\oval(8,5)} \put(84,19){\oval(8.5,5.5)}

\put(94,16){\oval(7,5)} \put(94,16){\oval(7.5,5.5)}

\put(86,12){\oval(5,4)} \put(86,12){\oval(5.5,4.5)}


\put(74.5,37){\makebox(0,0)[bl]{Clustering solution}}
\put(74.5,34){\makebox(0,0)[bl]{for ``basic'' objects}}

\put(88,33.5){\line(-2,-1){04.4}} \put(89,33.5){\line(2,-1){04.4}}

\put(99,20){\vector(1,0){04}}


\put(103,10){\line(1,0){17}} \put(103,30){\line(1,0){17}}
\put(103,10){\line(0,1){20}} \put(120,10){\line(0,1){20}}

\put(105.6,26){\makebox(0,0)[bl]{Joining}}
\put(105.6,23){\makebox(0,0)[bl]{of other}}
\put(105.6,20){\makebox(0,0)[bl]{objects }}
\put(104,17){\makebox(0,0)[bl]{to ``basic''}}
\put(106,14){\makebox(0,0)[bl]{object}}
\put(105,11){\makebox(0,0)[bl]{clusters}}

\put(120,20){\vector(1,0){04}}



\put(129,29){\oval(6,6)} \put(129,29){\oval(6.5,6.5)}
\put(129,29){\oval(8.5,10)}

\put(140,25){\oval(6,9)} \put(140,25){\oval(6.5,9.5)}
\put(140,25){\oval(8.5,13.5)}

\put(130,19){\oval(8,5)} \put(130,19){\oval(8.5,5.5)}
\put(130,19){\oval(10.5,7.5)}

\put(141,13){\oval(7,5)} \put(141,13){\oval(7.5,5.5)}
\put(141,13){\oval(11.5,8.5)}

\put(130,11){\oval(5,4)} \put(130,11){\oval(5.5,4.5)}
\put(130,11){\oval(9.5,6.5)}

\put(122,37){\makebox(0,0)[bl]{Final clustering}}
\put(128,34.5){\makebox(0,0)[bl]{solution}}


\put(135,34){\line(0,-1){04}}

\end{picture}
\end{center}

\begin{center}
\begin{picture}(105,65)

\put(14.5,00){\makebox(0,0)[bl]{Fig. 3.22. ``Grid'' over object
 ``space''/network}}

\put(00,57){\makebox(0,0)[bl]{\(y\)}}

\put(00,07){\vector(0,1){49}} \put(00,07){\vector(1,0){101}}
\put(102,06){\makebox(0,0)[bl]{\(x\)}}

\put(18,56.5){\makebox(0,0)[bl]{``Grid''}}
\put(21,56){\vector(-2,-1){04}} \put(26,56){\vector(2,-1){04}}

\put(62,56.5){\makebox(0,0)[bl]{Object ``space''/network}}
\put(75,56.5){\vector(0,-1){05}} \put(89,56.5){\vector(0,-1){05}}

\put(51,32){\oval(90,38)}



\put(03,52){\line(1,0){96}} \put(03,43){\line(1,0){96}}
\put(03,33){\line(1,0){96}} \put(03,21){\line(1,0){96}}
\put(03,12){\line(1,0){96}}

\put(05,10){\line(0,1){44}} \put(16,10){\line(0,1){44}}
\put(31,10){\line(0,1){44}} \put(44,10){\line(0,1){44}}
\put(56,10){\line(0,1){44}} \put(71,10){\line(0,1){44}}
\put(86,10){\line(0,1){44}} \put(97,10){\line(0,1){44}}

\put(37.5,47.5){\oval(18,11)}

\put(32,50){\circle*{1.1}} \put(41,47){\circle*{1.1}}
\put(34,45){\circle*{1.1}} \put(40,44){\circle*{1.1}}
\put(37,49){\circle*{1.1}} \put(38,48){\circle*{1.1}}


\put(63.5,38){\oval(20,12)}

\put(58,38){\makebox(0,0)[bl]{Empty}}
\put(58.4,35){\makebox(0,0)[bl]{region}}

\put(29.5,27){\oval(32,17)}

\put(16,39){\makebox(0,0)[bl]{Extended}}
\put(18.5,36){\makebox(0,0)[bl]{region}}
 \put(28.5,37.5){\vector(1,-1){07}}

\put(23.5,27){\oval(19,16)}

\put(20.5,29.5){\oval(9,8)}

\put(18,32){\circle*{0.7}} \put(18,32){\circle{1.4}}
\put(23,32){\circle*{0.7}} \put(23,32){\circle{1.4}}
\put(18,27){\circle*{0.7}} \put(18,27){\circle{1.4}}
\put(18,32){\line(1,0){05}} \put(18,27){\line(1,1){05}}
\put(18,27){\line(0,1){05}}

\put(27,24){\oval(7,6)}

\put(25,26){\circle*{1}} \put(29,26){\circle*{1}}
\put(25,22){\circle*{1}} \put(29,22){\circle*{1}}
\put(25,22){\line(1,0){04}} \put(25,26){\line(1,0){04}}
\put(25,22){\line(0,1){04}} \put(29,22){\line(-1,1){04}}
\put(29,22){\line(0,1){04}}


\put(32.5,24.5){\oval(17,6.5)}


\put(36.5,24.5){\oval(7.5,6.5)}

\put(34,27){\circle*{1.2}} \put(39,27){\circle*{1.2}}
\put(34,22){\circle*{1.2}} \put(39,22){\circle*{1.2}}
\put(34,22){\line(1,0){05}} \put(34,27){\line(1,0){05}}
\put(34,22){\line(0,1){05}} \put(34,22){\line(1,1){05}}
\put(39,22){\line(0,1){05}} \put(39,22){\line(-1,1){05}}

\put(19,14.5){\makebox(0,0)[bl]{Joining }}
\put(32,15){\makebox(0,0)[bl]{cluster}}
\put(32.5,17.5){\vector(0,1){04.5}}

\put(79,27){\oval(19,16)}

\put(76,28){\circle*{1}} \put(84,30){\circle*{1}}
\put(78,22){\circle*{1}} \put(81,23){\circle*{1}}
\put(74,26){\circle*{1}} \put(85,27){\circle*{1}}
\put(72,32){\circle*{1}} \put(85,22){\circle*{1}}


\end{picture}
\end{center}

\begin{center}
{\bf Table 3.35.}  List of some fast local clustering algorithms \\
\begin{tabular}{| c |l|l|c| l |}
\hline
  No. &Fast scheme & Description&Complexity &Source(s)   \\
  & & &estimate&  \\
   & & &(running time)&  \\
\hline

 1. &Basic agglomerative  & Bottom-up joining the &\(O (n^{3})\)& \cite{jain99} \\
 &(hierarchical) algorithm & closest object pair &   &  \\

 2. &Balanced by cluster size & Bottom-up joining the  &\(O (n^{3})\)&
  \\
 & hierarchical algorithm & closest object pair under&   &  \\
 &  & constraints for cluster size  &   &  \\

 3.&Minimum spanning tree&Clustering the spanning&\(O (n\log n)\)&\cite{gow69,gry06,mul12} \\
    &based algorithm & tree nodes &   &  \cite{paiv05,peter10,srin94}\\
    &  &   &   &  \cite{wang09,xu01,zhong10}\\

 4. &Balanced by cluster size&Clustering the spanning&\(O (n\log n)\)&
 \\
    &minimum spanning tree & tree nodes under&   &  \\
  &based algorithm & constraints for cluster size &   &  \\
 5. &Graph clustering & Detection of network &\(O (p^{2} n)\)& \cite{girvan02} \\
 &algorithm & communities  (edge  &   &  \\
 & &betweeness of the graph) &   &  \\

 6. &Modularity graph & Modularity based detection&\(O ((p+n)n)\) or & \cite{new03} \\
 &clustering algorithm & of network communities  &  \(O (n^{2})\) &  \\
 7. &Algorihtms based on &Assignment of objects into&\(O (n + n' \times n'')\) &   \cite{tsai10}  \\
 &grid over ``space of  &local regions of  ``space& (\(n' \ll n, n'' \ll n \))  &
 \\
 & object coordinates''        & of object coordinates'' && \\
 &(partition space &&&\\
 &techniques)&&&\\

  8.& Clustering based on& Preliminary cores& \( O(n^{2}) +O(h)\) & \cite{bata03}\\
   &cores decomposition   &decomposition      &&\\
   & of networks          & of covering  graph&&\\


\hline
\end{tabular}
\end{center}

\newpage
\section{Some Combinatorial Clustering Problems}


\subsection{Clustering with interval multiset estimates}

\subsubsection{Some  combinatorial optimization problems with
  multiset estimates}

 Multiset estimates are a simplification of multicriteria (vector)
 estimates.
 As a result, a simple scale (a little more complicated as an
 ordinal scale) is used.
 On the other hand,
 multiset estimate is a simple generalization
 of well-known binary voting procedure.
 Thus, multiset estimate can be used for simplification of
 multicriteria (multi-parameter) measurement
 in various problems/procedures.
%







%



 In \cite{lev12multiset,lev15},
 basic operations over multiset estimates
 have been described:
 integration, vector-like proximity, aggregation, and alignment.


 Integration of estimates (mainly, for composite systems)
 can be considered as summarization of the estimates by components (i.e.,
 positions).
%
%
%
%
%
%


 Let us consider vector-like proximity of two multiset estimates
 \cite{lev12multiset,lev15}.
  Let \(A_{1}\) and \(A_{2}\) be two alternatives
 with corresponding
 interval multiset estimates
 \(e(A_{1})\), \(e(A_{2})\).
  Vector-like proximity for the alternatives above is:
 ~~\(\delta ( e(A_{1}), e(A_{2})) = (\delta^{-}(A_{1},A_{2}),\delta^{+}(A_{1},A_{2}))\),
 where vector components are:
%
 (i) \(\delta^{-}\) is the number of one-step changes:
 element of quality \(\iota + 1\) into element of quality \(\iota\) (\(\iota = \overline{1,l-1}\))
 (this corresponds to ``improvement'');
 (ii) \(\delta^{+}\) is the number of one-step changes:
 element of quality \(\iota\) into element of quality  \(\iota+1\) (\(\iota = \overline{1,l-1}\))
 (this corresponds to ``degradation'').
 It is assumed:
 ~\( | \delta ( e(A_{1}), e(A_{2})) | = | \delta^{-}(A_{1},A_{2}) | + |\delta^{+}(A_{1},A_{2})|
 \).
%


 Aggregation of multiset estimates can be defined as
  a median estimate
  for the
 specified set of initial estimates
 (traditional approach).
 Let \(E = \{ e_{1},...,e_{\kappa},...,e_{n}\}\)
 be the set of specified estimates
 (or a corresponding set of specified alternatives),
 let \(D \)
 be the set of all possible estimates
 (or a corresponding set of possible alternatives)
 (\( E  \subseteq D \)).
%
  Thus, the median estimates
  (``generalized median'' \(M^{g}\) and ``set median'' \(M^{s}\)) are:
%
 ~~\(M^{g} =   \arg \min_{M \in D}~
   \sum_{\kappa=1}^{n} ~  | \delta (M, e_{\kappa}) |; ~~
%
 M^{s} =   \arg \min_{M\in E} ~
   \sum_{\kappa=1}^{n} ~ | \delta (M, e_{\kappa}) |\).
%
%
%
%
%
%
%
%

~~

 {\bf Multiple choice problem}
 with multiset estimates can be considered as
 follows
  \cite{lev12multiset,lev15}.
%
%
 Basic multiple choice problem is:
%
(e.g., \cite{gar79,keller04}):
%
 \[\max\sum_{i=1}^{m} \sum_{j=1}^{q_{i}} c_{ij} x_{ij} ~~~~~
  s.t. ~ \sum_{i=1}^{m} \sum_{j=1}^{q_{i}} a_{ij} x_{ij} \leq b;
  ~ \sum_{j=1}^{q_{i}} x_{ij} \leq 1,~ i=\overline{1,m}; ~
   x_{ij} \in \{0, 1\}.\]
%
%

%

%
 In the case of multiset estimates of
 item ``utility'' \(e_{i}, i \in \{1,...,i,...,m\}\)
 (instead of \(c_{i}\)),
 the following aggregated multiset estimate can be used
 for the objective function  (``maximization'')
 \cite{lev12multiset,lev15}):
 (a) an aggregated multiset estimate as the ``generalized median'',
 (b) an aggregated multiset estimate as the ``set median'',
 and
 (c) an integrated  multiset estimate.

 A special case of multiple choice problem is considered:

 (1) multiset estimates of item ``profit''/``utility''
  \(e_{i,j}, ~ i \in \{1,...,i,...,m\}, j = \overline{1,q_{i}}\)
 (instead of \(c_{ij}\)),

 (2) an aggregated multiset estimate as the ``generalized median''
 (or ``set median'')
 is used for the objective function (``maximization'').

 The
  item set is: ~~~
   \(A= \bigcup_{i=1}^{m} A_{i}\),
%
%
 ~\( A_{i} = \{ (i,1),(i,2),...,(i,q_{i})\} \).

 Boolean variable \(x_{i,j}\) corresponds to selection of the
 item \((i,j)\).
 The solution  is a subset of the initial item set:
 \( S = \{ (i,j) | x_{i,j}=1 \} \).
  The problem is:
%
%
 \[ \max~ e(S) =   \max~ M = ~~~~~
  \arg \min_{M \in D}
 ~~
  \sum_{(i,j) \in S=\{(i,j)| x_{i,j}=1\}} ~ | \delta (M, e_{i,j}) |,\]
 \[ s.t. ~~ \sum_{i=1}^{m} \sum_{j=1}^{q_{i}}  a_{ij} x_{i,j} \leq b;
 ~~ \sum_{j=1}^{q_{i}} x_{ij} =  1;
%
%
 ~~ x_{ij} \in \{0, 1\}.\]
%
%
 Here the following algorithms can be used (as for basic multiple choice problem)
 (e.g., \cite{gar79,keller04,lev12multiset,lev15}):
 (i) enumerative methods including dynamic programming approach,
 (ii) heuristics (e.g,  greedy algorithms),
 (iii) approximation schemes
 (e.g., modifications of dynamic programming approach).
%
%
%

~~

%
{\bf Combinatorial synthesis}  (Hierarchical Multicriteria
 Morphological Design - HMMD) with ordinal estimates of
 design alternatives is examined as follows
 (\cite{lev06,lev09,lev12morph,lev12multiset,lev15}).
 A composite
 (modular, decomposable) system consists
 of components and their interconnection or compatibility (IC).
 Basic assumptions of HMMD are the following:
 ~{\it (a)} a tree-like structure of the system;
 ~{\it (b)} a composite estimate for system quality
     that integrates components (subsystems, parts) qualities and
    qualities of IC (compatibility) across subsystems;
 ~{\it (c)} monotonic criteria for the system and its components;
 ~{\it (d)} quality of system components and IC are evaluated on the basis
    of coordinated ordinal scales.
 The designations are:
  ~(1) design alternatives (DAs) for leaf nodes of the model;
  ~(2) priorities of DAs (\(\iota = \overline{1,l}\);
      \(1\) corresponds to the best one);
  ~(3) ordinal compatibility for each pair of DAs
  (\(w=\overline{1,\nu}\); \(\nu\) corresponds to the best one).
%
%
%
%
%
%

%
 Let \(S\) be a system consisting of \(m\) parts (components):
 \(R(1),...,R(i),...,R(m)\).
 A set of design alternatives
 is generated for each system part above.
 The problem is:

~~

 {\it Find a composite design alternative}
 ~~ \(S=S(1)\star ...\star S(i)\star ...\star S(m)\)~~
 {\it of DAs (one representative design alternative}
 ~\(S(i)\)
 {\it for each system component/part}
  ~\(R(i)\), \(i=\overline{1,m}\)
  {\it )}
 {\it with non-zero compatibility}
 {\it between design alternatives.}

~~

 A discrete ``space'' of the system excellence
 (a poset)
 on the basis of the following vector is used:
 ~~\(N(S)=(w(S);e(S))\),
 ~where \(w(S)\) is the minimum of pairwise compatibility
 between DAs which correspond to different system components
 (i.e.,
 \(~\forall ~R_{j_{1}}\) and \( R_{j_{2}}\),
 \(1 \leq j_{1} \neq j_{2} \leq m\))
 in \(S\),
 ~\(e(S)=(\eta_{1},...,\eta_{\iota},...,\eta_{l})\),
 ~where \(\eta_{\iota}\) is the number of DAs of the \(\iota\)th quality in \(S\).
 Further,
  the problem is described as follows:
%
%
%
%
 \[ \max~ e(S),~~~
%
%
  \max~ w(S),~~~
%
%
 s.t.
%
  ~~w(S) \geq 1  .\]
%
%
 Here,
 composite solutions
 which are nondominated by \(N(S)\)
 (i.e., Pareto-efficient) are searched for.
%
%
 ``Maximization''  of \(e(S)\) is based on the corresponding poset.
%
%
%
 The considered combinatorial problem is NP-hard
 and an enumerative solving scheme is used.
%

%
 Here, combinatorial synthesis is based on usage of multiset
 estimates of design alternatives for system parts.
%
 For the resultant system \(S = S(1) \star ... \star S(i) \star ... \star S(m) \)
 the same type of the multiset estimate is examined:
  an aggregated estimate (``generalized median'')
  of corresponding multiset estimates of its components
 (i.e., selected DAs).
%
%
 Thus, \( N(S) = (w(S);e(S))\), where
 \(e(S)\) is the ``generalized median'' of estimates of the solution
 components.
%
%
%
 Finally, the modified problem is:
%
 \[ \max~ e(S) = M^{g} =
  \arg \min_{M \in D}~~
  \sum_{i=1}^{m} ~ |\delta (M, e(S_{i})) |, ~~~~~
  \max~ w(S),~~~
 s.t.
%
  ~~w(S) \geq 1  .\]
 Here enumeration methods or heuristics are used
 (e.g., \cite{lev06,lev09,lev12morph,lev12multiset,lev15}).
%

~~~

 {\bf Assignment problem} with multiset estimates is formulated
 as follows.
 Estimates of ``profits''/``utilities'' of local assignments
 (i.e., item-position)
  \(\{c_{ij}\}\) can be replaced by multiset
 estimates \(\{e_{ij}\}\).
 Further,
 summarization in objective function can be implemented as
 summarization of multiset estimates or
 by searching for a median estimate
 (\(S\) is an assignment solution for all elements \(i=\overline{1,n}\)):
%
 \[ \max~ e(S) =   \max~ M = ~~~~~
  \arg \min_{M \in D}
 ~~
  \sum_{(i,j) \in S=\{(i,j)| x_{i,j}=1\}} ~ | \delta (M, e_{i,j}) |,\]
 \[s.t.~~ \sum_{i=1}^{m} x_{i,j} \leq 1, j=\overline{1,n};
 ~ \sum_{j=1}^{n} x_{i,j} \leq 1, i=\overline{1,m};
 ~ x_{i,j} \in \{0,1\}, i=\overline{1,m}, j=\overline{1,n}. \]
 In the case of generalized problem (e.g., it is possible to
 assign several items to each position), the problem is
 (i.e., change of constraint for each position \(j\)):
 \[ \max~ e(S) =   \max~ M = ~~~~~
  \arg \min_{M \in D}
 ~~
  \sum_{(i,j) \in S=\{(i,j)| x_{i,j}=1\}} ~ | \delta (M, e_{i,j}) |,\]
 \[s.t.~~ \sum_{i=1}^{m} x_{i,j} \leq b_{j}, j=\overline{1,n};
 ~ \sum_{j=1}^{n} x_{i,j} \leq 1, i=\overline{1,m};
 ~ x_{i,j} \in \{0,1\}, i=\overline{1,m}, j=\overline{1,n}, \]
 here \(b_{j}\) is constraint for number of assigned elements
 for each position \(j\)).
 Clearly, other analogical constraints
 for each positions can be used as well
 (i.e., by other types of resources).
  It is reasonable to use heuristics
  as solving schemes.



\subsubsection{Towards Clustering
  with interval multiset estimates}

 In agglomerative algorithms,
 the basic methodological problem consists in selection/design
 of  proximity measure for objects/clusters.
 Evidently,
 the measure is often based on many parameters
 and it is necessary to use vector proximity/distance.
 Here,
 it may be reasonable to simplify  of the solving procedure
 via transformation of  the vector proximity (proximities)
 into multiset estimate(s).
 Minimization of multiset estimates is a simple process
 (in some complex situations Pareto-efficient point(s) can be
 used).

 Analogical approach can be used in
 k-means clustering method by the usage of interval multiset estimates
 instead of proximity/distance of objects to cluster centroids.
%
%
 In the case of assignment based clustering,
 the above-mentioned model based on multiset estimates
 can be used.

~~~

 {\bf Example 4.1.}
 A simplified numerical example is based on
 data from example 3.2:
   set of 9 end users
 (\(A = \{1,...,i,...,9\}\))~
 (Fig. 4.1)
  and their quantitative estimates
  (Table 3.15, vector estimate \(x_{i},y_{i},z_{i}\)).

\begin{center}
\begin{picture}(68,35.5)
\put(09,00){\makebox(0,0)[bl]{Fig. 4.1.
 Example: 9 items}}


\put(05,22){\circle*{1.5}} \put(03.8,24){\makebox(0,8)[bl]{\(1\)}}
\put(10,25.5){\oval(16,12)}

\put(15,29){\circle*{1.5}} \put(11,28){\makebox(0,8)[bl]{\(2\)}}

\put(08,06){\line(1,0){33}} \put(08,06){\line(0,1){10}}
\put(08,16){\line(1,0){33}} \put(41,06){\line(0,1){10}}

\put(14,08){\circle*{1.5}} \put(10,07){\makebox(0,8)[bl]{\(7\)}}
\put(38,15){\circle*{1.5}} \put(35.5,11){\makebox(0,8)[bl]{\(9\)}}
\put(26,10){\circle*{1.5}} \put(22.5,09){\makebox(0,8)[bl]{\(8\)}}

\put(26,17){\line(-1,1){05}} \put(26,17){\line(2,1){25}}
\put(21,22){\line(2,1){25}} \put(51,29.5){\line(-1,1){05}}

\put(26,21){\circle*{1.5}} \put(28,20){\makebox(0,8)[bl]{\(3\)}}
\put(35,27){\circle*{1.5}} \put(33,23.5){\makebox(0,8)[bl]{\(4\)}}
\put(45,32){\circle*{1.5}} \put(42,29){\makebox(0,8)[bl]{\(5\)}}

\put(58,22){\circle*{1.5}} \put(57,17){\makebox(0,8)[bl]{\(6\)}}
\put(58,20){\oval(6,09)}

\end{picture}
\end{center}

%
%
 Table 4.1 contains pair vector proximity estimates:~
%
 \(\overline{D}(i_{1},i_{2})= (d_{x}(i_{1},i_{2}),d_{y}(i_{1},i_{2}),d_{z} (i_{1},i_{2}))\),
 where

 \(d_{x} (i_{1},i_{2}) = | x_{i_{1}} - x_{i_{2}} |\),
 \(d_{y} (i_{1},i_{2}) = | y_{i_{1}} - y_{i_{2}} |\),
 \(d_{z} (i_{1},i_{2}) = | z_{i_{1}} - z_{i_{2}} |\).

\begin{center}
 {\bf Table 4.1.} Vector quantitative proximity between end users~
 \(\overline{D}(i_{1},i_{2}) \) \\
%
\begin{tabular}{| c |c c c c c c c c|}
\hline
 \(i_{1}\)&\(i_{2}:\) &&&&&&& \\
            & \(2\)&\(3\)&\(4\)&\(5\)&\(6\) &\(7\)&\(8\)&\(9\) \\
\hline
 1&\((28,9,0)\)&\((58,9,5)\)&\((80,4,0)\)&\((115,16,2)\)&\((140,4,0)\)&\((22,31,0)\)&\((56,31,2)\)&\((90,25,1)\)\\
 2&&\((30,18,5)\)&\((52,5,0)\)&\((107,25,3)\)&\((112,13,0)\)&\((6,40,0)\)&\((28,40,2)\)&\((82,34,1)\)\\
 3&&&\((22,13,5)\)&\((58,25,3)\)&\((82,5,5)\)&\((36,22,5)\)&\((2,22,3)\)&\((32,16,6)\)\\
 4&&&&\((35,12,2)\)&\((60,8,0)\)&\((58,35,0)\)&\((24,35,2)\)&\((10,29,1)\)\\
 5&&&&&\((25,20,2)\)&\((93,48,2)\)&\((59,48,0)\)&\((25,41,1)\)\\
 6&&&&&&\((118,27,0)\)&\((84,27,2)\)&\((50,21,1)\)\\
 7&&&&&&&\((34,0,2)\)&\((68,6,1)\)\\
 8&&&&&&&&\((34,6,3)\)\\
\hline
\end{tabular}
\end{center}

 Table 4.2 contains corresponding vector ordinal proximity
 \(\overline{r}(i_{1},i_{2})=
 (r_{x}(i_{1},i_{2}),r_{y}(i_{1},i_{2}),r_{z} (i_{1},i_{2}))\)
 (ordinal scale \([1,2,3]\) is used,
 \(1\) corresponds to close values).
 Ordinal values are calculated as follows
 (for parameter \(x\), for other parameters calculation is
 analogical):
%
%
    \[r_{x}(i_{1},i_{2}) = \left\{ \begin{array}{ll}
               1, & \mbox{if $  ~d^{min}_{x} \leq d_{x}(i_{1},i_{2}) \leq  ~d^{min}_{x}+\frac{\Delta_{x}}{3}, $}\\
               2, & \mbox{if $  ~d^{min}_{x}+\frac{\Delta_{x}}{3} <  d_{x}(i_{1},i_{2})~ \leq  d^{min}_{x}+\frac{2 \Delta_{x}}{3}, $}\\
               3, & \mbox{if $ ~d^{min}_{x}+\frac{2\Delta_{x}}{3} <  d_{x}(i_{1},i_{2})~ \leq d^{min}_{x}+\Delta_{x}~,  $}
               \end{array}
               \right. \]
 where

 \(d^{min}_{x} = \min_{i\in A} d_{i}(x) \),
 \(d^{min}_{y} = \min_{i\in A} d_{i} (y)\),
 \(d^{min}_{z} = \min_{i\in A} d_{i} (z)\);

 \(d^{max}_{x} = \max_{i\in A} d_{i}(x) \),
 \(d^{max}_{y} = \max_{i\in A} d_{i} (y)\),
 \(d^{max}_{z} = \max_{i\in A} d_{i} (z)\);

 \(\Delta_{x} = d^{max}_{x} - d^{min}_{x}\),
 \(\Delta_{y} = d^{max}_{y} - d^{min}_{y}\),
 \(\Delta_{z} = d^{max}_{z} - d^{min}_{z}\).
%

 Table 4.3 contains corresponding multiset estimates.
%
 The used multiset scale (poset) is shown in Fig. 4.2
%
%
 (assessment over scale \([1,3]\) with three elements;
 analogue of poset in Fig. 2.16).

\newpage
\begin{center}
{\bf Table 4.2.} Vector ordinal proximity  between end users ~\(\overline{r}(i_{1},i_{2}) \) \\
\begin{tabular}{| c |c c c c c c c c|}
\hline
 \(i_{1}\)&\(i_{2}:\) &&&&&&& \\
            & \(2\)&\(3\)&\(4\)&\(5\)&\(6\) &\(7\)&\(8\)&\(9\) \\
\hline
 1&\((1,1,1)\)&\((2,1,3)\)&\((2,1,1)\)&\((3,1,1)\)&\((3,1,1)\)&\((1,2,1)\)&\((2,2,1)\)&\((2,2,1)\)\\
 2&&\((1,2,3)\)&\((2,1,1)\)&\((3,2,2)\)&\((3,1,1)\)&\((1,3,1)\)&\((1,3,1)\)&\((2,3,1)\)\\
 3&&&\((1,1,3)\)&\((2,2,2)\)&\((2,1,3)\)&\((1,2,3)\)&\((1,2,2)\)&\((1,1,3)\)\\
 4&&&&\((1,1,1)\)&\((2,3,1)\)&\((2,3,1)\)&\((1,3,1)\)&\((1,2,1)\)\\
 5&&&&&\((3,2,1)\)&\((2,3,1)\)&\((2,3,1)\)&\((1,3,1)\)\\
 6&&&&&&\((3,2,1)\)&\((2,2,1)\)&\((2,2,1)\)\\
 7&&&&&&&\((1,1,1)\)&\((2,1,1)\)\\
 8&&&&&&&&\((1,1,2)\)\\
\hline
\end{tabular}
\end{center}

\begin{center}
{\bf Table 4.3.} Multiset proximity between end users ~\(e(i_{1},i_{2})\) \\
\begin{tabular}{| c |c c c c c c c c|}
\hline
 \(i_{1}\)&\(i_{2}:\) &&&&&&& \\
            & \(2\)&\(3\)&\(4\)&\(5\)&\(6\) &\(7\)&\(8\)&\(9\) \\
\hline
 1&\((3,0,0)\)&\((1,1,1)\)&\((2,1,0)\)&\((2,0,1)\)&\((2,0,1)\)&\((2,1,0)\)&\((1,2,0)\)&\((1,2,0)\)\\
 2&&\((1,1,1)\)&\((2,1,0)\)&\((0,2,1)\)&\((2,0,1)\)&\((2,0,1)\)&\((2,0,1)\)&\((1,1,1)\)\\
 3&&&\((2,0,1)\)&\((0,3,0)\)&\((1,1,1)\)&\((1,1,1)\)&\((1,2,0)\)&\((2,0,1)\)\\
 4&&&&\((3,0,0)\)&\((1,1,1)\)&\((1,1,1)\)&\((2,0,1)\)&\((2,1,0)\)\\
 5&&&&&\((1,1,1)\)&\((1,1,1)\)&\((1,1,1)\)&\((2,0,1)\)\\
 6&&&&&&\((1,1,1)\)&\((1,2,0)\)&\((1,2,0)\)\\
 7&&&&&&&\((3,0,0)\)&\((2,1,0)\)\\
 8&&&&&&&&\((2,1,0)\)\\
\hline
\end{tabular}
\end{center}

\begin{center}
\begin{picture}(58,83)

\put(10,00){\makebox(0,0)[bl] {Fig. 4.2.
  Multiset scale}}
%



\put(12.5,75){\makebox(0,0)[bl]{\((3,0,0)\) }}
\put(18,77){\oval(16,5)} \put(18,77){\oval(16.5,5.5)}


\put(18,70){\line(0,1){4}}
\put(12.5,65){\makebox(0,0)[bl]{\((2,1,0)\)}}
\put(18,67){\oval(16,5)}


\put(18,58){\line(0,1){6}}
\put(12.5,53){\makebox(0,0)[bl]{\((2,0,1)\) }}
\put(18,55){\oval(16,5)}


\put(18,46){\line(0,1){6}}
\put(12.5,41){\makebox(0,0)[bl]{\((1,1,1)\) }}
\put(18,43){\oval(16,5)}

\put(18,34){\line(0,1){6}}
\put(12.5,29){\makebox(0,0)[bl]{\((1,0,2)\) }}
\put(18,31){\oval(16,5)}

\put(18,22){\line(0,1){6}}
\put(12.5,17){\makebox(0,0)[bl]{\((0,1,2)\) }}
\put(18,19){\oval(16,5)}

\put(18,12){\line(0,1){4}}

\put(12.5,07){\makebox(0,0)[bl]{\((0,0,3)\) }}
\put(18,09){\oval(16,5)}

\put(20.5,63.5){\line(3,-1){15}}

\put(35.5,52){\line(-3,-1){15}}

\put(32.5,53){\makebox(0,0)[bl]{\((1,2,0)\) }}
\put(38,55){\oval(16,5)}

\put(38,46){\line(0,1){6}}
\put(32.5,41){\makebox(0,0)[bl]{\((0,3,0)\) }}
\put(38,43){\oval(16,5)}
\put(20.5,39.5){\line(3,-1){15}}

\put(38,34){\line(0,1){6}}
\put(32.5,29){\makebox(0,0)[bl]{\((0,2,1)\) }}
\put(38,31){\oval(16,5)}
\put(35.5,27.5){\line(-3,-1){15}}

\end{picture}
\end{center}

 Application of hierarchical agglomerative balance by cluster size
 (\(\leq 3\)) algorithm
 (while taking into account multiset proximity estimates)
 leads to the following results
 (integration of multiset estimates is implemented as searching  for
 the  median estimate).

 {\it Iteration 1.}
 The smallest pair multiset proximity
 \((3,0,0)\)
  correspond to
 the following item node pairs
 (with next integration of the pairs):
 \((1,2) \Rightarrow J_{1,2}\),
 \((4,5) \Rightarrow J_{4,5}\), and
 \((7,8) \Rightarrow J_{7,8}\)
 (concurrent elements integration).

 {\it Iteration 2.}
 The smallest pair multiset proximity
 \((2,1,0)\)
  correspond to
 the following item node pair
 (with next integration of the pairs):
 \((J_{7,8},9) \Rightarrow J_{7,8,9}\).
 Cluster  \(X_{1} = \{7,8,9 \}\) is obtained.

 {\it Iteration 3.}
 The smallest pair multiset proximity
 \((2,0,1)\)
  correspond to
 the following item node pair
 (with next integration of the pairs):
 \((J_{4,5},3) \Rightarrow J_{3,4,5}\).
 Cluster \(X_{2} = \{3,4,5 \}\) is obtained.

 Finally, the following clustering solution can be considered:

 \(\widehat{X} = \{X_{1}, X_{2}, X_{3}, X_{4}, X_{5} \}\),
 \(X_{1} = \{7,8,9 \}\),
 \(X_{2} = \{3,4,5 \}\),
 \(X_{3} = \{1,2 \}\),
 \(X_{4} = \{6 \}\).

~~~

 Note,
 a numerical example for assignment based clustering
 can be considered analogically
 (e.g., on the basis of data from example 3.2).



\subsection{Restructuring in clustering}

 Restructuring approach in combinatorial optimization has been
 suggested
 by the author in \cite{lev11restr,lev15}.
 In this section, restructuring approach for clustering problems is
 briefly described.
%

\subsubsection{One-stage restructuring}

 Fig. 4.3 and Fig. 4.4 illustrate
 the restructuring process (one-stage framework)
 \cite{lev11restr,lev15}.
 Restructuring in clustering problem is depicted in Fig. 4.5.

\begin{center}
\begin{picture}(100,52)
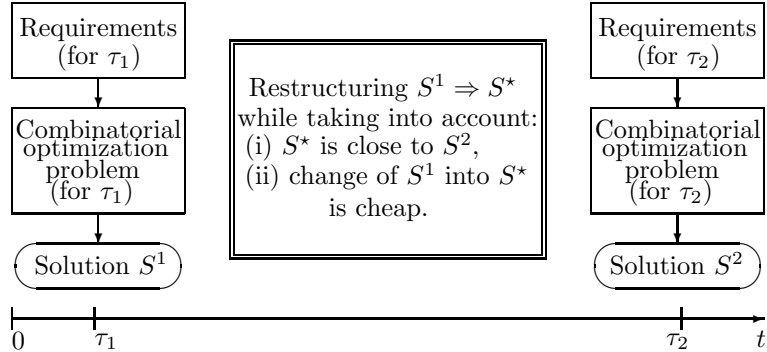
\put(09,00){\makebox(0,0)[bl]{Fig. 4.3. Framework of
  restructuring process
 \cite{lev11restr,lev15}
  }}

\put(00,09){\vector(1,0){100}}

\put(00,7.5){\line(0,1){3}} \put(11,7.5){\line(0,1){3}}
\put(89,7.5){\line(0,1){3}}

\put(00,05){\makebox(0,0)[bl]{\(0\)}}
\put(11,05){\makebox(0,0)[bl]{\(\tau_{1}\)}}
\put(87,05){\makebox(0,0)[bl]{\(\tau_{2}\)}}

\put(99,05.3){\makebox(0,0)[bl]{\(t\)}}


\put(00,41){\line(1,0){23}} \put(00,51){\line(1,0){23}}
\put(00,41){\line(0,1){10}} \put(23,41){\line(0,1){10}}

\put(1,46){\makebox(0,0)[bl]{Requirements}}
\put(6,42){\makebox(0,0)[bl]{(for \(\tau_{1}\))}}


\put(11.5,41){\vector(0,-1){4}}

\put(00,23){\line(1,0){23}} \put(00,37){\line(1,0){23}}
\put(00,23){\line(0,1){14}} \put(23,23){\line(0,1){14}}

\put(0.5,33){\makebox(0,0)[bl]{Combinatorial}}
\put(1.5,30){\makebox(0,0)[bl]{optimization}}
\put(4.5,27){\makebox(0,0)[bl]{problem}}
\put(5,24){\makebox(0,0)[bl]{(for \(\tau_{1})\)}}


\put(11.5,23){\vector(0,-1){4}}

\put(11.5,16){\oval(22,06)}

\put(03,15){\makebox(0,0)[bl]{Solution \(S^{1}\)}}


\put(29,17){\line(1,0){42}} \put(29,46){\line(1,0){42}}
\put(29,17){\line(0,1){29}} \put(71,17){\line(0,1){29}}

\put(29.5,17.5){\line(1,0){41}} \put(29.5,45.5){\line(1,0){41}}
\put(29.5,17.5){\line(0,1){28}} \put(70.5,17.5){\line(0,1){28}}

\put(31.5,38){\makebox(0,0)[bl]{Restructuring
 \(S^{1} \Rightarrow S^{\star } \)}}

\put(30,34){\makebox(0,0)[bl]{while taking into account:}}

\put(31,30){\makebox(0,0)[bl]{(i) \(S^{\star }\) is close to
 \(S^{2}\),}}

\put(31,26){\makebox(0,0)[bl]{(ii) change of \(S^{1}\)
 into \(S^{\star }\)}}

\put(41,22){\makebox(0,0)[bl]{ is cheap.}}


\put(77,41){\line(1,0){23}} \put(77,51){\line(1,0){23}}
\put(77,41){\line(0,1){10}} \put(100,41){\line(0,1){10}}

\put(78,46){\makebox(0,0)[bl]{Requirements}}
\put(83,42){\makebox(0,0)[bl]{(for \(\tau_{2}\))}}


\put(88.5,41){\vector(0,-1){4}}

\put(77,23){\line(1,0){23}} \put(77,37){\line(1,0){23}}
\put(77,23){\line(0,1){14}} \put(100,23){\line(0,1){14}}

\put(77.5,33){\makebox(0,0)[bl]{Combinatorial}}
\put(78.5,30){\makebox(0,0)[bl]{optimization}}
\put(81.5,27){\makebox(0,0)[bl]{problem }}
\put(82,24){\makebox(0,0)[bl]{(for \(\tau_{2})\)}}


\put(88.5,23){\vector(0,-1){4}}

\put(88.5,16){\oval(22,06)}

\put(79.5,15){\makebox(0,0)[bl]{Solution \(S^{2}\)}}


\end{picture}
\end{center}

\begin{center}
\begin{picture}(73,64)
\put(04,00){\makebox(0,0)[bl]{Fig. 4.4.
 Restructuring scheme \cite{lev11restr,lev15} }}

\put(00,05){\vector(0,1){51.5}} \put(00,05){\vector(1,0){69}}

\put(70,05){\makebox(0,0)[bl]{\(t\)}}

\put(00,60){\makebox(0,0)[bl]{Solution}}
\put(00,57){\makebox(0,0)[bl]{quality}}


\put(0.6,15){\makebox(0,0)[bl]{\(S^{1}\)}}
\put(5,15){\circle{1.7}}

\put(12,16.5){\makebox(0,0)[bl]{Initial}}
\put(12,13.5){\makebox(0,0)[bl]{solution}}
\put(12,10){\makebox(0,0)[bl]{(clustering)}}
\put(12,06){\makebox(0,0)[bl]{(\(t=\tau^{1}\))}}

\put(11,10){\line(-4,3){5}}

\put(6,16){\vector(1,1){23}}


\put(40,45){\circle*{2.7}}

\put(54,50.5){\makebox(0,0)[bl]{Goal}}
\put(54,47.5){\makebox(0,0)[bl]{solution}}
\put(54,44){\makebox(0,0)[bl]{(clustering)}}
\put(54,40){\makebox(0,0)[bl]{(\(t=\tau^{2}\)): \(S^{2}\)}}

\put(53,45){\line(-1,0){10.5}}

\put(40,45){\oval(12,10)} \put(40,45){\oval(17,17)}
\put(40,45){\oval(24,22)}


\put(40,45){\vector(-2,-1){9}} \put(30,40){\vector(2,1){9}}



\put(26,27){\makebox(0,0)[bl]{Proximity}}
\put(26,23){\makebox(0,0)[bl]{~\(\rho (S^{\star },S^{2})\)}}

\put(36,30){\line(0,1){12}}


\put(46,29){\makebox(0,0)[bl]{Neighborhoods }}
\put(49,26){\makebox(0,0)[bl]{of ~\(S^{2}\)}}

\put(56,32){\line(-1,1){10}}

\put(53,32){\line(-2,1){8}}



\put(30,40){\circle{2}} \put(30,40){\circle*{1}}

\put(10,53){\makebox(0,0)[bl]{Obtained}}
\put(10,50){\makebox(0,0)[bl]{solution}}
\put(10,46.5){\makebox(0,0)[bl]{(clustering)}}
\put(10,43.5){\makebox(0,0)[bl]{ ~~~~~ \(S^{\star }\)}}

\put(20,42.8){\line(4,-1){7}}

\put(1,37.5){\makebox(0,0)[bl]{Solution }}
\put(1,33.5){\makebox(0,0)[bl]{change cost }}
\put(1,29.5){\makebox(0,0)[bl]{\(H(S^{1}
 \rightarrow S^{\star })\)}}

\put(8,29){\line(1,-1){5}}
\end{picture}
\end{center}

 Let \(P\) be a combinatorial optimization problem with a solution as
 structure
 \(S\)
 (i.e., subset, graph),
 \(\Omega\) be initial data (elements, element parameters, etc.),
 \(f(P)\) be objective function(s).
 Thus \(S(\Omega)\) be a solution for initial data \(\Omega\),
 \(f(S(\Omega))\) be the corresponding objective function.
 Let \(\Omega^{1}\) be initial data at an initial stage,
  \(f(S(\Omega^{1}))\) be the corresponding objective function.
 \(\Omega^{2}\) be initial data at next stage,
  \(f(S(\Omega^{2}))\) be the corresponding objective function.
%
%
 As a result,
 the following solutions can be considered:
 ~(a) \( S^{1}=S(\Omega^{1})\) with \(f(S(\Omega^{1}))\) and
 ~(b) \( S^{2}=S(\Omega^{2})\) with \(f(S(\Omega^{2}))\).

 In addition it is reasonable to examine a cost of changing
 a solution into another one:~
 \( H(S^{\alpha} \rightarrow  S^{\beta})\).
 Let \(\rho ( S^{\alpha}, S^{\beta} )\) be a proximity between solutions
  \( S^{\alpha}\) and \( S^{\beta}\),
  for example,
 \(\rho ( S^{\alpha}, S^{\beta} ) = | f(S^{\alpha}) -  f(S^{\beta}) |\).
 Clearly, function \(f(S)\) can be a vector function.
 Thus, the following version of  restructuring problem is considered:

~~

 Find a solution \( S^{\star }\) while taking into account the
 following:

  (i) \( H(S^{1} \rightarrow  S^{\star }) \rightarrow \min \),
%
  ~(ii) \(\rho ( S^{\star }, S^{2} )  \rightarrow \min  \) ~(or constraint).


\begin{center}
\begin{picture}(80,56)
\put(05,00){\makebox(0,0)[bl]{Fig. 4.5. Restructuring in
 clustering problem}}

\put(00,10){\vector(1,0){80}}

\put(00,8.5){\line(0,1){3}} \put(8,8.5){\line(0,1){3}}
\put(72,8.5){\line(0,1){3}}

\put(00,06){\makebox(0,0)[bl]{\(0\)}}
\put(8,06){\makebox(0,0)[bl]{\(\tau_{1}\)}}
\put(70,06){\makebox(0,0)[bl]{\(\tau_{2}\)}}


\put(79,06.3){\makebox(0,0)[bl]{\(t\)}}






\put(00,28){\line(1,0){17}} \put(00,43){\line(1,0){17}}
\put(00,28){\line(0,1){15}} \put(17,28){\line(0,1){15}}

\put(0.5,38){\makebox(0,0)[bl]{Clustering}}
\put(2.0,34){\makebox(0,0)[bl]{problem}}
\put(2.5,30){\makebox(0,0)[bl]{(\(t=\tau_{1})\)}}


\put(8,28){\vector(0,-1){4}}

\put(8,19){\oval(16,10)}

\put(02,20){\makebox(0,0)[bl]{Solution}}

\put(07,16){\makebox(0,0)[bl]{\(\widehat{X}^{1}\)}}


\put(19,14){\line(1,0){42}} \put(19,43){\line(1,0){42}}
\put(19,14){\line(0,1){29}} \put(61,14){\line(0,1){29}}

\put(19.5,14.5){\line(1,0){41}} \put(19.5,42.5){\line(1,0){41}}
\put(19.5,14.5){\line(0,1){28}} \put(60.5,14.5){\line(0,1){28}}


\put(20.4,38){\makebox(0,0)[bl]{Restructuring
 (\(S^{1} \Rightarrow S^{\star }\)):}}

\put(21,34){\makebox(0,0)[bl]{1. Change of cluster set }}
\put(25,31){\makebox(0,0)[bl]{(if needed) }}

\put(21,28){\makebox(0,0)[bl]{2. Element reassignment: }}
\put(21,25){\makebox(0,0)[bl]{(ii) deletion of some}}
\put(21,22){\makebox(0,0)[bl]{elements from clusters,}}
\put(21,19){\makebox(0,0)[bl]{(iii) addition of some}}
\put(21,16.5){\makebox(0,0)[bl]{elements into clusters}}


\put(39.5,51){\oval(76,08)}

\put(05,49){\makebox(0,0)[bl]{Initial set of elements \(A =
 \{A_{1},...,A_{i},...,A_{n}\}\)}}


\put(8,47){\vector(0,-1){4}} \put(39.5,47){\vector(0,-1){4}}
\put(72,47){\vector(0,-1){4}}







\put(63.5,28){\line(1,0){16.5}} \put(63.5,43){\line(1,0){16.5}}
\put(63.5,28){\line(0,1){15}} \put(80,28){\line(0,1){15}}

\put(64,38){\makebox(0,0)[bl]{Clustering}}
\put(65.5,34){\makebox(0,0)[bl]{problem }}
\put(66,30){\makebox(0,0)[bl]{(\(t=\tau_{2})\)}}


\put(72,28){\vector(0,-1){4}}

\put(72,19){\oval(16,10)}

\put(66,20){\makebox(0,0)[bl]{Solution}}

\put(71,16){\makebox(0,0)[bl]{\(\widehat{X}^{2}\)}}

\end{picture}
\end{center}

 The basic optimization model can be examined as the following:

%
   \[\min \rho ( S^{\star }, S^{2} ) ~~~s.t.
 ~~ H(S^{1} \rightarrow  S^{\star })  \leq \widehat{h}, \]
 where \(\widehat{h}\) is a constraint for cost of the solution
 change.
%
%
 In a simple case, this problem can be
 formulated as knapsack problem for selection of a subset of
 change operations \cite{lev11restr,lev15}:
%
%
 \[\max\sum_{i=1}^{n} c^{1}_{i} x_{i}
 ~~~ s.t.~ \sum_{i=1}^{n} a^{1}_{i} x_{i} ~\leq~ b^{1},
 ~~~ x_{i} \in \{0,1\}.  \]
%
%
%
%
%
%
 In the case of interconnections between change operations,
 it is reasonable to consider combinatorial synthesis problem
 (i.e., while taking into account compatibility between the operations).

~~~

 {\bf  Example 4.2.}
 Initial information involves the following:

 (i) set of elements  \(A = \{1,2,3,4,5,6,7,8,9\}\);

 (ii) initial solution 1 (\(t=\tau_{1}\)):   \(\widehat{X}^{1} \ \{X^{1}_{1},X^{1}_{2},X^{1}_{3}\}\),
  clusters
 \(X^{1}_{1} = \{1,3,8\}\),
 \(X^{1}_{2} = \{2,4,7\}\),
 \(X^{1}_{3} = \{5,6,9\}\);

 (iii)  solution 2 (\(t=\tau_{2}\)):
  \(\widehat{X}^{2} = \{X^{2}_{1},X^{1}_{2},X^{2}_{3}\}\),
  clusters
 \(X^{2}_{1} = \{2,3\}\),
 \(X^{2}_{2} = \{5,7,8\}\),
 \(X^{2}_{3} = \{1,4,6,9\}\);

 (v) general set of considered possible change operations
 (each element can be replaced,
 the number of solution clusters is not changed):

 \(O_{11}\): none,
 \(O_{12}\):
 deletion of element \(1\) from  cluster \(X^{1}\),
 addition of element \(1\) into cluster \(X^{2}\),
 \(O_{13}\):
 deletion of element \(1\) from  cluster \(X^{1}\),
 addition of element \(1\) into cluster \(X^{3}\);

  \(O_{21}\): none,
  \(O_{22}\):
 deletion of element \(2\) from  cluster \(X^{2}\),
 addition of element \(2\) into cluster \(X^{1}\),
  \(O_{23}\):
 deletion of element \(2\) from  cluster \(X^{2}\),
 addition of element \(2\) into cluster \(X^{3}\);

  \(O_{31}\): none,
 \(O_{32}\):
 deletion of element \(3\) from cluster \(X^{1}\),
 addition of element \(3\) into cluster \(X^{2}\);
  \(O_{33}\):
 deletion of element \(3\) from cluster \(X^{1}\),
 addition of element \(3\) into cluster \(X^{3}\);

  \(O_{41}\): none,
 \(O_{42}\):
 deletion of element \(4\) from cluster \(X^{2}\),
 addition of element \(4\) into cluster \(X^{1}\),
 \(O_{43}\):
 deletion of element \(4\) from cluster \(X^{2}\),
 addition of element \(4\) into cluster \(X^{3}\);

  \(O_{51}\): none,
 \(O_{52}\):
 deletion of element \(5\) from cluster \(X^{3}\),
 addition of element \(5\) into cluster \(X^{1}\),
 \(O_{53}\):
 deletion of element \(5\) from cluster \(X^{3}\),
 addition of element \(5\) into cluster \(X^{2}\);

  \(O_{61}\): none,
 \(O_{62}\):
 deletion of element \(6\) from cluster \(X^{3}\),
 addition of element \(6\) into cluster \(X^{1}\),
 \(O_{63}\):
 deletion of element \(6\) from cluster \(X^{3}\),
 addition of element \(6\) into cluster \(X^{2}\);

  \(O_{71}\): none,
 \(O_{72}\):
 deletion of element \(7\) from cluster \(X^{2}\),
 addition of element \(7\) into cluster \(X^{1}\),
 \(O_{73}\):
 deletion of element \(7\) from cluster \(X^{2}\),
 addition of element \(7\) into cluster \(X^{3}\);

  \(O_{81}\): none,
 \(O_{82}\):
 deletion of element \(8\) from cluster \(X^{1}\),
 addition of element \(8\) into cluster \(X^{2}\),
 \(O_{83}\):
 deletion of element \(8\) from cluster \(X^{1}\),
 addition of element \(8\) into cluster \(X^{3}\);

  \(O_{91}\): none,
 \(O_{92}\):
 deletion of element \(9\) from cluster \(X^{3}\),
 addition of element \(9\) into cluster \(X^{1}\),
 \(O_{93}\):
 deletion of element \(9\) from cluster \(X^{3}\),
 addition of element \(9\) into cluster \(X^{2}\).

 In this case, optimization model (multiple choice problem) is:
%
%
%
 \[\max ~\sum_{i=1}^{n} ~\sum_{j=1}^{3}   c(O_{ij}) x_{ij}
 ~~~ s.t.~ \sum_{i=1}^{n}  ~\sum_{j=1}^{3}  a(O_{ij}) x_{ij}  ~\leq~ b,
 ~~~ x_{ij} \in \{0,1\},  \]
 where
 \(a(O_{ij}) \) is the cost of operation \(O_{ij}\),
 \(c(O_{ij}) \) is a ``local'' profit of operation \(O_{ij}\)
  as influence on closeness of obtained solution \(X^{\star}\)
   to clustering solution \(X^{2}\).
 Generally, it is necessary to examine quality
 parameters of clustering solution as basis for
 objective function(s).

 Evidently, the compressed
 set of change operations can be analyzed:

 \(O_{1}\):
 deletion of element \(1\) from cluster \(X^{1}\),
 addition of element \(1\) into cluster \(X^{3}\);

 \(O_{2}\):
 deletion of element \(2\) from cluster \(X^{2}\),
 addition of element \(2\) into cluster \(X^{1}\);

 \(O_{3}\):
 deletion of element \(4\) from cluster \(X^{2}\),
 addition of element \(4\) into cluster \(X^{3}\);

 \(O_{4}\):
 deletion of element \(5\) from cluster \(X^{3}\),
 addition of element \(5\) into cluster \(X^{2}\);

 \(O_{5}\):
 deletion of element \(8\) from cluster \(X^{1}\),
 addition of element \(8\) into cluster \(X^{2}\).

 In this case,
 optimization model is knapsack problem:
%
%
 \[\max ~\sum_{j=1}^{9}    c(O_{j}) x_{j}
 ~~~ s.t.~ \sum_{j=1}^{9}    a(O_{j}) x_{j}  ~\leq~ b,
 ~~~ x_{j} \in \{0,1\},  \]
 where
 \(a(O_{j}) \) is the cost of operation \(O_{j}\),
 \(c(O_{j}) \) is a ``local'' profit of operation \(O_{j}\)
  as influence on closeness of obtained solution \(X^{\star}\)
   to clustering solution \(X^{2}\).

 Finally, let us point out
 an illustrative example of clustering solution
 (Fig. 4.6):

  \(\widehat{X}^{\star} \ \{X^{\star}_{1},X^{\star}_{2},X^{\star}_{3}\}\),
  clusters
 \(X^{\star}_{1} = \{1,2,3\}\),
 \(X^{\star}_{2} = \{7,8\}\),
 \(X^{\star}_{3} = \{4,5,6,9\}\).

\begin{center}
\begin{picture}(34,45)
\put(05.5,00){\makebox(0,0)[bl]{Fig. 4.6.
 Example:  restructuring of clustering solution}}

\put(10,40){\makebox(0,0)[bl]{\(\widehat{X}^{1}\)}}
\put(12,22){\oval(24,34)}

\put(04,33.5){\makebox(0,0)[bl]{\(X^{1}_{1}\)}}
\put(06,25){\oval(08,15)}

\put(07,20){\circle*{1.3}} \put(04,19){\makebox(0,8)[bl]{\(8\)}}
\put(07,25){\circle*{1.3}} \put(04,24){\makebox(0,8)[bl]{\(3\)}}
\put(07,30){\circle*{1.3}} \put(04,29){\makebox(0,8)[bl]{\(1\)}}

\put(16,33.5){\makebox(0,0)[bl]{\(X^{1}_{2}\)}}
\put(18,25){\oval(08,15)}

\put(20,20){\circle*{1.0}} \put(16,19){\makebox(0,8)[bl]{\(7\)}}
\put(20,25){\circle*{1.0}} \put(16,24){\makebox(0,8)[bl]{\(4\)}}
\put(20,30){\circle*{1.0}} \put(16,29){\makebox(0,8)[bl]{\(2\)}}

\put(10,05.5){\makebox(0,0)[bl]{\(X^{1}_{3}\)}}
\put(12,13){\oval(20,6)}

\put(06,12){\circle*{0.7}} \put(06,12){\circle{1.4}}
\put(05,13){\makebox(0,8)[bl]{\(5\)}}

\put(12,12){\circle*{0.7}} \put(12,12){\circle{1.4}}
\put(11,13){\makebox(0,8)[bl]{\(6\)}}

\put(18,12){\circle*{0.7}} \put(18,12){\circle{1.4}}
\put(17,13){\makebox(0,8)[bl]{\(9\)}}


\put(25,28){\vector(1,0){8}} \put(25,22){\vector(1,0){8}}
\put(25,16){\vector(1,0){8}}

\end{picture}
%
\begin{picture}(34,45)

\put(10,40){\makebox(0,0)[bl]{\(\widehat{X}^{\star}\)}}
\put(12,22){\oval(24,34)}

\put(04,33.5){\makebox(0,0)[bl]{\(X^{\star}_{1}\)}}
\put(06,25){\oval(08,15)}

\put(07,20){\circle*{1.3}} \put(04,19){\makebox(0,8)[bl]{\(3\)}}
\put(07,25){\circle*{1.0}} \put(04,24){\makebox(0,8)[bl]{\(2\)}}
\put(07,30){\circle*{1.3}} \put(04,29){\makebox(0,8)[bl]{\(1\)}}

\put(16,33.5){\makebox(0,0)[bl]{\(X^{\star}_{2}\)}}
\put(18,25){\oval(08,15)}

\put(20,20){\circle*{1.0}} \put(16,19){\makebox(0,8)[bl]{\(7\)}}
\put(20,25){\circle*{1.3}} \put(16,24){\makebox(0,8)[bl]{\(8\)}}


\put(10,05.5){\makebox(0,0)[bl]{\(X^{\star}_{3}\)}}
\put(12,13){\oval(20,6)}

\put(04,12){\circle*{1.0}} \put(03,13){\makebox(0,8)[bl]{\(4\)}}

\put(09,12){\circle*{0.7}} \put(09,12){\circle{1.4}}
\put(08,13){\makebox(0,8)[bl]{\(5\)}}

\put(14,12){\circle*{0.7}} \put(14,12){\circle{1.4}}
 \put(13,13){\makebox(0,8)[bl]{\(6\)}}

\put(19,12){\circle*{0.7}} \put(19,12){\circle{1.4}}
\put(18,13){\makebox(0,8)[bl]{\(9\)}}



\put(28,27){\makebox(0,0)[bl]{\(\sim\)}}
\put(28,21){\makebox(0,0)[bl]{\(\sim\)}}
\put(28,14){\makebox(0,0)[bl]{\(\sim\)}}

\end{picture}
%
\begin{picture}(24,45)

\put(10,40){\makebox(0,0)[bl]{\(\widehat{X}^{2}\)}}
\put(12,22){\oval(24,34)}

\put(04,33.5){\makebox(0,0)[bl]{\(X^{2}_{1}\)}}
\put(06,25){\oval(08,15)}

\put(07,25){\circle*{1.3}} \put(04,24){\makebox(0,8)[bl]{\(3\)}}
\put(07,30){\circle*{1.0}} \put(04,29){\makebox(0,8)[bl]{\(2\)}}

\put(16,33.5){\makebox(0,0)[bl]{\(X^{2}_{2}\)}}
\put(18,25){\oval(08,15)}

\put(20,20){\circle*{1.0}} \put(16,19){\makebox(0,8)[bl]{\(7\)}}
\put(20,25){\circle*{1.3}} \put(16,24){\makebox(0,8)[bl]{\(8\)}}

\put(20,30){\circle*{0.7}} \put(20,30){\circle{1.4}}
\put(16,29){\makebox(0,8)[bl]{\(5\)}}

\put(10,05.5){\makebox(0,0)[bl]{\(X^{2}_{3}\)}}
\put(12,13){\oval(20,6)}

\put(04,12){\circle*{1.3}} \put(03,13){\makebox(0,8)[bl]{\(1\)}}

\put(09,12){\circle*{1.0}} \put(08,13){\makebox(0,8)[bl]{\(4\)}}

\put(14,12){\circle*{0.7}} \put(14,12){\circle{1.4}}
 \put(13,13){\makebox(0,8)[bl]{\(6\)}}

\put(19,12){\circle*{0.7}} \put(19,12){\circle{1.4}}
\put(18,13){\makebox(0,8)[bl]{\(9\)}}



\end{picture}
\end{center}

\subsubsection{Multistage restructuring,
 cluster/element trajectories}

 This kind of clustering (or classification) model/problem is close to multistage system design
  \cite{fortun10,hopcroft04,lev12clique,lev13tra,lev15}.
 Fig. 4.7 and Fig. 4.8 illustrate
 multistage classification and multistage clustering problems:

\begin{center}
\begin{picture}(150,71)
\put(34,00){\makebox(0,0)[bl]{Fig. 4.7. Illustration of multistage
 classification}}

\put(00,10){\vector(1,0){150}}

\put(00,8.5){\line(0,1){3}} \put(15,8.5){\line(0,1){3}}
\put(55,8.5){\line(0,1){3}} \put(95,8.5){\line(0,1){3}}
\put(135,8.5){\line(0,1){3}}

\put(15,06){\makebox(0,0)[bl]{\(0\)}}
\put(55,06){\makebox(0,0)[bl]{\(\tau_{1}\)}}
\put(95,06){\makebox(0,0)[bl]{\(\tau_{2}\)}}
\put(135,06){\makebox(0,0)[bl]{\(\tau_{3}\)}}

\put(148,06.3){\makebox(0,0)[bl]{\(t\)}}


\put(01.5,66){\makebox(0,0)[bl]{Initial element set}}

\put(15,40){\oval(30,50)}

\put(15,40){\oval(25,44)} \put(15,40){\oval(24,43)}

\put(15,57){\circle*{0.8}}  \put(15,57){\circle{1.4}}

\put(12,56){\makebox(0,0)[bl]{\(1\)}}

\put(15,45){\circle*{1.2}}

\put(12,44){\makebox(0,0)[bl]{\(2\)}}

\put(15,35){\circle*{0.7}}  \put(15,35){\circle{1.1}}
\put(15,35){\circle{1.7}}

\put(12,34){\makebox(0,0)[bl]{\(3\)}}



\put(32,38){\makebox(0,0)[bl]{\(\Longrightarrow \)}}
\put(72,38){\makebox(0,0)[bl]{\(\Longrightarrow \)}}
\put(112,38){\makebox(0,0)[bl]{\(\Longrightarrow \)}}


\put(41,66){\makebox(0,0)[bl]{Classification for \(\tau_{1}\)}}

\put(55,40){\oval(30,50)}

\put(55,57){\circle*{0.8}}  \put(55,57){\circle{1.4}}

\put(55,45){\circle*{1.2}}

\put(55,35){\circle*{0.7}}  \put(55,35){\circle{1.1}}
\put(55,35){\circle{1.7}}

\put(43,25){\makebox(0,0)[bl]{\(L^{4}\)}}
\put(43,35){\makebox(0,0)[bl]{\(L^{3}\)}}
\put(43,45){\makebox(0,0)[bl]{\(L^{2}\)}}
\put(43,55){\makebox(0,0)[bl]{\(L^{1}\)}}

\put(55,25){\oval(10,08)} \put(55,35){\oval(16,08)}
\put(55,45){\oval(12,08)} \put(55,55){\oval(09,08)}


\put(81,66){\makebox(0,0)[bl]{Classification for \(\tau_{2}\)}}

\put(95,40){\oval(30,50)}

\put(95,57){\circle*{0.8}}  \put(95,57){\circle{1.4}}

\put(95,53){\circle*{1.2}}

\put(95,25){\circle*{0.7}}  \put(95,25){\circle{1.1}}
\put(95,25){\circle{1.7}}

\put(95,25){\oval(10,08)} \put(95,35){\oval(16,08)}
\put(95,45){\oval(12,08)} \put(95,55){\oval(09,08)}


\put(121,66){\makebox(0,0)[bl]{Classification for \(\tau_{3}\)}}

\put(135,40){\oval(30,50)}

\put(135,57){\circle*{0.8}}  \put(135,57){\circle{1.4}}

\put(135,45){\circle*{1.2}}

\put(135,35){\circle*{0.7}}  \put(135,35){\circle{1.1}}
\put(135,35){\circle{1.7}}

\put(135,25){\oval(10,08)} \put(135,35){\oval(16,08)}
\put(135,45){\oval(12,08)} \put(135,55){\oval(09,08)}

\end{picture}
\end{center}

 {\it 1.} Multistage classification (Fig. 4.7):
 the same set of classes at each time stage (here: four classes \(L^{1}\), \(L^{2}\), \(L^{3}\), \(L^{4}\)),
 elements can belong to different classes at each stage.
 Here: elements \(1\), \(2\), \(3\);
  trajectory for element \(1\): \(J(1) = <L^{1},L^{1},L^{1}>\),
  trajectory for element \(2\): \(J(2) = <L^{2},L^{1},L^{2}>\),
  trajectory for element \(3\): \(J(3) = <L^{3},L^{4},L^{3}>\)).

 {\it 2.} Multistage clustering (Fig. 4.8):
 different set of clusters at each time stage can be examined,
  elements can belong to different clusters at each stage.
 Here: elements \(1\), \(2\), \(3\);
  trajectory for element \(1\): \(J(1) = <L^{1}_{1},L^{1}_{2},L^{1}_{3}>\),
  trajectory for element \(2\): \(J(2) = <L^{2}_{1},L^{1}_{2},L^{2}_{3}>\),
  trajectory for element \(3\): \(J(3) = <L^{3}_{1},L^{4}_{2},L^{5}_{3}>\)).

\begin{center}
\begin{picture}(150,71)
\put(36,00){\makebox(0,0)[bl]{Fig. 4.8. Illustration of multistage
 clustering}}

\put(00,10){\vector(1,0){150}}

\put(00,8.5){\line(0,1){3}} \put(15,8.5){\line(0,1){3}}
\put(55,8.5){\line(0,1){3}} \put(95,8.5){\line(0,1){3}}
\put(135,8.5){\line(0,1){3}}

\put(15,06){\makebox(0,0)[bl]{\(0\)}}
\put(55,06){\makebox(0,0)[bl]{\(\tau_{1}\)}}
\put(95,06){\makebox(0,0)[bl]{\(\tau_{2}\)}}
\put(135,06){\makebox(0,0)[bl]{\(\tau_{3}\)}}

\put(148,06.3){\makebox(0,0)[bl]{\(t\)}}


\put(01.5,66){\makebox(0,0)[bl]{Initial element set}}

\put(15,40){\oval(30,50)}

\put(15,40){\oval(25,44)} \put(15,40){\oval(24,43)}

\put(15,57){\circle*{0.8}}  \put(15,57){\circle{1.4}}

\put(12,56){\makebox(0,0)[bl]{\(1\)}}

\put(15,45){\circle*{1.2}}

\put(12,44){\makebox(0,0)[bl]{\(2\)}}

\put(15,35){\circle*{0.7}}  \put(15,35){\circle{1.1}}
\put(15,35){\circle{1.7}}

\put(12,34){\makebox(0,0)[bl]{\(3\)}}




\put(32,38){\makebox(0,0)[bl]{\(\Longrightarrow \)}}
\put(72,38){\makebox(0,0)[bl]{\(\Longrightarrow \)}}
\put(112,38){\makebox(0,0)[bl]{\(\Longrightarrow \)}}


\put(42.5,66){\makebox(0,0)[bl]{Clustering for \(\tau_{1}\)}}

\put(55,40){\oval(30,50)}

\put(43,25){\makebox(0,0)[bl]{\(L^{4}_{1}\)}}
\put(43,35){\makebox(0,0)[bl]{\(L^{3}_{1}\)}}
\put(43,45){\makebox(0,0)[bl]{\(L^{2}_{1}\)}}
\put(43,55){\makebox(0,0)[bl]{\(L^{1}_{1}\)}}

\put(55,25){\oval(10,08)} \put(55,35){\oval(16,08)}
\put(55,45){\oval(12,08)} \put(55,55){\oval(09,08)}

\put(55,57){\circle*{0.8}}  \put(55,57){\circle{1.4}}

\put(55,45){\circle*{1.2}}

\put(55,35){\circle*{0.7}}  \put(55,35){\circle{1.1}}
\put(55,35){\circle{1.7}}


\put(82.5,66){\makebox(0,0)[bl]{Clustering for \(\tau_{2}\)}}

\put(95,40){\oval(30,50)}

\put(83,25){\makebox(0,0)[bl]{\(L^{3}_{2}\)}}
\put(83,45){\makebox(0,0)[bl]{\(L^{2}_{2}\)}}
\put(83,55){\makebox(0,0)[bl]{\(L^{1}_{2}\)}}

\put(95,25){\oval(10,08)} \put(95,40){\oval(16,18)}


\put(95,55){\oval(09,08)}

\put(95,57){\circle*{0.8}}  \put(95,57){\circle{1.4}}

\put(95,53){\circle*{1.2}}

\put(95,25){\circle*{0.7}}  \put(95,25){\circle{1.1}}
\put(95,25){\circle{1.7}}


\put(122.5,66){\makebox(0,0)[bl]{Clustering for \(\tau_{3}\)}}

\put(135,40){\oval(30,50)}

\put(123,20){\makebox(0,0)[bl]{\(L^{5}_{3}\)}}
\put(123,30){\makebox(0,0)[bl]{\(L^{4}_{3}\)}}
\put(123,40){\makebox(0,0)[bl]{\(L^{3}_{3}\)}}
\put(123,50){\makebox(0,0)[bl]{\(L^{2}_{3}\)}}
\put(123,60){\makebox(0,0)[bl]{\(L^{1}_{3}\)}}

\put(135,20){\oval(10,08)} \put(135,30){\oval(16,08)}
\put(135,40){\oval(12,08)} \put(135,50){\oval(14,08)}
\put(135,60){\oval(09,08)}

\put(135,60){\circle*{0.8}}  \put(135,60){\circle{1.4}}

\put(135,50){\circle*{1.2}}

\put(135,20){\circle*{0.7}}  \put(135,20){\circle{1.1}}
\put(135,20){\circle{1.7}}

\end{picture}
\end{center}

 In this problem,
 it is necessary to examine
 a set of change trajectories for each element.
 As a result,
 multi-stage restructuring problem
 has to be based on
 multiple choice model.
 Generally, this problem is very prospective.

\subsubsection{Restructuring in sorting}

 One-stage restructuring for sorting problem
 can be considered as well.
 Let \(A = \{A_{1},...,A_{i},...,A_{n}\}\) be a initial element
 set.
 Solution is a result of dividing set \{A\}
 into \(k\) linear ordered subsets (ranking):
 \(\widehat{R} = \{R_{1},...,R_{j},...,R_{k} \}\),
 \(R_{j} \subseteq A ~ \forall j=\overline{1,k} \),
 \(|R_{j_{1}} \& R_{j_{2}}|=0 ~ \forall j_{1},j_{2} \).
  Linear order is:
  \(R_{1} \rightarrow ... \rightarrow R_{j} \rightarrow ... \rightarrow  R_{k}\),
 \( A_{i_{1}} \rightarrow A_{i_{2}} \) if
 \(A_{i_{1}} \in R_{j_{1}}\), \(A_{i_{2}} \in R_{j_{2}}\),
 \( j_{1} < j_{2}  \).

 Generally, the sorting problem (or multicriteria ranking)
 consists in transformation of set \(A \)
 into ranking \(R\):~ \(A \Rightarrow R \)
 while taking into account multicriteria estimates of elements
 and/or expert judgment
 (e.g., \cite{roy96,zop02}).
 In Fig. 4.9,
 illustration for restructuring in sorting problem is depicted.
 The problem is:
 \[\min \delta (\widehat{R}^{2}, \widehat{R}^{\star})
  ~~~ s.t. ~~ a (\widehat{R}^{1} \rightarrow \widehat{R}^{\star}) < b,\]
 where
 \(\widehat{R}^{\star}\) is solution,
 \(\widehat{R}^{1}\) is initial (the ``first'') ranking,
 \(\widehat{R}^{2}\) is the ``second'' ranking,
 \(\delta (\widehat{R}^{\star}, \widehat{R}^{2})\) is proximity
 between solution \(\widehat{R}^{\star }\)
 and the ``second'' ranking \(\widehat{R}^{\star }\)
 (e.g., structural proximity or  proximity by quality parameters for rankings),
 \(a(\widehat{R}^{1} \rightarrow \widehat{R}^{\star})\) is the
 cost of transformation of the ``first'' ranking \( \widehat{R}^{1} \)
 into solution \(\widehat{R}^{\star}\)
 (e.g., editing ``distance''),
 \(b\) is constraint for the transformation cost.
 Evidently, multi-stage restructuring problems
 (with change trajectories of elements)
 are prospective as well.

\begin{center}
\begin{picture}(38,43)
\put(12,00){\makebox(0,0)[bl]{Fig. 4.9.
 Example:  restructuring in sorting problem}}

\put(12,38){\makebox(0,0)[bl]{\(\widehat{R}^{1}\)}}
\put(14,21){\oval(28,32)}

\put(01,29.5){\makebox(0,0)[bl]{\(R^{1}_{1}\)}}

\put(16,31){\oval(20,6)}

\put(10,30){\circle*{1.0}} \put(09,31){\makebox(0,8)[bl]{\(7\)}}
\put(16,30){\circle*{1.0}} \put(15,31){\makebox(0,8)[bl]{\(8\)}}
\put(22,30){\circle*{1.0}} \put(21,31){\makebox(0,8)[bl]{\(9\)}}

\put(16,28){\vector(0,-1){4}}

\put(01,19.5){\makebox(0,0)[bl]{\(R^{1}_{2}\)}}

\put(16,21){\oval(20,6)}

\put(10,20){\circle*{1.0}} \put(09,21){\makebox(0,8)[bl]{\(1\)}}
\put(16,20){\circle*{1.0}} \put(15,21){\makebox(0,8)[bl]{\(2\)}}
\put(22,20){\circle*{1.0}} \put(21,21){\makebox(0,8)[bl]{\(3\)}}

\put(16,18){\vector(0,-1){4}}

\put(01,09.5){\makebox(0,0)[bl]{\(R^{1}_{3}\)}}

\put(16,11){\oval(20,6)}

\put(10,10){\circle*{1.0}} \put(09,11){\makebox(0,8)[bl]{\(4\)}}
\put(16,10){\circle*{1.0}} \put(15,11){\makebox(0,8)[bl]{\(5\)}}
\put(22,10){\circle*{1.0}} \put(21,11){\makebox(0,8)[bl]{\(6\)}}


\put(29,28){\vector(1,0){8}} \put(29,22){\vector(1,0){8}}
\put(29,16){\vector(1,0){8}}

\end{picture}
%
\begin{picture}(38,43)

\put(12,38){\makebox(0,0)[bl]{\(\widehat{R}^{\star}\)}}
\put(14,21){\oval(28,32)}

\put(01,29.5){\makebox(0,0)[bl]{\(R^{\star}_{1}\)}}

\put(16,31){\oval(20,6)}

\put(10,30){\circle*{1.0}} \put(09,31){\makebox(0,8)[bl]{\(1\)}}
\put(16,30){\circle*{1.0}} \put(15,31){\makebox(0,8)[bl]{\(2\)}}
\put(22,30){\circle*{1.0}} \put(21,31){\makebox(0,8)[bl]{\(3\)}}

\put(16,28){\vector(0,-1){4}}

\put(01,19.5){\makebox(0,0)[bl]{\(R^{\star}_{2}\)}}

\put(16,21){\oval(20,6)}

\put(09,20){\circle*{1.0}} \put(08,21){\makebox(0,8)[bl]{\(6\)}}
\put(14,20){\circle*{1.0}} \put(13,21){\makebox(0,8)[bl]{\(7\)}}

\put(18.5,20){\circle*{1.0}}
\put(17.5,21){\makebox(0,8)[bl]{\(8\)}}

\put(23,20){\circle*{1.0}} \put(22,21){\makebox(0,8)[bl]{\(9\)}}

\put(16,18){\vector(0,-1){4}}

\put(01,09.5){\makebox(0,0)[bl]{\(R^{\star}_{3}\)}}

\put(16,11){\oval(20,6)}

\put(13,10){\circle*{1.0}} \put(12,11){\makebox(0,8)[bl]{\(4\)}}
\put(19,10){\circle*{1.0}} \put(18,11){\makebox(0,8)[bl]{\(5\)}}


\put(32,27){\makebox(0,0)[bl]{\(\sim\)}}
\put(32,21){\makebox(0,0)[bl]{\(\sim\)}}
\put(32,14){\makebox(0,0)[bl]{\(\sim\)}}

\end{picture}
%
\begin{picture}(28,43)
\put(12,38){\makebox(0,0)[bl]{\(\widehat{R}^{2}\)}}
\put(14,21){\oval(28,32)}

\put(01,29.5){\makebox(0,0)[bl]{\(R^{2}_{1}\)}}

\put(16,31){\oval(20,6)}

\put(10,30){\circle*{1.0}} \put(09,31){\makebox(0,8)[bl]{\(1\)}}
\put(16,30){\circle*{1.0}} \put(15,31){\makebox(0,8)[bl]{\(2\)}}
\put(22,30){\circle*{1.0}} \put(21,31){\makebox(0,8)[bl]{\(3\)}}

\put(16,28){\vector(0,-1){4}}

\put(01,19.5){\makebox(0,0)[bl]{\(R^{2}_{2}\)}}

\put(16,21){\oval(20,6)}

\put(10,20){\circle*{1.0}} \put(09,21){\makebox(0,8)[bl]{\(4\)}}
\put(16,20){\circle*{1.0}} \put(15,21){\makebox(0,8)[bl]{\(5\)}}
\put(22,20){\circle*{1.0}} \put(21,21){\makebox(0,8)[bl]{\(6\)}}

\put(16,18){\vector(0,-1){4}}

\put(01,09.5){\makebox(0,0)[bl]{\(R^{2}_{3}\)}}

\put(16,11){\oval(20,6)}

\put(10,10){\circle*{1.0}} \put(09,11){\makebox(0,8)[bl]{\(7\)}}
\put(16,10){\circle*{1.0}} \put(15,11){\makebox(0,8)[bl]{\(8\)}}
\put(22,10){\circle*{1.0}} \put(21,11){\makebox(0,8)[bl]{\(9\)}}


\end{picture}
\end{center}

 \subsection{Clustering with multi-type elements}

 \subsubsection{Basic problems}

%
 Our basic clustering problem can be considered as the follows:~
 initial set of elements \(A\) consists of several subsets:
 \(A =  \bigcup_{j=1}^{n} A^{j} \)
 where \(A^{j} = \{ a^{j}_{1}, ... , a^{j}_{1}, ... ,  a^{j}_{n_{j}}  \}
 \),
 \(j\) corresponds to a certain kind of element type, i.e.,
 there is a set of the types:  \( J =  \{1,...,j,...j_{t} \} \) .
 Special binary relation is defined over the set \(J\):
 \(R_{J}\).

 The first clustering strategy is:

 (i) to group the initial set of elements \(A\)
 without analysis of the element types,

 (ii) to examine a connection of
 the elements of different types at the next stage
 (e.g., for each obtained cluster).

 Here, the second clustering strategy is examined:

 (1) to obtain clustering for each subset
 \(A^{j}=\{ A^{j}_{1}, ... , A^{j}_{l}, ... , A^{j}_{k_{j}}\}\).

 (2) to find a correspondence between
 clusters of subset \(A^{j^{1}}\) and
 clusters of subset \(A^{j^{2}}\)
 (for the case \(R(j^{1},j^{2}) = 1\));

 (3) to find a correspondence between
 cluster elements of subsets \(A^{j^{1}}\) and
 cluster elements of subsets \(A^{j^{2}}\)
 (for the case \(R(j^{1},j^{2}) = 1\)).

 Here,
  three new clustering problems with multi-type elements
 are suggested and examined:
 (i) clustering with two-type elements,
 (ii) clustering with three-type elements, and
 (iii) clustering with four-type elements.
 Examples of the special binary relation(s)
 over the element types are depicted in Fig. 4.10.

 Fig. 4.11 and Fig. 4.12 illustrate the above-mentioned second clustering
 strategy (three-type elements):

 (i) clustering of subsets:

  \(G = G_{1} \bigcup G_{2} \bigcup G_{3} \) where
   \(G_{1} = \{g^{1}_{1},g^{1}_{2},g^{1}_{3} \}\),
   \(G_{2} = \{g^{2}_{1},g^{2}_{2},g^{2}_{3} \}\),
   \(G_{3} = \{g^{3}_{1},g^{3}_{2},g^{3}_{3} \}\);

  \(B = B_{1} \bigcup B_{2} \bigcup B_{3} \) where
   \(B_{1} = \{b^{1}_{1},b^{1}_{2},b^{1}_{3} \}\),
   \(B_{2} = \{b^{2}_{1},b^{2}_{2},b^{2}_{3} \}\),
   \(B_{3} = \{b^{3}_{1},b^{3}_{2},b^{3}_{3} \}\);

  \(H = H_{1} \bigcup H_{2} \bigcup H_{3} \) where
   \(H_{1} = \{h^{1}_{1},h^{1}_{2},h^{1}_{3} \}\),
   \(H_{2} = \{h^{2}_{1},h^{2}_{2},h^{2}_{3} \}\),
   \(H_{3} = \{h^{3}_{1},h^{3}_{2},h^{3}_{3} \}\);

 (ii) three-matching of the obtained clusters:
  \(< B_{1} \star G_{3} \star H_{2}  >\),
  \(< B_{2} \star G_{1} \star H_{3}  >\), and
  \(< B_{3} \star G_{2} \star H_{1}  >\).

 (iii) three-matching of cluster elements:

 \(< B_{1} \star G_{3} \star H_{2}  >\):~
  \(< b^{1}_{1} \star g^{3}_{1} \star h^{2}_{3} >\),
  \(< b^{1}_{2} \star g^{3}_{3} \star h^{2}_{2} >\),
  \(< b^{1}_{3} \star g^{3}_{2} \star h^{2}_{1} >\);

  \(< B_{2} \star G_{1} \star H_{3}  >\):~
  \(< b^{2}_{1} \star g^{1}_{3} \star h^{3}_{2} >\),
  \(< b^{2}_{2} \star g^{1}_{1} \star h^{3}_{1} >\),
  \(< b^{2}_{3} \star g^{1}_{2} \star h^{3}_{3} >\);

 \(< B_{3} \star G_{2} \star H_{1}  >\):~
  \(< b^{3}_{1} \star g^{2}_{2} \star h^{1}_{3} >\),
  \(< b^{3}_{2} \star g^{2}_{1} \star h^{1}_{1} >\),
  \(< b^{3}_{3} \star g^{2}_{3} \star h^{1}_{2} >\).

\begin{center}
\begin{picture}(30,37)

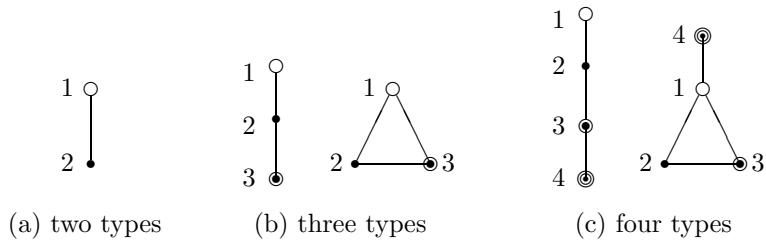
\put(05,00){\makebox(0,0)[bl] {Fig. 4.10. Illustration for binary
 relation over element types}}

\put(11,24){\line(0,-1){08.5}}

\put(00,05){\makebox(0,8)[bl]{(a) two types}}

\put(07,24){\makebox(0,8)[bl]{\(1\)}} \put(11,25){\circle{1.7}}

\put(11,15){\circle*{1}} \put(07,14){\makebox(0,8)[bl]{\(2\)}}

\end{picture}
%
\begin{picture}(40,35)


\put(01.5,05){\makebox(0,8)[bl]{(b) three types}}

\put(00,26){\makebox(0,8)[bl]{\(1\)}} \put(04.5,28){\circle{1.7}}

\put(04.5,21){\circle*{1}} \put(00,19){\makebox(0,8)[bl]{\(2\)}}

\put(04.5,13){\circle*{1}} \put(04.5,13){\circle{1.7}}
\put(00,12){\makebox(0,8)[bl]{\(3\)}}

\put(04.5,27){\line(0,-1){06}} \put(04.5,20.5){\line(0,-1){07.5}}


\put(15,15){\circle*{1}} \put(11.6,14){\makebox(0,8)[bl]{\(2\)}}

\put(25,15){\circle*{1}} \put(25,15){\circle{1.7}}
\put(26.5,14){\makebox(0,8)[bl]{\(3\)}}

\put(15,15){\line(1,0){10}}

\put(15,15){\line(1,2){4.5}} \put(25,15){\line(-1,2){4.5}}

\put(16,24){\makebox(0,8)[bl]{\(1\)}} \put(20,25){\circle{1.7}}

\end{picture}
%
\begin{picture}(34,35)


\put(03,05){\makebox(0,8)[bl]{(c) four types}}

\put(00,33){\makebox(0,8)[bl]{\(1\)}} \put(04.5,35){\circle{1.7}}

\put(04.5,28){\circle*{1}} \put(00,27){\makebox(0,8)[bl]{\(2\)}}

\put(04.5,20){\circle*{1}} \put(04.5,20){\circle{1.7}}
\put(00,19){\makebox(0,8)[bl]{\(3\)}}

\put(04.5,13){\circle*{0.8}} \put(04.5,13){\circle{1.3}}
\put(04.5,13){\circle{2.1}}

\put(00,12){\makebox(0,8)[bl]{\(4\)}}


\put(04.5,34){\line(0,-1){06}}

\put(04.5,28){\line(0,-1){07}} \put(04.5,20.5){\line(0,-1){07.5}}


\put(15,15){\circle*{1}} \put(11.6,14){\makebox(0,8)[bl]{\(2\)}}

\put(25,15){\circle*{1}} \put(25,15){\circle{1.7}}
\put(26.5,14){\makebox(0,8)[bl]{\(3\)}}

\put(15,15){\line(1,0){10}}

\put(15,15){\line(1,2){4.5}} \put(25,15){\line(-1,2){4.5}}

\put(16,24){\makebox(0,8)[bl]{\(1\)}} \put(20,25){\circle{1.7}}


\put(20,32){\circle*{0.8}} \put(20,32){\circle{1.3}}
\put(20,32){\circle{2.1}}

\put(16,31){\makebox(0,8)[bl]{\(4\)}}

\put(20,32){\line(0,-1){06.1}}

\end{picture}
\end{center}

\begin{center}
\begin{picture}(128,55)
\put(05,00){\makebox(0,0)[bl] {Fig. 4.11.
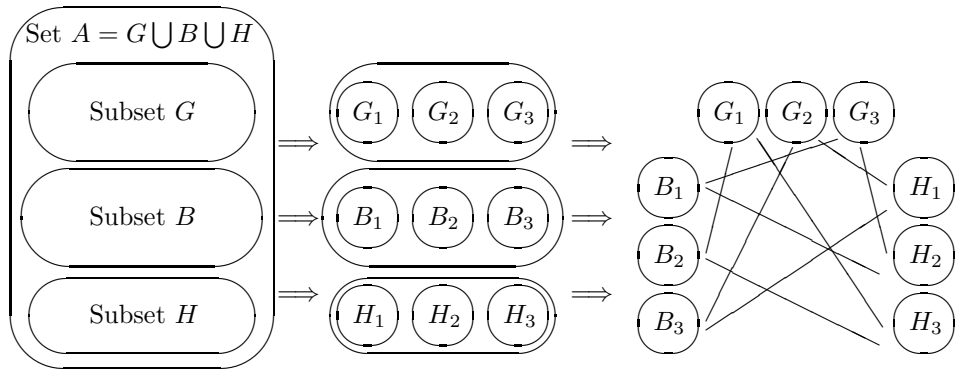
 Clustering strategy (three-type elements): clustering, three-matching}}

\put(38,35){\makebox(0,0)[bl]{\(\Longrightarrow \)}}
\put(38,25){\makebox(0,0)[bl]{\(\Longrightarrow \)}}
\put(38,15){\makebox(0,0)[bl]{\(\Longrightarrow \)}}


\put(20,30){\oval(35,48)}

\put(4.5,48.5){\makebox(0,8)[bl]{Set \(A = G \bigcup B \bigcup
H\)}}


\put(20,40){\oval(30,13)}

\put(013,39){\makebox(0,8)[bl]{Subset \(G\)}}


\put(20,26){\oval(32,13)}

\put(013,25){\makebox(0,8)[bl]{Subset \(B\)}}


\put(20,13){\oval(30,10)}

\put(013,12){\makebox(0,8)[bl]{Subset \(H\)}}


\put(77,35){\makebox(0,0)[bl]{\(\Longrightarrow \)}}
\put(77,25){\makebox(0,0)[bl]{\(\Longrightarrow \)}}
\put(77,15){\makebox(0,0)[bl]{\(\Longrightarrow \)}}


\put(60,40){\oval(30,13)}


\put(60,40){\oval(08,08)}
\put(58,38.5){\makebox(0,8)[bl]{\(G_{2}\)}}

\put(70,40){\oval(08,08)}
\put(68,38.5){\makebox(0,8)[bl]{\(G_{3}\)}}

\put(50,40){\oval(08,08)}
\put(48,38.5){\makebox(0,8)[bl]{\(G_{1}\)}}


\put(60,26){\oval(32,13)}


\put(60,26){\oval(08,08)}
\put(58,24.5){\makebox(0,8)[bl]{\(B_{2}\)}}

\put(70,26){\oval(08,08)}
\put(68,24.5){\makebox(0,8)[bl]{\(B_{3}\)}}

\put(50,26){\oval(08,08)}
\put(48,24.5){\makebox(0,8)[bl]{\(B_{1}\)}}


\put(60,13){\oval(30,10)}


\put(60,13){\oval(08,08)}
\put(58,11.5){\makebox(0,8)[bl]{\(H_{2}\)}}

\put(70,13){\oval(08,08)}
\put(68,11.5){\makebox(0,8)[bl]{\(H_{3}\)}}

\put(50,13){\oval(08,08)}
\put(48,11.5){\makebox(0,8)[bl]{\(H_{1}\)}}


\put(98,40){\oval(08,08)}
\put(96,38.5){\makebox(0,8)[bl]{\(G_{1}\)}}

\put(107,40){\oval(08,08)}
\put(105,38.5){\makebox(0,8)[bl]{\(G_{2}\)}}

\put(116,40){\oval(08,08)}
\put(114,38.5){\makebox(0,8)[bl]{\(G_{3}\)}}


\put(90,30){\oval(08,08)}
\put(88,28.5){\makebox(0,8)[bl]{\(B_{1}\)}}

\put(90,21){\oval(08,08)}
\put(88,19.5){\makebox(0,8)[bl]{\(B_{2}\)}}

\put(90,12){\oval(08,08)}
\put(88,10.5){\makebox(0,8)[bl]{\(B_{3}\)}}


\put(124,30){\oval(08,08)}
\put(122,28.5){\makebox(0,8)[bl]{\(H_{1}\)}}

\put(124,21){\oval(08,08)}
\put(122,19.5){\makebox(0,8)[bl]{\(H_{2}\)}}

\put(124,12){\oval(08,08)}
\put(122,10.5){\makebox(0,8)[bl]{\(H_{3}\)}}

\put(95,30.7){\line(3,1){17.5}}


\put(95,21.3){\line(1,4){03.5}}

\put(95,12.3){\line(1,2){11.5}}

\put(119,30.3){\line(-3,2){09}}


\put(119,21.3){\line(-1,4){03.5}}

\put(118.5,11.5){\line(-2,3){16.7}}

\put(95,30){\line(2,-1){23}}


\put(95,20.5){\line(2,-1){23}}

\put(95,11){\line(3,2){24}}

\end{picture}
\end{center}

 Problem solving frameworks are based on combinations of well-known
 combinatorial problems and corresponding algorithms:
 clustering (e.g., hierarchical clustering, k-means clustering),
 assignment/allocation, three-matching, for example:

 (a) clustering of elements for each element subset,

 (b) assignment of the obtained clusters
 (while taking into account the binary relation over element types),

 (c) assignment of cluster elements
 (while taking into account the binary relation over element types), and

 (d) analysis of the obtained solution and its correction/improvement.

\begin{center}
\begin{picture}(41,45)
\put(29,00){\makebox(0,0)[bl] {Fig. 4.12. Three-matching for
 elements}}

\put(03.6,05){\makebox(0,0)[bl] {(a)
 \(B_{1} \star G_{3} \star H_{2}\) }}


\put(00,31){\makebox(0,8)[bl]{\(B_{1}\)}}

\put(00,26){\makebox(0,8)[bl]{\(b^{1}_{1}\)}}
\put(04.5,28){\circle*{1}}

\put(04.5,21){\circle*{1}}
\put(00,19){\makebox(0,8)[bl]{\(b^{1}_{2}\)}}

\put(04.5,13){\circle*{1}}
\put(00,12){\makebox(0,8)[bl]{\(b^{1}_{3}\)}}

\put(05.7,28.3){\line(2,3){3.4}} \put(05.7,21.3){\line(3,2){19}}
\put(05.7,14){\line(2,3){12.5}}

\put(05.7,27.7){\line(3,-2){22.5}}

\put(05.7,20.5){\line(1,0){22.5}}

\put(05.8,12.5){\line(3,2){22.5}}


\put(16,40){\makebox(0,8)[bl]{\(G_{3}\)}}

\put(08,36){\makebox(0,8)[bl]{\(g^{3}_{1}\)}}
\put(10,35){\circle{1.3}}

\put(18,35){\circle{1.3}}
\put(16,36){\makebox(0,8)[bl]{\(g^{3}_{2}\)}}

\put(26,35){\circle{1.3}}
\put(24,36){\makebox(0,8)[bl]{\(g^{3}_{3}\)}}


\put(31,31){\makebox(0,8)[bl]{\(H_{2}\)}}

\put(32,26){\makebox(0,8)[bl]{\(h^{2}_{1}\)}}
\put(30.5,28){\circle*{0.8}} \put(30.5,28){\circle{1.7}}

\put(30.5,21){\circle*{0.8}} \put(30.5,21){\circle{1.7}}
\put(32,19){\makebox(0,8)[bl]{\(h^{2}_{2}\)}}

\put(30.5,13){\circle*{0.8}} \put(30.5,13){\circle{1.7}}
\put(32,12){\makebox(0,8)[bl]{\(h^{2}_{3}\)}}

\put(29,28.3){\line(-2,1){10}}

\put(29,21.3){\line(-1,4){3}}

\put(29,14){\line(-1,1){19}}

\end{picture}
%
\begin{picture}(41,45)

\put(03.6,05){\makebox(0,0)[bl] {(b)
 \(B_{2} \star G_{1} \star H_{3}\) }}


\put(00,31){\makebox(0,8)[bl]{\(B_{2}\)}}

\put(00,26){\makebox(0,8)[bl]{\(b^{2}_{1}\)}}
\put(04.5,28){\circle*{1}}

\put(04.5,21){\circle*{1}}
\put(00,19){\makebox(0,8)[bl]{\(b^{2}_{2}\)}}

\put(04.5,13){\circle*{1}}
\put(00,12){\makebox(0,8)[bl]{\(b^{2}_{3}\)}}

\put(06,28.7){\line(4,1){19}} \put(06.4,21.3){\line(1,4){3}}
\put(05.6,15){\line(2,3){12}}

\put(05.7,27.7){\line(3,-1){22.5}}
\put(05.7,20.5){\line(3,1){22.5}} \put(05.8,12){\line(1,0){22}}


\put(16,40){\makebox(0,8)[bl]{\(G_{1}\)}}

\put(08,36){\makebox(0,8)[bl]{\(g^{1}_{1}\)}}
\put(10,35){\circle{1.3}}

\put(18,35){\circle{1.3}}
\put(16,36){\makebox(0,8)[bl]{\(g^{1}_{2}\)}}

\put(26,35){\circle{1.3}}
\put(24,36){\makebox(0,8)[bl]{\(g^{1}_{3}\)}}


\put(31,31){\makebox(0,8)[bl]{\(H_{3}\)}}

\put(32,26){\makebox(0,8)[bl]{\(h^{3}_{1}\)}}
\put(30.5,28){\circle*{0.8}} \put(30.5,28){\circle{1.7}}

\put(30.5,21){\circle*{0.8}} \put(30.5,21){\circle{1.7}}
\put(32,19){\makebox(0,8)[bl]{\(h^{2}_{2}\)}}

\put(30.5,13){\circle*{0.8}} \put(30.5,13){\circle{1.7}}
\put(32,12){\makebox(0,8)[bl]{\(h^{3}_{3}\)}}

\put(28.6,28.7){\line(-4,1){18}} \put(29,21.3){\line(-1,4){3}}
\put(29.9,15){\line(-2,3){12}}

\end{picture}
%
\begin{picture}(37,45)

\put(03.6,05){\makebox(0,0)[bl] {(c)
 \(B_{3} \star G_{2} \star H_{1}\) }}


\put(00,31){\makebox(0,8)[bl]{\(B_{3}\)}}

\put(00,26){\makebox(0,8)[bl]{\(b^{3}_{1}\)}}
\put(04.5,28){\circle*{1}}

\put(04.5,21){\circle*{1}}
\put(00,19){\makebox(0,8)[bl]{\(b^{3}_{2}\)}}

\put(04.5,13){\circle*{1}}
\put(00,12){\makebox(0,8)[bl]{\(b^{3}_{3}\)}}

\put(06,28.1){\line(2,1){11}} \put(06.4,21.3){\line(1,4){3}}


\put(05.7,14){\line(1,1){19}}


\put(05.7,27.7){\line(3,-2){22.5}}

\put(05.7,20.5){\line(3,1){22.5}}

\put(05.8,12.5){\line(3,1){22.5}}


\put(16,40){\makebox(0,8)[bl]{\(G_{2}\)}}

\put(08,36){\makebox(0,8)[bl]{\(g^{2}_{1}\)}}
\put(10,35){\circle{1.3}}

\put(18,35){\circle{1.3}}
\put(16,36){\makebox(0,8)[bl]{\(g^{2}_{2}\)}}

\put(26,35){\circle{1.3}}
\put(24,36){\makebox(0,8)[bl]{\(g^{2}_{3}\)}}


\put(31,31){\makebox(0,8)[bl]{\(H_{1}\)}}

\put(32,26){\makebox(0,8)[bl]{\(h^{1}_{1}\)}}
\put(30.5,28){\circle*{0.8}} \put(30.5,28){\circle{1.7}}

\put(30.5,21){\circle*{0.8}} \put(30.5,21){\circle{1.7}}
\put(32,19){\makebox(0,8)[bl]{\(h^{1}_{2}\)}}

\put(30.5,13){\circle*{0.8}} \put(30.5,13){\circle{1.7}}
\put(32,12){\makebox(0,8)[bl]{\(h^{1}_{3}\)}}

\put(28.5,28.7){\line(-3,1){17}}

\put(29,21.3){\line(-1,4){3}}

\put(29,14){\line(-1,2){10}}

\end{picture}
\end{center}


\subsubsection{Example of Team Design}

 The problems of analysis, modeling and design of teams
 are widely used in many domains
 (R\&D, Start-Up companies, manufacturing, education, etc.).
  Some basic types of teams are briefly described in Table 4.4.

\begin{center}
 {\bf Table 4.4.} Types of teams \\
\begin{tabular}{| c | l| l|l |l|}
\hline
 No.  & Type &Purpose(s) &Domain(s) & Source(s)  \\
\hline

 1.& R\&D project team& Accomplishment of specific task
  & R\&D & \cite{ford92} \\

 2.& Multi-functional/  & Forming of multi-functional/
     & R\&D, design,
   &\cite{choud95,zaka99} \\
 &multi-disciplinary   &multi-disciplinary  task(s)& manufacturing, & \\
 & team & (system life cycle)&logistics, marketing, & \\
 && &  investment, etc.&\\

 3.& Formal work group & To deliver a product or service
   & R\&D, design, &\cite{stew06} \\
 &  &  & manufacturing, & \\
 &  &  & logistics, etc. & \\

 4.& Global virtual
   &Forming and management of &Integrated&\cite{lipn97,pras02}\\
 &  teams  &distributed team(s)&distributed&\\
 &   &&technologies&\\

 5. &Informal group &To collect and pass on business
   & Business &  \\
  &(e.g., friends)&information &  &  \\

 6. &Community of &Building and sharing/exchange
           &Professional   &\cite{less01,paas14,weng02}\\
     &practice &of knowledge, coordination&networks, &\\
  &(group of experts)&&organizations &\\

 7.& Start-Up team& Design of new product/service
   & Innovation & \cite{lev15}\\

 8.& Student team &Joint educational work,
  & Education & this paper\\
 &  &joint research project  &  & \\

 9. & Professors/lecturers & Design of/participation in
 & Education& \\
 &teams& new educational program&&\\
\hline
\end{tabular}
\end{center}

 Some additional description for two special types of teams are:

 (a) ``communities of practice'' are the groups of experts in applied domains
 (with interaction, exchange of knowledge and experience, etc.),

 (b) global virtual (distributed) teams
 involves four major elements:
 (i) virtual team structure,
 (ii) strategic objectives,
 (iii) work characteristics,
 (iv) situational constraints \cite{pras02}.

 The discipline of teams \cite{katz93} contains several basic
 research problems, for example:

 (1) design of teams (e.g., structure, elements)
 (e.g., \cite{lev15,lipn97,pras02}),


 (2) selection/forming of multi-functional teams
 (e.g., \cite{lev15,zaka99}),

 (3) analysis of relationships between team design and team performance
 (e.g., \cite{stew06}).

 (4) modeling of teams evolution and forecasting
 (e.g., \cite{lev15}).

~~~~

 {\bf Example 4.3.} A simplified numerical example for designing a multi-student teams for joint
 laboratory/project works is described.
 General solving framework consists of seven stages
 (Fig. 4.13):

~~

 {\it Stage 1.} Preliminary data analysis  (expert judgment).

  {\it Stage 2.} Analysis and generation/processing of criteria
  (expert judgment, databases).

 {\it Stage 3.} Assessment of objects upon criteria,
    processing of  estimates
  (expert judgment, computing).

 {\it Stage 4.} Grouping/classification of
 initial object set to obtain object subset for each type
 (filtering, classification/sorting).

  {\it Stage 5.} Evaluation of relationship between object pairs
  (e.g., ``friendship''/compatibility)
 (expert judgment, databases).

  {\it Stage 6.} Design of object configurations
  (special composite structures)
  (e.g., morphological clique problem).

  {\it Stage 7.} Analysis of results.


\begin{center}
\begin{picture}(126,43)
\put(40,00){\makebox(0,0)[bl]{Fig. 4.13. Solving framework}}

\put(08.5,16){\oval(17,22)}

\put(1,19){\makebox(0,0)[bl]{Initial set}}
\put(7,15.5){\makebox(0,0)[bl]{of }}

\put(02.5,10){\makebox(0,0)[bl]{objects}}

\put(17,24){\vector(1,0){04}} \put(17,19){\vector(1,0){04}}
\put(17,14){\vector(1,0){04}} \put(17,09){\vector(1,0){04}}


\put(21,30){\line(1,0){18}} \put(21,37){\line(1,0){18}}
\put(21,30){\line(0,1){07}} \put(39,30){\line(0,1){07}}

\put(21.5,34){\makebox(0,0)[bl]{Generation }}
\put(21.5,31){\makebox(0,0)[bl]{of criteria}}

\put(30,30){\vector(0,-1){04}}

\put(21,06){\line(1,0){18}} \put(21,26){\line(1,0){18}}
\put(21,06){\line(0,1){20}} \put(39,06){\line(0,1){20}}

\put(21.5,20){\makebox(0,0)[bl]{Evaluation}}
\put(21.5,16){\makebox(0,0)[bl]{of objects}}
\put(21.5,12){\makebox(0,0)[bl]{(upon}}
\put(21.5,08){\makebox(0,0)[bl]{criteria}}


\put(39,24){\vector(1,0){04}} \put(39,19){\vector(1,0){04}}
\put(39,14){\vector(1,0){04}} \put(39,09){\vector(1,0){04}}

\put(43,30){\line(1,0){10}} \put(43,36){\line(1,0){10}}
\put(43,30){\line(0,1){06}} \put(53,30){\line(0,1){06}}

\put(43.5,32){\makebox(0,0)[bl]{Rules}}

\put(48,30){\vector(0,-1){04}}
\put(43,06){\line(1,0){10}} \put(43,26){\line(1,0){10}}
\put(43,06){\line(0,1){20}} \put(53,06){\line(0,1){20}}

\put(43.5,20){\makebox(0,0)[bl]{Clas-}}
\put(43.5,16){\makebox(0,0)[bl]{sifi-}}
\put(43.5,12){\makebox(0,0)[bl]{cation}}
\put(43.5,08){\makebox(0,0)[bl]{}}

\put(53,24){\vector(1,0){04}} \put(53,19){\vector(1,0){04}}
\put(53,14){\vector(1,0){04}} \put(53,09){\vector(1,0){04}}


\put(60,26.5){\makebox(0,0)[bl]{Object subset \(1\)}}

\put(76,28){\oval(35,5)}


\put(60,20.5){\makebox(0,0)[bl]{Object subset \(2\)}}

\put(76,22){\oval(35,5)}


\put(72,18){\makebox(0,0)[bl]{{. . .}}}


\put(60,12.5){\makebox(0,0)[bl]{Object subset \((k-1)\)}}

\put(76,14){\oval(35,5)}


\put(60,06){\makebox(0,0)[bl]{Object subset \(k\)}}

\put(76,07.5){\oval(35,5)}


\put(98,30){\line(1,0){10}} \put(98,40){\line(1,0){10}}
\put(98,30){\line(0,1){10}} \put(108,30){\line(0,1){10}}

\put(98.5,37){\makebox(0,0)[bl]{Team}}
\put(98.5,34){\makebox(0,0)[bl]{struc-}}
\put(98.5,31){\makebox(0,0)[bl]{tures}}

\put(103,30){\vector(0,-1){04}}
\put(94,24){\vector(1,0){04}} \put(94,19){\vector(1,0){04}}
\put(94,14){\vector(1,0){04}} \put(94,09){\vector(1,0){04}}
\put(98,06){\line(1,0){10}} \put(98,26){\line(1,0){10}}
\put(98,06){\line(0,1){20}} \put(108,06){\line(0,1){20}}

\put(98.5,20){\makebox(0,0)[bl]{Confi-}}
\put(98.5,16){\makebox(0,0)[bl]{gura-}}
\put(98.5,12){\makebox(0,0)[bl]{ting}}
\put(98.5,08){\makebox(0,0)[bl]{teams}}

\put(108,24){\vector(1,0){04}} \put(108,19){\vector(1,0){04}}
\put(108,14){\vector(1,0){04}} \put(108,09){\vector(1,0){04}}

\put(113.5,22.5){\makebox(0,0)[bl]{Team \(1\)}}

\put(119,24){\oval(13,5)}


\put(113.5,16.5){\makebox(0,0)[bl]{Team \(2\)}}

\put(119,18){\oval(13,5)}


\put(115,13){\makebox(0,0)[bl]{{. . .}}}


\put(113.5,07.5){\makebox(0,0)[bl]{Team \(\mu\)}}

\put(119,09){\oval(13,5)}

\end{picture}
\end{center}

 Here example for 14 students (from Table 2.2)
 is considered: Table 4.5.
  The considered basic set of criteria (educational disciplines) is the following
 (ordinal scale \([3,4,5]\), \(5\) corresponds to the best level):
 (1) mathematics \(C_{1}\),
 (2) physics \(C_{2}\),
 (3) computer systems \(C_{3}\),
 (4) software engineering \(C_{4}\),
 (5) antenna devices \(C_{5}\),
 (6) signal processing \(C_{6}\),
 (7) receiver and sender systems \(C_{7}\),
 (8) information transmission \(C_{8}\),
 (9) measurement in radio engineering \(C_{9}\),
 (10) control systems \(C_{10}\).

 The following student types are
 examined (by professional inclination/skill):

   (1) formal methods (\(O_{1}\)),
  support disciplines by criteria: \(C_{1},C_{5},C_{6},C_{8},C_{10}\);

   (2) physical processes (\(O_{2}\)),
    support disciplines by criteria: \(C_{2},C_{9}\);

  (3) system design (\(O_{3}\)),
  support disciplines by criteria: \(C_{3},C_{6},C_{8},C_{10}\);

  (4) software development (\(O_{4}\)),
  support disciplines by criteria:     \(C_{3},C_{4}\);

 (5) simulation (\(O_{5}\)),
  support disciplines by criteria: \(C_{1},C_{3},C_{4}\);

 (6)  signal processing  (\(O_{6}\)),
  support disciplines by criteria: \(C_{1},C_{3},C_{4},C_{8},C_{9}\);

 (7) data transmission  (\(O_{7}\)),
  support disciplines by criteria:
    \(C_{1},C_{2},C_{3},C_{5},C_{6},C_{7},C_{8}\).

 Thus, it is necessary to transform initial estimates of students
 upon basic set of criteria
 (i.e., \(C_{1},...,C_{5}\))
 into estimates by inclination
(i.e., \(O_{1},O_{2},O_{3},O_{4}\))
  with selection of
 ``domain leader''/``quasi-domain leader''
  for each inclination type
 (i.e., by some rule(s),
  multicritria ranking/sorting problem,  expert judgment)
 (Table 4.6).
 Here,
 \(1\) corresponds to level of ``domain leader'',
 \(2\) corresponds to level of ``quasi-domain leader''.

 In the example,
 the problem purpose is to form
 the following laboratory/project student teams
 (joint laboratory works, student joint projects):

  (1)  monitoring system  project \(T_{1}\),
  support professional skills (as team structure):
  \( T_{1} = O_{1} \star O_{3} \star O_{4} \star O_{5}\star O_{7}\);

  (2) medical systems  (measurement and analysis)  \(T_{2}\),
 support professional skills (as team structure):
  \( T_{2} = O_{1} \star O_{3}\star O_{4} \star O_{6}\);

  (3) GIS and seismic modeling  \(T_{3}\),
 support professional skills (as team structure):
  \( T_{3} = O_{1} \star O_{2} \star O_{3}\star O_{4} \star O_{5} \star
  O_{6}\).


 Ordinal estimates of student friendship is presented in Table 4.7
 (expert judgment, ordinal scale \([0,1,2,3]\) is used,
 \(3\)) corresponds to the best friendship).


 The design process is based on morphological clique problem
 while taking into account
 ordinal quality of elements
 (Table 4.6)
 and
 elements compatibility (Table 4.7).
 Structures for configuration design of
 team \(T_{1}\), team \(T_{2}\), team \(T_{3}\)
 are depicted
 in Fig. 4.14,  in Fig. 4.15, and in Fig. 4.16.

 It is assumed that each student team consists of 2, 3, or 4 students.
 ``Domain leader(s)'' or ``quasi-domain leader(s)''
 have to be contained into each team.
 In educational process,
 elements which are not
``domain leaders''/``quasi-domain leaders'' have to be added
 into some some student teams (composite solutions).
 Numerical examples of composite solutions are:

 (a) for \(T_{1}\)  (Fig. 4.14):

 \(T_{1}^{1} = A_{6} \star A_{9} \star A_{2} \star A_{11}\),
 \(T_{1}^{2} = A_{6} \star A_{9} \star A_{2} \star A_{11}\),
 \(T_{1}^{3} = A_{6} \star A_{9} \star A_{2} \),
 \(T_{1}^{4} = A_{6} \star A_{9} \star A_{2} \);

 (b) for \(T_{2}\)  (Fig. 4.15):

 \(T_{2}^{1} = A_{6} \star A_{9} \star A_{2} \star A_{11}\),
 \(T_{2}^{2} = A_{6} \star A_{9} \star A_{2} \star A_{11}\),
 \(T_{2}^{3} = A_{6} \star A_{9} \star A_{2} \),
 \(T_{2}^{4} = A_{6} \star A_{9} \star A_{2} \);

 (c) for \(T_{3}\)  (Fig. 4.16):

 \(T_{3}^{1} = A_{6} \star A_{9} \star A_{2} \star A_{11}\),
 \(T_{3}^{2} = A_{6} \star A_{9} \star A_{2} \star A_{11}\),
 \(T_{3}^{3} = A_{6} \star A_{9} \star A_{2} \),
 \(T_{3}^{4} = A_{6} \star A_{9} \star A_{2} \).

\begin{center}
{\bf Table 4.5.} Students,  estimates upon criteria   \\
\begin{tabular}{| c |c|c|c|c|c||c|c|c|c|c|}
\hline
 Item (student)  &\(C_{1}\)&\(C_{2}\)&\(C_{3}\)&\(C_{4}\)&\(C_{5}\)&\(C_{6}\)&\(C_{7}\)&\(C_{8}\)&\(C_{9}\)&\(C_{10}\)  \\
\hline
 Student 1   & 3 & 3 & 4 & 4 & 4 & 4 & 5 & 5 & 5 & 4\\
 Student 2   & 3 & 3 & 3 & 3 & 3 & 4 & 4 & 3 & 4 & 3\\
 Student 3   & 4 & 4 & 4 & 4 & 4& 4 & 4 & 4 & 4 & 4 \\
 Student 4   & 5 & 5 & 4 & 4 & 4 & 4 & 4 & 5 & 3 & 4\\
 Student 5   & 3 & 3 & 3 & 4 & 3 & 4 & 4 & 3 & 4 & 3\\
 Student 6   & 5 & 4 & 5 & 4 & 5& 5 & 5 & 5 & 5 & 4 \\
 Student 7   & 3 & 3 & 3 & 4 & 3 & 4 & 4 & 3 & 4 & 3\\
 Student 8   & 4 & 4 & 4 & 4 & 3& 4 & 4 & 4 & 5 & 4 \\
 Student 9   & 5 & 5 & 5 & 5 & 5& 5 & 5 & 5 & 5 & 5 \\
 Student 10  & 5 & 5 & 5 & 4 & 5& 5 & 5 & 5 & 5 & 5 \\
 Student 11  & 3 & 3 & 3 & 3 & 3& 3 & 4 & 3 & 5 & 3 \\
 Student 12  & 3 & 4 & 4 & 4 & 4& 4 & 4 & 4 & 5 & 4 \\
 Student 13  & 5 & 5 & 5 & 5 & 5& 4 & 4 & 5 & 4 & 5 \\
 Student 14  & 3 & 4 & 4 & 4 & 4& 4 & 4 & 4 & 4 & 4 \\

\hline
\end{tabular}
\end{center}

\begin{center}
{\bf Table 4.6.} Students, evaluation by inclination types   \\
\begin{tabular}{| c | c|c|c|c|c|c|c|}
\hline
 Item (student)  &\(O_{1}\)&\(O_{2}\)&\(O_{3}\)&\(O_{4}\)&\(O_{5}\)&\(O_{6}\)&\(O_{7}\) \\
\hline
 Student 1   &  &   &   & 2 &   &   &  \\
 Student 2   &  &   &   &   &   &   &  \\
 Student 3   & 2&   &   &   &   & 2 & 2 \\
 Student 4   & 2& 2 &   &   &   & 2 & 2 \\
 Student 5   &  &   &   &   &   &   &  \\
 Student 6   & 1& 2 & 1 & 2 & 2 & 1 & 1\\
 Student 7   &  &   &   &   &   &   &  \\
 Student 8   &  & 2 & 2 &   &   & 2 &  \\
 Student 9   & 1& 1 & 1 & 1 & 1 & 1 & 1 \\
 Student 10  & 1& 1 & 1 & 2 & 2 & 1 & 1 \\
 Student 11  &  &   &   &   &   &   &  \\
 Student 12  &  & 2 & 2 & 2 &   &   &  \\
 Student 13  & 1& 2 & 1 & 1 & 1 & 1 & 2 \\
 Student 14  &  &   &   & 2 &   &   &  \\

\hline
\end{tabular}
\end{center}

\newpage
\begin{center}
{\bf Table 4.7.} Ordinal estimates of student friendship (compatibility) \\
\begin{tabular}{  |c|c|c|c| c|c|c|c |c|c|c|c | c |c| }
\hline
   Discipline
  &\(A_{2}\) &\(A_{3}\) &\(A_{4}\) &\(A_{5}\) &\(A_{6}\) &\(A_{7}\) &\(A_{8}\)
  &\(A_{9}\)&\(A_{10}\)&\(A_{11}\)&\(A_{12}\)&\(A_{13}\)&\(A_{14}\) \\
\hline

 \(A_{1}\) &\(0\)&\(3\)&\(1\)&\(3\)&\(3\)&\(1\)&\(1\)&\(2\)&\(2\)&\(3\)&\(3\)&\(1\)&\(1\)   \\
 \(A_{2}\) &&\(1\)&\(2\)&\(1\)&\(2\)&\(3\)&\(2\)&\(1\)&\(1\)&\(1\)&\(2\)&\(1\)&\(2\)   \\
 \(A_{3}\) &&&\(3\)&\(3\)&\(3\)&\(1\)&\(1\)&\(2\)&\(3\)&\(3\)&\(3\)&\(1\)&\(1\)   \\
 \(A_{4}\) &&&&\(2\)&\(3\)&\(1\)&\(1\)&\(2\)&\(1\)&\(3\)&\(3\)&\(0\)&\(2\)   \\
 \(A_{5}\) &&&&&\(3\)&\(1\)&\(1\)&\(3\)&\(2\)&\(2\)&\(3\)&\(1\)&\(0\)   \\
 \(A_{6}\) &&&&&&\(0\)&\(0\)&\(1\)&\(0\)&\(3\)&\(3\)&\(1\)&\(1\)   \\
 \(A_{7}\) &&&&&&&\(3\)&\(2\)&\(3\)&\(1\)&\(2\)&\(1\)&\(2\)   \\
 \(A_{8}\) &&&&&&&&\(3\)&\(3\)&\(1\)&\(2\)&\(1\)&\(0\)   \\
 \(A_{9}\) &&&&&&&&&\(3\)&\(2\)&\(3\)&\(2\)&\(1\)   \\
 \(A_{10}\)&&&&&&&&&&\(3\)&\(3\)&\(2\)&\(0\)   \\
 \(A_{11}\)&&&&&&&&&&&\(3\)&\(2\)&\(2\)   \\
 \(A_{12}\)&&&&&&&&&&&&\(2\)&\(2\)   \\
 \(A_{13}\)&&&&&&&&&&&&&\(2\) \\

\hline
\end{tabular}
\end{center}

\begin{center}
\begin{picture}(85,61.6)

\put(09.5,00){\makebox(0,0)[bl] {Fig. 4.14. 5-component team
  \(T_{1}\)}}

\put(00,09){\makebox(0,8)[bl]{\(A_{13}(1)\)}}
\put(00,13){\makebox(0,8)[bl]{\(A_{10}(1)\)}}
\put(00,17){\makebox(0,8)[bl]{\(A_{9}(1)\)}}
\put(00,21){\makebox(0,8)[bl]{\(A_{6}(1)\)}}
\put(00,25){\makebox(0,8)[bl]{\(A_{4}(2)\)}}
\put(00,29){\makebox(0,8)[bl]{\(A_{3}(2)\)}}

\put(15,05){\makebox(0,8)[bl]{\(A_{13}(2)\)}}
\put(15,09){\makebox(0,8)[bl]{\(A_{12}(2)\)}}
\put(15,13){\makebox(0,8)[bl]{\(A_{10}(1)\)}}
\put(15,17){\makebox(0,8)[bl]{\(A_{9}(1)\)}}
\put(15,21){\makebox(0,8)[bl]{\(A_{8}(2)\)}}
\put(15,25){\makebox(0,8)[bl]{\(A_{6}(2)\)}}
\put(15,29){\makebox(0,8)[bl]{\(A_{4}(2)\)}}

\put(30,09){\makebox(0,8)[bl]{\(A_{13}(1)\)}}
\put(30,13){\makebox(0,8)[bl]{\(A_{12}(2)\)}}
\put(30,17){\makebox(0,8)[bl]{\(A_{10}(1)\)}}
\put(30,21){\makebox(0,8)[bl]{\(A_{9}(1)\)}}
\put(30,25){\makebox(0,8)[bl]{\(A_{8}(2)\)}}
\put(30,29){\makebox(0,8)[bl]{\(A_{6}(1)\)}}

\put(45,17){\makebox(0,8)[bl]{\(A_{13}(1)\)}}
\put(45,21){\makebox(0,8)[bl]{\(A_{10}(2)\)}}
\put(45,25){\makebox(0,8)[bl]{\(A_{9}(1)\)}}
\put(45,29){\makebox(0,8)[bl]{\(A_{6}(2)\)}}

\put(60,09){\makebox(0,8)[bl]{\(A_{13}(2)\)}}
\put(60,13){\makebox(0,8)[bl]{\(A_{10}(1)\)}}
\put(60,17){\makebox(0,8)[bl]{\(A_{9}(1)\)}}
\put(60,21){\makebox(0,8)[bl]{\(A_{6}(1)\)}}
\put(60,25){\makebox(0,8)[bl]{\(A_{4}(2)\)}}
\put(60,29){\makebox(0,8)[bl]{\(A_{3}(2)\)}}

\put(3,35){\circle{2}} \put(18,35){\circle{2}}
\put(33,35){\circle{2}} \put(48,35){\circle{2}}
\put(63,35){\circle{2}}

\put(3,40){\line(0,-1){04}} \put(18,40){\line(0,-1){04}}
\put(33,40){\line(0,-1){04}} \put(48,40){\line(0,-1){04}}
\put(63,40){\line(0,-1){04}}

\put(3,40){\line(1,0){60}} \put(06,40){\line(0,1){22}}

\put(06,62){\circle*{2.4}}

\put(4,36.5){\makebox(0,8)[bl]{\(O_{1}\)}}
\put(13.5,36.5){\makebox(0,8)[bl]{\(O_{2}\)}}
\put(28.5,36.5){\makebox(0,8)[bl]{\(O_{3}\)}}
\put(43.5,36.5){\makebox(0,8)[bl]{\(O_{5}\)}}
\put(58.5,36.5){\makebox(0,8)[bl]{\(O_{7}\)}}

\put(09,61){\makebox(0,8)[bl]{\(T_{1} = O_{1}\star O_{2}\star
 O_{3} \star O_{5} \star O_{7}\)}}

\put(08,56){\makebox(0,8)[bl]{\(T^{1}_{1} =
 A_{1} \star A_{2} \star A_{2} \star A_{3} \)}}

\put(08,51){\makebox(0,8)[bl] {\(T^{2}_{1} =
 A_{1} \star A_{2} \star A_{2} \star A_{3} \)}}

\put(08,46){\makebox(0,8)[bl] {\(T^{3}_{1} =
 A_{1}\star A_{2}\star  A_{2} \)}}

\put(08,41){\makebox(0,8)[bl] {\(T^{4}_{1}=A_{1}\star A_{2}\star
 A_{2}\star A_{3} \star A_{3}\)}}

\end{picture}
%
\begin{picture}(54,65.6)

\put(02.5,00){\makebox(0,0)[bl] {Fig. 4.15. 4-component team
  \(T_{2}\)}}

\put(00,09){\makebox(0,8)[bl]{\(A_{13}(1)\)}}
\put(00,13){\makebox(0,8)[bl]{\(A_{10}(1)\)}}
\put(00,17){\makebox(0,8)[bl]{\(A_{9}(1)\)}}
\put(00,21){\makebox(0,8)[bl]{\(A_{6}(1)\)}}
\put(00,25){\makebox(0,8)[bl]{\(A_{4}(2)\)}}
\put(00,29){\makebox(0,8)[bl]{\(A_{3}(2)\)}}

\put(15,09){\makebox(0,8)[bl]{\(A_{13}(1)\)}}
\put(15,13){\makebox(0,8)[bl]{\(A_{12}(2)\)}}
\put(15,17){\makebox(0,8)[bl]{\(A_{10}(1)\)}}
\put(15,21){\makebox(0,8)[bl]{\(A_{9}(1)\)}}
\put(15,25){\makebox(0,8)[bl]{\(A_{8}(2)\)}}
\put(15,29){\makebox(0,8)[bl]{\(A_{6}(1)\)}}

\put(30,05){\makebox(0,8)[bl]{\(A_{14}(2)\)}}
\put(30,09){\makebox(0,8)[bl]{\(A_{13}(1)\)}}
\put(30,13){\makebox(0,8)[bl]{\(A_{12}(2)\)}}
\put(30,17){\makebox(0,8)[bl]{\(A_{10}(2)\)}}
\put(30,21){\makebox(0,8)[bl]{\(A_{9}(1)\)}}
\put(30,25){\makebox(0,8)[bl]{\(A_{6}(2)\)}}
\put(30,29){\makebox(0,8)[bl]{\(A_{1}(2)\)}}

\put(45,05){\makebox(0,8)[bl]{\(A_{13}(1)\)}}
\put(45,09){\makebox(0,8)[bl]{\(A_{10}(1)\)}}
\put(45,13){\makebox(0,8)[bl]{\(A_{9}(1)\)}}
\put(45,17){\makebox(0,8)[bl]{\(A_{8}(2)\)}}
\put(45,21){\makebox(0,8)[bl]{\(A_{6}(1)\)}}
\put(45,25){\makebox(0,8)[bl]{\(A_{4}(2)\)}}
\put(45,29){\makebox(0,8)[bl]{\(A_{3}(2)\)}}

\put(3,35){\circle{2}} \put(18,35){\circle{2}}
\put(33,35){\circle{2}} \put(48,35){\circle{2}}

\put(3,40){\line(0,-1){04}} \put(18,40){\line(0,-1){04}}
\put(33,40){\line(0,-1){04}} \put(48,40){\line(0,-1){04}}

\put(3,40){\line(1,0){45}} \put(06,40){\line(0,1){22}}

\put(06,62){\circle*{2.4}}

\put(4,36.5){\makebox(0,8)[bl]{\(O_{1}\)}}
\put(13.5,36.5){\makebox(0,8)[bl]{\(O_{3}\)}}
\put(28.5,36.5){\makebox(0,8)[bl]{\(O_{4}\)}}
\put(43.5,36.5){\makebox(0,8)[bl]{\(O_{6}\)}}

\put(09,61){\makebox(0,8)[bl]{\(T_{2} = O_{1}\star
 O_{3} \star O_{4} \star O_{6}\)}}

\put(08,56){\makebox(0,8)[bl] {\(T^{1}_{2}=
 A_{1}\star  A_{2}\star A_{3} \star A_{3}\)}}

\put(08,51){\makebox(0,8)[bl] {\(T^{2}_{2}=
 A_{1}\star A_{2}\star A_{3} \star A_{3}\)}}

\put(08,46){\makebox(0,8)[bl] {\(T^{3}_{2}=
 A_{1}\star A_{2}\star A_{3}\)}}

\put(08,41){\makebox(0,8)[bl] {\(T^{4}_{2}=
 A_{1}\star A_{2}\star A_{3}\)}}

\end{picture}
\end{center}

\begin{center}
\begin{picture}(90,65.6)
\put(15.5,00){\makebox(0,0)[bl] {Fig. 4.16. 6-component team
  \(T_{3}\)}}

\put(00,09){\makebox(0,8)[bl]{\(A_{13}(1)\)}}
\put(00,13){\makebox(0,8)[bl]{\(A_{10}(1)\)}}
\put(00,17){\makebox(0,8)[bl]{\(A_{9}(1)\)}}
\put(00,21){\makebox(0,8)[bl]{\(A_{6}(1)\)}}
\put(00,25){\makebox(0,8)[bl]{\(A_{4}(2)\)}}
\put(00,29){\makebox(0,8)[bl]{\(A_{3}(2)\)}}

\put(15,05){\makebox(0,8)[bl]{\(A_{13}(2)\)}}
\put(15,09){\makebox(0,8)[bl]{\(A_{12}(2)\)}}
\put(15,13){\makebox(0,8)[bl]{\(A_{10}(1)\)}}
\put(15,17){\makebox(0,8)[bl]{\(A_{9}(1)\)}}
\put(15,21){\makebox(0,8)[bl]{\(A_{8}(2)\)}}
\put(15,25){\makebox(0,8)[bl]{\(A_{6}(2)\)}}
\put(15,29){\makebox(0,8)[bl]{\(A_{4}(2)\)}}

\put(30,09){\makebox(0,8)[bl]{\(A_{13}(1)\)}}
\put(30,13){\makebox(0,8)[bl]{\(A_{12}(2)\)}}
\put(30,17){\makebox(0,8)[bl]{\(A_{10}(1)\)}}
\put(30,21){\makebox(0,8)[bl]{\(A_{9}(1)\)}}
\put(30,25){\makebox(0,8)[bl]{\(A_{8}(2)\)}}
\put(30,29){\makebox(0,8)[bl]{\(A_{6}(1)\)}}

\put(45,05){\makebox(0,8)[bl]{\(A_{14}(2)\)}}
\put(45,09){\makebox(0,8)[bl]{\(A_{13}(1)\)}}
\put(45,13){\makebox(0,8)[bl]{\(A_{12}(2)\)}}
\put(45,17){\makebox(0,8)[bl]{\(A_{10}(2)\)}}
\put(45,21){\makebox(0,8)[bl]{\(A_{9}(1)\)}}
\put(45,25){\makebox(0,8)[bl]{\(A_{6}(2)\)}}
\put(45,29){\makebox(0,8)[bl]{\(A_{1}(2)\)}}

\put(60,17){\makebox(0,8)[bl]{\(A_{13}(1)\)}}
\put(60,21){\makebox(0,8)[bl]{\(A_{10}(2)\)}}
\put(60,25){\makebox(0,8)[bl]{\(A_{9}(1)\)}}
\put(60,29){\makebox(0,8)[bl]{\(A_{6}(2)\)}}

\put(75,05){\makebox(0,8)[bl]{\(A_{13}(1)\)}}
\put(75,09){\makebox(0,8)[bl]{\(A_{10}(1)\)}}
\put(75,13){\makebox(0,8)[bl]{\(A_{9}(1)\)}}
\put(75,17){\makebox(0,8)[bl]{\(A_{8}(2)\)}}
\put(75,21){\makebox(0,8)[bl]{\(A_{6}(1)\)}}
\put(75,25){\makebox(0,8)[bl]{\(A_{4}(2)\)}}
\put(75,29){\makebox(0,8)[bl]{\(A_{3}(2)\)}}

\put(3,35){\circle{2}} \put(18,35){\circle{2}}
\put(33,35){\circle{2}} \put(48,35){\circle{2}}
\put(63,35){\circle{2}} \put(78,35){\circle{2}}

\put(3,40){\line(0,-1){04}} \put(18,40){\line(0,-1){04}}
\put(33,40){\line(0,-1){04}} \put(48,40){\line(0,-1){04}}
\put(63,40){\line(0,-1){04}} \put(78,40){\line(0,-1){04}}

\put(3,40){\line(1,0){75}} \put(06,40){\line(0,1){22}}

\put(06,62){\circle*{2.4}}

\put(4,36.5){\makebox(0,8)[bl]{\(O_{1}\)}}
\put(13.5,36.5){\makebox(0,8)[bl]{\(O_{2}\)}}
\put(28.5,36.5){\makebox(0,8)[bl]{\(O_{3}\)}}
\put(43.5,36.5){\makebox(0,8)[bl]{\(O_{4}\)}}
\put(58.5,36.5){\makebox(0,8)[bl]{\(O_{5}\)}}
\put(73.5,36.5){\makebox(0,8)[bl]{\(O_{6}\)}}

\put(09,61){\makebox(0,8)[bl]{\(T_{3} = O_{1}\star O_{2}\star
 O_{3}\star O_{4} \star O_{5} \star O_{6}\)}}

\put(08,56){\makebox(0,8)[bl] {\(T^{1}_{3}=
 A_{1}\star A_{2}\star A_{2}\star A_{2}\)}}

\put(08,51){\makebox(0,8)[bl] {\(T^{2}_{3}=
 A_{1}\star A_{2}\star A_{2}\star A_{2}\)}}

\put(08,46){\makebox(0,8)[bl] {\(T^{3}_{3}=
 A_{1}\star A_{2}\star A_{2}\)}}

\put(08,41){\makebox(0,8)[bl] {\(T^{4}_{3}=
 A_{1}\star A_{2}\star A_{2}\)}}

\end{picture}
\end{center}

 Note, the system problem
 to design a trajectory of composite teams
 (or multistage team design) is very prospective one
 \cite{lev15}.

\subsubsection{Analysis of Network}

 Multi-type clustering strategy can be applied
 for analysis in networks.
 The types of networks nodes can be obtained by analysis of their
 connections (the number and types of neighbors):
 (a) multi-connected nodes (type 1),
 (b) connected nodes (type 2),
 (c) outliers (type 3),
 (d) isolated nodes (type 4).

 Let \(G = (A,E)\) be the examined network (graph),
 where
  \(A = \{A_{1},...,A_{i},...,A_{n}\}\) is the set of nodes,
  \(E\) is the set of edges (\(|E| = h\)).
 The following clustering scheme can be considered:

~~~

 {\it Stage 1.} Building the list of nodes (with info on neighbors)
 \(O(n)\)

 {\it Stage 2.} Selection of multi-neighbor nodes
 (type 1)
 (complexity estimate equals \(O(n)\)).
 Result: subset \(B_{1} \subset A\).

 {\it Stage 3.} Selection of outlier nodes (i.e., leaves, type 3)
 (complexity estimate equals \(O(n)\)).
 Result: subset \(B_{3} \subset \{A \ B_{1}\}\).

 {\it Stage 4.} Selection of other nodes (type 2)
  (complexity estimate equals \(O(n)\)).
 Result: subset \(B_{2} \subset A\),
 \(B_{2} =  \{A\ \{B_{1}\bigcup B_{3} \}\).

 {\it Stage 5.} Clustering of multi-neighbor nodes
  \(B_{1}\) (complexity estimate equals about
  \( O ( | B_{1} |^{2} )\)
  (about \(O((n/3)^{2})\)).
  Thus, a preliminary clustering solution is:
  \(\widehat{X}^{1}=\{X^{1}_{1},...,X^{1}_{l},...,X^{1}_{q_{1}}\}\).
 Now, a macro-network can be built:
 \(G^{1} = (\widehat{X}^{1}, E^{1})\),
 where
 \(\widehat{X}^{1}\) is the node set that corresponds
 to the obtained clustering solution (i.e., set of clusters),
 \(E^{1}\) is a built set of edges.
 (Note,
 the obtained clusters can be used as centroids
 in k-means clustering at the next step(s)).

 {\it Stage 6.} Clustering of nodes of type 2,
 i.e., set \(B_{2}\) (if needed).
 The corresponding clustering solution is:
  \(\widehat{X}^{2}=\{X^{2}_{1},...,X^{2}_{l},...,X^{2}_{q_{2}}\}\).
 Now, a macro-network can be built:
 \(G^{2} = (\widehat{X}^{2}, E^{2})\),
 where
 \(\widehat{X}^{2}\) is the node set that corresponds
 to the obtained clustering solution (i.e., set of clusters),
 \(E^{2}\) is a built set of edges.

 {\it Stage 7.} Matching of two graphs
  \(G^{1} = (\widehat{X}^{1}, E^{1})\)
  and
  \(G^{2} = (\widehat{X}^{2}, E^{2})\).
 The matching process can be based
 on edges from initial network
 or
 on the usage of additional parameters
 (e.g., node coordinates).
 As a result,
 integrated clustering solution
 can be obtained
 \(\widehat{X}^{12}=\{X^{12}_{1},...,X^{12}_{l},...,X^{12}_{q_{12}}\}\).

 {\it Stage 8.} Joining outliers  (\(B_{3}\)) to
 clusters of solution
 \(\widehat{X}^{12}\).
 As a result,
 integrated clustering solution can be obtained
 \(\widehat{X}^{123} \).
%


\subsection{Scheduling in multi-beam antenna
 communication system}

 There are the following initial information
 (Fig. 4.17):

 (a) multi-beam antenna system,

 (b) number of antenna beams: \(\mu\),

 (c) set of communication nodes \(A =
 \{A_{1},...,A_{i},...,A_{n}\}\),

 (d) volume of transmitted data is about the same for each \(A_{i}\).

 The problem is:

~~~

 Design a schedule
 for connection of antenna system to communication nodes
 while taking into account the following:
 ~(i) minimization of total connection time,
 ~(ii) providing  the best communication quality (by the minimum
 interference between neighbor (by angle) connections, i.e.,
 \[\max_{i\in A} ~\min_{i_{1},i_{2} \in A} ~D^{angular} (A_{i_{1}},A_{i_{2}}),\]
 where
 \(D^{angular} (A_{i_{1}},A_{i_{2}})\)
 is  angular separation
 \(D^{angular} (A_{i_{1}},A_{i_{2}})\)
 (section 2.2.4)
 (or angle between beams to the nodes).

~~~

 The solving scheme is the following:

 {\it Stage 1.} Linear ordering of communication nodes by their angle
 (Fig. 4.17, node 1 is the 1st).

 {\it Stage 2.} Dividing of the obtain
 list of nodes
 into \(\mu \) equal by size groups
 (the last group can have less elements)
  and numeration as follows:

 group \(1\):
 \(\{(1,1),(1,2),...,(1,k)\}\),

 group \(2\):
 \(\{(2,1),(2,2),...,(2,k)\}\),

 ...

  group \(\mu\):
 \(\{(\mu ,1),(\mu,2),...,(\mu,k)\}\).

 Here
 \(k =\lceil  \frac {n}{\mu } \rceil \).

 {\it Stage 3.} Generation of scheduling
 by the rules:
 Slot \(j\) (\(j=\overline{1,k}\)):
 the \(j\)-th element from each group
 (\(\zeta = 1,2...\mu\)), i.e.,
 elements \(\{\zeta,j)\}\)
 (Fig. 4.17).

 {\it Stage 4.} Stop.

~~~

\begin{center}
\begin{picture}(75,85)
\put(05,0){\makebox(0,0)[bl]{Fig. 4.17.
 Multi-beam antenna system}}

\put(01.6,70){\makebox(0,0)[bl]{Slot}}
\put(04,67){\makebox(0,0)[bl]{\(1\)}}

\put(11.6,70){\makebox(0,0)[bl]{Slot}}
\put(14,67){\makebox(0,0)[bl]{\(2\)}}

\put(28,68){\makebox(0,0)[bl]{{\bf ...}}}

\put(41.6,70){\makebox(0,0)[bl]{Slot}}
\put(44,67){\makebox(0,0)[bl]{\(k\)}}


\put(00,40){\makebox(0,0)[bl]{\(0\)}}
\put(71,44){\makebox(0,0)[bl]{\(t\)}}

\put(00,45){\vector(1,0){70}}

\put(00,65){\line(1,0){10}}
\put(01,61){\makebox(0,0)[bl]{\((1,1)\)}}
\put(00,55){\line(1,0){10}} \put(00,60){\line(1,0){10}}

\put(01,56){\makebox(0,0)[bl]{\((2,1)\)}}
\put(00,55){\line(1,0){10}}

\put(02,52){\makebox(0,0)[bl]{. . .}}

\put(00,50){\line(1,0){10}}
\put(01,46){\makebox(0,0)[bl]{\((\mu,1)\)}}
\put(10,65){\line(1,0){10}}
\put(11,61){\makebox(0,0)[bl]{\((1,2)\)}}
\put(10,55){\line(1,0){10}} \put(10,60){\line(1,0){10}}

\put(11,56){\makebox(0,0)[bl]{\((2,2)\)}}
\put(10,55){\line(1,0){10}}

\put(12,52){\makebox(0,0)[bl]{. . .}}

\put(10,50){\line(1,0){10}}
\put(11,46){\makebox(0,0)[bl]{\((\mu,2)\)}}
\put(40,65){\line(1,0){10}}
\put(41,61){\makebox(0,0)[bl]{\((1,k)\)}}
\put(40,55){\line(1,0){10}} \put(40,60){\line(1,0){10}}

\put(41,56){\makebox(0,0)[bl]{\((2,k)\)}}
\put(40,55){\line(1,0){10}}

\put(42,52){\makebox(0,0)[bl]{. . .}}

\put(40,50){\line(1,0){10}}
\put(41,46){\makebox(0,0)[bl]{\((\mu,k)\)}}
\put(20,75){\makebox(0,0)[bl]{Period \(T\)}}
\put(25,74){\vector(-1,0){25}} \put(25,74){\vector(1,0){25}}

\put(00,43){\line(0,1){32}} \put(10,44){\line(0,1){22}}
\put(20,44){\line(0,1){22}} \put(40,44){\line(0,1){22}}
\put(50,43){\line(0,1){32}}



\put(02,34){\makebox(0,0)[bl]{Multi-beam}}
\put(04,31){\makebox(0,0)[bl]{antenna}}

\put(25,35){\circle*{1.5}} \put(25,35){\circle{2.5}}

\put(25,35){\vector(-1,-1){8}} \put(25,35){\vector(-1,-2){4}}
\put(25,35){\vector(0,-1){8}} \put(25,35){\vector(1,-2){4}}
\put(25,35){\vector(1,-1){8}} \put(25,35){\vector(2,-1){14}}

\put(16,26){\line(-1,-1){10}} \put(40,27.5){\line(2,-1){23.6}}

\put(05,15){\circle*{1.5}} \put(35,15){\circle*{1.5}}
\put(65,15){\circle*{1.5}}


\put(00,11){\makebox(0,0)[bl]{Node \(1\)}}

\put(17,14){\makebox(0,0)[bl]{{\bf ...}}}

\put(30,11){\makebox(0,0)[bl]{Node \(i\)}}

\put(47,14){\makebox(0,0)[bl]{{\bf ...}}}

\put(60,11){\makebox(0,0)[bl]{Node \(n\)}}


\put(09,08){\oval(12,4)} \put(05,08){\circle*{1.5}}
\put(07.5,07.7){\makebox(0,0)[bl]{...}} \put(13,08){\circle*{1.5}}
\put(22,08){\oval(12,4)} \put(18,08){\circle*{1.5}}
\put(20.5,07.7){\makebox(0,0)[bl]{...}} \put(26,08){\circle*{1.5}}
\put(35,08){\oval(12,4)} \put(31,08){\circle*{1.5}}
\put(33.5,07.7){\makebox(0,0)[bl]{...}} \put(39,08){\circle*{1.5}}

\put(46.5,07.7){\makebox(0,0)[bl]{{\bf ...}}}

\put(61,08){\oval(12,4)} \put(57,08){\circle*{1.5}}
\put(59.5,07.7){\makebox(0,0)[bl]{...}} \put(65,08){\circle*{1.5}}

\end{picture}
\end{center}

\newpage
\section{Conclusions}

 The paper addresses
%
 combinatorial modular viewpoints to
 clustering problems and procedures:
%
 (a) generalized modular clustering frameworks
 (i.e., typical combinatorial engineering frameworks);
 (b) main structural clustering models/methods
 (e.g., hierarchical clustering,
 minimum spanning tree based clustering,
 clustering as assignment,
 detection of clisue/quasi-clique based clustering,
 correlation clustering,
 network communities based clustering);
 (c)  main ideas for fast clustering schemes.
%
%
%
 Described problem solving frameworks are based on
 compositions
   of well-known
 combinatorial optimization problems and corresponding algorithms
 (e.g., assignment, partitioning, assignment,
 knapsack problem,
 multiple choice problem,
  morphological clique problem,
 searching for consensus/median for structures).

 A set of
  numerical examples
 illustrate all stages of clustering processes
 (problem statement,
 assessment and evaluation,
 design of solving algorithms/schemes,
 analysis of results).
%


%
 Future research directions can involve the following:

 {\it 1.} analysis and design of new composite problems and
  modular solving frameworks;

 {\it 2.} design of special decision support tools
 (modular solving environments) for structural clustering problems;

 {\it 3.} special study of dynamic structural clustering problems and
 and very large applications;

%

%
 {\it 4.} applications of  structural
 clustering problems in networking
 (design, covering, routing, etc.); and

 {\it 5.} application of the structural
 clustering problems in
 education and in educational data mining.

\section{Acknowledgments}

 This research (without sections 4.2)
 was partially supported by Russian Science Foundation
 grant 14-50-00150 ``Digital technologies and their applications''
 (project of Inst. for Information Transmission Problems).
 The research materials presented in  sections 4.2
 were partially supported by The Russian Foundation for
 Basic Research,
 project 15-07-01241
 ``Reconfiguration of Solutions in Combinatorial Optimization''
 (principal investigator: Mark Sh. Levin).

 The author thanks Prof. Andrey I. Lyakhov for preliminary
 engineering description of problem:
 design of communication schedule for multiple beam antenna
 (section 4.4).

\newpage



\begin{thebibliography}{800}

 \bibitem {abbas07} A.A. Abbasi, M. Younis,
 A survey on clustering algorithms for wireless sensor networks.
  Computer Communications 30(14), 2826--2841, 2007.

 \bibitem {abe02} J. Abello, M.G.C. Resende, S. Sudarsky,
 Massive quasi-clique detection. In:
 S. Rajsbaum (ed),
 Proc. of 5th Latin American Symp. on
 Theor. Inform.
 LATIN 2002,
 LNCS 2286, Springer,
  598--612, 2002.


 \bibitem {abu13} B. Abu-Jamous, R. Fa, D.J. Roberts, A.K. Nandi,
 Paradigm of tunable clustering using binarization of
 consensus partition matrices (Bi-CoPaM)
 for gene discovery.
 PLOS ONE 8(2), 1--14, 2013.

 \bibitem {achtert07} E. Achtert, C. Bohm, H.-P. Kriegel,
 P. Kroger, A. Zimek,
 Robust, complete, and efficient correlation clustering.
 In: Proc. SIAM Int. Conf. on Data Mining (SDM),
 413--418, 2007.

  \bibitem {achtert08} E. Achtert, C. Bohm, J. David,
 P. Kroger, A. Zimek,
 Global correlation clustering based on the hough transform.
 Statistical Analysis and Data Mining 1, 111--127, 2008.

 \bibitem {acker08} M. Ackerman, S. Ben-David,
 Measures of clustering quality:
 A working set of axioms for clustering.
 In:
 Advances in Neural Information Processing Systems (NIPS),
 MIT Press,
 121--128, 2008.

  \bibitem {agar08} G. Agarwal, D. Kempe,
  Modularity maximizing network communitites using
  mathematical programming.
  The European Physical Journal B, 66, 4009--418, 2008.

 \bibitem {aggar07} C.C. Aggarwal (ed.),
  Data Streams: Models and Algorithms.
  Springer, New York, 2007.

  \bibitem {agrawal05} R. Agrawal, J. Gehrke, D. Gunopulos,
  P. Raghavan,
  Automatic subspace clustering of high dimensional data.
  Data Mining and Knoweldge Discovery,
  11(5), 5--33, 2005.

  \bibitem {aho74} A.V. Aho, J.E. Hopcroft, J.D. Ullman,
  The Design and Analysis of Computer Algorihtms.
  Addison Welsey, Reading, MA, 1974.

 \bibitem {ailon08} N. Ailon, M. Charikar, A. Newman,
 Aggregating inconsistent information:
 Ranking and clustering.
 J. of the ACM, 55(5), art. No. 23, 2008.

  \bibitem {akatsu00} T. Akatsu, M.M. Halldorsson,
 On the approximation of largest common subtrees
 and largest common point sets.
 Theoretical Computer Science 233(1-2), 33--50, 2000.

  \bibitem {akko73} E. Akkoyunlu,
 The enumeration of maximal cliques of large graph.
 SIAM J. on Computing 2(1), 1-6, 1973.

 \bibitem {al95} K. Al-Sultan,
 A Tabu search algorithm to the clustering problem.
  Pattern Recogn.
 28(9), 1443--1451, 1995.

 \bibitem {alba07} E. Alba, G. Luque, J. Garcia-Nieto,
 MALLBA: a software library to design efficient optimization
 algorithms.
  Int. J. of Innovative Comput. \& Appl. 1(1), 74--85, 2007.

 \bibitem {alev09} D. Alevras,
 Assignment and matching.
 In:
  C.A. Floudas, P.M. Pardalos (eds.),
  Encyclopedia of Optimization. 2nd ed.,
  Springer, pp. 106--108, 2009.

 \bibitem {alexe06} G. Alexe, S. Alexe, P.L. Hammer,
 Pattern-based clustering and attribute analysis.
 Soft Computing 10(5), 442--452, 2006.

 \bibitem {alon98} N. Alon, M. Krivelevich, B. Sudakov,
 Finding a large hidden clique in a random graph.
 In: Proc. of the Ninth Annual ACM-SIAM Symp. on
 Discr. Algorithms, SIAM, 594--598, 1998.

 \bibitem {alp95} C.J. Alpert, A.B. Kahng,
 Recent directions in netlist partitioning: a survey.
  Integration, The VLSI Journal 19(1), 1--81, 1995.


  \bibitem {amir97} A. Amir, D.  Keselman,
   Maximum agreement subtree of a set of evolutionary trees -
   metrics and efficient algorithms.
    SIAM J. on Comp. 26(6), 1656--1669, 1997.

 \bibitem {an01} B. An, S. Papavassiliou,
 A mobility-based clustering approach to
 support mobility management and multicast routing in mobile
 ad-hoc wireless networks.
 Int. J. of Network Manag.
 11(6), 387--395, 2001.

 \bibitem {and73} M.R. Anderberg,
 Cluster Analysis for Applications.
 Academic Press, New York, 1973.

 \bibitem {arabie81} P. Arabie, J.D. Carrol,
 W.S. DeSarbo, J. Wind,
 Overlapping clustering:
 A new method for product positioning.
 J. of Marketing Research 18, 310-317, 1981.

 \bibitem {arabie94} P. Arabie, L.J. Hubert,
 Cluster analysis in marketing research. In:
 Advanced Methods in marketing Research.
 Blackwell, Oxford, 160--189, 1994.

  \bibitem {arabie96} P. Arabie, L.J. Hubert,
  G. De Soete (Eds.),
  Clustering and Classification.
 World Scientific, 1996.

 \bibitem {aug04} C.J. Augeri, H.H. Ali,
 New graph-based algorithms for partitioning
 VLSI circuits.
 In: Proc. of the 2004 Int. Symp. on Circuits and Systems
 ISDAS'04, vol. 4, pp. 521--524, 2004.

 \bibitem {aug70} J.G. Augston, J. Minker,
 An analysis of some graph theoretical clustering techniques.
 J. of the ACM 17(4), 571--588, 1970.

  \bibitem {ayad10} H. Ayad, M.S. Kamel,
 On voting-based consensus of cluster ensembles.
 Pattern Recognition 43(5), 1943--1953, 2010.

 \bibitem {babel94} L. Babel,
 A fast algorithm for the maximum weight
 clique problem.
 Computing 52, 31--38, 1994.

 \bibitem {babu94} G. Babu, M. Nurty,
 Clustering with evolution strategy.
 Pattern Recogn. Lett. 14(10), 763--769, 1993.

 \bibitem {bach06} F. Bach, M.I. Jordan,
  Learning spectral clustering with applications to
  speach separation.
  J. of Machine Learning Research 7, 1963--2001, 2006.

 \bibitem {bae99} R. Baeza-Yates, B. Ribeiro-Neto,
 Modern Information Retrieval.
 Addison-Wesley, 1999.

 \bibitem {bagon11} S. Bagon, M. Galun,
 Optimizing large scale correlation clustering.
   Electronic preprint, 9 p.,
 Dec. 13, 2011.
 http://arxiv.org/abs/1112.2903 [cs.CV]

 \bibitem {balas87} E. Balas, V. Chvatal, J. Nesetril,
 On teh maximum weight clique problem.
 Math. Oper. Res. 12(3), 522--535, 1987.

 \bibitem {baldi02} P. Baldi, G. Hatfield,
 DNA Microarrays and Gene Expression.
 Cambridge Univ. Press, 2002.


  \bibitem {band03} S. Bandyopadhyay, E.J. Coyle,
 An energy efficient hierarchical clustering algorithm for
 wireless sensor networks.
 In: Proc. of INFOCOM 2003,
 vol. 3, 1713--1723, 2003.

  \bibitem {band04} S. Bandyopadhyay, E.J. Coyle,
 Minimizing communication costs in hierarchically-clustered
 networks of wireless sensors.
 Computer Networks 44(1), 1--16, 2004.

  \bibitem {baner01} S. Banerjee, S. Khuller,
 A clustering scheme for hierarchical control in multi-hop
 wireless networks. In:
 Proc. of Twentieth Annual Joint Conf. of the IEEE
 Computer and Communications Societies INFOCOM 2001, vol. 2,
 1028--1037, 2001.


 \bibitem {bansal02} N. Bansal, A. Blum, S. Chawla,
 Correlation clustering.
 In: Proc. of Int. Conf. FOCS, 2002,
 238--250, 2002.

  \bibitem {bansal04} N. Bansal, A. Blum, S. Chawla,
 Correlation clustering.
 Machine Learning 56(1-3), 89--113, 2004.

 \bibitem {bansal09} M.S. Bansal, D. Fernandez-Baca,
 Computing distances between partial rankings.
  Inform. Proc. Lett. 109(4), 238--241, 2009.

 \bibitem {bar99} A. Baraldi, P. Blonda,
 A survey of fuzzy clustering algorithms for pattern
 recognition - Part I and II. IEEE Trans. SMC,
 Part B, 29(6), 778--801, 1999.   

 \bibitem {barb08} W. Barbakh, C. Fife,
 Online clustering algorithms.
 Int. J. of Neural Systems 18(03), 185--194, 2008.

 \bibitem {bar86} J.-P. Barthelemy, B. Leclerc,
 B. Monjardet,
 On the use of ordered sets in problems of comparison and
 consensus of classifications.
 J. of Classification, 3(2), 17--24, 1986.

 \bibitem {bata03} V. Batagelj, M. Zavershik,
 An O(m) algorithm for cores decomposition of networks.
 Electronic preprint, 10 p., Oct. 25, 2003.
   http://arxiv.org/abs/0310.0049 [cs.DS]

 \bibitem {bata14} V. Batagelj, P. Doreian, A. Ferlgoj,
 N. Kejzar, 
 Understanding Large Temporal Networks and Spatial Networks:
 Exploration, Pattern Searching, Visualization and Network
 Evolution.
 Wiley, 2014.

  \bibitem {beck83} M.P. Beck, B.W. Lin,
 Some heuristics for the consensus ranking problem.
 Computers and Operations Research
 10(1), 183, 1--7, 1983.

 \bibitem {bens11} A. Benslimane, T. Taleb, R. Sivaraj,
 Dynamic clustering-based adaptive mobile gateway management in
 integrated VANET -3G heterogeneous wireless networks.
 IEEE J. on Selected Areas in Communications
 29(3), 559--570, 2011.

 \bibitem {berry96} M.J.A. Berry, G. Linoff,
 Data Mining Techniques for Marketing, Sales and Customer
 Support.  Wiley, 1996.     

  \bibitem {berry99} M.W. Berry, M. Browne,
  Understanding Search Engines: Mathematical Modeling
  and Text Retrieval.
  SIAM, 1999.

  \bibitem {ben06} S. Ben-David, U. von Luxburg, D. Pal,
  A sober look at clustering stability.
  In:
  Proc. of 19th Annual Conf. on Machine Learning COLT 2006,
  Springer, Berlin, 5--19, 2006.

  \bibitem {ben99} A. Ben-Dor, R. Shamir, Z. Yakhini,
 Clustering gene expression patterns.
 J. of Computational Biology 6(3–4), 281--292, 1999.

 \bibitem {ben01} A. Ben-Hur, D. Horn, H. Siegelman, V. Vapnik,
 Support vector clustering.
 J. Mach. Learn. Res. 2(0), 125--137, 2001.

 \bibitem {berin06} J. Beringer, E. Hullermeier,
 Online clustering of parallel data streams.
 Data \& Knowledge Engineering 58(2), 180--204, 2006.

  \bibitem {berk06} P. Berkhin,
 A survey of clustering data mining techniques.
 In: Grouping Multidimensional Data, Springer, 25--71, 2006.

 \bibitem {bezdek98} J. Bezdek, N. Pal,
 Some new indexes of cluster validity.
 IEEE Trans. SMC, Part B, 28(3), 301--315, 1998.

 \bibitem {bid09} M. Biddick,
 Cluster grouping for the gifted and talented: It works!
 APEX 15(4), 78--86, 2009.

 \bibitem {billard07} L. Billard, E. Diday,
 Symbolic Data Analysis. Wiley, 2007.

  \bibitem {blond08} V.D. Blondel, J.-L. Guillaume, R. Lambiotte,
 E. Lefebvre,
 Fast unfolding of communities in large networks.
  Electronic preprint. 12 p., July 25, 2008.
    http://arxiv.org/abs/0803.0476 [physics.soc-ph]

  \bibitem {blond12} V.D. Blondel,
 M. Esch, C. Chan, F. Clerot, P. Deville, E. Huens,
 F. Morlot, Z. Smoreda, C. Ziemlicki,
 Data for development the d4d challenge on mobile phone data.
  Electronic preprint. 10 p., Jan. 28, 2012.
    http://arxiv.org/abs/1210.0137 [cs.CY]

 \bibitem {boc06} S. Boccaletti, V. Latora,
 Y. Moreno, M. Chavez, D.-U. Hwang,
 Complex netwoprks:
 Structure and dynamics.
 Physics Reports 424, 175---208, 2006.

  \bibitem {bock96} H. Bock,
  Probabilistic models in cluster analysis.
  Comput. Statist. Data Anal. 23, 5--28, 1996.

 \bibitem {bock00} H.H. Bock, E. Diday (eds),
 Analysis of Symbolic Data.
 Springer, Heidelberg, 2000.

  \bibitem {boker09} S. Boker, S. Briesemeister, Q.B.A. Bui,
 A. Truss,
 Going weighted:
 Parametrized algorithms for cluster editing.
 Theor. Comput. Sci. 410, 5467--5480, 2009.

  \bibitem {boker11a} S. Boker, P. Damschke,
  Even faster parametrized clsuter deletion and clsuter editing.
  Inf. Process. Lett. 111(14), 717--721, 2011.

 \bibitem {boker11} S. Boker, S. Briesemeister, G.W. Klau,
 Going weighted:
 Exact algoroithms for cluster editing:
 Evaluation and experiments.
 Algorithmica, 60(2), 316--334, 2011.

 \bibitem {boley99} D. Boley, M. Gini, R. Gross,
 S. Han, K. Hastings, G. Kapyris, V. Kumar,
 B. Mobasher, J. Moor,
 Partitioning-based clustering of web document categorization.
 DSS 25(3), 329--341, 1999.

  \bibitem {bomze99} I.M. Bomze, M. Budinich,
 P.M. Pardalos,  M. Pelillo,
 The maximum clique problem.
 In: D.-Z. Du, P.M. Pardalos (Eds.),
 {\it Handbook of Combinatorial Optimization}.
 (Supp. vol. A), Springer, New York, 659-729, 1999.

 \bibitem {boor73} S.A. Boorman, D.C. Oliver,
 Metrics on spaces of finite tress.
 J. of Math. Psychology 10(1), 26--59, 1973

 \bibitem {borg05} I. Borg, P.J.F. Groenen,
 Modern Multidimensional Scaling:
 Theory and Applications. 2nd ed.,
 Springer, New York, 2005.

  \bibitem {bou09} D. Bouyssou,
 Outranking methods.
 In:
  C.A. Floudas, P.M. Pardalos (Eds.),
  Encyclopedia of Optimization. 2nd ed., Springer,
   2887-2893, 2009.


 \bibitem {bra91} V.L. Brailovsky,
  A probabilistic approach to clustering.
  Pattern Recogn. Lett. 12(4), 193--198, 1991.

 \bibitem {brand08} U. Brandes, D. Delling, M. Gaertler,
 R. Gorke, M. Hoefer, Z. Nikolosk, D. Wagner,
 On modularity clustering.
 IEEE Trans KDE 20, 172--188, 2008.

 \bibitem {brin98} S. Brin, L. Page,
 The anatomy of a large-scale hypertextual Web search engine.
 Computer Networks and ISDN Systems 30(1-7),
  107--117, 1998.

 \bibitem {brod97}
  A.Z. Broder, S.C. Glassman,
 M.S. Manasse, G. Zweig,
 Syntactic Clustering of the Web.
 SRC Technical Note, 1997-015, DEC, 1997.

  \bibitem {bron73} C. Bron, J. Kerbosch,
 Algorithm 457: Finding all cliques of an undirected graph.
 Commun. of the ACM 16(9), 575--577, 1973.

 \bibitem {bron06} M.M. Bronstein, A.M. Bronstein,
 R. Kimmel, I. Yavneh,
 Multigrid multidimensional scaling.
 Numerical Linear Algebra with Applicaitons 13(2-3),
 149--171, 2006.


 \bibitem {brown92} D. Brown, C. Huntley,
  A practical application of simulated annealing to clustering.
 Pattern Recognit. 25(4), 401--412, 1992.




 \bibitem {burkard09} R. Burkard, M. Dell'Amico, S. Martello,
 Assignment Problems.
 SIAM, Providence, 2012. 

 \bibitem {burke94} E.K. Burke, D.G. Elliman, R.F. Weare,
 The automation of the timetabling process in higher education.
 J. of Educational Technology Systems 23(4), 353--362, 1994.

 \bibitem {but06} S. Butenko, W. Wilhelm,
 Clique-detection models
 in computational biochemistry and genomics.
 EJOR 173(1), 1--17, 2006.

 \bibitem {butt09} L. Buttyan, T. Holczer,
 Private cluster head election in wireless sensor networks.
 In: Proc. of IEEE 6th Int. Conf. on Mobile Adhoc and Sensor
 Systems MASS'09, 1048--1053, 2009.

 \bibitem {cai00} F. Cai, N.-A. LeKhac, M-T. Kechadi,
 Toward a new classificaiton model for analyzing financial
 datasets.
 In:
 S. Fong, P. Pichappan, S. Mohammed, P. Hung, S. Asghar (eds),
 Seventh Int. Conf. on Digital Information Management
 (ICDIM), IEEE Press,
 22-24 Aug., 2012, Macau, 1--6, 2012.

 \bibitem {camp13} T. Campbell, M. Liu, B. Kulis,
 J.P. How, L. Carin,
 Dynamic clustering via asymptotics of dependent
 Dirichlet process mixture.
     Electr. prepr. 9 pp., Nov. 1, 2013.
    http://arxiv.org/abs/1305.6659 [cs.LG]

 \bibitem {carp09} C. Carpineto, S. Osinski, R. Romano, D. Weiss,
 A survey of Web clustering engines.
 ACM Comput. Surv. 41, 1--38, 2009.

 \bibitem {carrol80} J.D. Carrol, P. Arabie,
 Multidimensional scaling.
  Annu. Rev. Psychol. 31, 607--649, 1980.

 \bibitem {catt92} D.G. Cattrysse, L.N. Van Wassenhove,
 A survey of algorithms for the generalized assignment
 problem. EJOR 60(3), 260--272, 1992.

  \bibitem {catt94} D.G. Cattrysse, M. Salomon,
  L.N. Van Wassenhove,
  A set partitioning heuristic
  for the generalized assignment problem.
  EJOR 72(00), 167--174, 1994.

 \bibitem {cavendish06} D. Cavendish, M. Gerla,
 Routing optimization in communication networks.
 In: M.X. Cheng, Y. Li, D.-Z. Du (eds),
 Combinatorial Optimization in Communication Networks.
 Springer, 505--547, 2006.

 \bibitem {cela98} E. Cela,
  The Quadratic Assignment Problem: Theory and Algorithms.
  Kluwer Academic Publishers, Dordrecht, 1998.

 \bibitem {cham09} A. Chamam, S. Pierre,
 On the planning of wireless sensor networks:
 Energy-efficient clustering under the joint routing and coverage
 constraints.
 IEEE Trans. Mobile Computing 8(8), 1077--1086, 2009.

 \bibitem {chan07} T.M. Chan, H. Zarrabi-Zadeh,
 A randomized algorithm for online unit clustering.
 In: Approximation and Online Algorithms. Springer, 121--131,
 2007.

 \bibitem {cha09} V. Chandola, A. Banerjee, V. Kumar,
 Anomaly detection: A survey.
 ACM Computing Surveys 41(3),
 Article no. 15, 2009.

 \bibitem {cgw03} M. Charikar, V. Guruswami,
 A. Wirth,
 Clustering with quantitative information.
 In: Int. Conf. FOCS 2003,524--533, 2003.

 \bibitem {cgw05} M. Charikar, V. Guruswami,
 A. Wirth,
 Clustering with quantitative information.
 J. of Comput. Syst, Sci.
 71(3), 360--383, 2005.

 \bibitem {char08} I. Charon, O. Hundry,
  Optimal clustering in multipartite graph.
  Discrete Applied Mathematics 156(8), 1330--1347, 2008.

   \bibitem {cha05} F. Chataigner,
   Approximating the maximum agreement forest on k trees.
 Information Processing Letters 93(5), 239--244, 2005.

 \bibitem {chatt02} M. Chatterjee, S.K. Das, D.A. Turgut,
 WCA: A weighted clustering algorithm for mobile Ad Hoc networks.
 Cluster Computing 5(2), 193--204, 2002.


  \bibitem {cha88} B. Chazelle,
 A functional approach to data structures and
 its use in multidimensional scaling.
 SIAM J. on Computing 17, 427--462, 1988.

 \bibitem {chen01} W. Chen,
 New algorithm for ordered tree-to-tree correction problem.
  J. of Algorithms 40(2), 135--158, 2001.

 \bibitem {chen02} Y.P. Chen, A.L. Liestman,
 Approximating minimum size weakly-connected dominating sets for
 clustering mobile ad hoc networks. In:
 Proc. of the 3rd ACM Int. Symp. on Mobile Ad Hoc Networking \&
 Computing, 165--172, 2002.

 \bibitem {chenc04} C.-Y. Chen, F. Ye,
 Particle swam optimization algorithm and its
 application to cluster analysis.
 In: Proc. of 2004 IEEE Int. Conf. on Networking, Sensing and
 Control, vol. 2, 789-794, 2004.

 \bibitem {chenh06} H. Chen, S. Megerian,
 Cluster sizing and head selection for efficient
 data aggregation and routing in sensor networks.
 In:
 Proc. of Wireless Communicaitons and Netowrking Conf. WCNC 2006,
 vol. 4, 2318--2323, 2006. 

 \bibitem {chenw04} W.-P. Chen, J.C. Hou, L. Sha,
 Dynamic clustering for acoustic target tracking in wireless
 sensor networks.
 IEEE Trans. Mobile Computing 3(3), 258--271,
 2004. 


 \bibitem {cheny05} Y.P. Chen, A.L. Liestman,
 Maintaining weakly-connected dominaitng sets for clustering ad
 hoc networks.
 Ad Hoc Networks 3, 629--642, 2005.

 \bibitem {chenp10} P. Chen, S. Redner,
 Community structure of the physical review
 citation network.
 J. of Informetrics 4(3), 278--290, 2010. 

  \bibitem {chen11} Y. Chen, J.H. Lv, F.L. Han, X.H. Yu,
 On the cluster consensus of discrete-time multi-agent systems.
 Systems and Control lettes 60, 517--523, 2011.

 \bibitem {nchen13} N. Chen, Z. Xu, M. Xia,
 Correlation coefficients of hesitant fuzzy sets
 and their applications to clustering analysis.
 Applied Mathematical Modelling 37(4), 2197--2211, 2013.

 \bibitem {chen13} Z. Chen, S.X. Xia, B. Liu,
 A robust fuzzy kernel clustering algorithm.
 Applied Mathematics \& Information Sciences
 7(3), 1005--1012, 2013. 

  \bibitem {nchen14} N. Chen, Z. Xu, M. Xia,
 Hierarchical hesitant fuzzy K-means clustering algorithm.
 Applied Mathematics 29(1), 1--17, 2014.

  \bibitem {cheng95} C.H. Cheng,
 A branch-and-bound clustering algorithm.
 IEEE Trans. SMC 25, 895--898, 1995.

  \bibitem {cheng98} T.W. Cheng, D.B. Goldgof, L.O. Hall,
 Fast fuzzy clustering.
 Fuzzy Sets and Systems 93(1), 49--56, 1998.

 \bibitem {cheng05} S.Y. Cheng, C.S. Lin, H.H. Chen, J.S. Heh,
 Learning and diagnosis of individual and class conceptual
 perspectives:
 an intelligent systems approach using clustering techniques.
 Computers \& Education 44(3), 257--283, 2005.

  \bibitem {cheng06} M.X. Cheng, Y. Li, D.-Z. Du (Eds.),
 Combinatorial Optimization in Communication Networks.
 Springer, 2006.

 \bibitem {cheu04} E.Y. Cheu, C. Keongg, Z. Zhou,
 On the two-level hybrid clustering algorithm.
 In: Proc. of Int. Conf. on Artificial Intelligence
 in Science and Technology, 138--142, 2004.

 \bibitem {cheung05} K.W. Cheung, H.C. So,
 A multidimensional scaling framework for
 mobile location using time-of-arrival measurements.
 IEEE Trans. Signal Processing 53(2), 460--470, 2005.

 \bibitem {cheungy05} Y. Cheung,
 Maximum weighted likelihood via rival
 penalized EM for density mixture clustering
 with automatic modle selection.
 IEEE Trans. KDE 17(6), 750--761, 2005. 

 \bibitem {chia03} J. Chiang, P. Hao,
 A new kernel-based fuzzy clustering approach:
 Support vector clustering with cell growing.
 IEEE Trans. Fuzzy Syst. 11(4), 518--527, 2003.

 \bibitem {chias04} C.-F. Chiasserini, I. Chlamtac,
 P. Monti, A. Nucci,
 An enrgy-fficient method for node assignment in cluster-based
  Ad Hoc networks.
 Wireless Networks 10(3), 223--231, 2004.


 \bibitem {chiu02}  C. Chiu,
 A case-based customer classification approach for direct marketing.
 ESwA
   22(2), 163--168, 2002.

 \bibitem {choud95} D. Choudron,
 Organizational development:
 how to improve cross-functional teams.
 HR Focus 72(8), 1--4, 1995. 

 \bibitem {chu89} C.H. Chu,
 Cluster analysis in manufactruring cellular formation.
 Omega 17(3), 289--295, 1989.

 \bibitem {chu97} P.C. Chu, J.E. Beasley,
 A genetic algorithm for the generalized
 assignment problem.
 Computers and Oper. Res. 24(1), 17--23, 1997.

 \bibitem {chu12} C.-W. Chu, J.D. Holliday, P. Willett,
 Combining multiple classification of chemical structures using
 consensus clustering.
 Bioorganic \& Medicinal Chemistry 20(18), 5366-5371, 2012.

 \bibitem {clark94} M. Clark, L. Hall, C. Li, D. Goldgof,
 Knowledge based  (re-)clustering.
 In: Proc. 12th IAPR Int. Conf. Pattern Recognition,
  245--250, 1994.

 \bibitem {clark98} M.C. Clark, L.O. Hall, D.B. Goldgof,
 R. Velthuizen, F.R. Murtagh, M.S. Silbiger,
 Automatic timor segmentation using knowledge-based techniques.
 IEEE Trans. on Medical Imaging 17(2), 187-201, 1998 

 \bibitem {clay04} A. Clauset, M. E. J. Newman, and C. Moore,
 Finding community structure in very large networks.
  Physical Review E, vol. 70, no. 066111, 2004.

 \bibitem {coble06}  J. Coble, D.J. Cook, L.B. Holder,
 Structure discovery in sequentially-connected data streams.
 Int. J. on Artificial Intelligence Tools 15(6), 917--944, 2006.

 \bibitem {cobos11} C. Cobos, M. Mendoza, E. Leon,
 A hyper-heuristic approach to design and tuning
 heuristic methods for web document clustering.
 In: Proc. of 2011 IEEE Congress on Evol. Comput.
 (CEC), 1350--1358, 2011.

 \bibitem {cohen06} R. Cohen, L. Katzir, D. Raz,
 An efficient approximation for the generalized
 assignment problem,
 Information Proc. Lett. 100(4), 162--166, 2006.

 \bibitem {cok06} D. Cokuslu, K. Erciyes, O. Dagdeviren,
 A dominating set based clustering algorithm for
 mobile ad hoc networks. In:
 V.N. Alexandrov, G.D. van Albada, P.M.A. Sloot, J. Dongarra
 (eds),
 Proc. of Int. Conf. on Computational Sciences ICCS 2006,
 LNCS 3991, Springer, 571--578, 2006.

 \bibitem {cok07} D. Cokuslu, K. Erciyes,
 A hierarchical connected dominating set based
 clustering algorithm for
 mobile ad hoc networks. In:
 Proc. of 15th Int. Symp. on
 Modeling, Analysis and Simulation of Computer and
 Telecommunication Systems MASCOTS'07,
  60--66, 2007.

 \bibitem {comm88} C.L. Comm, D.F.X. Mathaisel,
 College course scheduling. A market computer software support.
 J. of Research of Computing in Education, 21, 187--195, 1988.


 \bibitem {condon01} A. Condon, R.M. Karp,
 Algorithms for graph partitining on the planted partition model.
 Random Structures and Algorithms 18, 116--140, 2001.

 \bibitem {cook92} W.P. Cook, M. Kress,
 Ordinal Information and Preference Structures:
 Decision Models and Applications.
  Prentice-Hall, Englewood Cliffs, 1992.

  \bibitem {cook96} W.D. Cook, L.M. Seiford, M.Kress,
  A general framework for distance-based consensus in
  ordinal ranking models.
  EJOR
   96(2) 392--397, 1996.

 \bibitem {cook00} D.J. Cook, L.B. Holder,
 Graph-based data mining.
 IEEE Intelligent Systems 15(2), 32--41, 2000.

  \bibitem {coppi02} R. Coppi, P. D'Urso,
  Fuzzy K-means clustering models for triangular
  fuzzy time trajectories.
  Statist. Methods Appl. 11(1), 21--40, 2002.

 \bibitem {cormen90} T.H. Cormen, C.E. Leiserson, R.L. Rivest,
 Introduction to Algorithms. 3rd ed.,
 MIT Press and McGraw-Hill, 2009. 

 \bibitem {corn84} D.G. Corneil, Y. Perl,
 Clustering and domination in perfect graphs.
 Discrete Applied Mathematics 9(1), 27--39, 1984.

 \bibitem {costa06} J.A. Costa, N. Patwari, A.O. Hero III,
 Distributed weighted multidimensional scaling for
 node localization in sensor networks.
 ACM Trans. on Sensor Networks 2(1), 39--64, 2006.

 \bibitem {cow99} M.C. Cowgill, R.J. Harvey, L.T. Watson,
 A genetic algorithm approach to cluster analysis.
 Comput. Math. Appl. 37(7), 99--108, 1999.

 \bibitem {cox00}  T.F. Cox, M.A.A. Cox,
  Multidimensional Scaling.
 CRC Press, 2000.


 \bibitem {crespo05} F. Crespo, R. Weber,
 A methodology for dynamic data mining based
 on fuzzy clustering.
 Fuzzy Sets and Systems 150(2), 267--284, 2005.

 \bibitem {cur90} J. Current, H. Min, D. Schilling,
 Multiobjective analysis of facility location decisions.
  EJOR 49(3), 295--300, 1990.

 \bibitem {dahl00} E. Dahlhaus,
  Parallel algorithms for hierarchical clustering and applications
  to split decomposition and party graph recognition.
  J. Algorithms, 36(2), 205--240, 2000.

 \bibitem {dama10} P. Damaschke,
 Fixed-parameter enumerability of cluster editing and related
 problems.
 Theory Computing Sys. 46, 261--283.

 \bibitem {das05} S. Dasgupta, P.M. Long,
 Performance guarantees for hierarchical clustering.
 J. of Computer and System Sciences 70(4),
  555---569, 2005.


 \bibitem {dave92} R. Dave, R. Krishnapuram,
 Robust clustering methods:
 A unified view.
 IEEE Trans. Fuzzy Syst. 5(2), 270--293, 1997.

 \bibitem {david83} M.L. Davidson,
 Multidimensional Scaling.  Wiley, 1983.

  \bibitem {dawande00} M. Dawande, J. Kalagnanam,
  P. Keskinocak, R. Ravi, F.S. Salman,
  Approximation algorithms for the multiple knapsack problem
  with assignment restrictions.
  J. of Combinatorial Oprtimization 4(00),
  171--186, 2000.

 \bibitem {dawande01} M. Dawande, P. Keskinocak,
 J.M. Swaminathan, S. Tayur,
 On bipartite and multipartite clique problems.
  J. of Algorithms  41(2) (2001) 388-403.

  \bibitem {day86} W.H.E. Day,
 Fireword: comparison and consensus of classifications.
 J. of Class.
  3(2), 183--185, 1986.


 \bibitem {deamor92} S.G. de Amorim,
 J.-P. Barthelemy, C.C. Ribeiro,
 Clustering and clique partitioning:
 Simulated anealing and tabu search approaches.
 J. of Classification 9(1), 17--41, 1992.

 \bibitem {debo05} P.T. de Boer, D.K. Kroese, S. Mannor,
 R.Y. Rubinstein.
 A tutorial on the cross-entropy method.
 Annals of Operations Research 134, 19--67, 2005.


  \bibitem {desmet09} Y. De Smet, S. Eppe,
 Multicriteria relational clustering:
 The case of binary outranking matrices.
 In:
 M. Ehrgott et al. (eds),
 Proc. of 5th Int. Conf. on Evolutionary Multi-Criterion
 Optimization EMO 2009, LNCS 5467, Springer,
  380--392, 2009.

 \bibitem {dehne06} F. Dehne, M.A. Langston, X. Luo,
 S. Pitre, P. Shaw, Y. Zhang,
 The cluster editing problem: implementation and experiments.
 In: H.L. Bodlaender, M.A. Langston (eds),
  Proc. of Int. Workshop on parameterized and Exact Comp.
  IWPEC 2006, LNCS 4169, Springer,
  13-24, 2006.

  \bibitem {demaine03} E.D. Demaine, N. Immorlica,
 Correlation clusteirng with partial information.
 In:
 Approximation, Randomization, and Combinatorial Optimization.
 Algorithms and Techniques,
 Springer, 1--13. 2003.

  \bibitem {dem06} E.D. Demaine, D. Emanuel, A. Fiat,
  N. Immorlica,
  Correlation clustering in general weighted graphs.
  Theoretical Computer Science 361(2), 172--187, 2006.

 \bibitem {dewa00} M. Dewatripont, I. Jewitt, J. Tirole,
 Multitask agency problems:
 Focus and task clustering.
 European Economic Review 44(4), 869--877, 2000.

 \bibitem {dhil05} I. Dhillon, Y. Guan, B. Kulis,
 A unified view of kernel k-means, spectral clustering, and
 graph partitioning.
 Tehcnical Report No. UTCS TR-0425, U. of Texas at Austin, 2005.

 \bibitem {diaz01} J.A. Diaz, E. Fernandez,
 A tabu search heuristic for the generalized assignment problem.
 EJOR, 132, 22--38, 2001.

  \bibitem {dick87} P.R. Dickson, J.L. Ginter,
  Market segmentation, product differentiation, and marketing strategy.
  J. of Marketing 51, 1--10, 1987.

 \bibitem {diday73} E. Diday,
 The dynamic cluster method in non-hierarchical
 clustering.
 J. Comput. Inf. Sci., 2, 61--88, 1973.

  \bibitem {diday88} E. Diday,
  The symbolic approach in clustering.
  In: K.S. Fu (ed), Digital Pattern Recongnition,
  Springer,
  47--94, 1988.

  \bibitem {ding01} C.H.Q. Ding, X. He, H. Zha, M. Gu,
  H.D. Simon,
 A min-max algorithm for graph partitioning and
 data clustering.
 In: Proc. IEEE Int. Conf. on Data Mining (ICDM 2001),
 107--114, 2001.

 \bibitem {dinic70} E.A. Dinic,
 An algorithm for the solution of the max-flow
 problem with the polynomial estimation.
 Soviet mathematics Doklady 11, 1277-1280, 1970.

 \bibitem {dji06} H.N. Djidjev,
 A scalable multilevel algorithm for
 graph clustering and community structure detection.
 In: W. Aiello, A. Broder, J. Janssen, E. Milios (eds),
 Proc. of the 4th Int. Workshop on Algorithms and Models for
 Web-Graph (WAW 2006), LNCS 4936, Springer, 117--128, 2008.

 \bibitem {dore04} P. Doreian, V. Batagelj, A. Ferligoj,
  Generalized Blockmodeling.
  Cambridge Univ. Press, 2004.

 \bibitem {dorn94} U. Dorndorf, E. Pesch,
 Fast clustering algorithms.
 ORSA J. of Computing 6(2), 141-153, 1994.

 \bibitem {doro71} A.A. Dorofeyuk,
 Methods for automatic classification: A review.
 Automation and Remote COntrol 32(12), 1928--1958, 1971.


 \bibitem {down02} G.M. Downs, J.M. Barnard,
 Clustering methods and their uses in computational chemistry.
 In: K.B. Lipkowitz, D.B. Boyd (eds),
 Reviews in Computatiopnal Chemistry, vol. 18,
 Wiley, 1--40, 2002.

 \bibitem {drezner09} T. Drezner,
 Competitive facility location.
 In:
  C.A. Floudas, P.M. Pardalos (Eds.),
  Encyclopedia of Optimization. 2nd ed., Springer,
   pp. 396--401, 2009.


 \bibitem {duhand99} D.-Z. Du, P.M. Pardalos (Eds.),
 Handbook of Combinatorial Optimization.
 Volumes 1,2,3, Springer, New York, 1999.

  \bibitem {du09} D.-Z. Du, B. Lu, H. Ngo, P.M. Pardalos,
 Steiner tree problems.
 In:
  C.A. Floudas, P.M. Pardalos (Eds.),
  Encyclopedia of Optimization. 2nd ed., Springer,
  pp. 3723--3743, 2009.

 \bibitem {duan12} D. Duan, Y. Li, R. Li, Z. Lu,
 Incremantal K-clique clustering in dynamic social networks.
 Artificial Intelligence Review 38(2), 129--147, 2012.


 \bibitem {dubes79} R.C. Dubes, A.K. Jain,
 Validity studies in clustering methodologies.
 Pattern Recogn.
 11(4), 235--254, 1979.

 \bibitem {duch05} J. Duch, A. Arenas,
 Community detection in complex networks using extremal
 optimization. Physical Review E, vol. 72, no. 027104, 2005.

 \bibitem {duran74} B. Duran, P. Odell,
 Cluster Analysis: A Survey.
 Springer, New York, 1974.

 \bibitem {dur06} P. D'Urso, P. Giordani,
 A weighted fuzzy c-means clustering model for fuzzy data.
 Computational Statistics \& Data Analysis
 50, 1496--1523, 2006.

 \bibitem {dutt96} S. Dutt, W. Deng,
 A probability-based approach to VLSI circuit partitioning.
 In: Proc. of the 33rd Annual Design Automation Conf.,
 ACM, 100--105, 1996.


 \bibitem {dy04} J.G. Dy, C.E. Brodley,
 Feature selection for unsupervised learning.
 The J. of Machine Learning Research 5, 845--889, 2004.


 \bibitem {edmo72} J. Edmonds, R.M. Karp,
 Theoretical improvements in algorithmi  efficiency
 for network flow problems.
 J. of the ACM 19, 248--264, 1972.

 \bibitem {el98} Y. El-Sonbaty, M.A. Ismail,
 Fuzzy clustering for symbolic data.
 IEEE Trans. Fuzzy Syst. 6(2), 195--204, 1998.

 \bibitem {els09} M. Elsner, W. Schudy,
 Bounding and comparing methods for correlation
 clustering beyond ILP.
 In:
 Proc. of the NAACL HLT Workshop on Integer Linear Programming for
 Natural Language Processing,
 19--27, 2009.

 \bibitem {elt98} T. Eltoft, R. de Figueiredo,
 A new neural netowrk for cluster-detection-and-labeling.
 IEEE Trans. Neural Netw. 9(5), 1021--1035, 1998.

 \bibitem {ema03} D. Emanuel, A. Fiat,
 Correlation clustering - minimizing
 disagreements on arbitrary weighted graphs.
 In: Proc. Algorihtms-ESA 2003,
 Springer, 208--220, 2003.

 \bibitem {enr00} A.J. Enright, C.A. Ouzounis,
 GeneRAGE: A robust algorithm
 for sequence clustering and domain detection.
 Bioinformatics 16(5), 451--457, 2000.


 \bibitem {even99} G. Even, J. Naor, S. Rao, B. Schieber,
 Fast approximate graph partitioning algorithms.
 SIAM J. on Computing 28(6), 2187--2214, 1999.

   \bibitem {fagin06} R. Fagin, R. Kumar, M. Mahdian,
  D. Sivakumar, E. Vee,
 Comparing partial rankings.
  SIAM J. Discrete Math. 20(3), 628--648, 2006.

 \bibitem {fan09} N. Fanizzi, C. d'Amato, F. Esposito,
 Metric-based stochastic conceptual clustering for ontologies.
 Information Systems 34(8), 792--806, 2009.

  \bibitem {farach95} M. Farach, T. Przytycka, M. Thorup,
 On the agreement of many trees.
  Inf. Process. Lett. 55(6), 297--301, 1995.

 \bibitem {feig00} U. Feige, R. Krauthgamer,
 Finding and certifying a large clique
 in a semi-random graph.
 Random Struc. Algorithms 16(2), 195--208, 2000.

 \bibitem {feld11} D. Feldman, M. Langberg,
 A unified framework for approximating and clustering data.
 In:
 Proc. of 47th Annual ACM Symp. on Theory of Computing
 STOC 2011, 569--578, 2011.

 \bibitem {feld15} A.E. Feldman, L. Foschini,
 Balanced partitions of trees and applications.
 Algorithmica 71(2), 354--376, 2015.

 \bibitem {fel11} M.R. Fellows, J. Guob, C. Komusiewicz,
 R. Niedermeier, J. Uhlmann,
 Graph-based data clustering with overlaps.
 Discrete Optimization 8(1), 2–-17, 2011.

 \bibitem {fer92} A. Ferligoj, V. Batagelj,
 Direct multicriteria clustering algorithms.
 J. of Classification 9(1), 43--61, 1992.

 \bibitem {fis70} P.C. Fishburn,
 Utility Theory for Decision Making.
  Wiley, New York, 1970.

 \bibitem {fish87} D.H. Fischer,
 Knowledge acquisition via incremantal conceptual clustering.
 Machine Learning 2(2), 139--172, 1987.


 \bibitem {fle06} L. Fleischer, M.X. Goemans, V.S. Mirrokni,
 M. Sviridenko,
 Tight approximation algorithms for maximum general
 assignment problems.
 In: SODA 2006, 611--620, 2006.

 \bibitem {flores14} M. Flores-Carrido, J.A. Carrasco-Ochoa, J. F.
 Martinez-Trinidad,
 Graph clustering via inexact patterns.
 In:
 E. Bayro-Correchano, E. Hancock (eds),
 Proc. of 19th Iberoamerican Congress on
 Progress in Pattern Recognition, Image Analysis,
 Computer Vision, and Applications CIARP 2014,
 LNCS 8827, Springer, 391--398, 2014.

 \bibitem {flor98} D. Florescu, A.Y. Levy, A.O. Mendelzon,
 Database techniques for the world-wide-web: A survey.
 SOGMOD Record 27(3), 59--74, 1998.

  \bibitem {floudasenc09} C.A. Floudas, P.M. Pardalos (Eds.),
  Encyclopedia of Optimization. 2nd ed.,  Springer, 2009.

 \bibitem {ford92} R. Ford, F. McLaughlin,
  Successful project teams:
  a study of MIS managers.
  IEEE Trans. Engineering Management
   39(4), 312--317, 1992. 

 \bibitem {for09} A. Forster, A.L. Murphy,
 CLIQUE: Role-free clustering with
 Q-learning for Wireless Sensor Networks. In:
 29th IEEE Int. Conf. on Distributed Computing Systems ICDCS'09,
 441-449, 2009.

 \bibitem {fortun10} S. Fortunato,
 Community detection in graphs.
  Electronic preprint, 103 p.,
 Jan. 25, 2010.
 http://arxiv.org/abs/0906.0612v2 [physics.soc-ph]

  \bibitem {fow83} E.B. Fowlkes, C.L. Mallows,
  A method for comparing two hierarchical clusterings.
  J. of thr American Statistical Association 78(383),
  553--569, 1983.

 \bibitem {fra05} G. Frahling, C. Sohler,
 Coresets in dynamic geometric data streams.
 In: Proc. of 37th Annual ACM Symp. on Theory of Computing
 STOC 2005, 209--217, 2005.

 \bibitem {fraley98} C. Fraley, A.E. Raftery,
 How many clusters? Which clustering method?
 Answers via model-based cluster analysis.
 The Computer J. 41(8), 578--588, 1998.

 \bibitem {fred87} M.L. Fredman, R.E. Tarjan,
  Fibonacci heaps and their uses in improved network
  optimization algorithms.
   J. of the ACM 34(3) (1987) 596--615.

  \bibitem {frilast07} M. Friedman, M. Last, Y. Makover, A. Kandel,
 Anomaly detection in web documents using crisp and fuzzy-based
 cosine clustering methodology.
 Inf. Sci. 177(2), 467--475, 2007.

 \bibitem {frig99} H. Frigui, R. Krishnapuram,
 A robust competitive clustering algorithm
 with applications in computer vision.
 IEEE Trans. PAMI 21(5), 450--465, 1999.

  \bibitem {fu05} Z. Fu, W. Hu, T. Tan,
  Similarity based vehicle trajectory clustering and anomaly
  detection. In:
  Int. Conf. on Image Processing, vol. 2, 602--605, 2005.



  \bibitem {fur13} E.M. Furems,
 Dominance-based extension of STEPCLASS for multiattribute
 nominal classification.
 Int. J. of Information Technology \& Decision Making
 12(5), 905--925, 2013.

  \bibitem {fursok11} E.M. Furems, L. Sokolova,
 Expert's knowledge acquisition for differential diagnostics
 of bronchial astma in childfren in STEPCLASS.
 Int. J. of Technology and Management
 11(1), 68--85, 2011.

  \bibitem {gab00} B. Gabrys, A. Bargiela,
  General fuzzy min-max neural network
  for clustering and classification.
  IEEE Trans. Neural Netw. 11(3), 769--783, 2000.



 \bibitem {gabow86} H.N. Gabow, Z. Galil, T. Spencer, R.E. Tarjan,
 Efficient algorithms for finding
 minimum spanning trees in undirected and directed graphs.
 Combinatorica 6(2), 109--122, 1986.

 \bibitem {gar79} M.R. Garey, D.S. Johnson,
   Computers and Intractability. The
  Guide to the Theory of NP-Completeness.
   W.H.Freeman and Company, San Francisco, 1979.

 \bibitem {gavish91} B. Gavish, H. Pirkul,
 Algorithms for the multi-resource generalized assigment problem.
 Manag. Science 37(6), 695--713, 1991


  \bibitem {gent09} M. Gentry, J. MacDougall,
  Total school cluster grouping: Model, research, and practice.
  In: J.S. Renzulli, E.J. Gubbins, K.S. McMillen,
  R.D. Eckert, C.A. Little (eds),
  Systems and Models for Developing Programs for the Gifted and
  Talented. 2nd ed., Creative Learning Press, Mansfield Center,
  CT,  211--234, 2009.

\bibitem {gerla95} M. Gerla, J.T.C. Tsai,
 Multicluster, mobile, multimedia radio networks.
 Wireless networks 1(3), 255--265, 1995.

 \bibitem {ghah04}  Z. Ghahramani,
 Unsupervised learning.
 In: O. Bousquet, U. von Luxburg, G. Raersch (eds),
 Machine Learning, LNCS 3176, Springer,
 72--112, 2004.

  \bibitem {ghia02} S. Ghiasi, A. Srivastava, X. Yang,
  M. Sarrafzadeh,
  Optimal energy aware clustering in sensor networks.
  Sensors 2(7), 258--269, 2002.

  \bibitem {ghosh11} J. Ghosh, A. Acharya,
 Cluster ensembles.
 Data Mining \& Knowledge Discovery, 1(4), 305--315, 2011.

 \bibitem {giotis06} I. Giotis, V. Guruswami,
 Correlation clustering with a fixed number of clusters.
 In: Proc. of the Seventeenth Annual ACM-SIAM
 Symp. on Discrete Algorithms, SIAM, 1167--1176, 2006.

  \bibitem {girvan02} M. Girvan, M.E.J. Newman,
 Community structure in social an biological networks.
  Community structure in social and biological networks.
  PNAS 99(12), 7821--7826, 2002.


 \bibitem {goldb08}  J. Goldberger, T. Tassa,
 A hierarchical clustering algorithm based on
 the Hungarian method.
 Pattern Recognition Letters 29(11), 1632--1638, 2008.

 \bibitem {gold13} B. Goldengorin, D. Krushinsky, P.M. Pardalos,
 Cell Formation in Industrial Engineering:
 Theory, Algorithms and Experiments. Springer, 2013.


  \bibitem {gonz85} T.F. Gonzalez,
  Clustering to minimize the maximum intercluster distance.
  Theoret. Comput. Sci. 38, 293--306, 1985.

 \bibitem {gop13} P.K. Gopalan, D.M. Blei,
 Efficient discovery of overlapping communites in massive
 networks.
 PNAS 110(36), 14534--14539, 2013. 

 \bibitem {gord99} A. Gordon,
 Classification. 2nd ed., Chapman and Hall, London, 1999.

 \bibitem {gowda77} K.C. Gowda, G. Krishna,
 Agglomerative clustering using the concept of mutual nearest
 neighborhood.
 Pattern Recognition 10(2), 105--112, 1977.

 \bibitem {gowda91} K.C. Gowda, E. Diday,
 Symbolic clustering using a new dissimalirity measure.
 Pattern Recognition 24(6), 567--578, 1991.

  \bibitem {gowda92} K.C. Gowda, E. Diday,
 Symbolic clustering using a new simalirity measure.
 IEEE Trans. SMC 22(2), 368--378, 1992.

 \bibitem {gow69} J. Gower, G. Ross,
 Minimum spanning trees and single linkage cluster analysis.
 J. of the Royal Statistical Society,
 Series C (Applied Statistics) 18(1), 54--64, 1969.


 \bibitem {gra05} J. Gramm, J. Guo, F. Huffner, R. Niedermeier,
 Graph-modeled data clustering: Fixed-parameter algorithm for
 clique generation.
 Theory of Computing Systems 38(4), 373--392, 2005.

 \bibitem {gran89} C.W.J. Granger,
 Combining forecasts - twenty years later.
 J. of Forecasting 8(3), 167--173, 1989.

 \bibitem {grave10} D. Graves, W. Pedrycz,
 Kernel-based fuzzy clustering and fuzzy clustering:
 A comparative experimental study.
 Fuzzy Sets and Systems 161(4), 522--543, 2010.

  \bibitem {green72} P.E. Green, V.R. Rao,
 Configural synthesis in multidimensional scaling.
 J. of Marketing Research 9, 65--68, 1972.

 \bibitem {green89} P.E. Green, F.J. Carmone Jr., S.M. Smith,
 Multidimensional Scaling: Concepts and Applications. Allyn and
 Bacon, Boston, 1989.

 \bibitem {groen96} P.J.F. Groenen, W.J. Heiser,
 The tunneling method for global optimization
 in multidimensional scaling.
 Psychometrica 61(3), 529--550, 1996.

 \bibitem {groen99} P.J.F. Groenen, W.J. Heiser,
 J.J. Meulman,
 Global optimization in least-squares multidimensional
 scaling by distance smoothing.
 J. of Classification
  16(2), 225--254, 1999.


  \bibitem {gross97} S. Grossman, M. Miller, K. Cone,
 D. Fischel, D. Ross,
 Clustering and competition in asset markets.
 J. of Low and Economics 40, 23--60, 1997.

 \bibitem {gry06} O. Grygorash, Y. Zhou, Z. Jorgensen,
 Minimum spanning tree based clustering algorithms.
 In: Proc. of 18th IEEE Int. Conf. on Tools with
 Artificial Intelligence
 ICTAI'06, 73--81, 2006.

 \bibitem {gue11} A. Guenoche,
 Consensus partitions: a constructive approach.
 Adv. Data Analysis and Classification 5(3),
 215--229, 2011.

 \bibitem {guha00} S. Guha, R. Rastogi, K. Shim,
 ROCK: A robust clustering algorithm for categorical
 attributes.
 Inf. Syst. 25(5), 345--366, 2000.

 \bibitem {guha00a} S. Guha, N. Mishra, R. Motwani, L.
 O'Callagham,
 Clustering data streams.
 In:
 Proc. of the Annual Symp. on Foundations of CS FOCS 2000,
 359--366, 2000.

  \bibitem {gui03} R. Guimera, L. Dadon,
 A. Diaz-Guilera, F. Giralt, A. Arenas,
 Self-similar community structure
 in a network of human interactions.
  Physical Review E, vol. 68, no. 065103, 2003.

  \bibitem {gui04} R. Guimera, M. Sales-Pardo,  L.A.N. Amaral,
 Modularity from fluctuations in random graphs and complex networks.
  Physical Review E, vol. 70, no. 025101, 2004.

 \bibitem {gunes06} I. Gunes, H. Bingol,
 Coomunity detection in complex networks
 using agents.
  Electronic preprint, 5 p.,
 Oct. 23, 2006,
  arXiv:cs/0610129 [cs.MA]

   \bibitem {guo09} J. Guo,
  A more effective linear kernelization for cluster editing.
  Theor. Comput. Sci. 410, 718--726, 2009.


 \bibitem {gus97} D. Gusfield,
 Algorithms on Strings, Trees, and Sequences:
 Computer Science and Computational Biology.
 Cambridge Univ. Press, Cambridge, Cambridge, UK, 1997.

 \bibitem {had04} S. Haddadi, H. Ouzia,
 Effective algorthm and heuristic for the
 generalized assignment problem.
 EJOR 153, 184--190, 2004.

 \bibitem {hag92} L. Hagen, A. Kahng,
 New spectral methods for ratio cut partitioning and
 clustering.
 CAD  11(9), 1074--1085, 1992.

 \bibitem {halg05} S.K. Halgamuge, L. Wang (eds),
 Classification and Clustering for Knowledge Discovery.
 Springer, 2005.

  \bibitem {hal01} M. Halkidi, Y. Batistakis, M. Vazirgiannis,
 On clustering validation techniques.
 J. of Intelligent Information Systems 17, 107--145, 2001.

 \bibitem {hal02} M. Halkidi, Y. Batistakis, M. Vazirgiannis,
 Cluster validity methods: Part I.
 SIGMOD Record 31(2), 40-45, 2002.

 \bibitem {hal02a} M. Halkidi, Y. Batistakis, M. Vazirgiannis,
 Clustering validity checking methods: Part II.
 SIGMOD Record 31(3), 19-27, 2002.
rrf
  \bibitem {hallett07} M.T. Hallett, C. McCartin,
 A faster FPT algorithm for the maximum agreement forest problem.
  Theory of Computing Systems 41(3), 539--550, 2007.

 \bibitem {hallez09} A. Hallez, A. Bronselaer, G. De Tre,
 Comparison of sets and multisets.
 Int. J. Uncertainty, Fuzziness and Knowledge-Based Systems
  17, 
   153--172, 2009.

  \bibitem {hamel96} A.M. Hamel, M.A. Steel,
 Finding a maximum compatible tree is NP-hard for sequences
 and trees.
 Applied Mathematical Letters 9(2), 55--59, 1996.

 \bibitem {han01} J. Han, M. Kamber,
 Data Mining: Concepts and Techniques. 2nd ed.,
 Morgan Kaufmann, 2005.

 \bibitem {han07} B. Han, W. Jia,
 Clustering wireless ad hoc networks  with weakly connected
 dominating set.
 J. of Parallel and Distributed Computing 67(6), 727--737, 2007.

  \bibitem {han13} Y. Han, W. Lu, T. Chen,
 Cluster consensus in discrete-time networks of
 multi-agents with inter-cluster nonidentical inputs.
  Electronic preprint, 13 p., Mar. 9, 2013.
 http://arxiv.org/abs/1201.2803 [math.OC]

 \bibitem {hand01} D.J. Hand, H. Mannila, P. Smyth,
 Principles of Data Mining.
 The MIT Press, 2001.

 \bibitem {handc07} M.S. Handcock, A.E. Raftery, J.M. Tantrum,
 Model-based clustering for social networks.
 J. of the Royal Statistical Soociety:
 Series A (Statistics in Society)
 170(2), 301--354, 2007.

 \bibitem {hansen97} P. Hansen, B. Jaumard,
 Cluster analysis and mathematical programming.
 Mathematical Programming: Series A and B 79(1-3), 191--215, 1997.

  \bibitem {hansen97a} P. Hansen, N. Mladenivic,
 Variable neighborhood search for the p-median.
 Location Sceince 5(4), 207--226, 1997.

 \bibitem {hansen07} P. Hansen, J. Brimberg, D. Urosevic, N. Mladenovic,
 Data Clustering using Large p-Median
 Models and Primal-Dual Variable Neighborhood Search.
 TR G-2007-41, 2007 GERAD, June 2007.

 \bibitem {hansen07b} P. Hansen, J. Brimberg, D. Urosevic, N. Mladenovic,
 Primal-Dual Variable Neighborhood Search
 for the simple plant-location problem.
 INFORMS J. on Computing 19, 552-564, 2007.

 \bibitem {hansen09} P. Hansen, J. Brimberg, D. Urosevic,
 N. Mladenovic,
 Solving large p-median clustering problems by
 primal-dual variable neighborhood search.
 Data Mining and Knowledge Discovery
 19(3), 351--375,2009.

 \bibitem {har04} S. Har-Peled, S. Mazumdar,
 On coresets for k-mean and k-median clustering.
 In: Proc. of 36th Annual ACM Symp. on Foundations of
 Computer Science STOCS 2004, 291--300, 2004.

 \bibitem {har85} K.R. Harrigan,
 An application of clustering for strategic group analysis.
 Startegic Management J. 6(1), 55--73, 1985.

  \bibitem {harris91} L. Harris,
 Stock price clustering and discreteness.
 Review of Financial Studies 4, 389--415, 1991.

 \bibitem {har75} J.A. Hartigan,
 Clustering algorithms. Wiley, New York, 1975.

 \bibitem {har79} J.A. Hartigan, M.A. Wong,
  A K-mean clustering algorithm.
  J. of the Royal Statistical Society, Ser. C,
  28(1), 100--108, 1979.

 \bibitem {has14} M. Hassani, P. Spaus, T. Seidl,
 Adaptive multi-resolution stream clustering.
 In: Petra Perner (ed),
 Proc. of 10th Int. Conf. Machine Learning and Data Mining
 in Pattern Recognition, LNCS 8556, Springer,
 134--148, 2014.

 \bibitem {he05} Z. He, X. Xu, S. Deng,
 A cluster ensemble method for clustering categorical data.
 Information Fusion 6(2), 143--151, 2005.

 \bibitem {he07} L. He, L.D. Wu, Y.C. Cai,
 Survey of clustering algorithms in data mining.
 Application Research in Computers 1, 55--57, 2007.

 \bibitem {hein02} W.B. Heinzelman, A.P. Chandrakasan, H.
 Balakrishnan,
 An applciaiton-specific protocol architecture for wireless
 microsensor networks.
 IEEE Trans on Wireless Communicaitons
 1(4), 660-670, 2002 

 \bibitem {higg01} A.J. Higgins,
 A dynamic tabu search for large-scale generalized assignment
 problems.
 Comput. and Oper. Res. 28, 1039--1048, 2001.

 \bibitem {hir00} S. Hirano, S. Tsumoto,
 Rough clustering and its applicaiton to medicine.
 J. of Information Science 124, 125--137, 2000.

  \bibitem {hof94} T. Hofmann, J.M. Buhmann,
 Multidimensional scaling and data clustering.
  In: Proc. NIPS 1994,  459-466, 1994.

  \bibitem {holman78} E.W. Holman,
 Completely nonmetric multidimensional scaling.
 J. of Math. Psychol.
 18, 39--51, 1978.

  \bibitem {hopcroft03} J. Hopcroft, O. Khan,
  B. Kulis, B. Selman,
 Natural communities in large linked networks.
 In:
  Proc. of the Ninth ACM SIGKDD Int. Conf. on Knowledge Discovery
 and Data Mining KDD'03,  541--546, 2003.

  \bibitem {hopcroft04} J. Hopcroft, O. Khan,
  B. Kulis, B. Selman,
  Tracking evolving communities in large linked networks.
   PNAS 101(Suppl 1), 5249--5353, 2004.

 \bibitem {hop99} F. Hoppner, F. Klawonn, R. Kruse,
 Fuzzy Cluster Analysis:
 Methods for Classification, Data Analysis, and Image Recognition.
 Wiley, New York, 1999.


 \bibitem {hru09} E.R. Hruschka, R.G.B. Campello,
 A.A. Freitas, A.P.L. Carvalho,
 A survey of evolutionary algorithms for clustering.
 IEEE Trans SMC, Part C 39(2), 133-–155, 2009.

 \bibitem {huang98} Z. Huang,
 Extensions to the k-means algorithm for clustering large data
 sets with categorical values.
 Data Mining Knowl. Discov 2(3), 283--304, 1998.

 \bibitem {huang99} Z. Huang, M.K. Ng,
 A fuzzy k-modes algorithm for clustering categorical data.
 IEEE Trans. on Fuzzy Systems 7(4),
 446--452, 1999. 

 \bibitem {huang05} J.Z. Huang, M.K. Ng, H. Rong, Z. Li,
 Automated variable weighting in k-means type clustering.
 IEEE Trans. PAMI 27(5), 657--668,  2005 

 \bibitem {hub87} L.J. Hubert,
 Assignment Methods in Combinatorial Data Analysis.
 M. Dekker, New York, 1987.

 \bibitem {hub85} L. Hubert, P. Arabie,
 Comparing partitions.
 J. of Classification 2, 193--218, 1985.

  \bibitem {huss14} Z. Hussain, M. Meila,
 Graph-sensitive indices for comparing clusterings.
 Electronic preprint, 15 p.,
 Nov. 27, 2014.
 http://arxiv.org/abs/1411.7582 [cs.LG]

 \bibitem {ibarra75} O.H. Ibarra, C.E. Kim.
 Fast approximation algorithms for the knapsack and
 sum of subset problems.
 J. of the ACM  22, 463--468, 1975.


 \bibitem {indu11} R.K.R. Indukuri, S.V. Penumathsa,
 Dominating sets and spanning tree based clustering algorithms
 for mobile ad hoc networks.
 Int. J. of Advanced Computer Science and Applications
 2(2), 75--81, 2011.

 \bibitem {ismail86} M.A. Ismail, S.Z. Selim,
 Fuzzy c-means: Optimality of solutions and
 effective termination of the problem.
 Pattern Recognition 19(6), 481--485, 1986.

 \bibitem {ismail89} M.A. Ismail, M.S. Kamel,
 Multidimensional data clustering utilizing hybrid search
 strategies.
 Pattern Recognition, 22(1), 75--89, 1989.

 \bibitem {jain10} A.K. Jain,
  Data clustering: 50 years beyond k-means.
  Pattern Recogn. Lett.
   31(8), 651--666, 2010.

 \bibitem {jain88} A.K. Jain, R.C. Dubes,
 Algorithms for clustering data.
 Prentice Hall, Upper Saddle River, NJ,  1988.

 \bibitem {jain00} A.K. Jain, R. Duin, J. Mao,
  Statistical pattern recognition: A review.
  IEEE Trans. PAMI 22(1), 4-37, 2000.

  \bibitem {jain99} A.K. Jain,  M.N. Murty, P.J. Flynn,
  Data clustering: a review.
  ACM Comput. Surv.
 31(3) (1999) 264--323.

 \bibitem {jans09} F. Janssens, L. Zhang, B. De Moor, W. Glanzel,
 Hybrid clustering for validation and improvement of
 subject-classification schemes.
 Information Processing and Management 45(6), 683--702, 2009.


 \bibitem {ji04} X. Ji, H. Zha,
 Sensor positioning in wireless ad-hoc sensor networks
 using multidimensional scaling.
 In: Proc. of Twenty-third Annual Joint Conf. of the
 IEEE Computer and Communications Societies INFOCOM 2004,
 vol. 4, 2652--2661, 2004. 

 \bibitem {jiang95b} T. Jiang, L.  Wang, K. Zhang,
 Alignment of trees - an alternative to tree edit.
 Theoretical Computer Science 143(1), 137--148, 1995.

 \bibitem {jia04} D. Jiang, C. Tang, A. Zhang,
 Cluster analysis for gene expression data: A survey.
 IEEE Trans. KDE 16(11), 1370--1386, 2004.

 \bibitem {jin05} S. Jin, C. Park, D. Choi, K. Chung,
 H. Yoon,
 Cluster-based trust evaluation scheme in an Ad Hoc network.
 ETRI J. 27(4), 465--468, 2005. 

 \bibitem {joa05} T. Joachims, J. Hopcroft,
 Error bounds for correlation clustering.
 In: Proc. of the 22rd Int. Conf. on Machine Learning
 ICML'05,
  ACM, 385--392, 2005. 

  \bibitem {johnson96} D.S. Johnson, M.A. Trick (Eds.),
    Cliques, Coloring, and Satisfiability.
    DIMACS Series in Discrete Mathematics and
    Theoretical Computer Science,
    Vol. 26, AMS, Providence, 1996.


 \bibitem {jolion91} J.-M. Jolion, P. Meer, S. Bataouche,
 Robust clustering with applications in computer vision.
 IEEE Trans. PAMI, 22(8), 1025--1034, 1991.



 \bibitem {jov15} R. Jovanovic, M. Tuba, S. Voss,
 An ant colony optimization algorithm for partitioning
 graphs with supply and demand.
  Electr. prep. 21 p., March 3, 2015.
 http://arxiv.org/abs/1503.00899 [cs.AI]

 \bibitem {jung07} C.J. Jung, H.Y. Cho, Y.-H. Oh,
 Data-driven subvector clustering using
 the cross-entropy method.
 In:
  Proc. of the IEEE Int. Conf. on Acoustics, Speech, and Signal
  Processing   ICASSP 2007,
 vol. 4, 977 -- 980, 2007.

 \bibitem {kae96} L.P. Kaelbling, M.L. Littman, A.W. Moore,
 Reinforcement learning: A survey.
 J. of Artificial Intelligence Research, 4, 237--285, 1996.

 \bibitem {kamal81} A.A. Kamal, S.M. Alyeid, M.S> Mahmoud,
 Multistage clustering: An efficient technique in socioeconomic
 field experiments.
 IEEE Trans. SMC 11(12), 779--785, 1981.

 \bibitem {kam90} B. Kamgar-Parsi, J.A. Gualtieri, J.A. Devaney,
 K. Kamgar-Parsi,
 Clustering with neural networks.
 Biol. Cybern. 63(3), 201--208, 1990.

 \bibitem {kaneko94} K. Kaneko,
 Relevance of dynamic clustering to
 biological networks.
 Physica D: Nonlinear Phenomena 75(1), 55--73, 1994.

 \bibitem {kannan04} R. Kannan, S. Vempala, A. Vetta,
  On clustering: Good, bad and spectral.
  J. of the ACM 51(3), 497--515, 2004.

 \bibitem {kanu00} T. Kanungo, D. Mount, N. Netanyahu,
 C. Piatko, R. Silverman, A. Wu,
 An efficient K-means clustering algorithm: Analysis and
 implementation.
 IEEE Trans. PAMI 24(7), 881--892, 2000.

  \bibitem {karp72} R.M. Karp,
 Reducibility among combinatorial problems.
 In: R.E. Miller, J.W. Thatcher (eds),
 Complexity of Computer Computations.
 Plenum,  pp. 85--103, 1972.


 \bibitem {karyp99} G. Karypis, R. Aggarwal, V. Kumar, S. Shekhar,
 Multilevel hypergraph partitioning: Application in VLSI domain.
 IEEE Trans.  Very Large Scale Integration Systems (VLSI)
 7(1), 69--79, 1999.

 \bibitem {kau90} L. Kaufman, P.J. Rousseeuw,
 Finding Groups in Data:
 An Introduction to Cluster Analysis. Wiley, 1990.

  \bibitem {katz93} J.R. Katzenbach, D.K. Smith,
 The discipline of teams.
 Harvard Business Review 71, 111--120, 1993.

 \bibitem {kearns98} M. Kearns, Y. Mansour, A.Y. Ng,
 An information-theoretic analysis of hard and soft assignment
 methods for clustering.
 In: Learning in Graphical Models, Springer,
 495--520, 1998.

 \bibitem{kee76} R.L. Keeny, H. Raiffa,
   Decisions with Multiple Objectives:
   Preferences and Value Tradeoffsþ
  J.Wiley \& Sons,  Wiley, New York, 1976.

 \bibitem {keller04}  H. Kellerer, U. Pferschy, D. Pisinger,
   Knapsack Problems. Springer, Berlin, 2004.

 \bibitem {kem72} R.L. Kemeny, J.L. Snell,
 Mathematical Models in Social Sciences.
 MIT Press, Canbridge, Mass, 1972.


 \bibitem {kendall62} M. Kendall,
 Rank correlation methods. 3rd ed., Hafner, New York,
 1962.

 \bibitem {ker70} B. Kernigham, S. Lin,
 An efficient heuristic procedure for partitioning graphs.
 Bell Systems Technical J. 49, 291--307, 1970.

 \bibitem {ket96} D.J. Ketchen, C.L. Shook,
 The application of cluster analysis in strategic management
 research and critique.
 Strategic Management J. 17(6), 441--458, 1996.

 \bibitem {kim04} D.-W. Kim, K.H. Lee, D. Lee,
 Fuzzy clustering of categorical data using fuzzy centroids.
 Pattern Recognition Letters
 25, 1263--1271, 2004.

 \bibitem {kim11} S. Kim, S. Nowozin, P. Kohli, C.D. Yoo,
 Higher-order correlation clustering for image segmentation.
 In: Advances in Neural Information Processing Systems,
 1530--1538, 2011.

 \bibitem {kin07} G.W. Kinney, J.W. Barnes, B.W. Colletti,
 A reactive tabu search algorithm with variable clustering
 for the unicost set covering problem.
 Int. J. of Operational Research 2(2), 156--172, 2007.


 \bibitem {klein02} J.M. Kleinberg,
 An impossibility theorem for clustering.
 In: Advances in Neural Information Processing
 Systems (NIPS 2002), 15, 446--453, 2002.

  \bibitem {klein98} J.M. Kleinberg, C. Papadimitriou,
  P. Raghavan,
 Segmentation problems.
 In: Proc. of the Thirtieth Annual ACM Symp.
 on the Theory on Computing STOC'98,
 473--482, 1998.

 \bibitem {klein77} L. Kleinrock, F. Kamoun,
 Hierarchical routing for large networks performance evaluation
 and optimization.
 Computer Networks  1(3), 155--174, 1977.

 \bibitem {klein80} L. Kleinrock, F. Kamoun,
 Optimal clustering structures for hierarchical
 topological design of large computer networks.
 Networks  10, 221--248, 1980.

 \bibitem {klim13} M. Klimenta, U. Brandes,
 Graph drawing by classical multidimensional scaling:
 New perspectives,
 In:
 W. Didimo, M. Patrignani (eds),
 Proc. of Int. Conf. GD 2012, LNCS 7704, Springer,
 55--66, 2013.

   \bibitem {knuth98} D.E. Knuth,
  The Art of Computer Programming.
 Vol. 2, Seminumerical Algorithms.
 Addison Wesley, Reading, 1998.

 \bibitem {knu92} D.E. Knuth, A. Raghunathan,
 The problem of compatible representatives.
 SIAM J. on Disc. Math. 5(3) (1992) 422--427.

 \bibitem {koch05} G. Kochenberg, F. Glover, B. Alidaee, H. Wang,
 Clustering of microarray data via clique partitioning.
 J. of Combinatorial Optimization 10(1), 77--92, 2005.

 \bibitem {kor04} J. Korbitz, J. Koscielny, Z. Kowalczuk,
 W. Cholewa (eds),
 Fault Diagnosis: Models, Artifical Intelligence, Applications.
 Springer, 2004.


 \bibitem {kot04} S. Kotsiantis, P. Pintelas,
 Recent advances in clustering:
 A brief survey.
 WSEAS Trans. on Information Science and Applications
 1(1), 73-81, 2004.

  \bibitem {kot07} S.B. Kotsiantis,
 Supervised machine learning:
 A review of classification techniques.
 Informatica 31(3), 249--2658, 2007.

 \bibitem {krie08}  H.-P. Kriegel, P. Kroger,
 E. Schubert, A. Zimek,
 A general framework for increasing the robustness of PCA-based
 correlation clustering algorithms.
 In:
 Proc. of 20th Int. Conf. on
 Scientific and Statistical Database Management (SSDBM),
  Springer, 418--435, 2008. 

 \bibitem {kri09} H.-P. Kriegel, P. Kroger, A. Zimek,
  Clustering high dimensional data: A survey on subspace
  clustering, pattern-based clustering, and correlation
  clustering. ACM Trans. KDD, 3(1), 1--58, 2009.

 \bibitem {krish99} K. Krishna, M. Murty,
 Genetic K-means algorithm.
 IEEE Trans. SMC, Part B 29(3), 433--439, 1999.

  \bibitem {kris95} R. Krishnapuram, H. Frigui, O. Nasraoui,
  Fuzzy and probabilistic shell clustering algorithms and
  their applications to boundary detection and surface
  approximation.
  IEEE Trans. Fuzzy Systems 3(1), 29--60, 1995.

 \bibitem {kroe07} D.P. Kroese, R.Y. Rubinstein, T. Taimre,
 Application of the cross-entropy method for
 clustering and vector quantization.
 J. of Global Optimization 37(1), 137--157, 2007.

  \bibitem {krus64} J.B. Kruskal,
  Multidimensional scaling by optimizing goodness of
  fit to a nonmetric hypothesis.
  Psychometrica 29(1), 1--27, 1964.

  \bibitem {krus64a} J.B. Kruskal,
 Nonmetric multidimensional scaling: A numerical method.
 Psychometrica 29, 115--129, 1964.

  \bibitem {kuh57} H.W. Kuhn,
 The Hungarian Method for the assignment problems.
  Naval Research Logistics  2, 83--97, 1957.


 \bibitem {kumari13} A.C. Kumari, K. Srinivas,
 Software module clustering using
 a fast multi-objective hyper-heuristic evolutionary algorithm.
 Int. J. of Applied Information Systems 5(6), 12--18, 2012.

 \bibitem {kumar05} V. Kumar, M. Steinbach, P.-N. Tan,
 Introduction to Data Mining.
 Addison-Wesley, 2005.

 \bibitem {kumar14} V. Kumar, S. Minz,
 Feature selection: a literature review.
 Smart Comput.
  Review 4(3), 211--229, 2014.


 \bibitem {kurita91} T. Kurita,
 An efficient agglomerative clustering algorithm using a heap.
 Pattern Recognition 24(3), 205--209, 1991.


 \bibitem  {kwa93} R.K. Kwatera, B. Simeone,
 Clustering heuristics for set covering.
 Annals of Operations Research
 43(5), 295--308, 1993.

  \bibitem {kyper08} M. Kyperountas, A. Tefas, I. Pitas,
 Dynamic training using multistage clustering
 for face recognition.
 Pattern Recognition 41(3), 894--905, 2008.

 \bibitem {lafon06} S. Lafon, A.B. Lee,
 Diffusion maps and coarse-graining:
 A unified framework for dimensionality
 reduction, graph partitioning, and data set parametrization.
 IEEE Trans. PAMI 28(9), 1393-1403,
 2006. 

 \bibitem {lange04} T. Lange, M. Braun, V. Roth, J.M. Buhmann,
 Stability-based validation of clustering solutions.
 Neural Computation 16y(6), 1299--1323, 2004. 

  \bibitem {last02} M. Last,
 Online classification of nonstationary data streams.
 Intell. Data Anal. 6(2), 129--147, 2007.

 \bibitem {last01} M. Last, A. Kandel, O. Maimon,
 Information-theoretic algorithm for feature selection.
 Pattern Recognition 22(6), 799--811, 2001.

 \bibitem {lat07} G. Latsoudas, N.D. Sidiropoulos,
 A fast and effective multidimensional scaling approach
 for node localization in wireless sensor netwokrs.
 IEEE Trans. on Signal Processing 55(10), 5121--5127,  2007.

 \bibitem {lawler85} E.L. Lawler, J.K. Lenstra,
 A.H.G. Rinnooy Kan, D.B. Shmoys (Eds.),
 The Traveling Salesman Problem.
 Wiley, New York, 1985.

 \bibitem {lech06} Y. Lechevallier, R. Verde, F. de A.T. Carvalho,
 Symbolic clustering of large datasets.
 In: V. Batagelj, H.-H. Bock, A. Ferligoj, Z. Ziberna (eds),
 Data Science and Classification.
 Springer, 193--201, 2006.

   \bibitem {lee07} J.-G. Lee, J. Han, K.-Y. Whang,
 Trajectory clustering: A partition-and group framework.
 In: Proc. 2007 ACM SIGMOD Int. Conf. on Management
 of Data,
  593--604, 2007.  


 \bibitem {leew77} J. de Leeuw,
 Applications of convex analysis in multidimensional scaling.
 In:
 Recent Developments in Statistics, 133--145, 1977.

  \bibitem {leew88} J. de Leeuw,
 Convergence of the majorization methods for multidimensional
 scaling.
 J. of Classification 5(2), 163--180, 1988.

  \bibitem {lef85} L.P. Lefkovitch,
 Multi-criteria clustering in genotype environment interaction
 problems.
 Theoretical and Applied Genetics 70, 585--589, 1985.

 \bibitem {lei08} Y. Lei, Z. He, Y. Zi, X. Chen,
 New clustering algorithm-based fault diagnosis using compensation
 distance evaluation technique.
 Mechanical Systems and Signal Processing 22(2), 419--435, 2008.

 \bibitem {les09} J. Leskovec, K.J. Lang, A. Dasgupta,
 M.W. Mahoney,
 Community structure in large networks:
 Natural cluster sizes and the absence of large well-defiend
 clusters.
 Internet Math. 6, 29--123, 2009.

 \bibitem {less01} E. Lesser, J. Storck,
 Communities of practice and organizational performance.
 IBM Syst. J. 40(4), 831--841, 2001.


 \bibitem {lev98} M.Sh. Levin,
  Combinatorial Engineering of Decomposable Systems.
  Kluwer Academic Publishers, Dordrecht, 1998.




  \bibitem {lev03} M.Sh. Levin,
  Common part of preference relations.
  Foundations of Computing \& Dec. Sciences
  28(4), 223--246, 2003.




 \bibitem {lev06} M.Sh. Levin,
   Composite Systems Decisions. Springer,  2006.



 \bibitem {lev07e} M.Sh. Levin,
 Towards hierarchical clustering. In:
 V. Diekert, M. Volkov, A. Voronkov (Eds.),
 Proc. of Int. Conf.  Computer Science in Russia
  CSR-2007, LNCS 4649, Springer, pp. 205--215, 2007.


   \bibitem {lev09} M.Sh. Levin,
 Combinatorial optimization in system configuration design.
  Autom. \& Remote Control 70(3) (2009) 519--561.

  \bibitem {lev09edu} M.Sh. Levin,
 Student research projects in system design.
 In: Proc. of 1st Int. Conf.  CSEDU-2009,
 Lisbon, Portugal,  vol. 2, pp. 67--72, 2009.


  \bibitem {lev09bi} M.Sh. Levin,
 Combinatorial scheme for design of marketing strategy,
  Business Infomatics 2(08), 42--51, 2009 (in Russian)



  \bibitem {lev11eng} M.Sh. Levin,
  k-Set frameworks in multicriteria combinatorial optimisation.
  J. of Technology, Policy and Management, 11(1), 85--95, 2011.

 \bibitem {levcsedu11} M.Sh. Levin,
  Towards k-set problem frameworks in education.
 The 3rd Int. Conf. on Computer Supported Education CSEDU-2011,
 vol. 2, The Netherlands, pp. 99--104, 2011.

\bibitem {lev11restr} M.Sh. Levin,
 Restructuring in combinatorial optimization.
 Electronic preprint, 11 p., Febr. 8, 2011.
 http://arxiv.org/abs/1102.1745 [cs.DS]


  \bibitem {lev11ed} M.Sh. Levin,
  Course on system design (structural approach).
  Electronic preprint, 22 p.,
 Febr. 19, 2011.
 http://arxiv.org/abs/1103.3845 [cs.SE]
%


 \bibitem {lev11ADES} M.Sh. Levin,
 Four-layer framework for combinatorial optimization
 problems domain.
 Advances in Engineering Software 42(12), 1089--1098, 2011.

  \bibitem {lev11agg} M.Sh. Levin,
 Aggregation of composite solutions:
 strategies, models, examples.
 Electronic preprint, 72 p., Nov. 29, 2011.
   http://arxiv.org/abs/1111.6983 [cs.SE]

 \bibitem{lev12morph} M.Sh. Levin,
 Morphological methods for design of modular systems (a survey).
  Electronic  preprint. 20 p., Jan. 9, 2012.
  http://arxiv.org/abs/1201.1712 [cs.SE]

 \bibitem {lev12multiset} M.Sh. Levin,
  Multiset estimates and combinatorial synthesis.
   Electronic preprint. 30 p., May 9, 2012.
  http://arxiv.org/abs/1205.2046 [cs.SY]

   \bibitem {lev12b} M.Sh. Levin,
  Composite strategy for multicriteria ranking/sorting
 (methodological issues, examples).
    Electronic preprint. 24 p., Nov. 9, 2012.
    http://arxiv.org/abs/1211.2245 [math.OC]

  \bibitem {lev12hier} M.Sh. Levin,
  Towards design of system hierarchy (research survey).
    Electronic preprint. 36 p., Dec. 7, 2012.
    http://arxiv.org/abs/1212.1735 [math.OC]

 \bibitem {lev12clique} M.Sh. Levin,
   Clique-based fusion of graph streams in multi-function system testing.
    Informatica 23(3), 391--404, 2012.

  \bibitem {lev13eva} M.Sh. Levin,
 Note on evaluation of hierarchical modular systems.
 Electronic preprint, 15 p., May 21, 2013.
 http://arxiv.org/abs/1305.4917 [cs.AI]

 \bibitem {lev13tra} M.Sh. Levin,
  Towards multistage design of modular systems.
 Electronic preprint, 13 p., June 19, 2013.
 http://arxiv.org/abs/1306.4635 [cs.AI].


 \bibitem {lev15} M.Sh. Levin,
 Modular System Design and Evaluation, Sprigner, 2015.



 \bibitem {levlast06} M.Sh. Levin, M. Last,
 Design of test inputs and their sequences
 in multi-function system testing,
  Applied Intelligence 25(1) (2006) 544-553.

  \bibitem {lev08b} M.Sh. Levin, A.O. Merzlyakov,
 Composite combinatorial scheme of test plannig
 (example for microprocessor systems). In:
 V. Diekert, M. Volkov, A. Voronkov (Eds.),
 Proc. of IEEE Region 8 Int. Conf. Sibircon-2008,
  Novosibisrsk, pp. 291--295, 2008.



 \bibitem {levfim09} M.Sh. Levin, A.V. Fimin,
 Combinatorial scheme for analysis of political candidates
 and their strategies.
 Information Processes
 9(2), 83--92, 2009 (in Russian)

   \bibitem{levpet10} M.Sh. Levin, M.V. Petukhov,
  Multicriteria assignment problem (selection of access points).
  In: N. Garcia-Pedrajas et al. (Eds.),
 {\it Proc. of 23rd Int. Conf.  IEA/AIE 2010},
 LNCS 6097, part II,  Springer,  277--287, 2010.

 \bibitem {levpet10a}  M.Sh. Levin, M.V. Petukhov,
 Conneciton of users with telecommunications network:
  Multicriteria assignment problem.
 J. of Telecommunications Technology and Electronics
  55(12), 1532--1541, 2010.





  \bibitem {levnur11} M.Sh. Levin, R.I. Nuriakhmetov,
 Multicriteria Steiner tree problem for communication network.
  Electronic preprint, 11 p.,
 Febr. 8, 2011.
 http://arxiv.org/abs/1102.2524 [cs.DS]



 \bibitem {li02} D. Li, K.D. Wong, Y.H. Hu, A.M. Sayeed,
 Detection, classification, tracking of targets
 in micro-sensor networks.
 IEEE Signal Processing Magazine
 19(2), 17--29, 2002.

 \bibitem {li06} X. Li, W. Hu, W. Hu,
  A coarse-to-fine strategy for vehicle motion
  trajectory clustering.
 In: Proc. 18th Int. Conf. on Pattern Recognition ICPR 2006,
 vol. 1, 591--594, 2006.

 \bibitem {li15} H. Li, Y. Ping,
 Recent advances in support vector clustering:
  theory and applications.
 Int. J. of Pattern Recognition and Artificial Intelligence
 29(01), 16--43, 2015 

 \bibitem {lin08} N.P. Lin, C.-I. Chang,
  H.-E. Chueh, H.-J. Chen, W.-H. Hao,
 A deflected grid-based algorithm
 for clustering analysis.
 WSEAS Trans. on Computers 3(7), 125--132, 2007.

  \bibitem {lipn97} J. Lipnack, J. Stamps,
 Virtual Teams: Reaching across Space,
 Time and Organizations with Technology.
 Wiley, New York, 1997.

 \bibitem {liu10} X. Liu, S. Yu, F.A.L. Janssens,
 W. Glanzel, Y. Moreau, B. De Moor,
 Weighted hybrid clustering by combining text mining and
 bibliometrics on a large-scale journal database.
 JASIST 61(6), 1105--1119, 2010.

 \bibitem {liu07} X. Liu,  D. Li, S. Wang, Z. Tao,
 Effective algorithm for detecting community structure in
 complex networks based on GA and clustering.
 In:
 Y. Shi, G. Dick van Albada, J. Dongarra, P.M. Sloot (eds),
  Proc. of the 7th Int. Conf. on Computational
 Science ICCS'07, Part II, LNCS 4488, Springer, 657-664, 2007.

 \bibitem {liu11} X. Liu, T. Murata,
 Detecting communities
 in k-partite k-uniform (hyper)networks.
 J. of Computer Science and Technology 26(5), 778--791, 2011.

  \bibitem {liu13} X. Liu, T. Murata, K. Wakita,
 Extending modularity by capturing the similarity
 attraction feature in the null model.
  Electronic preprint. 10 p., Feb. 12, 2013.
    http://arxiv.org/abs/1210.4007 [cs.SI]

 \bibitem {lorena96} L.A.N. Lorena, M.G. Narciso,
 Relaxation heuristics for a general assignment problem.
 EJOR 91(3), 600--610, 1996.

 \bibitem {lov90} L. Lovasz, M. Simonovits,
 The mixing rate of Markov chains, an isoperimetric inequality,
 and computing the volume.
 In:
 Proc. 31st IEEE Annual Symp. on Foundations of Computer Science
 (FOCS), vol. 1, 346--354, 1990.

  \bibitem {lov93} L. Lovasz, M. Simonovits,
 Random walks in a convex body and an improved volume algorithm.
 RSA: Random Structures \& Algorithms 4, 359--412, 1993.

 \bibitem {luy05} Y. Lu, Y. Sun, G. Xu, G. Liu,
 A grid-based clustering algorithm for high-dimensional
 data streams.
 In: Advanced Data Mining and Applications,
 Springer, 824--831, 2005.

 \bibitem {lu07} Z. Lu, T.K. Leen,
 Penalized probabilistic clustering.
 Neural Networks 19(6), 1528--1567, 2007.

 \bibitem {lucas14} N. Lucas, B. Zalik, K.R. Zalik,
 Sweep-hyperplan clustering algorithm using dynamic model.
 Informatica 25(4), 563--580, 2014.



 \bibitem {lux07}  U. von Luxburg,
 A tutorial on spectral clustering.
  Electronic preprint, 32 p., Nov. 2007.
  http://arxiv.org/abs/0711.0189 [cs.DS]

 \bibitem {mad04} S.C. Madeira, A.L. Oliveira,
 Biclustering algorithms for biological data analysis: A survey.
 IEEE/ACM Trans. Computat. Biol. Bioinformatics
 1(1), 24-45, 2004.

  \bibitem {maji07} P. Maji, S.K. Pal,
 RFCM: A hybrid clustering algorithm using rough and fuzzy sets.
 Fundamenta Informaticae 80(), 475--496, 2007.


 \bibitem {man10} B. Mannaa,
 Cluster editing problem for points on the real time:
 A polynomial time algorithm.
 Inform. Proc. Lett. 110, 961--965, 2010.

 \bibitem {man08} C.D. Manning, P. Raghavan, H. Schutze,
 Introduction to Information Retrieval.
 Cambridge Univ. Press, 2008.

 \bibitem {margush82} T. Margush,
 Distance between trees.
 Discr. Appl. Math. 4, 281--290, 1982.

 \bibitem {mar90} S. Martello, P. Toth,
  Knapsack Problem: Algorithms and Computer Implementation.
   Wiley, New York, 1990.


 \bibitem {marx02} Z. Marx, I. Dagan, J.M. Buhmann, E. Shamir,
 Coupled clustering: A method for detecting structural
 correspondence.
 J. of Machine Learning Research 3, 747--780, 2002.


 \bibitem {mas99} F. Massuli, A. Schenone,
 A fuzzy clustering based segmentation system
 as support to diagnosis in medical imaging.
 Artificial Intelligence in Medicine 16(00),
 129--147, 1999.

 \bibitem {mat10} C. Mathieu, W. Schudy,
 Correlation clustering with noisy input.
 In: Proc. of the Twenty-first Annual
 ACM-SIAM Symp. on Discrete Algorithms, SIAM,
 712--728, 2010.

 \bibitem {mau02} U. Maulik, S. Bandyopadhyay,
 Performance evaluation of some clustering algorithms and
 validity indices.
 IEEE Trans. PAMI 24(12), 1650--1654, 2002.



 \bibitem {med05} A. Medius, G. Acuna, C.O. Dorso,
 Detection of community strcuture in networks
 via global optimization.
  Physica A 358,  396--405, 2005.


 \bibitem {mehrot98} A. Mehrotra, M.A. Trick,
 Cliques and clustering: A combinatorial approach.
 Operations Research Lett. 22(1), 1--12, 1998.

 \bibitem {meila01} M. Meila, D. Heckerman,
  An experimental comparison of model-based clustering methods.
  Machine Learning 42(1-2), 9--29,  2001.

 \bibitem {meila12} M. Meila
  Local equivalences of distances between
  clusterings - a geometric prespective.
  Machine Learning 86(3), 369--389,  2012.


 \bibitem {mill81} G.W. Milligan,
 A Monte-Carlo study of 30 internal criterion measures
 for cluster-analysis.
 Psychometrica 46, 187--195, 1981.

 \bibitem {millig96} G.W. Milligan,
 Clustering validation: Results and implifications
 for applied analyses.
 In: P. Arabie, L. Hubert, G. DeSoete (eds),
 Clustering and Classification.
 World Scientific Publishers,
 River Edge, NJ, 341--375, 1996.

   \bibitem {mim12} S. Mimaroglu, M. Yagci,
 CLICOM: Cliques for combining multiple clusterings.
 ESwA 39(2), 1889--1901, 2012.

  \bibitem {minoux89} M. Minoux,
 Network synthesis and optimum network design problems:
 models, solutions, applications,
  Networks 152(3), 530--554, 1989.

  \bibitem {mirkin79} B.G. Mirkin,
 Group Choice.
 Winston, New York, 1979.

 \bibitem {mirkin96} B.G. Mirkin,
 Mathematical Classification and Clustering.
 Kluwer, 1996. 

  \bibitem {mirkin05} B.G. Mirkin,
 Clustering for Data Mining: A Data Recovery Approach.
  Chapman \& Hall/CRC,  Boca Raton, FL, 2005.

 \bibitem {mirkin70} B.G. Mirkin, L.B. Chernyi,
 On a distance measure between partitions of a finite set.
 Automation and Remote Control
 31(5), 786--792, 1970.

 \bibitem {mirkin99} B. Mirkin, I. Muchnik,
 Combinatorial optimization in clustering.
 In:
 D.-Z. Du, P.M. Pardalos (Eds.),
 Handbook of Combinatorial Optimization.
 volume 2, Springer, New York,
  261--329, 1999.

 \bibitem {mis07} N. Mishra, R. Schreiber, I. Stanton,
 R.E. Tarjan,
 Clustering in social networks. In:
 A, Bonato, F.R.K. Chuing (eds),
 Proc. of 5th Int. Workshop Algorithms and Models for the
 Web-Graph  WAW 2007,
 LNCS 4863, Springer,
  56--67, 2007.

 \bibitem  {mitchell06} M. Mitchell,
 Complex systems: Network thinking.
 Artificial Intelligence 170, 
 1194--1212, 2006.

 \bibitem {miy90} S. Miyamoto,
 Fuzzy Sets in Information Retrieval and Cluster Analysis.
 Kluwer, Dordrecht, 1990.

 \bibitem {mohri12} M. Mohri, A. Rostamizadeh, A. Talwalkar,
 Foundations of Machine Learning.
 The MIT Press,
 2012.

 \bibitem {mola04} S. Mola, T. Loughram,
 Discounting and clustering inseasoned equity offering
 prices.
 J. of Financial and Quantitative Analsis
 39(1), 1--23, 2004.

  \bibitem {monti03} S. Monti, P. Tamayo, J. Mesirov, T. Golub,
 Consensus clustering:
 a resampling-based method for class discovery and visualization of
 gene expression microarray data.
 Machine Learning
 52(1-2), 91--118, 2003.

 \bibitem {moon65} J.W. Moon, L. Moser,
 On cliques in graphs.
 Israel J. of Mathematics 3(1), 23--28, 1965.

 \bibitem {mul12} A.C. Muller, S. Nowozin, C.H. Lampert,
 Information theoretic clustering using minimum spanning
 trees.
 In:  A. Pinz et al. (eds.),
 Proc. of Joint 34th DAGM and 36th OAGM Symp.
 Pattern Recognition, LNCS 7476,
 Springer, 205--2015, 2012.

 \bibitem {mul09} E. Muller, I. Assent, S. Gunnemann,
 R. Krieger, T. Seidl,
 Relevant subspace clustering: Mining the most interesting
 non-redundent concepts in high dimensional data. In:
 Proc. of Ninth IEEE Int. Conf. on Data Mining ICDM'09,
 377-386, 2009.

 \bibitem {mull15} E. Muller, I. Assent, S. Gunnermann, T. Seidl,
 J. Dy (eds),
 MultiClust special issue on discovering, summarizing and using
 multiple clusterings.
 Machine Learning 98(1-2), 2015.

 \bibitem {mur10a} T. Murata,
 Detecting communities from tripartite networks. In:
 World Wide Web Conf. (WWW'2010), 1159--1160, 2010.

 \bibitem {mur10b} T. Murata,
 Modularity for heterogeneous networks. In:
 Proc. of the 21th ACM Conf. on Hypertext and Hypermedia
 (HyperText2010), 129--134, 2010.


 \bibitem {mur09} R. Murphey,
 Frequency assignment problem.
 In:  C.A. Floudas, P.M. Pardalos (Eds.),
  Encyclopedia of Optimization. 2nd ed., Springer,
   pp. 1097--1101, 2009.

  \bibitem {murtagh85} F. Murtagh,
 Multidimensional Clustering Algorithms.
 Physica-Verlag, Vienna, 1985.

 \bibitem {murtagh92} F. Murtagh,
 Comments on ``Parallel Algorithms for Hierarchical Clustering and
 Cluster Validity''.
 IEEE Trans. PAMI 14(10), 1056--1057,  1992.

 \bibitem {mur95} M.N. Murty, A.K. Jain,
 Knowledge-based clustering scheme for collection management
 and retrieval of library books.
 Pattern Recognition 28(7), 949--964, 1995.

 \bibitem {naeni15} L.M. Naeni, R. Berretta, P. Moscano,
 MA-Net: A reliable memetic algorithm for community
 detection by modularity optimization.
 In: H. Handa, H. Ishibuchi, HY.-S. Ong, K.C. Tan (eds),
 Proc. of the 18th Asia Pacific Symp.  on
 Intelligent and Evolutionary Systems, Springer
   vol. 1, 311-323, 2015.

 \bibitem {nauss03} R.M. Nauss,
 Solving the generalized assignment problem:
 an optimizing and heuristic approach.
 INFORMS J. on Computing 15, 249--266, 2003.


  \bibitem {nem98} G.I. Nemhauser, M.A. Trick,
 Scheduling a major college basketball conference.
 Operations Research 46(1), 1--8, 1998.

 \bibitem {new03} M.E.J. Newman,
 Fast algorithm for detecting community structure in networks.
   Electronic preprint. 5 p., Sep. 22, 2003.
  http://arxiv.org/abs/0309508    [cond-mat.stat-mech]

 \bibitem {new04a} M.E.J. Newman,
 Detecting community structure in networks.
 Eur. Phys. J. B 38(2), 321--330, 2004.

 \bibitem {new06} M.E.J. Newman,
 Modularity and community structure in networks.
 PNAS 103(23), 8577--8582, 2006. 

 \bibitem {new10} M.E.J. Newman,
 Networks: an Introduction.
 Oxford Univ. Press, Oxford, 2010.

  \bibitem {new04} M.E.J. Newman, M. Girvan,
 Finding and evaluating community structure in networks.
   Electronic preprint. 16 pp., Aug. 11, 2003.
  http://arxiv.org/abs/0308217  [cond-mat.stat-mech]


 \bibitem {ng94} Y.-H. Ng, S.-F. Chin,
  Problem Solving in a Dynamic Environment.
  World Scientific Publishing Co., Singapore, 1994.

 \bibitem {ngai09} E.W.T. Ngai, L. Xiu, D.C.K. Chau,
 Application of data mining techniques in customer relationship
 management: A literature review and classification.
 ESwA
 36(2), 2592--2602, 2009.

 \bibitem {nguyen13} T.M. Nguyen, Q.M.J. Wu,
  Dynamic fuzzy clustering and its application
  in motion segmentation.
  IEEE Trans. on Fuzzy Systems 21(6),
  1019--1031,  2013.

 \bibitem {noack08} A. Noack, R. Rotta,
 Multi-level algorithms for modularity clustering.
  Electronic preprint. 12 p., Dec. 22, 2008.
    http://arxiv.org/abs/0812.4073 [cs.DC]

 \bibitem {nutov06} Z. Nutov, I. Beniaminy, R. Yuster,
 A (1-1/e)-approximation algorithm for the
 generalized assignment problem.
 Oper. Res. lett. 34(3), 283--288, 2006.

 \bibitem {oli07} J.V. de Oliveira, W. Pedrycz,
 Advances in Fuzzy Clustering and Its Applications.
 Wiley, 2007.

  \bibitem {olson95} C. Olson,
  Parallel algorithms for hierarchical clustering.
  Parallel Comput. 21, 1313--1325, 1995.


 \bibitem {oost01} M. Oosten, J.G.C. Rutten, F.C.R. Spieksma,
 The clique partioning problem:
 Facets and patching facets.
 Networks 38(4), 209--226, 2001.

 \bibitem {ops09} T. Opsahl, P. Panzarasa,
 Clustering in weighted networks.
 Social Networks 31(2), 155--163, 2009.

 \bibitem {osman94} I.H. Osman, N. Christofides,
 Capacitated clustering problems by hybrid simulated
 annealing and tabu search.
 Int. Trans. on Operationa Research
  1(3), 317--336, 1994.

  \bibitem {ost73} R.E. Osteen, J.T. Tou,
 A clique-detection algorithm based on neighborhoods
 in graphs.
  Int. J. of Computer \& Inform. Sciences
  2(4), 257--268, 1973.

 \bibitem {oster99} P.R.J. Ostergard,
 A new algorithm for the maximum-weight clique problem.
 In: Electronic Notes in Discrete Mathematics,
 6th Twente Workshop on Graphs and Combinatorial Optimization,
 vol. 3, 153--156, 1999.

  \bibitem {osuagwu08} L. Osuagwu,
 Political marketing:
 Conceptualization, dimensions and research agenga.
  Marketing Intelligence \& Planning 26(7),  793--810, 2008.

 \bibitem {ovel12} M. Ovelgonne, A. Geyer-Schulz,
 A comparison of agglomerative hierarchical algorithms
 for modularity clustering.
 In: Proc. of Conf. on
 Challenges at the Interface of Data Analysis, Computer Science,
 and Optimization.
 Springer, 225--232, 2012.

 \bibitem {ozdal04} M.M. Ozdal, C. Ayakznat,
 Hypergraph models and algorithms for data-pattern-based
 clustering.
 Data Mining and Knowledge Discovery 9(1), 29--57, 2004.

 \bibitem {ozy09} T. Ozyer, R. Alhajj,
 Parallel clustering of high dimensional data by integrating
 multi-objective genetic algorithm with divide and conquer.
 Applied Intelligence 31(3), 318-–331, 2009.

  \bibitem {paas14} M. Paasivaara, C. Lassenius,
 Communities of practice in a large distributed agile software
 development organization - Case Ericsson.
 Information and Software Technology
 56(12), 1556--1577, 2014. 

 \bibitem {paiv05} N. Paivinen,
 Clustering with a minimum spanning tree of
 scale-free-like structure.
 Pattern Recogn. Lett. 26(7), 921--930, 2005.

 \bibitem {pal90} S.K. Pal, S. Mitra,
 Fuzzy dynamic clustering algorithm.
 Pattern Recogn. Lett.
 11(8), 525--535, 1990. 


 \bibitem {pard94} P.M. Pardalos, J.  Xue,
 The maximum clique problem.
  J. of Global Optimization,
  4(3), 301--328, 1994.

 \bibitem {pardalos94} P.M. Pardalos, H. Wolkowicz (Eds.),
  Quadratic Assignment and Related Problems.
  AMS, Providence, 1994.

 \bibitem {pardalos00} P.M. Pardalos, L. Pitsoulis (Eds.),
 {\it Nonlinear Assignment Problems}, Kluwer, Boston,
     2000.

 \bibitem {pardalos13} P. Pardalos, M.  Batzyn, E. Maslov,
  Cliques and quasi-cliques in large graphs:
  theory and applications. In:
  Proc. of Int. Conf. on Discrete Optimization and Operations
  Research DOOR-2013.
  Novosibirsk, Sobolev Inst. of Mathematics, 24--28, 2013.

 \bibitem {pareto71} V. Pareto,
  Mannual of political economy.
 (English translation), A. M. Kelley Publishers,
 New York, 1971.

 \bibitem {park04} N.H. Park, W.S. Lee,
 Statistical grid-based clustering over data strams.
 In: ACM SIGMOD Record 33(1), 32-37, 2004.


 \bibitem {pat07} A. Patcha, J.M. Park,
 An overview of anomaly detection techniques:
 Existing solutions and latest technological trends.
 Computer Networks 51(12), 3448--3470, 2007.

 \bibitem {pat89} R. Patton,P. Frank, R. Clark,
 Fault Diagnosis in Dynamic Systems.
 Theory and Application.
 Prentice Hall, Hertfordshire, 1989.


 \bibitem {pav07} M. Pavan, M. Pelillo,
 Dominant sets and pairwise clustering.
 IEEE Trans. PAMI 29(1), 167--172, 2007.

 \bibitem {pedr05} W. Pedrycz,
 Knowledge-Based Clustering:
 From Data to Information Granules.
 Wiley, Hoboken, NJ, 2005.

  \bibitem {pei03} J. Pei, X. Zhang, M. Cho, H. Wang, P.S. Yu,
 MaPle: A fast algorithm for maximal pattern-based clustering.
 In:
 Proc. of the 3rd IEEE Int. COnf. on Data Mining (ICDM 2003),
 259--266, 2003.


 \bibitem {peter10} S.J. Peter, S.P. Victor,
 A novel algorithm for dual similarity clusters using
 minimum spanning tree.
 J. of Theoretical and Applied Information Tehcnology
  14(1), 60--66, 2010.

 \bibitem {pet02} S. Pettie, V. Ramashandran,
 An optimal minimum spanning tree algorithm.
 J. of the ACM 49(1), 16--34, 2002.

  \bibitem {pit09} L. Pitsoulis, P.M. Pardalos,
  Quadratic assignment problem.
  In:  C.A. Floudas, P.M. Pardalos (Eds.),
  Encyclopedia of Optimization. 2nd ed.,
  Springer, pp. 3119--3149, 2009.

 \bibitem {pons06} P. Pons, M. Latapy,
 Computing communities in large networks
 using random works.
 J. of Graph Algorithms and Applications 10,
  191--218, 2006.

  \bibitem {port09} M.A. Porter, J.-P. Onnela, P.J. Mucha,
 Communities in networks.
 Notices of the AMS 56(9), 1082--1097,1164, 2009.

  \bibitem {pras02}  K. Prasad, K.B. Akhilesh,
 Global virtual teams:
 what impacts their design and performance?
 Team Performance Management 8(5/6),
 102--112, 2002.

 \bibitem {puz00} J. Puzicha, T. Hofmann, J.M. Buhmann,
 Theory of proximity based clustering:
 Structure detection by optimization.
 Pattern Recognition 33(4), 617-634, 2000.

  \bibitem {qu09} R. Qu, E.K. Burke, B. McCollum,
  L.T.G. Merlot, S.Y. Lee,
 A survey of search methodologies and automated system development
 for examination timetabling.
 J. of Scheduling 12(1), 55--89, 2009.

 \bibitem {rah07} S. Rahman, T. Wittkop, J. Baumbach,
  M. Martin, A. Truss, S. Boker,
 Exact and heuristic algorithms
 for weighted cluster editing.
 In: Comput. Syst. Bioinformatics Conf., 6(1), 391--401, 2007.

 \bibitem {rand71} W. Rand,
 Objective criteria for the evaluation of clustering methods.
  J. of the American Statistical Association
  66, 846--850, 1971.

 \bibitem {rask99} B. Raskutti, C. Leckie,
 An evaluaiton of criteria for measuring the quality
 of clusters.
 In: Proc. of the 16th Int. Joint Conf. on Artificial
 Intelligence IJCAI'99, vol. 2, 905-910, 1999.

 \bibitem {raj13} F. Rajabi-Alni,
 Two exact algorithms for the gereralized
 assignment problem.
   Electronic  preprint. 13 p., Mar. 17, 2013.
  http://arxiv.org/abs/1303.4031 [cs.DS]

 \bibitem {reich06} J. Reichardt,  S. Bornholdt,
  Statistical mechanics of community detection.
 Physical Review E, vol. 74, no. 016110, 2006.

 \bibitem {rice96} J.R. Rice, R.F. Boisvert,
  From scientific software libraries to problem-solving
  environments,
  IEEE Comput. Sci. \& Eng.  3(3) (1996) 44--53.


 \bibitem {roberts84} F.S. Roberts,
  Applied Combinatorics.
  Prentice Hall, Englewood Cliffs, NJ, 1984.

  \bibitem {rob76} F.S. Roberts,
  Discrete Mathematical Models with Applications to
  Social, Biological and Environmental Problems.
  Prentice Hall, Englewood Cliffs, NJ, 1976.

 \bibitem {rocha13a} C. Rocha, L.C. Dias, I. Dimas,
 Multicriteria classification with unknown categories:
 A clustering-sorting approach and an application to conflict
 management.
 J. of Multi-Criteria Decision Analysis 20(1-2), 13--27, 2013.

 \bibitem {rocha13} C. Rocha, L.C. Dias,
 MPOC - an agglomerative algorithm for multicriteria
 partially ordered clustering.
 4OR  11(3), 253--273, 2013.

 \bibitem {romei00} H.E. Romeijn, D.R. Morales,
 A class of greedy algorithms for the generalized
 assignemnt problem.
 Discrete Applied mathematics 103(00), 209--235, 2000.

 \bibitem {romero07} C. Romero, S. Ventura,
 Educational data mining:
 A survey from 1995 to 2005.
 ESwA
 33(00), 125--146, 2007.

 \bibitem {romero10} C. Romero, S. Ventura,
 Educational data mining:
 A review of the state of the art.
 IEEE Trans. SMC, Part C 40(6), 601--618, 2010.

 \bibitem {romero10a} C. Romero, S. Ventura,
 M. Pechenizkiy, R.S.J.d. Baker (eds.),
 Handbook of Educational Data Mining.
 Chapman \& Hall/CRC Press, 2010.


  \bibitem {ronen85} S. Ronen, O. Shenkar,
 Clustering countries on attitudinal dimensions:
 A review and synthesis.
 Academy of Management Review, 435--454, 1985.

 \bibitem {ross75} G.T. Ross, R.M. Soland,
 A branch-and-bound algorithm for the generalized
  assignment problem.
 Math., Programming 8(1), 91--103, 1975.

 \bibitem {ros07} M. Rosvall, C.T. Bergstrom,
 An information-theoretic framework for resolving
 community structure in complex networks.
 PNAS 104(18), 7327--7331, 2007.

 \bibitem {rotta08} R. Rotta,
 A multi-level algorithm for modularity clustering.
 MS thesis, Brandenburg Univ. of Technology, 2008.

 \bibitem {roy96} B. Roy,
 Multicriteria Methodology for Decision Aiding.
  Kluwer,
  Dordrecht, 1996.

 \bibitem {roy06} B. Roy, R. Slowinski,
 Multi-criteria assignment problem with incompatibility
 and capacity constraints.
 Annals of Operations Research, 147(1) (2006) 287-–316.

 \bibitem {rub02} R.Y. Rubinstein,
 Cross-entropy and rare-events for
 maximal cut and partition problems.
 ACM Trans. on Modeling and Computer Simulaiton
 12(1), 27--53, 2002. 



  \bibitem {saeed12} F. Saeed, N. Salim, A. Abdo, H. Hentabli,
 Combining multiple individual clusterings
 of chemical structures using
  cluster-based similarity partitionig algorithm.
  In: A.E. Hasssanien, A.-B.M. Salem, R. Ramadan, T.-h. Kim (eds),
  Proc. of 1st Int. Conf.
 Advanced Machine learning Technologies and Applications
  AMLTA 2012, CCIS 322, Springer, 276--284, 2012.

 \bibitem {saeed12a} F. Saeed, N. Salim, A. Abdo,
 Voting-based consensus clustering for combining
 multiple clusterings of chemical structures.
 J. of Cheminformatics 4(37), 1--8, 2012.

 \bibitem {sahni75} S. Sahni,
  Approximate algorithms for the 0-1 knapsack problem.
  J. of the ACM 22(1), 115--124, 1975.

 \bibitem {sahni76} S. Sahni, T. Gonzalez,
 P-complete approximaiton problems.
  J. of the ACM 23(00), 555--565, 1976.

 \bibitem {salz09} J. Salzmann, R. Behnke, M. Gag, D. Timmermann,
 4-MASCLE - improved coverage aware clustering with self healing
 abilities.
 In: IEEE Symposia and Workshops on Ubiquitous, Autonomic and Trusted
 Computing UIC-ATC'09, 537--543, 2009.

  \bibitem {sasha89} D. Sasha, K. Zhang,
 Simple fast algorithms for editing distance
 between trees and related problems.
 SIAM J. on Computing 18(6), 1245--1262, 1989.

  \bibitem {sas14} A.A.V. Sastry, K. Netti,
 A parallel sampling based clustering.
  Electronic preprint, 3 pp., Dec. 5, 2014.
 http://arxiv.org/abs/1412.1947 [cs.LG]

 \bibitem {sato06a} M. Sato-Ilic,
 Dynamic fuzzy clustering using fuzzy cluster loading.
 Int. J. General Systems 35(2), 209--230, 2006.

  \bibitem {sato06} M. Sato-Ilic, L.C. Jain,
 Kernel based fuzzy clustering. In:
 M. Sato-Ilic,
 Innovations in Fuzzy Clustering,
  Springer, 89--104, 2006.

 \bibitem {sato11} M. Sato-Ilic,
 Symbolic clustering with interval-valued data.
 Procedia Computer Science 6, 358--363, 2011.

 \bibitem {savel97} M. Savelsbergh,
 A branch-and-price algorithm for the generalized
 assignment problem.
 Operations Research 45, 831--841, 1997.

  \bibitem {scar02} A. Scarelli, S.C. Narula,
 A multicriteria assignment problem,
  J. of Multi-Criteria Dec. Anal. 11(2), 65--74, 2002.

  \bibitem {scha07} S.E. Schaeffer,
  Graph clustering.
  Computer Science Review 1(1), 27--64, 2007.

 \bibitem {schen04} A. Schenker, M. Last, H. Bunke, A. Kandel,
 Classification of web documents using graph matching.
 Int. J. of Pattern Recognition and Artificial Intelligence
 18(3), 475--496, 2004.


  \bibitem {schuf07} A. Schuffenhaer, N. Brown, P. Ertl,
   J.L. Jenkins, P. Selzer, J. Hamon,
  Clustering and rule-based classificaitons of chemical structures
  evaluated in the biological activity space.
  J. Chem. Inf. Model. 47(2), 325--336, 2007.

  \bibitem {selim91} S. Selim, K. Alsultan,
 A simulated annealing algorithm for the clustering problems.
 Pattern Recognition 24(10), 1003--1008, 1991.

  \bibitem {selim98} H.M. Selim, R.G. Askin, A.J. Vakharia,
  Cell formation in group technology:
  review, evaluation and direction for future research.
  Computers \& Industrial Engineering 34(1), 3--20, 1998.

 \bibitem {shama04} S. Shama, K. Gopolan, S. Nanda,
 Viking: A multi-spanning-tree Ethernet architecture
 for metropolian area and cluster networks.
 In: Proc. of Twenty-third Annual Joint Conf. of the IEEE Computer and
 Comunication Societies INFOCOM 2004,
 vol. 4, 2283--2294, 2004.

 \bibitem {shamir02} R. Shamir, R. Sharan, D. Tsur,
 Cluster graph modification problems. In:
 Proc. of 28th WG, Springer, LNCS 2573, 379--316, 2002.

 \bibitem {sham04} R. Shamir, R. Sharan, D. Tsur,
  Cluster graph modification problems.
  Discrete Applied Mathematics 144 (1–2),  173–-182, 2004.


  \bibitem {shamiro10} O. Shamir, N. Tishby,
  Stability and model selection in k-means clustering.
 Machine Learning 80(2-3), 213--243, 2010.

 \bibitem {shap01} L.G. Shapiro, G.C. Stockman,
 Computer Vision. Prentice-Hall, NJ, 2001.

 \bibitem {sheik00} G. Sheikholeslami, C. Chattterjee,
  A. Zhang,
  WaveCluster: a wavelet-based clustering approach for
  spatial data in very large databases.
  The VLDB J. 8(3-4), 289--304, 2000.

 \bibitem {shekar87} B. Shekar, N.M. Murty, G. Krishna,
 A knowledge-based clustering scheme.
 Pattern Recogn. Lett. 5(4), 253--259, 1987.


 \bibitem {shep80} R.N. Shepard,
 Multidimensional scaling: Tree-fitting, and clustering.
 Science 210(4468), 390--398, 1980. 

 \bibitem {shio13} H. Shiokawa, Y. Fujiwara, M. Onizuka,
 Fast algorithm for modularity-based graph clustering.
 In: Proc. of the Twenty-Seventh AAAI Conf. on Artificial
 Intelligence,  1170--1176, 2013.

 \bibitem {shmo93} D.B. Shmoys, E. Tardos,
 An approximation algorithm for the generalized
 assignment problem.
 Math. progam. 62(3), 461--474, 1993.

 \bibitem {simcha85} R. Simcha, O. Shenkar,
 Clustering countries on attitudinal dimensions:
 A review and synthesis.
 Academy of Managgement Review, 435--454, 1985.

 \bibitem {sim73} H.A. Simon,
 The structure of ill-structured problems,
 Artificial Intelligence 4(3) (1973) 181--201.



 \bibitem {sinka02} M.P. Sinka, D.W. Corne,
 A large banchmark dataset for web ducument clustering.
 Soft Computing Systems: Design, Management and Applications
  87, 881--890, 2002.

\bibitem {skou10} D. Skoutas, D. Sacharidis, A. Simitsis,
 T. Sellis,
 Ranking and clusteirng web services using multicriteria dominance
 relationships.
 IEEE Trans. on Service Computing 3(3), 163--177, 2010.



  \bibitem {smith80} S.P. Smith, R.C. Dubes,
 Stability of hierarchical clustering.
 Pattern Recognition 12(3), 177--187, 1980.


 \bibitem {spiel04} D.A. Spielman, S.-H. Teng,
 Nearly-linear time algorithms for graph partitioning,
 graph sparsification, and solving linear systems.
 In: Proc. of the 36th
 Ann. ACM Symp. on Theory of Comput.,
  81--90, 2004.

 \bibitem {spiel13} D.A. Spielman, S.-H. Teng,
 A local clustering algorithm for massive graphs
 and its application to nearly linear time graph partitioning.
 SIAM J. Comput. 42(1), 1--26, 2013.

 \bibitem {srin94} G. Srinivasan,
 A clustering algorithm for machine cell formation
 in group technology using minimum spanning tree.
 The Int. J. of Production Research 32(9), 2149--2158, 1994.

 \bibitem {sri05} A. Srivastava, S.H. Joshi, W. Mio, X. Liu,
  Statistical shape analysis: Clustering, learning, and testing.
  IEEE Trans. PAMI 27(4), 590--602, 2005.



   \bibitem {stein00} M. Steinbach, G. Karypis, V. Kumar,
 A comparison of document clsutering techniques.
 TR 00-034, Dept. of CS, U. of Minnesota, May 2000.

 \bibitem {stepp86} R.E. Stepp, R.S. Michalski,
 Conceptual clustering of structured objects:
 A goal-oriented approach.
 Artif. Aintell. 28(1), 43--69, 1986.

  \bibitem {stew06} G.L. Stewart,
 A meta-analytic review of relationships between team
 design and team performance.
 J. of Management 32(1), 29--55, 2006.

 \bibitem {str02} A. Strehl, J. Ghosh,
 Cluster ensembles - a knowledge reuse framework for combining
 multiple partitions.
 J. of Machine Learning Research 3, 583--617, 2002.

 \bibitem {sung00} C.S. Sung, H.W. Jin,
  A Tabu-search-based heuristic for clustering.
  Pattern Recognition 33(5), 849--858, 2000.

  \bibitem {surd05} M. Surdeanu, J. Turmo, A. Ageno,
 A hybrid unsupervised approach
 for document clustering.
 In: Proc. of the 11th ACM SIGKDD Int. Conf. on Knowledge
 Discovery in Data Mining, 685--690, 2005.

 \bibitem {sut98} R.S. Sutton, A.G. Barto,
 Reinforcement Learning: An Introduction.
 The MIT Press, Boston, 1998.

 \bibitem {swam04} C. Swamy,
 Correlation clustering: maximizing agreements
 via semidifinite programming.
 In: Proc. of the Fifteenth Annual ACM-SIAM
 Symp. on Discrete Algorithms,
 SIAM, 526--527, 2004.

 \bibitem {szo82} P. Szolovits (ed),
 Artificial Intelligence in Medicine.
 Westview Press, Boulder, CO, 1982.

 \bibitem {szu01} M. Szummer, T. Jaakkola,
 Partially labeled classification with Markov random walks.
 In:
 T.G. Dietrich, S. Becker, Z. Ghahramani (eds),
 Advances in Neural Information Processing Systems (NIPS 2001),
 vol. 14, 945--952, 2001.

 \bibitem {tabor13} J. Tabor, K. Misztal,
 Detection of elliptical shapes via cross-entropy clustering.
 In:
 J.M. Sanches, L. Mico, J.S. Cardoso (eds), Proc. of
 6th Iberian Conf. Pattern Recognition and Image Analysis
  IbPRIA 2013, LNCS 7887, Springer, 656--663, 2013.

 \bibitem {tabor14} J. Tabor, P. Spurek,
 Cross-entropy clustering.
 Pattern Recognition 47(9), 3046--3059, 2014.

  \bibitem {tai79} K.-C. Tai,
 The tree-to-tree correction problem.
  J. of the ACM 26(3), 422--433, 1979.


 \bibitem {taksar01} B. Taksar, E. Segal, D. Koller,
 Probabilsitic clustering in relational data. In:
 Proc. Seventeenth Int. Joint Conf. on Artificial
 Intelligence (IJCAI), 870--887, 2001.


 \bibitem {tanaka94} E. Tanaka,
 A metric between unrooted and unordered trees and
 its bottom-up computing method.
 IEEE Trans. PAMI 16(12), 1233--1238, 1994.

 \bibitem {tanaka88} E. Tanaka, K.  Tanaka,
 The tree-to-tree editing problem.
 Int. J. Pattern Recogn. and Art. Intell.
  2(2), 221--240, 1988.



  \bibitem {tarata07} Z. Tarapata,
 On a multicriteria shortest path problem.
  Int. J. Appl. Math. Comput. Sci.,
  17(2) (2007) 269--287.


  \bibitem {tavana03} M. Tavana,
 CROSS: a multicriteria group-decision-making model for evaluating
 and prioritizing advanced-technology projects at NASA.
  Interfaces 33(3), 40-56, 2003.

 \bibitem {tec12} P. Tecuanhuehue-Vera,
 J.A. Carrasco-Ochoa, J.F. Martinez-Trinidad,
 genetic algorithm for multidimensional scaling over mixed and
 incomplete data.
 In:
 J.A. Carrasco-Ochoa, J.F. Martinez-Trinidad,J.A.O. Lopez,
 K.L. Boyer  et al. (eds),
 Proc. of 4th Mexican Conf. on Pattern Recognition MCPR,
 LNCS 7329, Springer,
  3226--235, 2012.

 \bibitem {tilla05} P. Tillapart, S. Thammarojsakul,
 T. Thumthawatworn, P. Santiprabhob,
 An approach to hybrid clustering and routing in wireless sensor
 networks.
 In: Proc. of 2005 IEEE Aerospace Conf., 1--8, 2005.

 \bibitem {till02} J. Tillet, R. Rao, F. Sahin,
 Cluster-head identification in ad hoc sensor networks
 using particle swam optimization.
 In:
 2002 IEEE Int. Conf. on Personal Wireless Communications,
 201--205, 2002.

  \bibitem {tomita06} E. Tomita, A. Tanaka, H. Takahashi,
 The worst-case time complexity for generating all maximal cliques
 and computational experiments. Theoretical Computer Science
 363(1), 28--42, 2006.


 \bibitem {topchy04} A.P. Topchy, M.H.C. Law,
 A.K. Jain, A.L. Fred,
 Analysis of consensus partition in cluster ensemble.
 In: Proc. of Fourth IEEE Int. Conf. on Data Mining ICDM 2004,
 225--232, 2004.

 \bibitem {torg52} W.S. Torgerson,
 Multidimensional scaling: I. Theory and method.
 Psychometrica 17, 401--419, 1952.

 \bibitem {trick92} M.A. Trick,
 A linear relaxation heuristic for the generalized
 assignemnt problem. Nav. Res. Ligist. 39(00),
 137--151, 1992.

 \bibitem {trif08} A. Trifunovic, W.J. Knottenbelt,
 Parallel multilevel algorithms for hypergraph partitioning.
 J. of Parallel and Distributed Computing 68(5),
 563--581, 2008.

 \bibitem {tripa12} R.K. Tripathi, Y.N. Singh, N.K. Verma,
 N-leach, a balanced cost ckuster-heads seleciton algorithm
 for wireless sensor networks.
 In: Proc. of 2012 Nat. Conf. on Communicaitons (NCC),
 1--5, 2012.

 \bibitem {tsai07} C.-F. Tsai, C.-C. Yen,
 ANGEL: a new effective and efficient hybrid clustering
 techniques for large databases. In:
 Z.-H. Zhou, H. Li, Q. Yang (eds),
 Proc. of 11th Pacific-Asia Conf. on
  Advances in Knowledge Discovery and Data Mining
  PAKDD 2007, LNCS 4426, Springer,
  Springer, Berlin, 817--824, 2007.

  \bibitem {tsai10} C.-F. Tsai, H.-F. Yeh, J.-F. Chang, N.-H. Liu,
  PHD: an efficient data clustering scheme using partition space technique
  for knowledge discovery in large databases.
  Appl. Intell. 33(1), 39--53, 2010.

 \bibitem {tsai12} C.-W. Tsai, H.-J. Song, M.-C. Chiang,
 A hyper-heuristic clustering algorithm.
 In: Proc. of 2012 IEEE Int. Conf. on Systems, Man, and
 Cybernetics, 2839--2844, 2012.

 \bibitem {tseng01} L.Y. Tseng, S.B. Yang,
 A genetic approach to the automatic clustering problem.
 Pattern Recognit. 34(2), 415--424, 2001.

 \bibitem {tsuda06}  K. Tsuda, T. Kudo,
 Clustering graphs by wieghted substructure mining.
 In: Proc. 23rd Int. Conf. on Machine Learning,
 953--960, 2006.

 \bibitem {tuba03} M. Tubaishat, S.K. Madria,
 Sensor networks: an overview.
 Potentials, IEEE 22(2), 20-23, 2003.

  \bibitem {tumer08} K. Tumer, A.K. Agogino,
 Ensemble clustering with voting active clusters.
 Pattern Recognition Lett. 29(14), 1947--1953, 2008.


  \bibitem {valiente02} G. Valiente,
  Algorithms on trees and graphs. Springer, Berlin, 2002.

 \bibitem {vand03} D.W. Van der Merwe, A.P. Engelbrecht,
 Data clustering using particle swam optimization.
 In: The 2003 Congress on Evolutionary Computation
 CEC'03, vol. 1, 215--220, 2003.

 \bibitem {van79} C.J. Van Rijsbergen,
 Information Retrieval, 2nd ed.,
 Dept. of CS, Univ. of Glasgow, 1979.

 \bibitem {vanr77} J. Van Ryzin (Ed.),
  Classification and Clustering.
  Academic Press, New York, 1977.

 \bibitem {vap00} V.N. Vapnik,
 The Nature of Statistical Learning Theory.
 2nd ed., Springer, Berlin, 2000.

 \bibitem {vas07} A. Vashist, C.A. Kulikowsky, I. Muchnik,
 Orthlog clustering on a multipartite graph.
 IEEE/ACM Trans. Comput. Biology and Bioinformatics
 4(1), 17--27, 2007.

 \bibitem {vega11} S. Vega-Pons, J. Ruiz-Schulcloper,
 A survey of clustering ensemble algorithms.
 Int. J. of Pattern Recogn. Artificial Intelligence
 25(11), 337--372, 2011.

 \bibitem {voor86} E.M. Voorhees,
 Implementing agglomerative hierarchic clustering algorithms
 for use in document retrieval.
 Information Processing \& Management
 22(6), 465--476, 1986.



   \bibitem {voss09} S. Voss,
 Capacitated minimum spanning trees.
 In:  C.A. Floudas, P.M. Pardalos (Eds.),
  Encyclopedia of Optimization. 2nd ed., Springer,
   pp. 347--357, 2009.

 \bibitem {wak07} K. Wakita, T. Tsusumi,
 Finding community structure in mega-scale social networks.
 Electronic preprint. 9 p., Fev. 8, 2007.
    http://arxiv.org/abs/0702.2048 [cs.CY]


 \bibitem {wall68} C.S. Wallace, D.M. Boulton,
 An information measure for classification.
 The Computer J. 11, 185--194, 1968.


 \bibitem {wal06} D. Walsh, L. Rybicki,
 Symptom clustering in advanced cancer.
 Supportive Care in Cancer 14(8), 831--836, 2006.

 \bibitem {wang01} J.T.-L. Wang, K. Zhang,
 Finding similar consensus between trees: an algorithm and
 distance hierarchy.
  Pattern Recognition 34(1), 127--137, 2001.

 \bibitem {wang02} H. Wang, W. Wang, J. Yang, P.S. Yu,
 Clustering by pattern similarity in large data sets.
 In: M.J. Franklin, B. Moon, A. Ailamaki (eds),
 Proc. of the 2002 ACM SIGMOD Int. Conf. on Management of Data,
 394--405, 2002.

  \bibitem {wang09a} C.-H. Wang,
  Outlier identification and market segmentation using kernel-based clustering techniques.
  Expert Systems with Applications 36(2), Part 2, 3744--3750, 2009.

 \bibitem {wang09} X. Wang, X. Wang, D.M. Wikes,
 A divide-and-conquer approach for minimum spanning
 tree-based clustering.
 IEEE Trans. KDE 21(7), 945--958, 2009.

 \bibitem {wang12} S. Wang, W. Chaovalitwongse, R. Babuska,
 Machine learning algorithms in bipedal robot control.
 IEEE Trans. SMC, Part A 42(5), 728--743, 2012.

 \bibitem {wang13} Q. Wang, E. Fleury,
 Overlapping community structure and modular overlaps
 in complex networks. In:
 T. Ozyer, Z. Erdem, J. Rokne, S. Khoury (eds.),
 Mining Social Networks and Security
 Informatics,
 Lectire Notes in Social Networks, Springer,
  15--40, 2013.

 \bibitem {was94} S. Wassserman, K. Faust,
 Social Network Analysis: Methods and Applications.
 Cambridge Univ. Press, Cambridge, 1994.

  \bibitem {white05} S. White, P. Smyth,
 A spectral clustering approach to
 finding communities in graph.
 In: SIAM Data Mining Conference,
 76--84, 2005. 

 \bibitem {wells12} A.F. Wells,
 Structural Inorganic Chemistry.
 Oxford Univ. Press, 2012.

  \bibitem {weng02} E. Wenger, R. McDermott, W.M. Snyder,
 Cultivating Communities of Practice.
 Harvard Business Review Press, Cambridge, MA, 2002


  \bibitem {willett86} P. Willett,
  Similarity and Clustering in Chemical Information Systems.
  Research Studies Press, Letchworth, 1987.

  \bibitem {willet88} P. Willett,
 Recent trends in hierarchical document clustering:
  a critical review.
 Information Processing \& Management  24(5), 577--597, 1988.


 \bibitem {wilson72} R.J. Wilson,
  Inroduction to Graph Theory. Academic Press, New York, 1972

 \bibitem {wineb08} S. Winebrenner, D. Brulles,
 The Cluster Grouping Handbook: How to Challenge
 the Gifted Students and Improve Achiviement for All.
 Free Spirit Publishing, Minneapolis, MN, 2008.


  \bibitem {wongm82} M.A. Wong,
 A hybrid clustering method for identifying high-density
 clusters.
 J. of the American Statistical Association 77(380),
 841--847, 1982.

 \bibitem {wong99} C.C. Wong, C.C. Chen,
 A hybrid clustering and gradient descent approach for fuzzy
 modeling.
 IEEE Trans. SMC, Part B 29(6), 686--693, 1999.

 \bibitem {wus05} S. Wu, T.W.S. Chow,
 PRSOM: A new visualization mehtod by
 hybridizing multidimensional scaling and self-organizing map.
 IEEE Trans. on Neural Networks 16(6), 1362--1380, 2005.

 \bibitem {wu93} Z. Wu, R. Leahy,
 An optimal graph theoretic approach to data clustering:
 Theory and its application to image segmentation.
 IEEE Trans. PAMI 11(00), 1101--1113, 1993.

  \bibitem {wynants01} C. Wynants,
   Network Synthesis Problems.
    Kluwer Academic Publishers, Dordrecht, 2001.

 \bibitem {xie13} J. Xie, S. Kelley, B.K. Szymanski,
 Overlapping community detection in networks:
 The state-of-the-art and comparative study.
 ACM Computing Surveys 45(4), art. 443, 2013

 \bibitem {xu05} R. Xu, D. Wunsch II,
  Survey on clustering algorithms.
  IEEE Trans. Neural Networks 16(3), 645--678, 2005.

 \bibitem {xu09} R. Xu, D. Wunsch,
 Clustering.
 Wiley-IEEE Press, 2009.

 \bibitem {xux07} X. Xu, N. Yuruk, Z. Feng, T.A.J. Schweiger,
 SCAN: a structural clustering algorithm for networks.
 In:
 Proc. of the 13th ACM SIGKDD Int. Conf. on Knowledge Discovery
 and Data Mining
 SIGKDD'07,
 824--833, 2007.

 \bibitem {xu01} Y. Xu, V. Olman, D. Xu,
 Minimum spanning trees for gene expression data clustering.
 Genome Informatics 12, 24--33, 2001.

 \bibitem {xuz11} Z. Xu, M. Xia,
 Distance and similarity measures fir hesitant fuzzy sets.
 Information Sciences 181(11), 2128--2138, 2011. 

  \bibitem {xu14a} Z. Xu,
 Distance, similarity, correlation, entropy measures and
 clustering algorithms for hesitant fuzzy information.
  In: Z. Xu, Hesitant Fuzzy Sets Theory. Springer,
   165--279, 2014.

   \bibitem {yager86}  R.R. Yager,
 On the theory of bags.
 Int. J. of General Systems 13(1), 23--37, 1986.

 \bibitem {yag00} R. Yager,
 Intelligent control of the hierarchical agglomerative
   clustering process.
  IEEE Trans. SMC 30(6), 835--845, 2000.

  \bibitem {yan09} B. Yan, S. Gregory,
 Detecting communities in networks by merging cliques.
 In: Proc. 2009 IEEE Int. Conf. on Intelligent Computing and
 Intelligent Systems (ICIS 2009), 832--836, 2009.

 \bibitem {yang00} Q. Yang, J. Wu,
 Keep it simple:
 A case-base maintenance policy based on clustering and
 information theory.
 In:
 Advances in Artificial Intelligence,
 Springer, 102--114, 2000.

 \bibitem {yangj13} J. Yang, J. Leskovec,
 Overlapping community detection at scale:
  A nonnegative matrix factorization approach.
 In: Proc. of Sixth ACM Int. Conf. on Web Search and Data Mining
 WSDM 2013, 587--596, 2013. 

 \bibitem {yangj14} J. Yang, J. Leskovec,
 Overlapping communities explain core-periphery organization of
 networks.
 Proc. of the IEEE  102(12), 1892--1902, 2014.

  \bibitem {yangj14a} J. Yang, J. Leskovec,
 Structure and overlaps of ground-truth communities
 in networks.
 ACM Trans. on Intelligent Systems and Technology (TIST)
   15(2), art. 26, 2014. 

 \bibitem {yangj15} J. Yang, J. Leskovec,
 Designing and evaluation network communities based on
 ground-truth.
 Knowl. Inf. Syst. 42(1), 181--213, 2015.


 \bibitem {yang06} Y. Yang, M.S. Kamel,
 An aggregated clustering approach using multi-ant colonies
 algorithms.
 Pattern Recognition, 39(7), 1278--1289, 2006.

 \bibitem {yang10} G. Yang, M. Xiao, E. Cheng,
 J. Zhang,
 A cluster-head selection scheme for underwatger
 acoustic sensor networks.
 In:
 Proc. of 2010 Int. Conf. on Communicaitons and
 Mobile Computing (CMC),
 vol. 3, 188--191, 2010.

  \bibitem {yao75} A.C. Yao,
 An
 \(O(|E|log log |V|)\)
  algorithm for finding
 minimum spanning trees.
 Inform. Process. Lett. 4(1), 21--23, 1975.

  \bibitem {yeh86} D.Y. Yeh.
  A dynamic programming approach to the complete set partitioning
  problem.
  BIT Numerical Mathematics, 26(4), 467-–474, 1986.

 \bibitem {yeu01} K.Y. Yeung, W.L. Ruzzo,
 Principal components analysis for clustering gene expression
 data.
 Bioinformatics 17(9), 763--774, 2001.


  \bibitem {youn04} O. Younis, S. Fahmy,
  HEED: A hybrid, energy-efficient, distributed
  clustering approach
  for Ad Hoc sensor networks.
 IEEE Trans. Mob. Comput. 3(4), 366--379, 2004.

 \bibitem {youn06} O. Younis, M. Krunz, S. Ramasubramanian,
 Node clustering in wireless sensor networks:
 Recent developments and deployment challenges.
 IEEE Networks 20(3), 20--25, 2006.

 \bibitem {yu07} M. Yu, K.K. Leung, A. Malvankar,
 A dynamic clustering and energy efficient routing
 technique for sensor networks.
 IEEE Trans. Wireless Communications 6(8), 3069--3079, 2007.

 \bibitem {yuang13} F.W. Yuang,
 Multidimensional Scaling: History, Theory, and Applications.
 Psychology Press, 2013.

 \bibitem {zach77} W.W. Zachary,
 An information flow model for conflict and fission
 in small groups.
 J. of Anthropological Research 33, 452--473, 1977.

   \bibitem {zadeh65} L.A. Zadeh,
 Fuzzy sets.
 Information and Control 8(3), 338--353, 1965.

  \bibitem {zah07} D. Zaharie, F. Zamfirache, V. Negru,
 D. Pop, H. Popa,
 A comparison of quality criteria for unsupervised
 clustering of documents based on differential evolution.
 In: Proc. of Int. Conf. on Knowledge Engineering,
 Principles and Techniques KEPT2007,
 25--32, 2007.

  \bibitem {zahn71} C.T. Zahn,
 Graph-theoretical methods for detecting and describing gestalt
 clusters.
 IEEE Trans.  Computers 20(1), 68-86, 1971.

 \bibitem {zait97} M. Zait, H. Messatfa,
 A comparative study of clustering methods.
 Future Generation Computer Systems 13(2-3),
 149--159, 1997.

  \bibitem {zaka99} A. Zakarian, A. Kusiak,
 Forming teams: an analytical approach.
 IIE Transactions  31, 85--97, 1999.


 \bibitem {zamir99} O. Zamir, O. Etzioni,
  Grouper: a dynamic clustering interface to Web
  search results.
  Computer Networks 31(11), 1361--1374, 1999.


 \bibitem {zhang96} T. Zhang, R. Ramakrishnan, M. Livny,
 BIRCH: An efficient data clustering method
 for very large databases. In:
 Proc. 1996 ACM SIGMOD Int. Conf. on Management of Data,
 vol. 25, 103--114, 1996.


 \bibitem {zhang04} D. Zhang, Y. Dong,
 Semantic, hierarchical, online clustering of
 web search results. In:
 Advanced Web Technologies and Applications,
 Springer, 69--78, 2004.

 \bibitem {zhang07} C.W. Zhang, H.L. Ong,
 An efficient solution to biobjective generalized assignment
 problem.
 Advances in Engineering Software 38(00), 50--58, 2007.

 \bibitem {zhang15} X. Zhang, Z. Xu,
 Hesitant fuzzy agglomerative hierarchical clustering
 algorithm.
 Int. J. of Systems Science 46(3), 562--576, 2015.

 \bibitem {zhong03} S. Zhong, J. Ghosh,
 A unified framework for model-based clustering.
 J. of Machine Learning Research 4, 1001--1037, 2003.

\bibitem {zhong05} S. Zhong, J. Ghosh,
 Generative model-based document clustering:
 a comparative study.
 Knowl. Inf. Syst. 8(3), 374--384, 2005.

 \bibitem {zhong05a} S. Zhong, T.M. Khoshgoftaar,
 S.V. Nath,
 A clustering approach to wireless network intrusion detection.
 In:
 Proc. of 17th IEEE Int. Conf. on Tools with Artificial
 Intelligence ICTAI 2005,
 190--196, 2005.

 \bibitem {zhong10} C. Zhong, D. Miao, R. Wang,
 A graph-theoretical clustering method based on two rounds of
 minimum spanning trees.
 Pattern Recognition 43(3), 752--766, 2010.

 \bibitem {zhou05} D. Zhou, J. Li, H. Zha,
  Ding, Zhou, Jia Li, Hongyuan Zha,
 A new Mallows distance based metric for comparing clusterings.
 In:
 Proc. of the 22nd Int. Conf. on Machine Learning ICML 2005,
 1028--1035, 2005.

 \bibitem {zimek08} A. Zimek,
  Correlation Clustering.
  PhD Thesis,
  Faculty of Mathematics, Informatics, and Statistics,
  Univ., of Munchen, 2008.

 \bibitem {zin83} J. Zinnes, D. MacKay,
 Probabilistic multidimensional scaling:
 complete and incomplete data.
 Psychometrica 48, 27--48, 1983.

 \bibitem {ziv05} E. Ziv, M. Middendorf, C. Wiggins,
 Information-theoretic approach to network modularity.
 Physical Review E, vol. 71, no. 046117, 2005.

 \bibitem {zogg06} D. Zogg, E. Shafai, H.P. Geering,
 Fault diagnosis for heat pumps with
 parameters identification and clustering.
 Control Engineering Practice 14(12), 1435--1444, 2006.


  \bibitem {zop02} C. Zopounidis, M. Doumpos,
 Multicriteria classification and sorting methods:
 a literature review.
  EJOR
 138(2), 229--246, 2002.

  \bibitem {zup82} J. Zupan,
  Clustering of Large Data Sets.
 Research Studies Press Ltd., Taunton, UK, 1982.




 \end{thebibliography}
\end{document}